\documentclass[runningheads]{llncs}
\usepackage{graphicx}
\usepackage{amsmath,amssymb} % define this before the line numbering.
\usepackage{color}
\usepackage{caption}
\usepackage{array}
\newcolumntype{H}{>{\setbox0=\hbox\bgroup}c<{\egroup}@{}}
\usepackage{amsmath}
\usepackage{multirow}
\usepackage{subcaption}
\usepackage{rotating}
\usepackage{float}
\usepackage[export]{adjustbox}
\usepackage[dvipsnames]{xcolor}

\definecolor{turquoise}{cmyk}{0.65,0,0.1,0.3}
\definecolor{purple}{rgb}{0.65,0,0.65}
\definecolor{dark_green}{rgb}{0, 0.5, 0}
\definecolor{orange}{rgb}{0.8, 0.6, 0.2}
\definecolor{red}{rgb}{0.8, 0.2, 0.2}
\definecolor{darkred}{rgb}{0.6, 0.1, 0.05}
\definecolor{blueish}{rgb}{0.0, 0.3, .6}
\definecolor{light_gray}{rgb}{0.7, 0.7, .7}
\definecolor{pink}{rgb}{1, 0, 1}
\definecolor{greyblue}{rgb}{0.25, 0.25, 1}
\usepackage[dvipsnames]{xcolor}
\definecolor{teal}{rgb}{0, 0.39, 0.39 }
\definecolor{brown}{rgb}{0.6, 0.3, 0.39 }

\newcommand\tstrut{\rule{0pt}{2.4ex}}
\usepackage{algpseudocode}
\usepackage{algorithm}

\usepackage{amssymb}% http://ctan.org/pkg/amssymb
\usepackage{pifont}% http://ctan.org/pkg/pifont
\newcommand{\cmark}{\ding{51}}%
\newcommand{\xmark}{\ding{55}}%

\newcommand{\modelname}{BodyMetric\xspace}

\def\ECCV16SubNumber{9945}  % Insert your submission number here
\newcommand{\st}{}
\renewcommand\st[1]{}
\newcommand{\rmstd}{}
\renewcommand\rmstd[1]{}

\makeatletter
\DeclareRobustCommand\onedot{\futurelet\@let@token\@onedot}
\def\@onedot{\ifx\@let@token.\else.\null\fi\xspace}

\def\etal{\emph{et al}\onedot}
\makeatother

\usepackage{xspace}
\usepackage{amssymb}
\usepackage{multirow}
\usepackage{booktabs}

\usepackage{algorithm}
\usepackage{algpseudocode}
\usepackage[pagebackref,breaklinks,colorlinks]{hyperref}
\usepackage{hyperref}

\begin{document}
\title{BodyMetric: Evaluating the Realism of Human Bodies in Text-to-Image Generation} 

\titlerunning{BodyMetric}
\author{Nefeli Andreou \and
Varsha Vivek \and
Ying Wang \and 
Alex Vorobiov \and 
Tiffany Deng \and 
Raja Bala \and 
Larry Davis \and 
Betty Mohler Tesch
}

\authorrunning{N.~Andreou et al.}

\institute{Amazon}

\maketitle

\begin{abstract}
Accurately generating images of human bodies from text remains a challenging problem for state of the art text-to-image models. Commonly observed body-related artifacts include extra or missing limbs, unrealistic poses, blurred body parts, etc. Currently, evaluation of such artifacts relies heavily on time-consuming human judgments, limiting the ability to benchmark models at scale. We address this by proposing BodyMetric, a learnable metric that predicts body realism in images. BodyMetric is trained on realism labels and multi-modal signals including 3D body representations inferred from the input image, and textual descriptions. In order to facilitate this approach, we design an annotation pipeline to collect expert ratings on human body realism leading to a new dataset for this task, namely, BodyRealism. Ablation studies support our architectural choices for BodyMetric and the importance of leveraging a 3D human body prior in capturing body-related artifacts in 2D images. In comparison to concurrent metrics which evaluate general user preference in images, BodyMetric specifically reflects body-related artifacts. We demonstrate the utility of BodyMetric through applications that were previously infeasible at scale. In particular, we use BodyMetric to benchmark the generation ability of text-to-image models to produce realistic human bodies. We also demonstrate the effectiveness of BodyMetric in ranking generated images based on the predicted realism scores.

\keywords{computer vision, generative models, text-to-image, benchmark, dataset, virtual humans, body metric}
\end{abstract}
\section{Introduction}
\label{sec:intro}
\begin{figure}
    \centering
      \begin{subfigure}[b]{0.18\textwidth}
         \includegraphics[width=\textwidth]{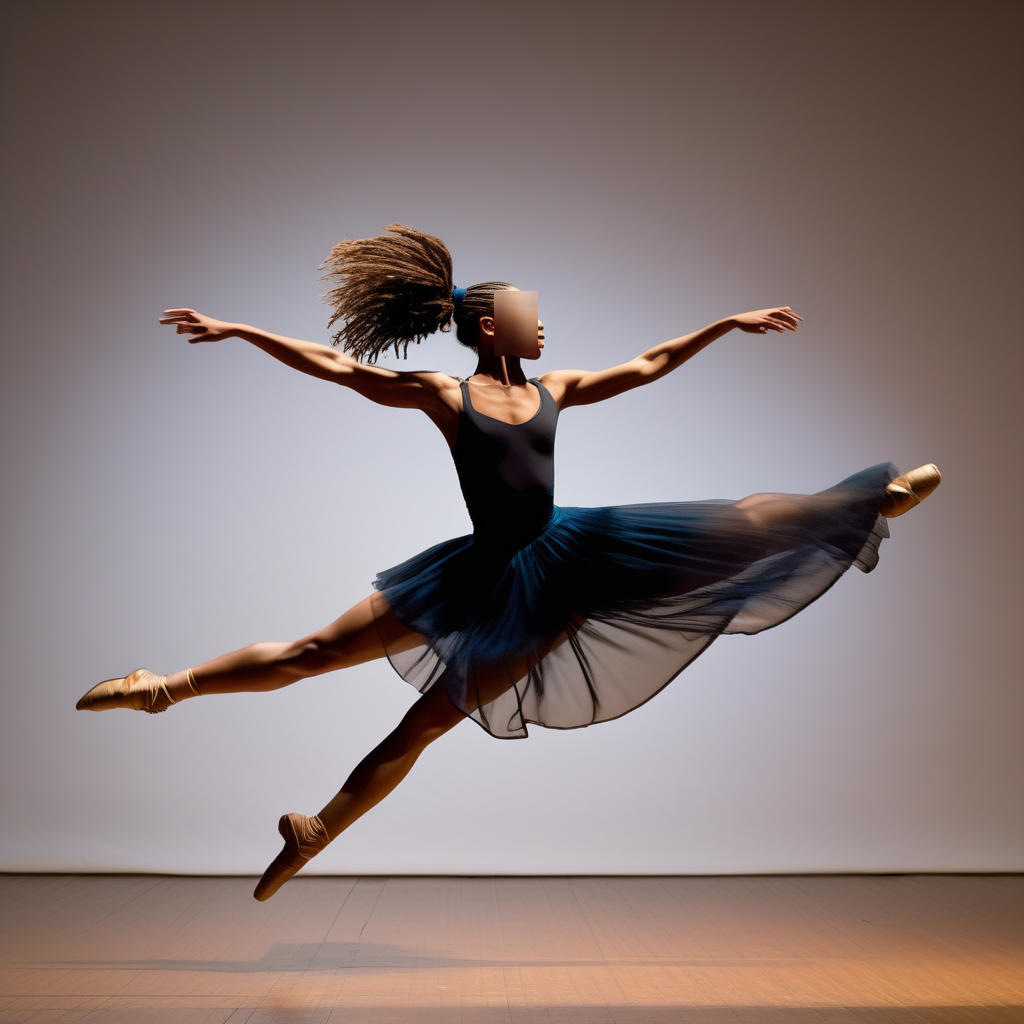}
    \end{subfigure}
        \begin{subfigure}[b]{0.18\textwidth}
         \includegraphics[width=\textwidth]{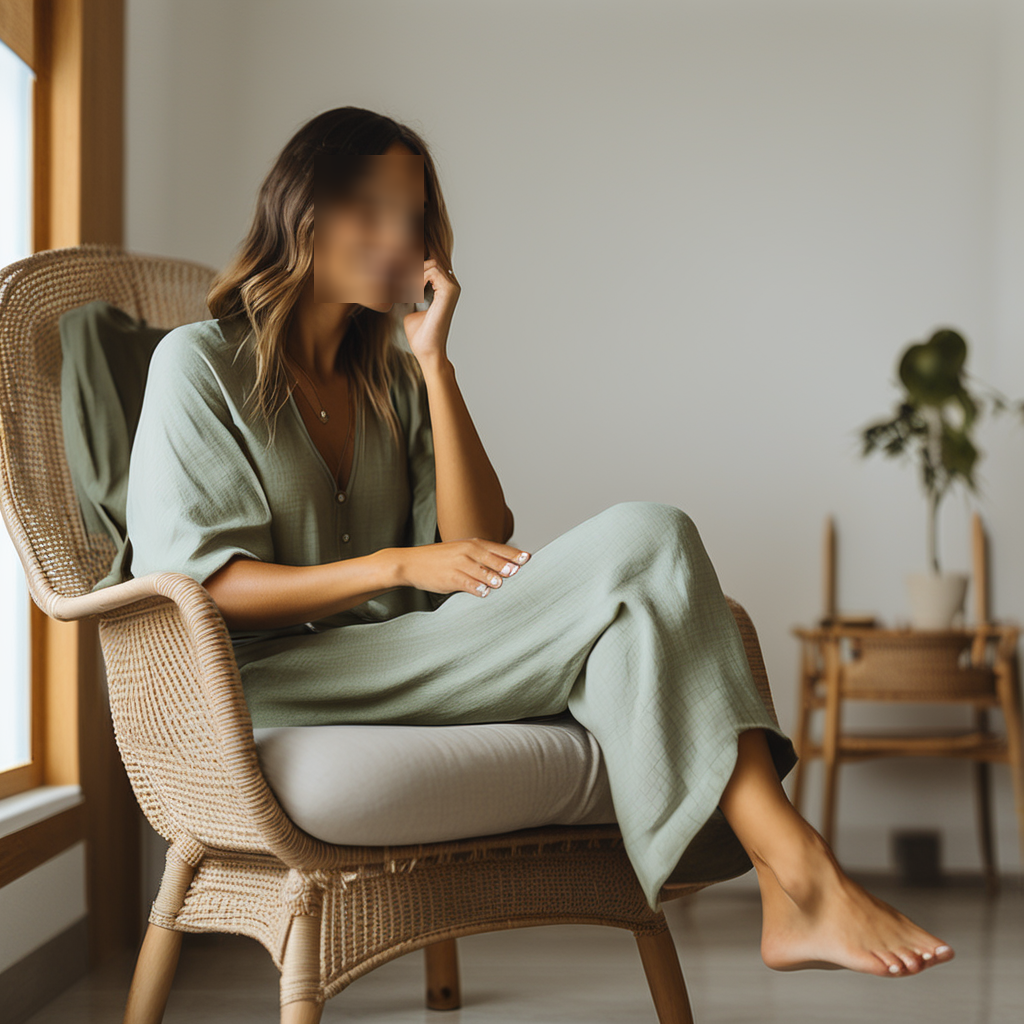}
    \end{subfigure}
      \begin{subfigure}[b]{0.18\textwidth}
       \includegraphics[width=\textwidth]{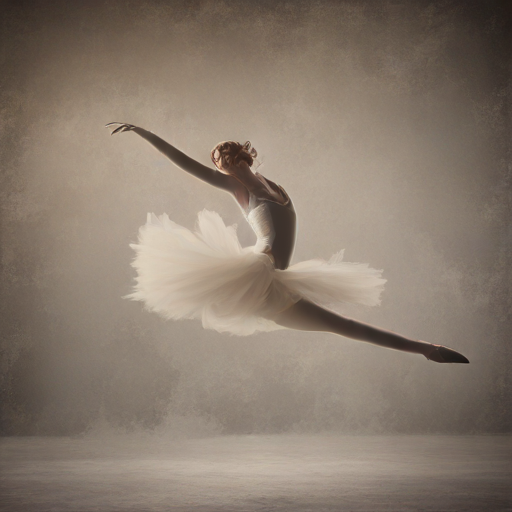}
    \end{subfigure}
    \begin{subfigure}[b]{0.18\textwidth}
         \includegraphics[width=\textwidth]{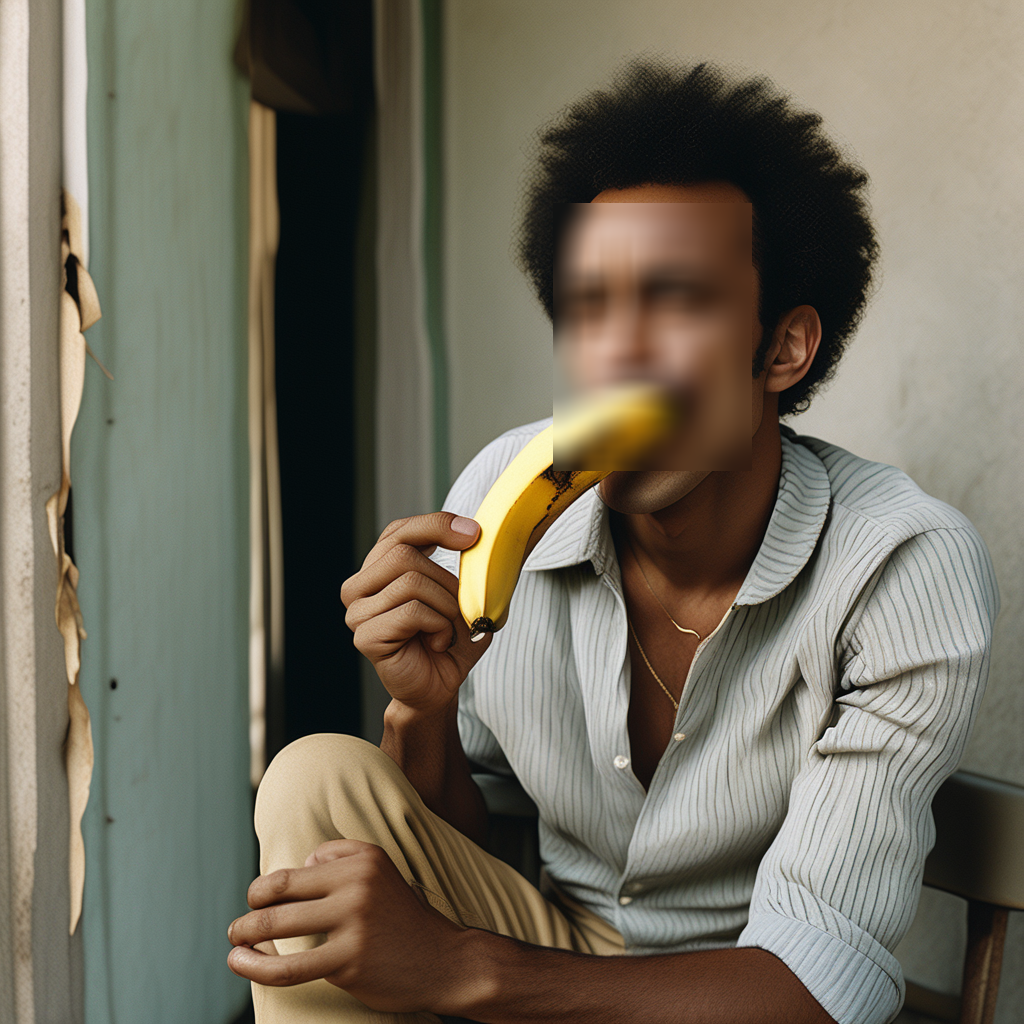}
    \end{subfigure}
     \begin{subfigure}[b]{0.18\textwidth}
      \includegraphics[width=\textwidth]{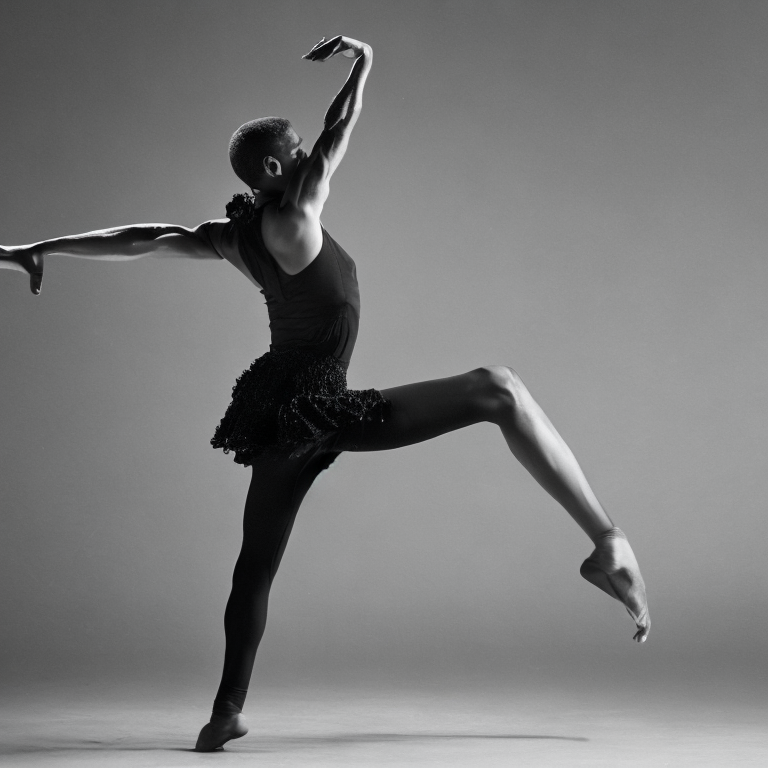}
    \end{subfigure}
  \caption{Common artifacts observed in T2I. Images were generated using SOTA models such as Stable Diffusion 2.1~\cite{Roombach:2022}, XL~\cite{Podel:2023:SDXL}, XL-Turbo~\cite{Sauer:2023:SDXLTurbo}. Faces are blurred for privacy considerations.}       \label{fig:main}
\end{figure}

% text-to-image generation 
Advances in generative modeling over recent years have enabled impressive progress across many domains of image generation~\cite{Zhan:2023,SOTA_Diffusion,Text2Live,Roombach:2022,ImaGen}. However, one area continues to present unique challenges - producing photorealistic human images directly from text. State-of-the-art generative models have excelled at artistic synthesis tasks but face additional difficulties when attempting to depict the complexity of human form and appearance to meet human standards of perceived authenticity. As seen in Fig.~\ref{fig:main}, this issue is notably manifested in the frequent generation of human figures with unrealistic body characteristics such as additional limbs, or abnormal body poses. While minor irregularities may go unnoticed in other generative domains, accurately depicting the human form poses a unique challenge. Even subtle deviations from typical human anatomy can negatively impact the perceived realism. This high standard arises from humans' deep-rooted ability to discern abnormal facial and anatomical features~\cite{Uncanny2,Uncanny}.

\vspace{-5mm}
Developing appropriate evaluation metrics is crucial for benchmarking progress in generative AI. Existing metrics mostly focus on evaluating the overall image quality \cite{FID,Salimans:2016}, alignment to the prompt ~\cite{Hessel:2021:CLIPScore,Hu:2023:TIFA} or the overall user preference with respect to image aesthetics \cite{Xu:2023:ImageReward,Wu:2023:HPSv1,Wu:2023:HPSv2}. Our empirical analysis, as illustrated in Fig.~\ref{fig:traditional_metrics}, demonstrates that standard assessment techniques have limited sensitivity to human body related artifacts, such as extra/missing limbs or unnatural poses. Therefore, most existing works resort to extensive human evaluation for this purpose - which is highly time consuming, acting as a bottleneck for model improvements.
To alleviate this bottleneck we propose BodyMetric, a novel trainable metric specialized on the realism of human bodies in images. 
% We propose a solution to this problem with a new automated evaluation paradigm that is specialized on the demanding requirements of human body rendering.
\begin{table}
    \centering
    \begin{tabular}{c|cccc}
        \multicolumn{1}{c}{} & \multicolumn{2}{c}{a ballerina in tight clothing dancing.} &  \multicolumn{2}{c}{a photo of a hip-hop dancer.} \\
        \multicolumn{1}{c}{}& \includegraphics[width=2.5cm]{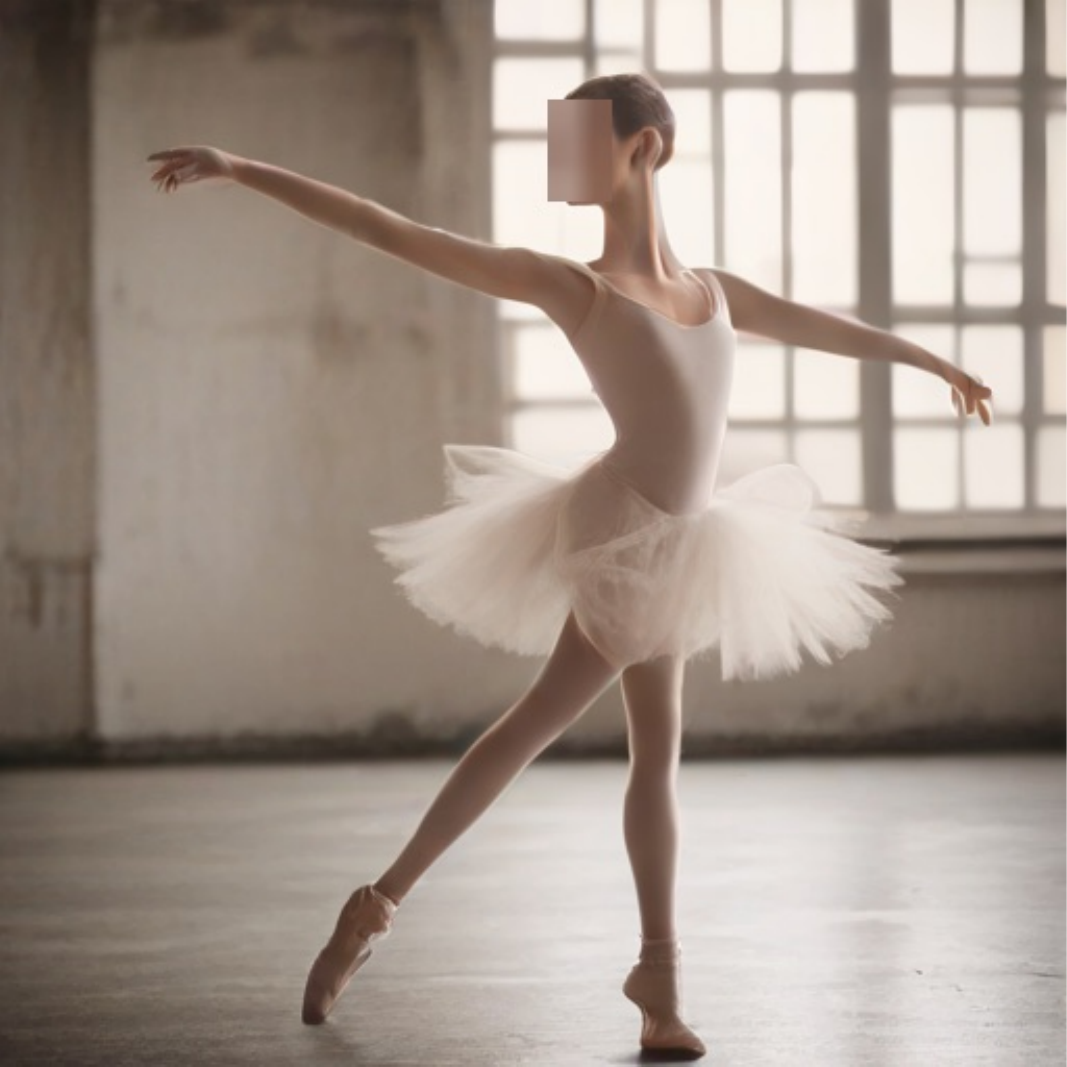} & \includegraphics[width=2.5cm]{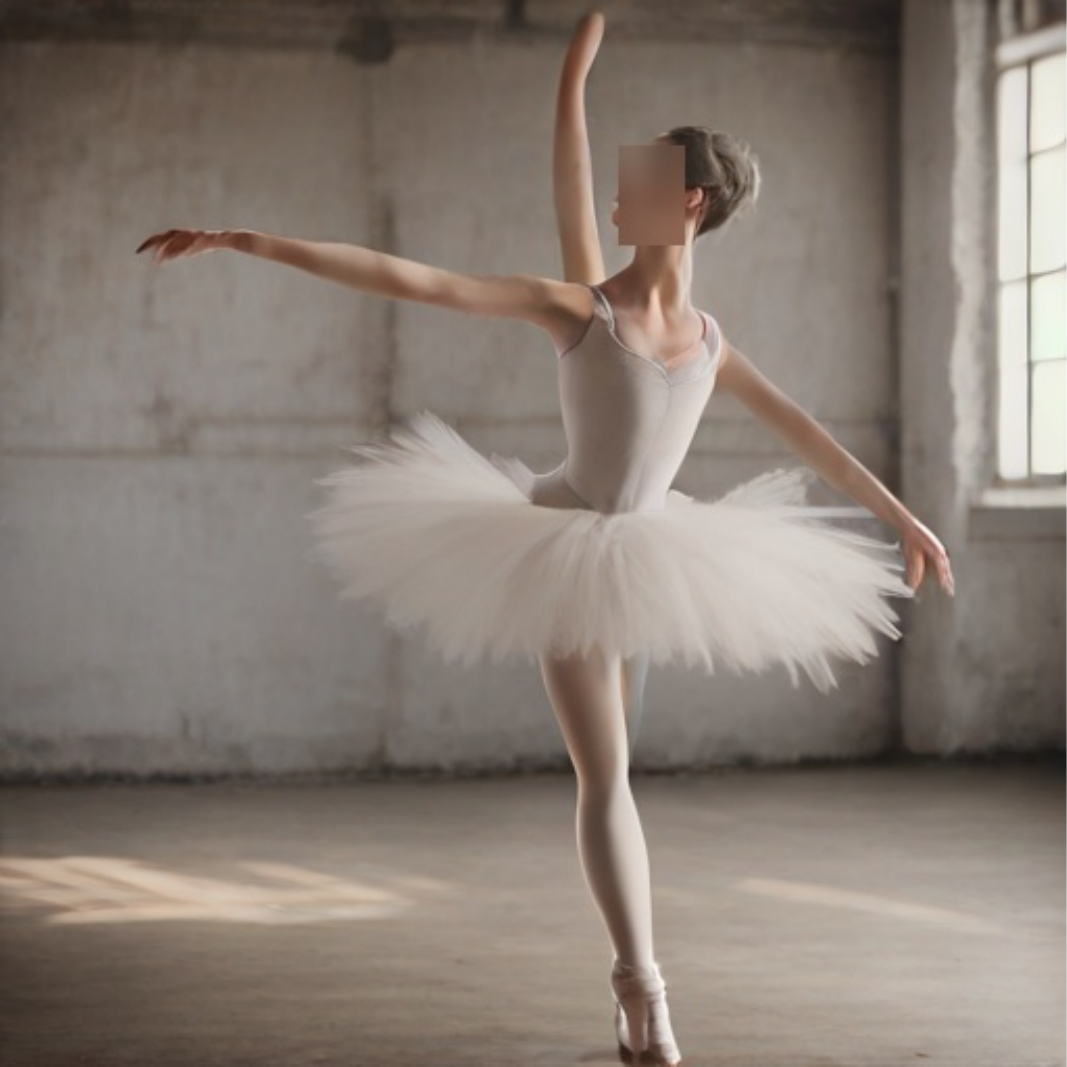} & \includegraphics[width=2.5cm]{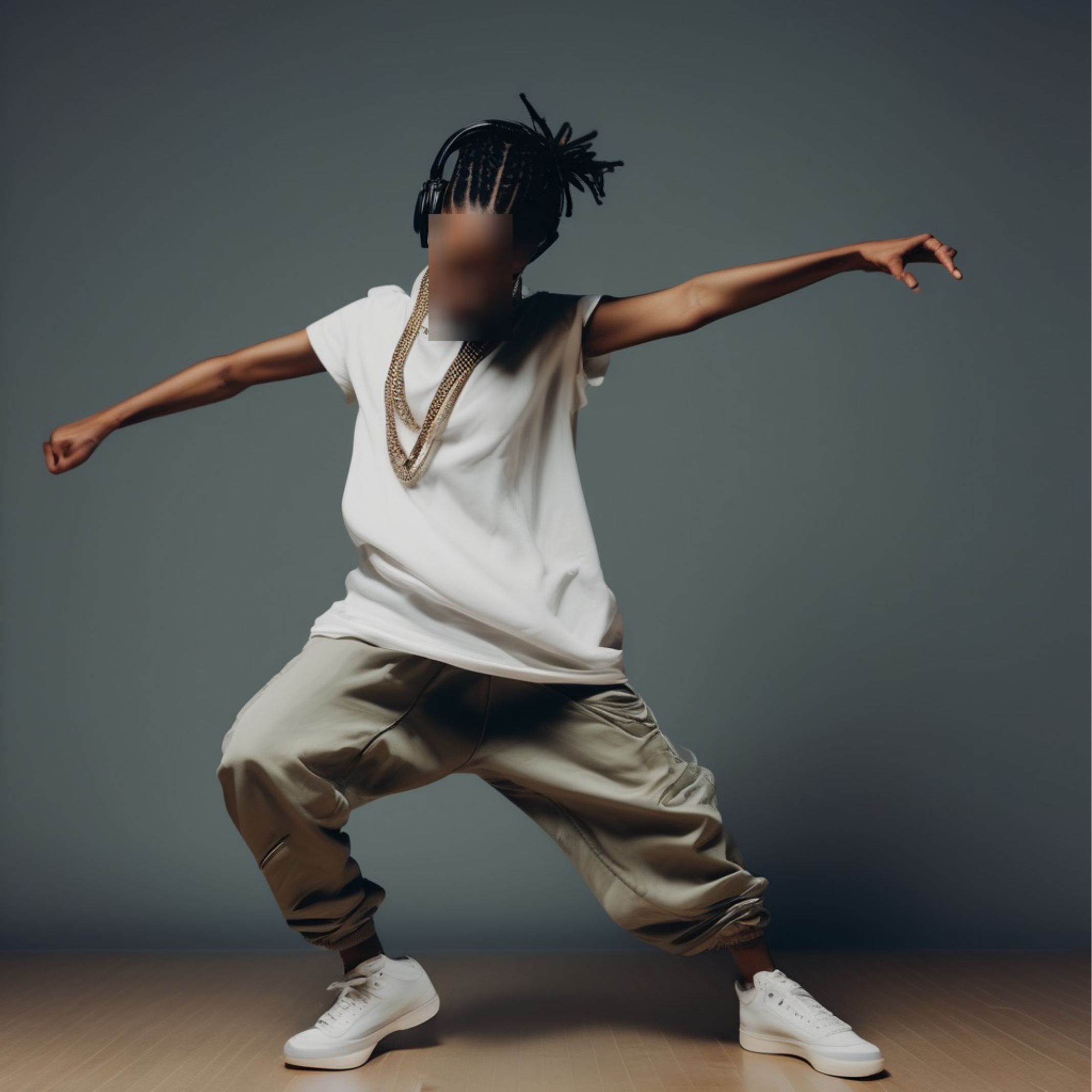} & \includegraphics[width=2.5cm]{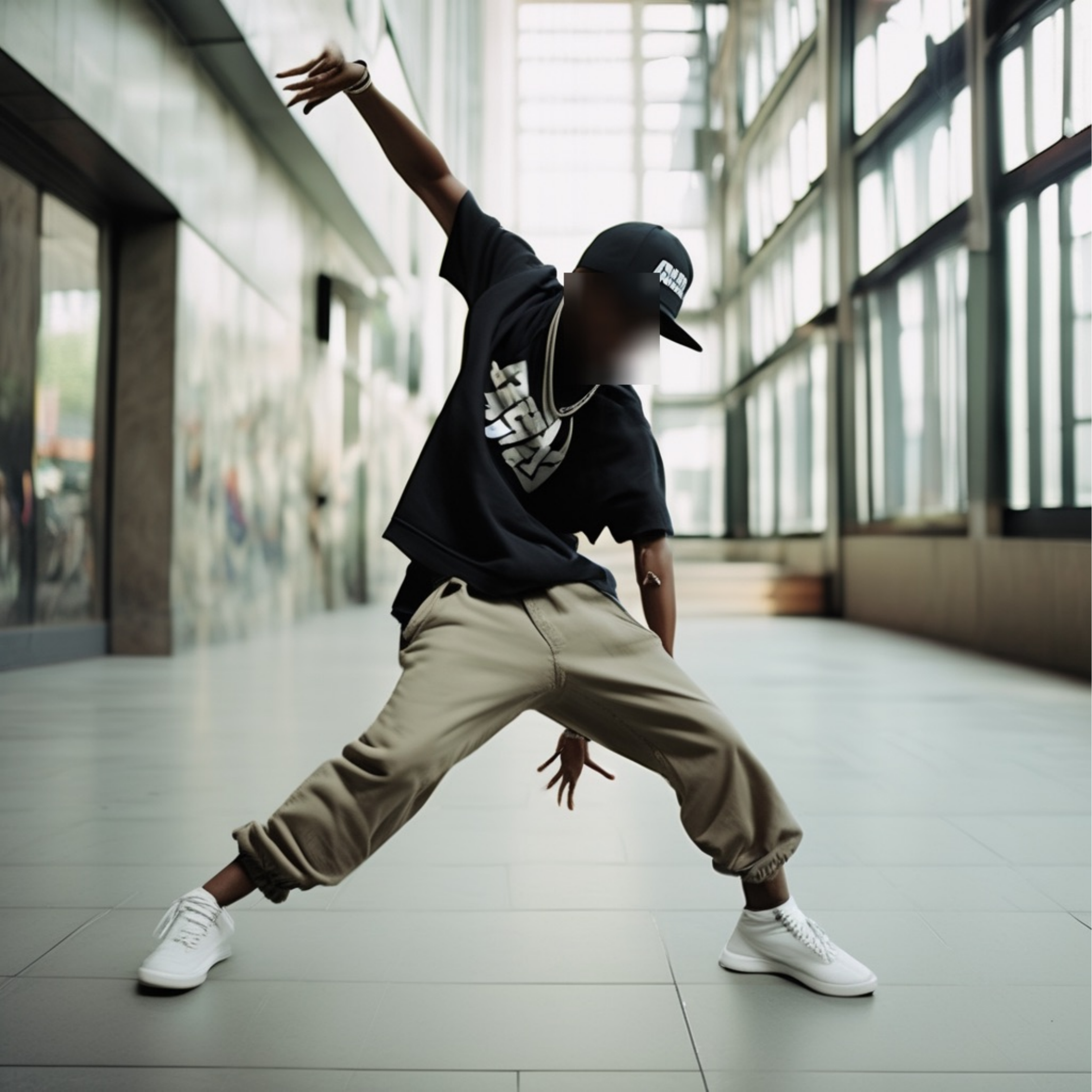} \\
        \hline
       \hline
        \scriptsize{HPS\cite{Wu:2023:HPSv2}}&\xmark &\cmark & \xmark& \cmark \\
        \scriptsize{IR\cite{Xu:2023:ImageReward}} &\xmark &\cmark & \xmark& \cmark \\
        \scriptsize{PS\cite{Kirstain:2023:PickScore}}&\xmark &\cmark & \xmark& \cmark \\
        \hline
        \scriptsize{\textbf{BodyMetric}} & \cmark &\xmark &\cmark &\xmark \\
    \end{tabular}
    \captionsetup{type=figure}
    \vspace{6pt}
    \caption{Traditional metrics e.g., Human Preference Score (HPS), ImageReward (IR), PickScore (PS) favor images despite unrealistic body features. Instead, \modelname captures unrealistic body features and successfully selects images with realistic bodies.}
    \label{fig:traditional_metrics}
\end{table}
We ground the definition of body realism on a 3D human body model learnt using thousands of 3D real body scans~\cite{SMPLx:2019}. Realistic human body structures encompass a wide range of body shapes, sizes, and proportions, and may include imperfections and asymmetries. Human bodies have articulated joints at the shoulders, elbows, wrists, hips, knees, and ankles, with specific degrees of freedom and ranges of motion. We consider as unrealistic, structures which do not exhibit a recognizable humanoid shape and morphology with distinct body parts such as head, torso, arms, and legs, or structures which exhibit articulations beyond the specific ranges of motion. Trained with multi-modal information and a novel architecture leveraging a 3D body prior, BodyMetric is designed to capture artifacts such as extra or missing limbs, deformed limbs and unnatural poses. We systematically identify such artifacts and carefully design BodyRealism, a richly curated multi-modal dataset that contains images of humans along with their corresponding text descriptions, high-quality body-realism scores, and 3D body representations. 
\noindent Our contributions can be summarised as follows:
\begin{itemize}
     \item BodyRealism, a dataset of $\sim$30k generated and real images paired with multi-modal signals such as text descriptions, high-quality body realism annotations collected in collaboration with expert annotators, and 3D body representations.
     \item BodyMetric, a learnable metric that evaluates the realism of human bodies in 2D images. BodyMetric leverages a 3D body prior to obtain its ``body-aware'' capabilities.
   
    \item Applications of \modelname such as benchmarking text-to-image models, or ranking generated images based on human body realism. 
    
\end{itemize}

\section{Related Work}
\label{sec:related_work}

% \subsection{Generative Models for Image Synthesis}
%  1¶ for VAEs, VQ-VAEs, GANs, normalizing flows
\subsection{Text-to-Image Generation}
Image generation has drawn considerable attention in recent years, with significant improvements in the capabilities of the state-of-the-art models. Existing image generation models utilise architectures such as (Vector-Quantized) Variational Autoencoders (VAEs, VQ-VAEs)~\cite{Kingma:2013:VAE,Oord:2021}, GANs~\cite{Bengio:2015}, and normalizing flows (NFs)~\cite{Kingma:2018}.

% . Even though they have demonstrated impressive results, GANs suffer from convergence issues and vanishing gradients, which make training challenging. These models suffer from mode collapse, struggling to generate diverse outputs. Oord~\etal~\cite{Oord:2021} introduced which addresses some of the shortcomings of VAEs by utilising a discrete latent space and a learnt prior to enhance the learnt representations in variational inference. While VAEs and VQ-VAEs only approximate the value of the latent variables, normalizing flow architectures~\cite{Kingma:2018} enable exact latent-variable inference due to a series of transformations which make calculation of log-likelihood tractable. Normalizing flows demonstrate good capabilities in capturing rich and complex distribution, while their design leads to increased efficiency both during training as well as inference time.

% \noindent \textbf{Diffusion Models}
Recently, diffusion models are used to generate images from text ~\cite{Dhariwal:2021,Ho:2022,Roombach:2022,Song:2020,Roombach:2022,pernias2023wuerstchen}. These models can capture complex data distributions, leading to better sampling quality compared to GANs. 
%They leverage iterative denoising processes to generate high-quality images from noised latents. 
Diffusion models can be conditioned on a variety of control signals such as text or pose, using classifier~\cite{Dhariwal:2021} or classifier-free~\cite{Ho:2022} guidance. Despite significant advances, these generative models can still fail to generate realistic humans in images.

\subsection{Datasets}
For our task, datasets of both real and generated images are valuable. Real image datasets like ImageNet~\cite{ImageNet} and MS COCO~\cite{mscoco}, contain images of humans with a diverse range of backgrounds along with textual descriptions. Additionally, several synthetic image datasets (DiffusionDB~\cite{Wang:2022:DiffusionDB}, OpenParti, HPSv2~\cite{Wu:2023:HPSv2}, Pick-a-Pic~\cite{Kirstain:2023:PickScore}) are created using generative text-to-image models. BodyRealism includes a subset of text prompts from existing datasets which are used to generate images of humans. In addition, it contains a subset of in-the-wild images of humans, curated from MS COCO. In contrast,to the above datasets covering a wide range of categories, BodyRealism only contains images of humans. Similar to OpenParti, Pick-a-Pic and HPSv2 BodyRealism contains preference scores for each image. Different from these datasets, the scores in BodyRealism are collected from expert annotators and are designed to reflect the realism of the human body in terms of anatomy.

\subsection{Evaluation Metrics}
Commonly adopted evaluation techniques for images can be classified to image fidelity, text-image alignment and user preference. The most reliable process to assess image quality is to train human annotators to rate the images in terms of visual quality or text alignment. However, this can be a cumbersome and time demanding process and the quality heavily depends on the expertise, background and clarity of annotation instructions.
\vspace{6pt} \\
\noindent \textbf{Image Fidelity}
Image fidelity metrics can be broadly characterised to those utilising low-level image features or deep features. Metrics based on low-level image features, such as SSIM~\cite{Zhou:2004:SSIM} and PSRN, lack semantic context or may assume pixel-wise independence, thus, failing to properly capture image fidelity. Since deep visual representations obtained from large pre-trained models have been found to carry semantic knowledge~\cite{Zhang:2018:LPIPS}, they were used to define metrics such as the Inception Score(IS)~\cite{Salimans:2016} and Fréchet Inception Distance (FID)~\cite{FID}. Yet, existing works challenge the credibility of such metrics for non-ImageNet datasets~\cite{Ali:2019,Zhou:2019:HYPE}. We argue that the diverse information encoded by visual features (background, shadows, and objects), hinders body realism scoring, and hypothesize that a targeted method focusing on the body features would likely yield superior results for this task.
\vspace{6pt} \\
\noindent \textbf{Text-Image Alignment}
 Widely adopted metrics, utilise CLIP as their backbone, since it has been found to be a good candidate for reference-free evaluation of images. In particular, CLIPScore~\cite{Hessel:2021:CLIPScore} defines the text-image alignment based on the cosine similarity of the CLIP text and image embeddings without using any reference images, while RefCLIPScore~\cite{Hessel:2021:CLIPScore} achieves a higher text-image correlation by utilising reference images when available. Instead of using CLIP, TIFA~\cite{Hu:2023:TIFA} utilises visual question answering to capture text-image relevance. In particular a large language model (LLM) is used to create question-answer pairs based on the text prompt and assumes that a good image candidate should be able to provide accurate answers to the questions using VQA. While this technique works well for coarse image features, it might struggle to capture finer details which relate to fingers and blurriness. Furthermore, since TIFA is bounded by the LLM it can be difficult to formulate questions relevant to human anatomy, making it less suitable in evaluating the realism of bodies in images. 
\vspace{8pt} \\ \\ 
\noindent \textbf{Learnable User Preference}
Recent works~\cite{Wu:2023:HPSv1,Kirstain:2023:PickScore,Xu:2023:ImageReward} show that existing metrics do not align with human preferences, making them unreliable for evaluating text-to-image models. Wu~\etal~\cite{Wu:2023:HPSv1,Wu:2023:HPSv2} and Kirstain~\etal~\cite{Kirstain:2023:PickScore} take inspiration from Natural Language Processing (NLP) tasks and collect a dataset of human preferences in order to train a learnable user preference metric based on CLIP. The learnable metric is trained with an objective function similar to InstructGPT's~\cite{Ouyang:2022:InstructGPT}. However, the collected annotations describe the image as a whole and are not robust when it comes to evaluating human bodies. Xu~\etal~\cite{Xu:2023:ImageReward} learn a Reward Model (RM) similar to those for language models, by using BLIP~\cite{Li:2022:BLIP} as the preference model backbone. The learnt RM can be used to improve text-to-image generation using Reward Feedback Learning (ReFL). Inspired by these works, we design \modelname particularly tailored towards the assessment of body realism in images. 

\section{BodyRealism Dataset}
\label{subsec:dataset}

We strive for a dataset consisting of a diverse range of human poses and actions by formulating the prompts accordingly. Other aspects of human diversity (such as ethnicity, gender etc) largely depend on generative capabilities of the models that are used to produce the images. In order to train a learnable metric specifically tailored to assess anatomy-related artifacts, we associate each image with a combination of human annotations and multi-modal signals. Specifically, we tag each text-image pair with body realism annotations, and body-prior information that is leveraged to explicitly introduce the notion of human anatomy into our metric. Formally, BodyRealism Dataset is defined as a set of tuples $\mathcal{D} = ({x}, {y}, {r}, {b})$ where ${x, y, r, b}$ correspond respectively to text prompt, images, realism scores and 3D body representation. These are described in detail next.

\subsection{Text-Image Pairs}
\label{sec:image_gen}
\vspace{6pt} 
\noindent \textbf{Text Prompts} To generate the images we define a set of text prompts curated from existing text-image datasets (e.g., ImageNet, MS COCO, TiFa, Pick-a-Pic, DiffusionDB, OpenParti). Since we are interested in generating images with humans, we filter the text prompts accordingly.

\vspace{6pt} 
\noindent \textbf{Image Generation} We utilise SOTA models such as stable diffusion variants\cite{Roombach:2022,Podel:2023:SDXL,Sauer:2023:SDXLTurbo} to generate images from text prompts.  We incorporate negative prompts (such as ``black and white'', ``sepia'') during generation to overcome the biases of models to predominantly generate images of a particular style. While negative prompts improve the overall aesthetics of generated images, we find that they are not successful in improving body related artifacts. We show this by generating images with negative prompts related to body realism (such as ``deformed body'') and find that even with highly crafted prompts, most T2I models consistently produce unrealistic human bodies (see Fig.~\ref{fig:main}). Including such images in our dataset ensures that our model is trained to handle difficult/persistent body-related artifacts which cannot be resolved with prompt engineering. While we make efforts to limit the text prompts to a subset which will lead to the generation of individual humans in images, inevitably some generated images might contain more than one human or no humans at all. Thus, we use instance segmentation to filter out such images. Since human annotators are involved in the process, we remove NSFW images using the Amazon Rekognition Content Moderation~\cite{Rekognition}. Due to privacy concerns, we blur all faces in the dataset using MTCNN~\cite{MTCNN}. 
 
\subsection{Body Realism Annotation}
\label{sec:realism_annotation}
The quality of annotations heavily depends on the expertise of annotators and the instructions provided to them. To ensure consistency and quality, all annotators were  trained and instructed to follow a Standard Operating Procedure (SOP) including representative images (see Sup. Mat.). As illustrated in Fig.~\ref{fig:BodyRealism}, images are annotated on a 1-10 scale. We opt for a 1-10 scale, instead of forcing a choice between pairwise image comparisons for multiple reasons: first, since both images could display artifacts, forcing a choice would lead to incorrect labels. Second, independent realism scores per image give us the flexibility to experiment with the formation of training pairs. Finally, a scale provides a higher margin of error especially for multiple annotators, allowing us to devise a tailored strategy for score aggregation (Alg.~\ref{alg:cap}). The annotators are instructed to utilise a mental three-tiered severity scale to categorise body artifacts as: (A) scores 1-3 corresponding to highly unrealistic images, (B) scores 7-10 corresponding to realistic images; (C) corresponding to moderate artifacts. The scores in each bucket are further mapped to fine-grained descriptors and exemplar images ensuring adequate characterization of occurring body artifacts. Images with noticeable inaccuracies in the larger limbs such as arms, legs or a highly deformed body pose, are considered as severe. Images with less conspicuous errors such as those in the smaller limbs - extra/missing fingers, slightly blurred body parts - are considered as moderate. High-scores (8 or more) correspond to negligible or no artifacts in the human bodies.
Annotators are instructed to label images which contain more than one human or no human at all as invalid. Since the annotation focuses purely on the realism of bodies the corresponding text is not shown to the annotator and the faces in the displayed images are blurred.

After examining the collected human annotations, we devise a tailored strategy to distill them into robust singular realism scores per image. Given image $y$ with annotations $r=\{r_j\} \text{ for } j \in [1,N]$ provided by $N=5$ annotators, we use the median and interquantile range (IQR) to filter outliers (see Sup. Mat.), ensuring high-quality body realism scores. We consider data samples as invalid when 3 or more annotators agree on the ``invalid-image'' label.

\subsection{Human Body Prior}
\label{sec:3dprior}
We ground our definition of realistic body on SMPL-X~\cite{SMPLx:2019}, a 3D body model of human body pose, hand pose, and facial expression, learnt using thousands of 3D real body scans. We obtain the SMPL-X parameters for each image using PIXIE~\cite{PIXIE:2021}. Grounded by a parametric 3D body model, PIXIE's body reconstructions confine to the real body structure regardless of body artifacts in the image. We enhance our dataset with 3D body parameters, including 3D mesh vertices, and 3D keypoints or pose parameters. In Sec.~\ref{sec:method}, we elaborate on the ways in which we leverage the body information as a prior in BodyMetric. 

\subsection{Statistics and Analysis}
\label{sec:statistics}
After filtering out invalid images, we end up with $\sim$30k images generated using $\sim$2k unique text prompts describing more than 200 diverse actions. Out of 30,622 generated images, 12,107 are labelled with score less than 3, and 11,178 with score higher than 7, ensuring adequate representation across the full quality range.
In addition to generated images, we include 1,705 real in-the-wild text-image pairs of humans from MS COCO, in order to offset any domain biases that the model might learn from generated images. Real images are consistently assigned a high score of 9. We do not assign a score of 10 for real images in order to account for any potential image related artifacts such as blur, obfuscation etc. To our knowledge, BodyRealism is the first dataset to provide scores focusing purely on the realism of human bodies in images. We aim to periodically update the dataset and metric, as we continue with our efforts to collect more annotations covering a wider span of generative models and conditioning prompts. 
\begin{table}
    \centering
    \begin{tabular}{cc}
         \includegraphics[width=5.5cm]{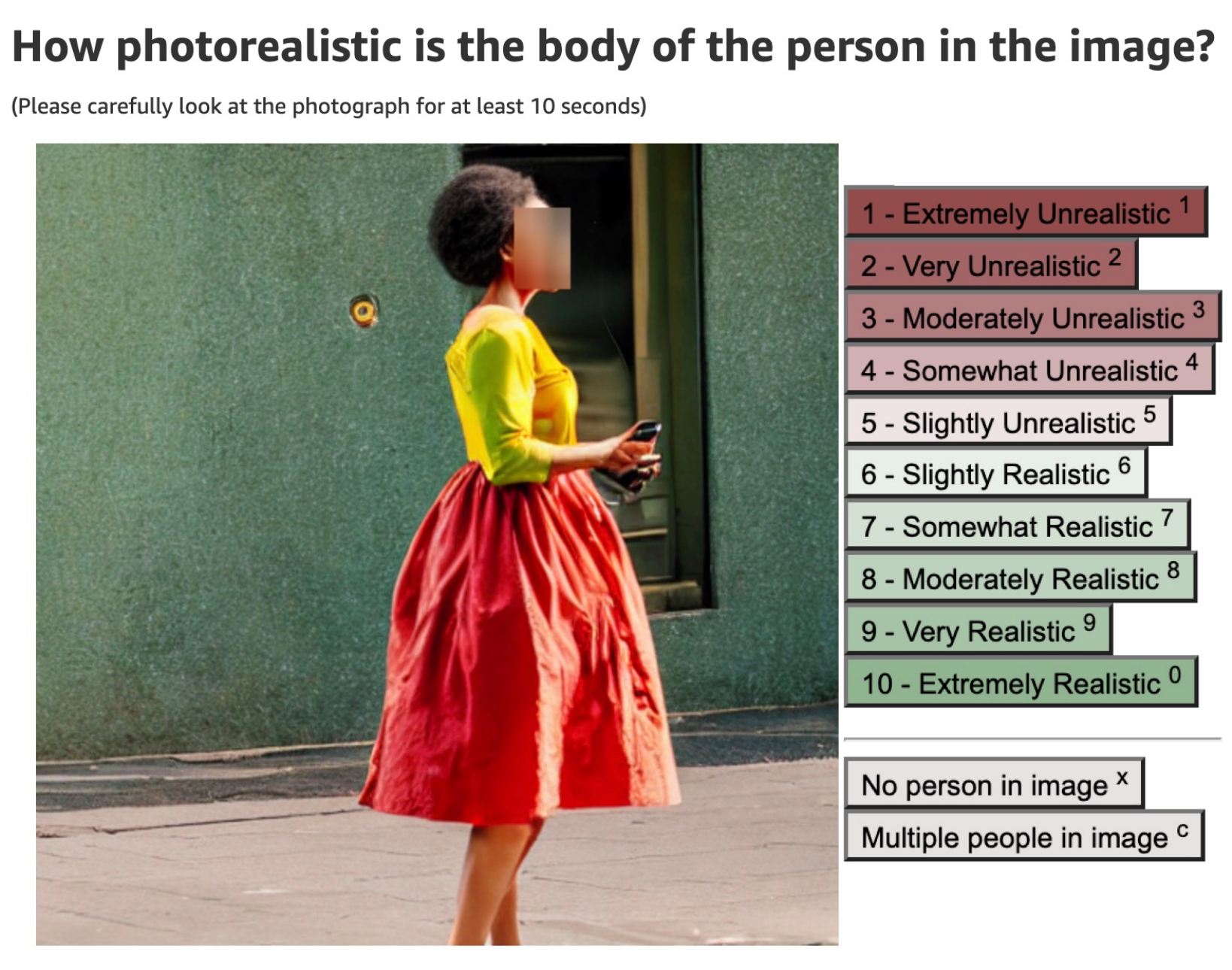}  & \includegraphics[width=6.7cm]{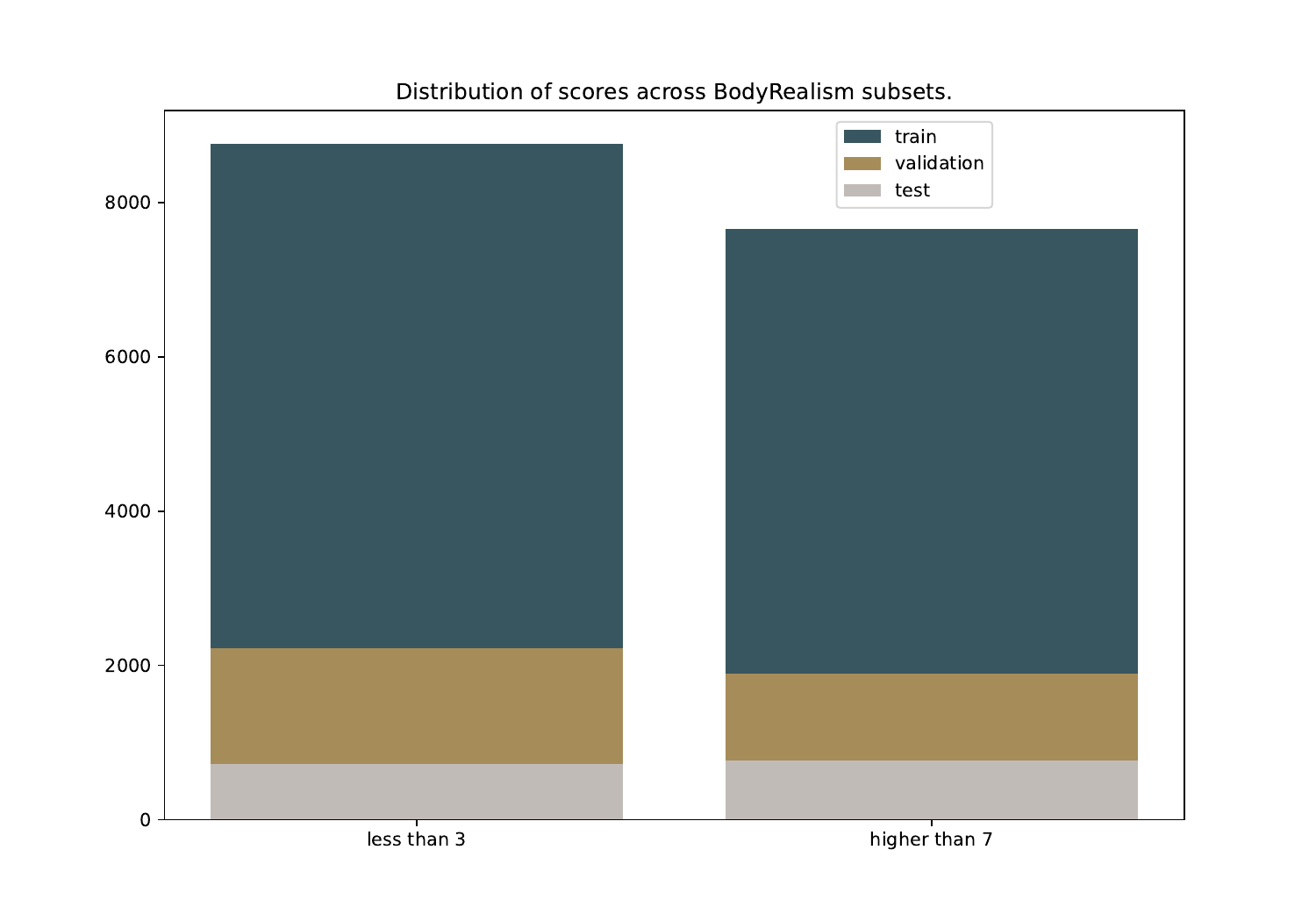} \\
    (a) & (b) \\
    \end{tabular}
    \captionsetup{type=figure}
    \caption{(a) Annotation template; (b) Body realism scores across BodyRealism subsets.}
    \label{fig:BodyRealism}
\end{table}
\section{BodyMetric}
\label{sec:method}
Given the collected BodyRealism Dataset, we train BodyMetric, a scoring function which measures the body realism in the images. We elaborate on how BodyMetric jointly leverages the multi-modal signals in BodyRealism. Realism annotations provide supervision during training to mimic human judgement. In addition to human annotations, we argue that anchoring BodyMetric on a 3D human body representation further strengthens the model's ``body-awareness''.

\subsection{Model Design and Training}
\vspace{-7mm}
 \begin{figure}[h]
    \centering
    \includegraphics[scale=0.41]{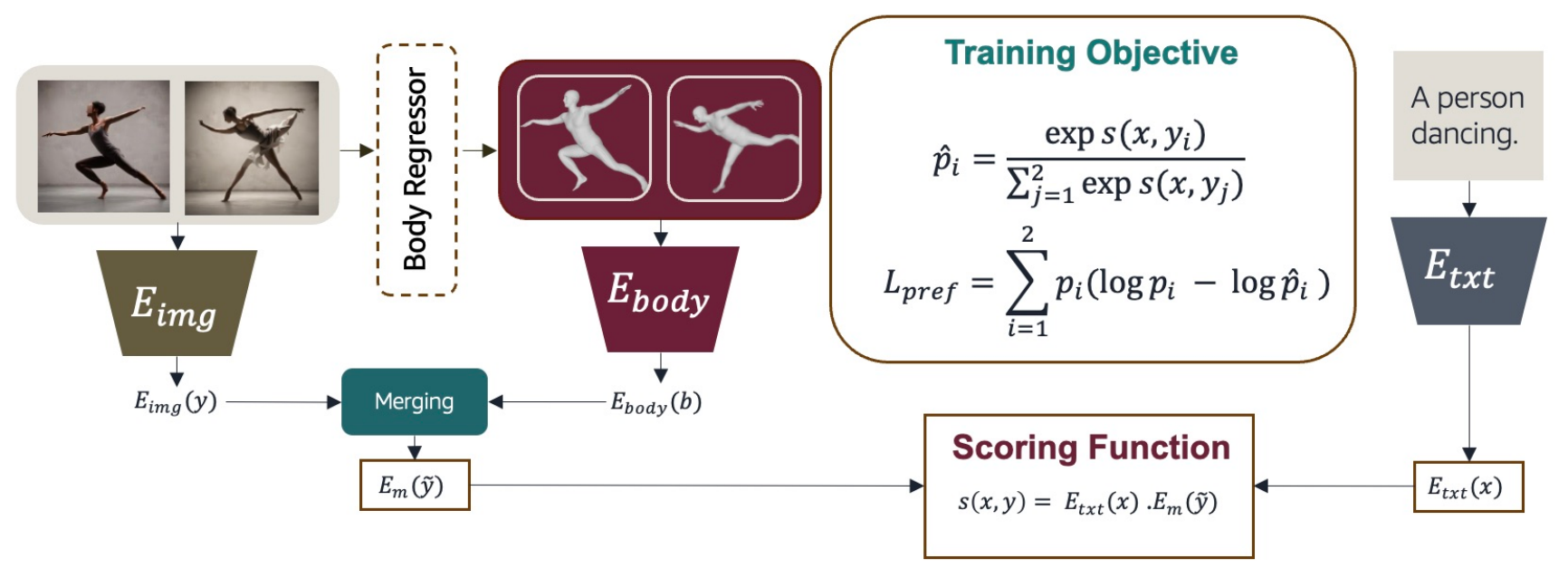}
    \caption{BodyMetric Architecture.}
    \label{fig:architecture}
\end{figure}
\vspace{-5mm}
\noindent \textbf{Model} Given an input image $x$, we infer the 3D keypoints of the SMPL-X body $b \in \mathbb{R}^{1 \times 435}$ using a body regressor~\cite{PIXIE:2021}. \modelname receives as input an image $x$, a text prompt $y$, and 3D keypoints of the SMPL-X body $b$, and outputs a body realism score $s \in \mathbb{R}$. As shown in Fig.\ref{fig:architecture}, the image and text are encoded through the corresponding CLIP encoders, yielding text and image embeddings $\mathop{E}_{txt}(x), \mathop{E}_{img}(y)$. The body representation is projected to the body latent space using a Multi-Layer Perceptron (MLP), yielding body features $\mathop{E}_{body}(b)$. The image and body features are merged to an enhanced feature $\mathop{E}_{m}(\Tilde{y})$ using another MLP. BodyMetric follows a CLIP-based architecture; the score is calculated using the inner product of embeddings, i.e., 
\vspace{-0.5mm}
 \begin{equation}
     s(x,\Tilde{y})=E_{txt} (x)  \cdot {E}_{m}(\Tilde{y}).
     \label{eq:intructGPT}
 \end{equation}
\vspace{-0.5mm}
\noindent \textbf{Objective} Following previous works, we formulate the objective of \modelname analogously to the reward model objective used in InstructGPT. In particular, given a prompt $x$, a pair of images $\{y_1,y_2\}$, and a preference distribution vector $p$ over the two images, the goal is to optimize the parameters of the scoring function $s$ by minimizing the KL-divergence between the preference $p$ and the softmax-normalized scores of $\Tilde{y}_1$ and $\Tilde{y}_2$, i.e.,
\vspace{-0.7mm}
\begin{equation}
    L_{pref} = \sum_{i=1}^{2}p_i(\log p_i - \log \hat{p}_i),
\end{equation}
\vspace{-0.7mm}
where
\vspace{-0.7mm}
\begin{equation}
    \hat{p}_i = \frac{\exp s(x,\Tilde{y}_i)}{\sum_{j=1}^2\exp s(x,\Tilde{y}_j)}.
\end{equation}
\vspace{-0.7mm}
 We follow existing preference metrics (e.g. HPS, PickScore) for the design of BodyMetric, without employing text dropout. We argue that although the text modality may not be necessary for most cases, it can provide an important signal for more diverse body representation beyond the typical anatomy captured by the SMPL-x structure.
 
To generate more distinct training examples, we only pair images with realism scores less than 3 and greater than 7, excluding pairs with intermediate scores. Further, we balance the distribution of data points across ties and non-ties. This results in $\sim$160k pairs for training, $\sim$10k for validation, and $\sim$1k for testing. In this case, $p$ takes a value of [1,0] if $y_1$ is preferred in terms of body realism, [0,1] if $y_2$ is preferred, or [0.5, 0.5] for ties. 
To minimize the risk of overfitting, we follow~\cite{Kirstain:2023:PickScore} and apply a weighted average across the batch with a weight inversely proportional to the frequency of each prompt in the dataset.  
\subsection{Implementation Details}
We train \modelname end-to-end on the BodyRealism dataset and initialize $\mathop{E}_{img}$ and $\mathop{E}_{txt}$ from PickScore~\cite{Kirstain:2023:PickScore}. \modelname is trained for 4,000 steps, with a learning rate of 3e-6, a total batch size of 64, and a warmup period of 500 steps, which follows a linearly decaying learning rate; the experiment takes around 3 hours with 8 A100 GPUs. We evaluate the model’s accuracy on the BodyRealism validation set in intervals of 100 steps, and keep the best-performing checkpoint. On an A100 GPU, BodyMetric runs at 0.08s per image pair with an additional 0.12s per image for PIXIE reconstructions. 
\section{Experiments}
\label{sec:experiments}
We evaluate BodyMetric's ability to select from a pair of images the one with the most realistic body. We ablate on our design choices, and perform comparisons with SOTA image evaluation metrics.
\subsection{Preference Prediction}
\noindent \textbf{Evaluation Protocol} We perform our experiments on the BodyRealism test set consisting of $\sim$1k pairs, generated from an independently curated subset of 181 MS COCO text prompts, and 54 synthetic captions describing complex and diverse actions (e.g. running, sitting, dancing, pirouetting). We compare pairs of images corresponding to the same prompt, and measure the performance based on the number of correct guesses. We consider a tied outcome when $| \hat{p}_1 - \hat{p}_2 |< t$ where $t$ is the tie threshold, defined separately for each model based on the accuracy on the validation set. 

\noindent \textbf{Comparison to the State-of-the-Art} We consider as baselines CLIPScore~\cite{Hessel:2021:CLIPScore} and PickScore~\cite{Kirstain:2023:PickScore}. Closest to ours, PickScore was trained on image preference annotations from real users. However, since PickScore's training data is not explicitly designed to reflect body realism in images, we create a stronger baseline by fine-tuning PickScore on the BodyRealism dataset, denoted as Base. Fig.~\ref{fig:base_vs_body} illustrates the improved performance of Base relative to PickScore, confirming the superiority of the former approach and importance of the BodyRealism dataset when evaluating body realism in images.

Fig.~\ref{fig:val_thresholds} shows the validation curve across tie thresholds for the different models. This illustrates that fine-tuning on the BodyRealism dataset is consistently effective for the task of evaluating human body realism across different tie thresholds. Tab.~\ref{tab:base_test} shows the accuracy on the BodyRealism test set, demonstrating the superiority of BodyMetric in comparison to Base as well as existing metrics such as PickScore and CLIPScore.

Fig.~\ref{fig:qual} demonstrates the performance of BodyMetric in pairs of images. Given a prompt and two images BodyMetric consistently identifies the image with fewer body related artifacts as the most realistic. The significance of having an explicit body prior as part of \modelname becomes more evident by looking at the correlation of predicted scores to the degree of body realism in each image. BodyRealism design choices ensure that artifact levels in images are reflected in their probability scores. As shown in Fig~\ref{fig:base_vs_body}, pairs where one of the images has major artifacts, such as extra/missing limbs (Col. 1), have significantly different BodyRealism scores, while those with a similar degree of artifacts (Col. 6) show a smaller difference in scores. This emphasises BodyMetric's ability to make more confident predictions when there is a noticeable difference in the body realism between the two images as well as in a wide range of diverse actions (Fig. ~\ref{fig:traditional_metrics}, ~\ref{fig:qual}). In addition, while any 3D shape inference will encounter minor errors, our observations show that the reconstruction errors are minimal and that Bodymetric is robust to these errors (see Fig.~\ref{fig:qual} Col. 1, 2).

\begin{table}[h]
\begin{minipage}[b]{0.32\linewidth}
\begin{tabular}{c|cccccc}
 \multicolumn{1}{c}{}&\multicolumn{1}{p{1.5cm}}{\scriptsize{a woman preparing a plate.}
}&\multicolumn{1}{p{1.5cm}}{\scriptsize{a man holding a fuchsia umbrella.}}
&\multicolumn{1}{p{1.5cm}}{\scriptsize{a woman sitting on a bench.}} 
&\multicolumn{1}{p{1.5cm}}{\scriptsize{a man in a white suit.}} & \multicolumn{1}{p{1.5cm}}{ \scriptsize{a woman sitting on a bench.}} & \multicolumn{1}{p{1.5cm}}{\scriptsize{a barefoot woman sitting in a chair.}} \\
 \multicolumn{1}{c}{A} & \includegraphics[width=1.8cm]{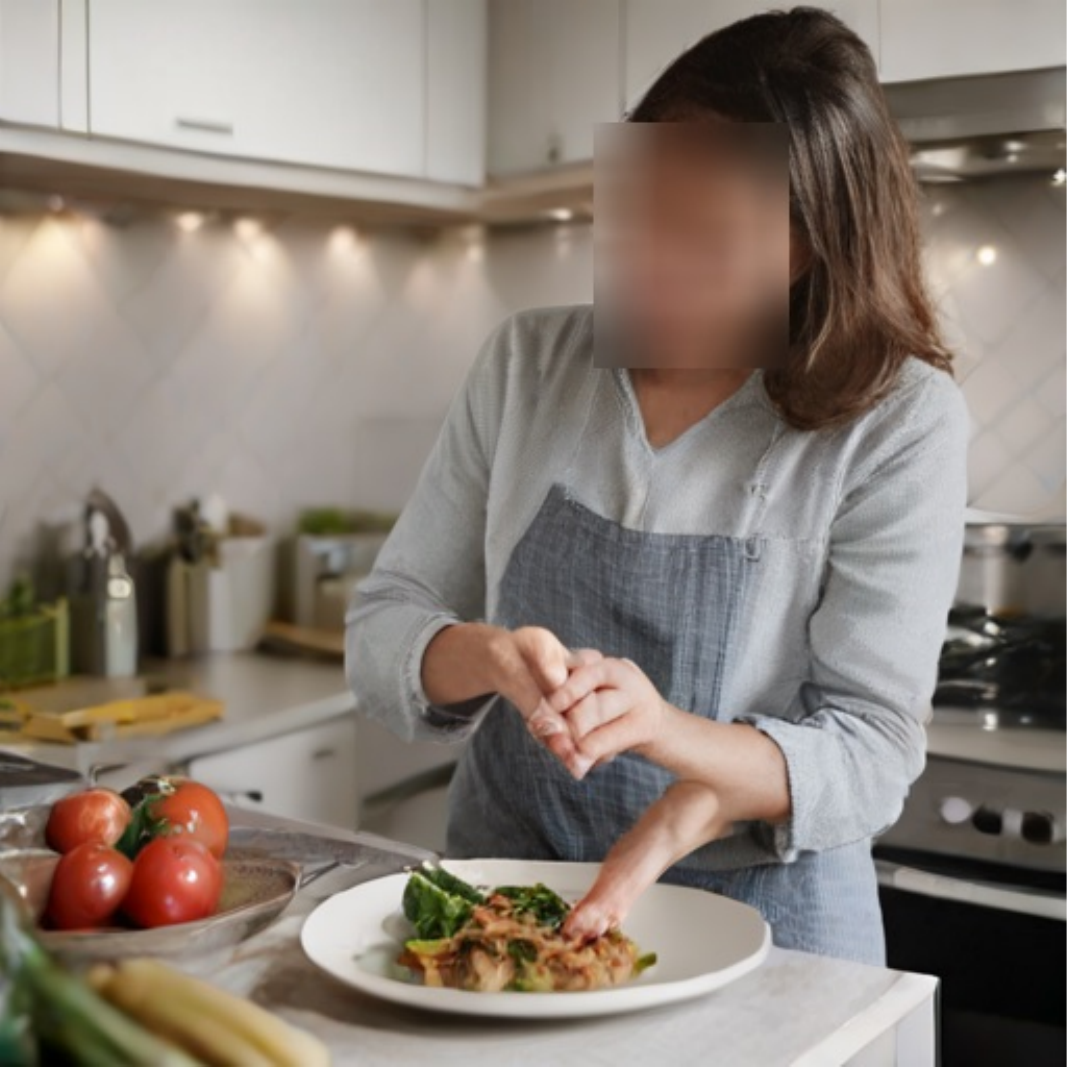} &  \includegraphics[width=1.8cm]{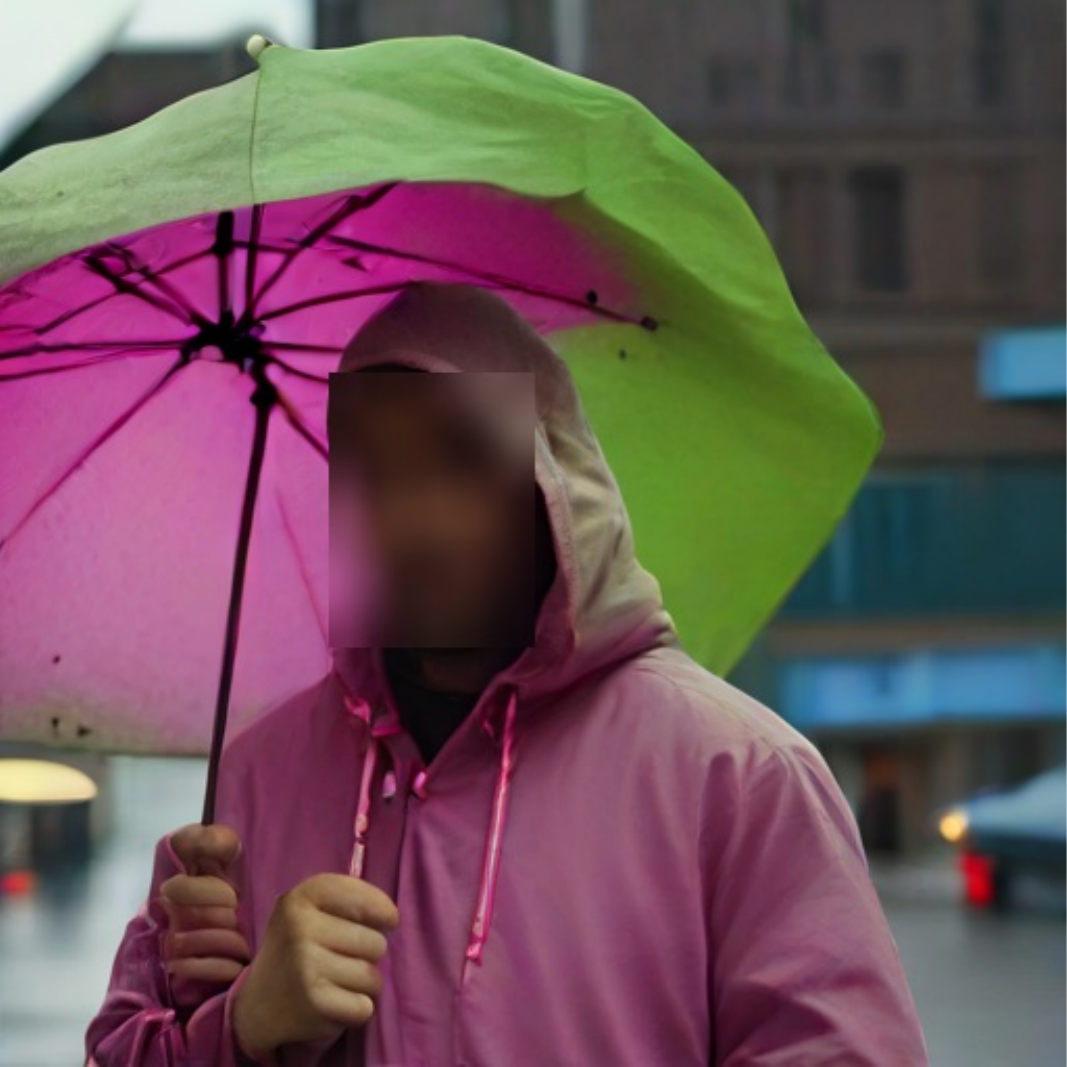} & \includegraphics[width=1.8cm]{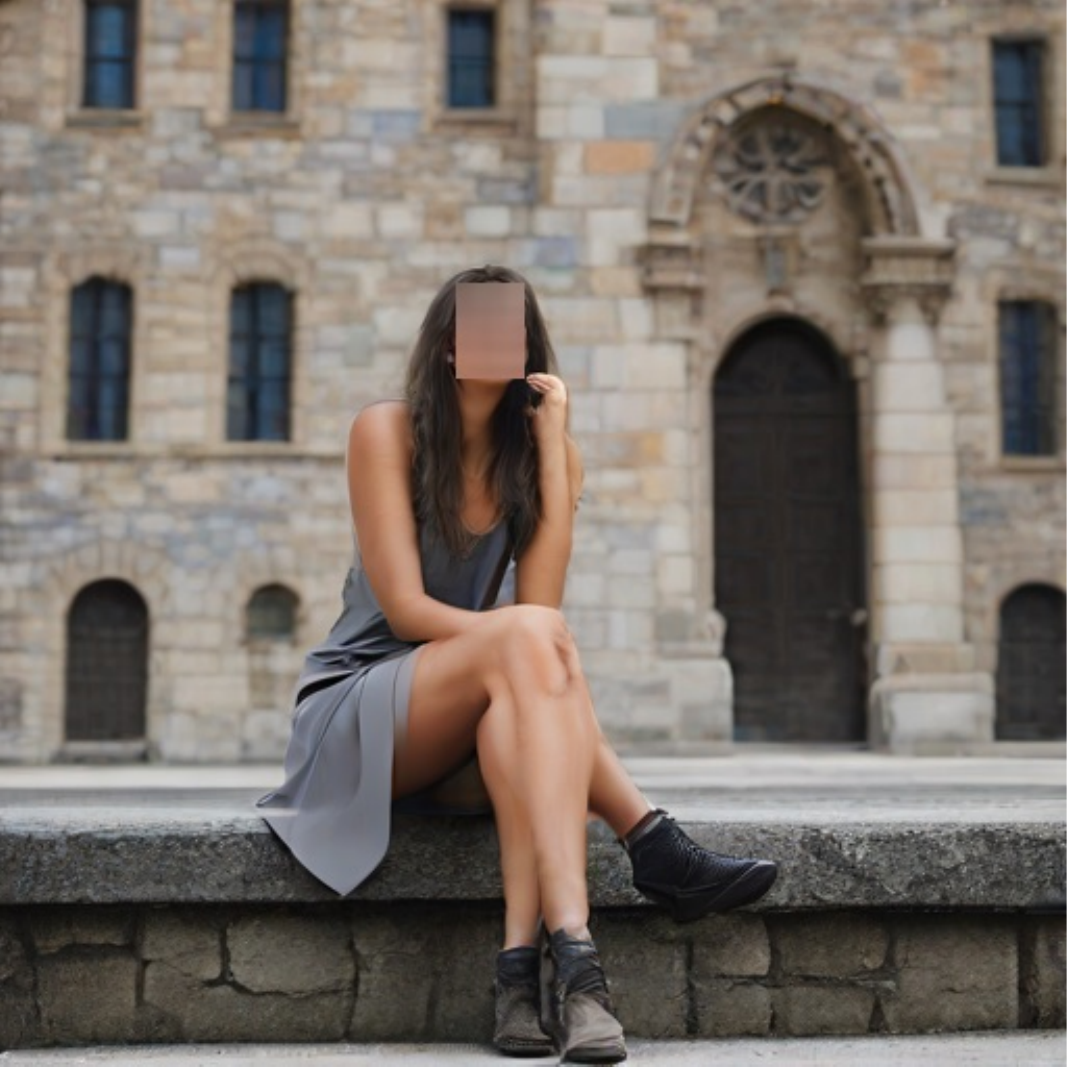} & \includegraphics[width=1.8cm]{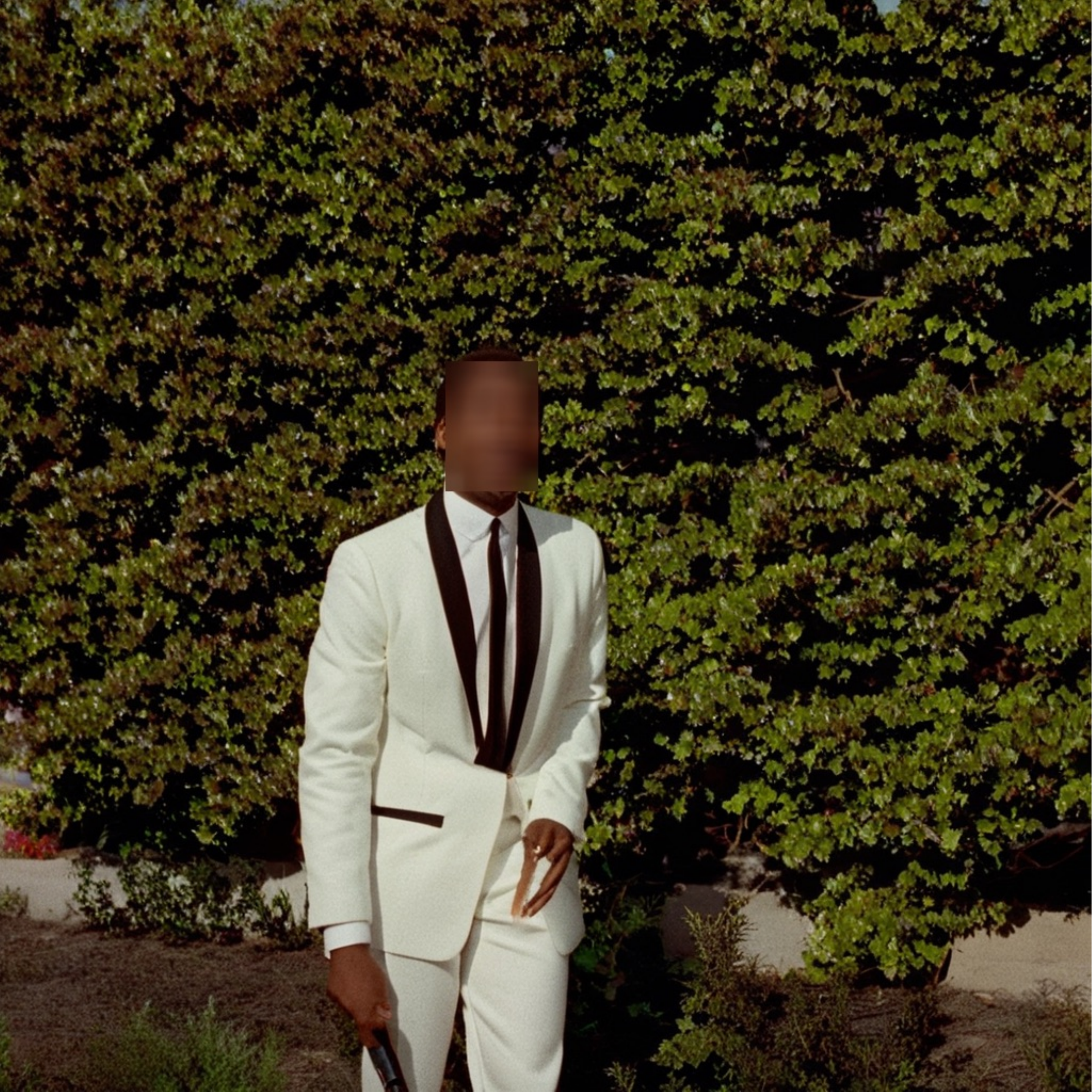} & \includegraphics[width=1.8cm]{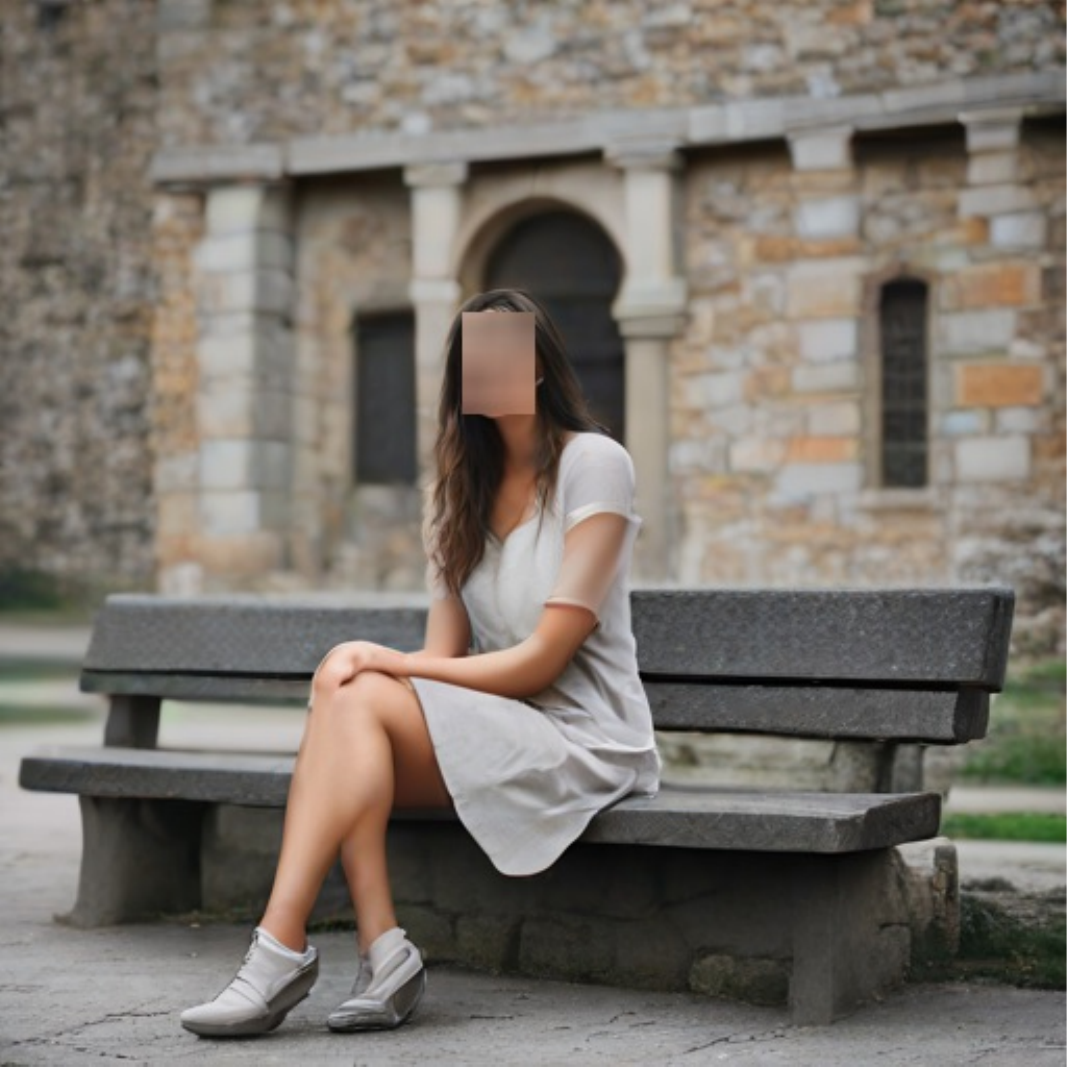} & \includegraphics[width=1.8cm]{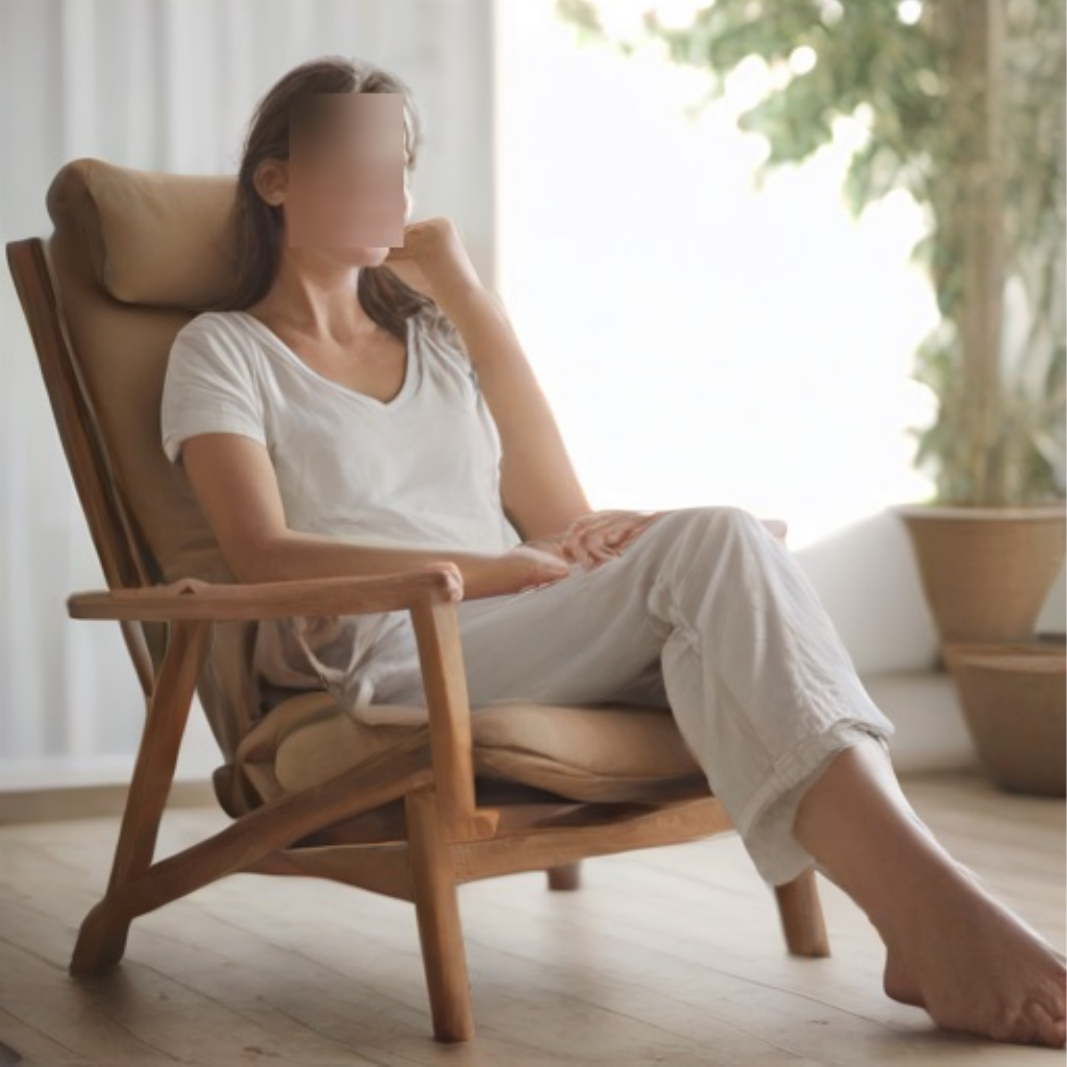}   \\
   \multicolumn{1}{c}{B} & \includegraphics[width=1.8cm]{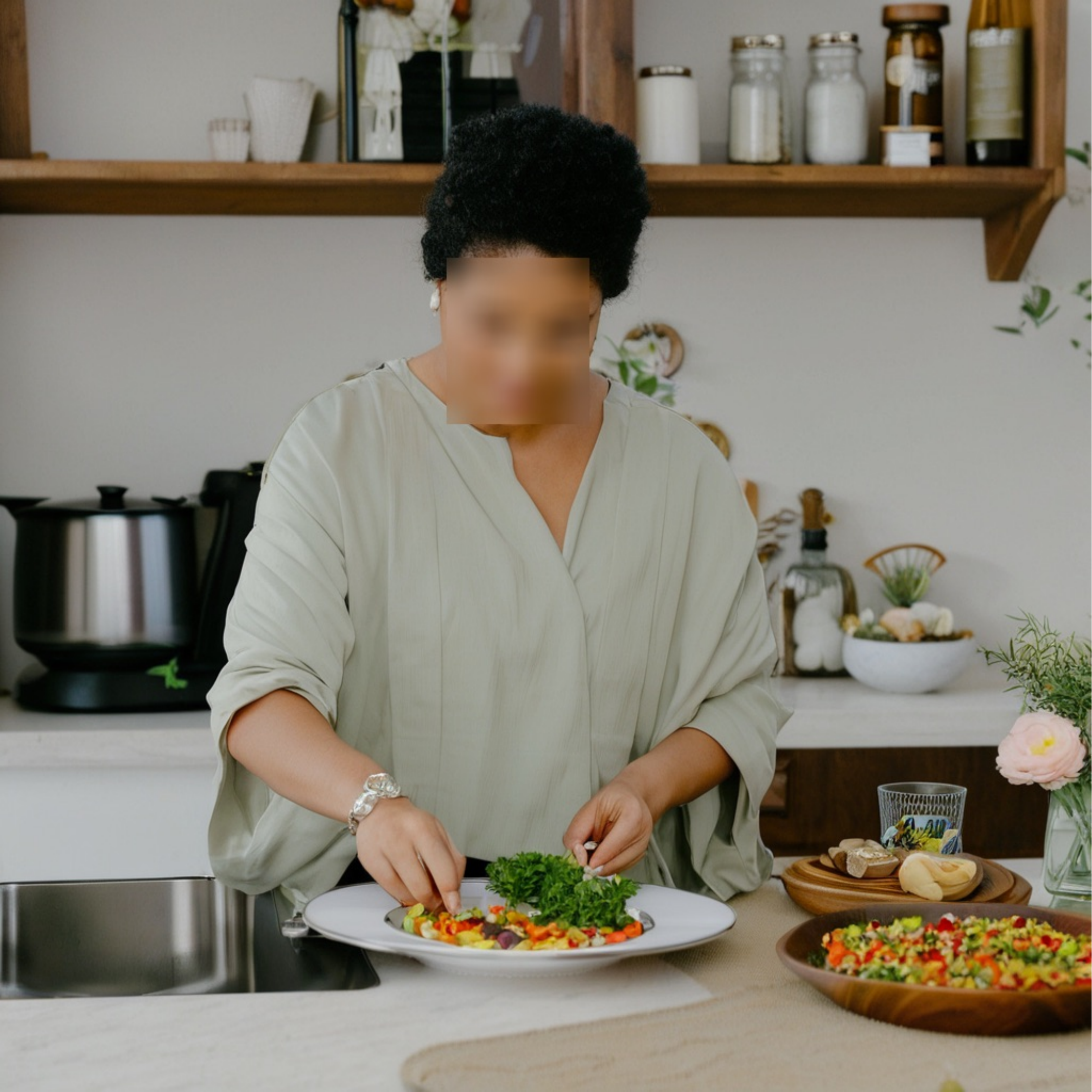}& \includegraphics[width=1.8cm]{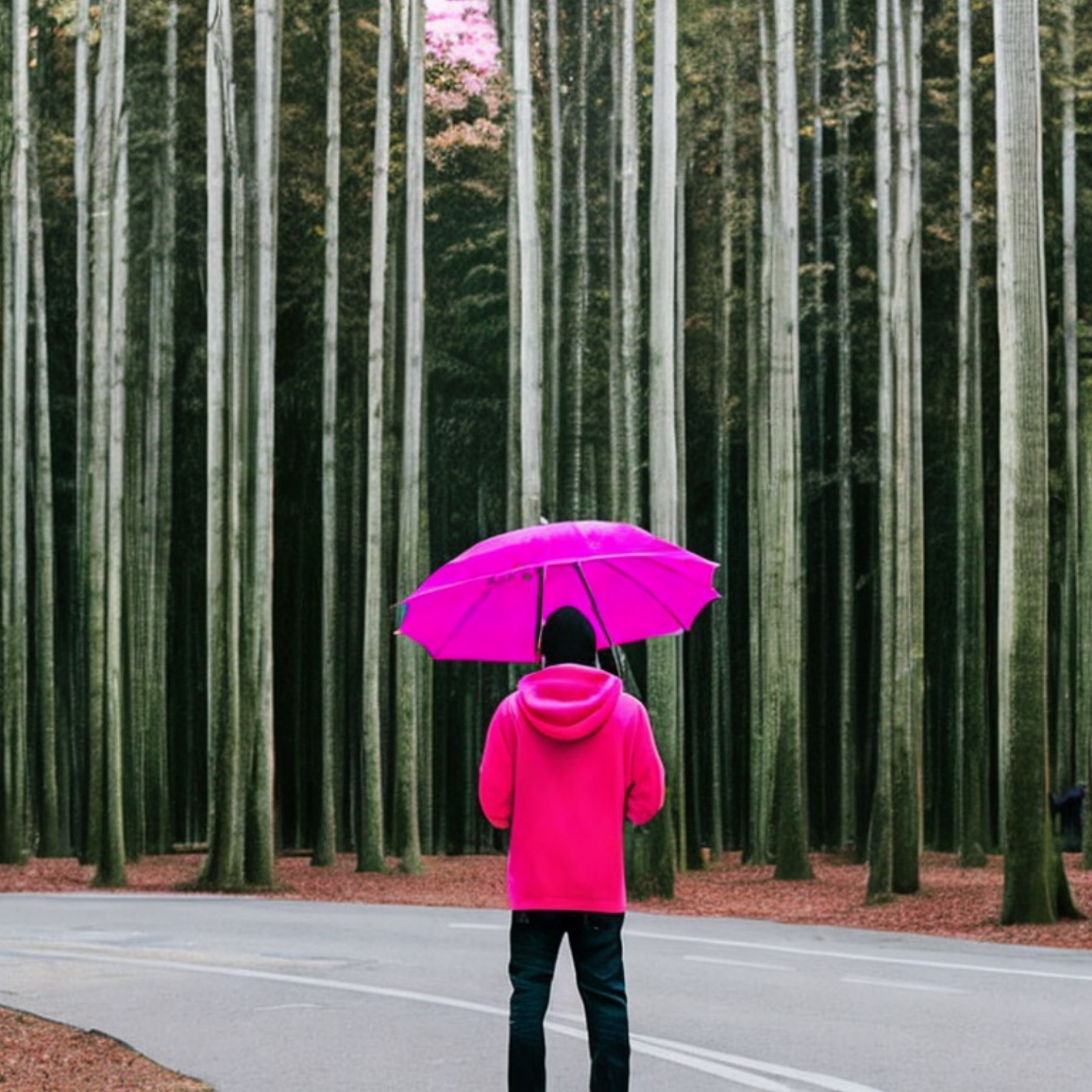} & \includegraphics[width=1.8cm]{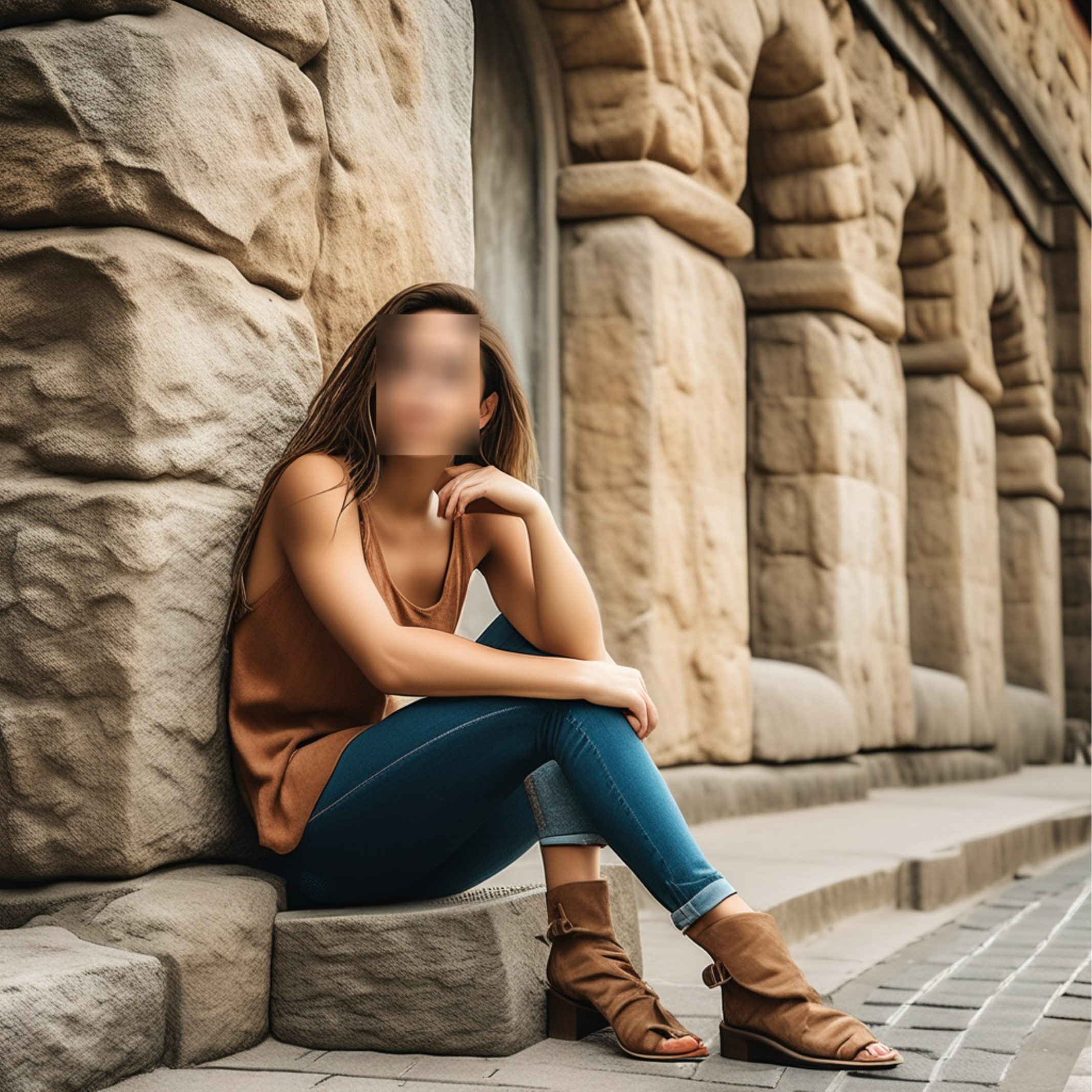} & \includegraphics[width=1.8cm]{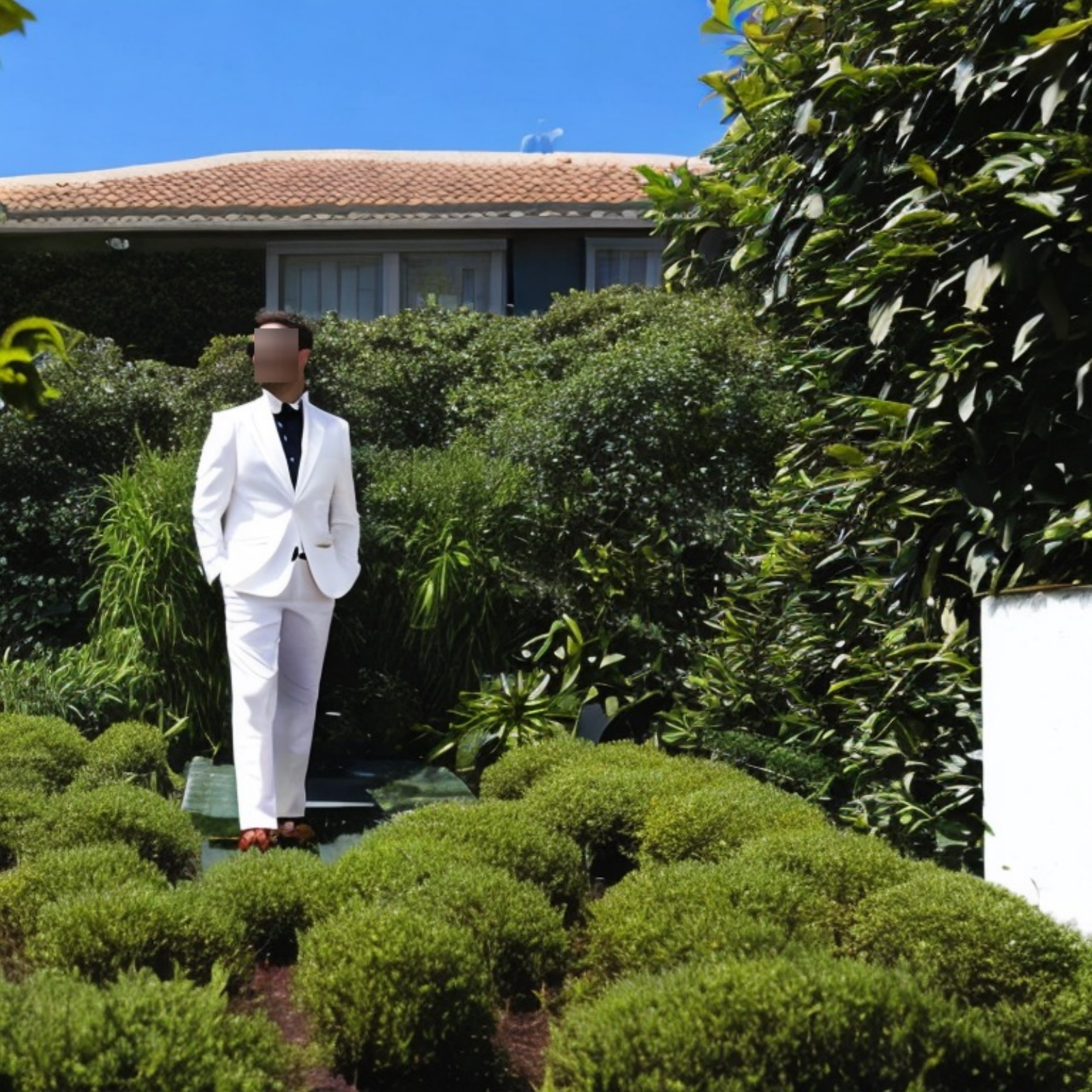} & \includegraphics[width=1.8cm]{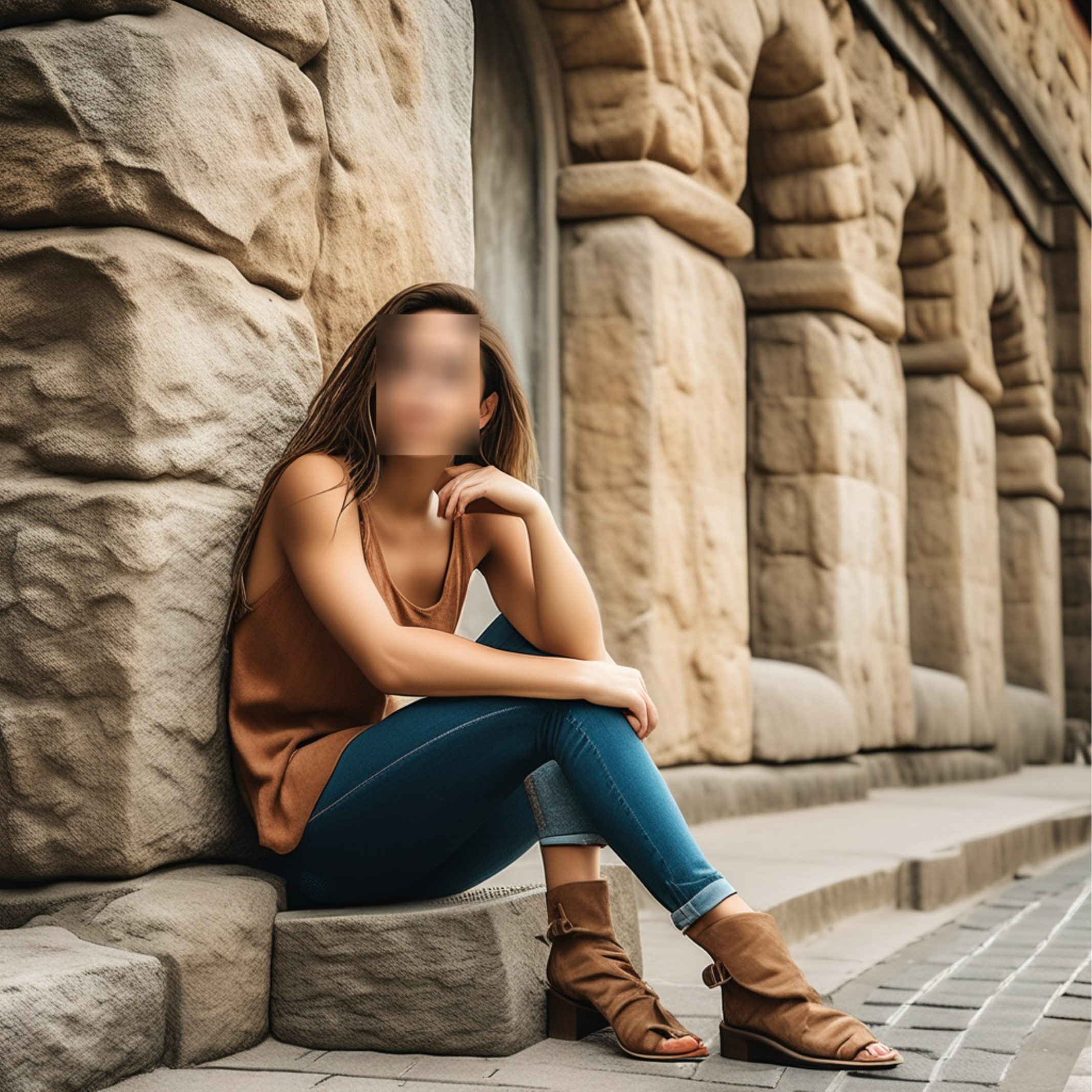} & \includegraphics[width=1.8cm]{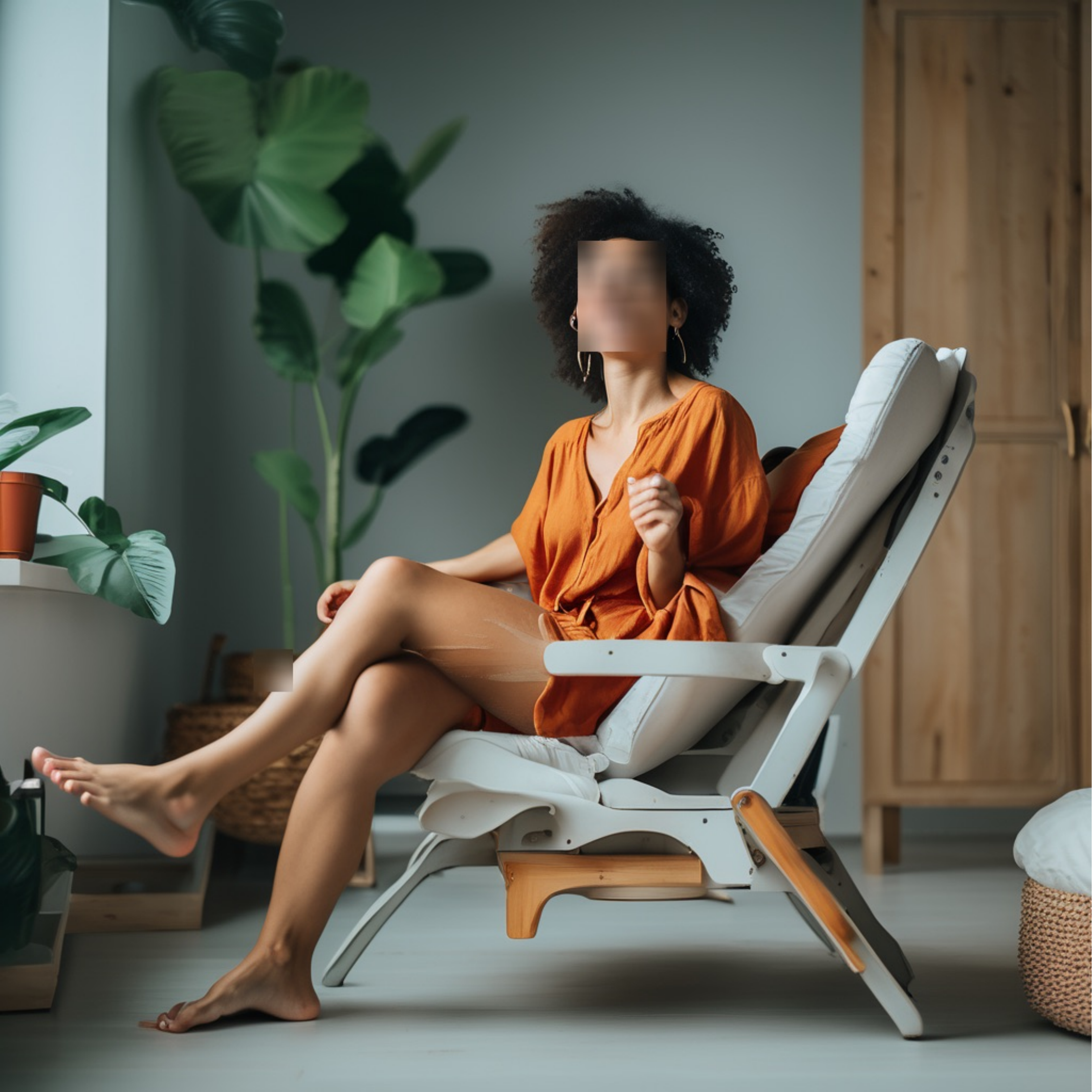}   \\
  \multicolumn{7}{c}{\textbf{PickScore~\cite{Kirstain:2023:PickScore}}}  \\
  \hline
  \hline
  A & {\textcolor{brown}{0.70}}  &  {\textcolor{brown}{0.70}}  &  {\textcolor{brown}{0.51}} &  {\textcolor{brown}{0.59}}  &  {\textcolor{brown}{0.53}} &  {\textcolor{brown}{0.57}} \\
  B &  {\textcolor{brown}{0.30}} &  {\textcolor{brown}{0.30}}   &  {\textcolor{brown}{0.49}} &  {\textcolor{brown}{0.41}}  &  {\textcolor{brown}{0.47}} &  {\textcolor{brown}{0.43}} \\ 
  \multicolumn{7}{c}{{Base}}  \\
  \hline
  \hline
   A & {\textcolor{teal}{0.04}}  & {\textcolor{teal}{0.34}}  &  {\textcolor{brown}{0.81}} &  {\textcolor{brown}{0.50}} &  {\textcolor{brown}{0.82}} &  {\textcolor{brown}{0.69}} \\
   B & {\textcolor{teal}{0.96}}  & {\textcolor{teal}{0.66}}  &  {\textcolor{brown}{0.19}} &  {\textcolor{brown}{0.50}} &  {\textcolor{brown}{0.18}} &  {\textcolor{brown}{0.31}} \\
  \multicolumn{7}{c}{{BodyMetric}}  \\
  \hline
  \hline
   A & {\textcolor{teal}{0.02}}  & {\textcolor{teal}{0.08}}  & {\textcolor{teal}{0.48}} & {\textcolor{teal}{0.34}}  & {\textcolor{teal}{0.40}} & {\textcolor{teal}{0.49}} \\
   B & {\textcolor{teal}{0.98}}  & {\textcolor{teal}{0.92}}  & {\textcolor{teal}{0.52}} & {\textcolor{teal}{0.66}}  & {\textcolor{teal}{0.60} }&{ \textcolor{teal}{0.51}} \\
\end{tabular}
\end{minipage}
\captionsetup{type=figure}
\caption{Pair-wise image preference using PickScore, Base and BodyMetric. Images in row A are annotated by human experts as less realistic than B. We display the scores per pair for each model and highlight the \textbf{\textcolor{teal}{correct}} and \textbf{\textcolor{brown}{incorrect}} predictions.}
\label{fig:base_vs_body}
\end{table}
\noindent \textbf{Can image preference metrics identify unrealistic bodies?} To address this question, we consider a comparison between BodyMetric, against concurrent works that learn user preference for text-to-image generation. Fig.~\ref{tab:photoshop_images} depicts a qualitative comparison to HPS~\cite{Wu:2023:HPSv2}, ImageReward (IR)~\cite{Xu:2023:ImageReward}, and PickScore (PS)~\cite{Kirstain:2023:PickScore}. We have selected a small and representative set of images with various degrees of body related artifacts and have asked human experts to improve such images by removing artifacts based on our notion of realistic bodies. For a fair comparison, we compute the logits for each pair and pass them through a softmax layer to obtain a probability distribution over the pair, where a higher probability corresponds to the preferred image. In the above setup, we expect edited images to receive higher realism scores across all metrics. However, we observe that baseline metrics demonstrate an inconsistent behaviour which does not align with the perceived quality of body realism. Instead, BodyMetric scores appear more robust, capturing different levels of body related artifacts. 

\begin{table}[h]
\begin{minipage}[b]{0.25\linewidth}
\centering
    \caption{Accuracy on BodyRealism test.}
     \vspace{10pt}
    \resizebox{\linewidth}{!}{
    \begin{tabular}{c|c}
        \textbf{Model} & \textbf{Accuracy}\\
        \hline
        \hline
        CLIP-H  & 0.50 \\
        PickScore  & 0.50 \\
        Base  & 0.58 \\
        {\textcolor{teal}{BodyMetric}}  & {\textcolor{teal}{0.61}} 
    \end{tabular}}
    \vspace{30pt}
    \label{tab:base_test}
  \end{minipage}\hfill
    \begin{minipage}[b]{0.35\linewidth}
    % \vspace{0pt}
        \centering
    \begin{tabular}{c}
         \includegraphics[width=\linewidth]{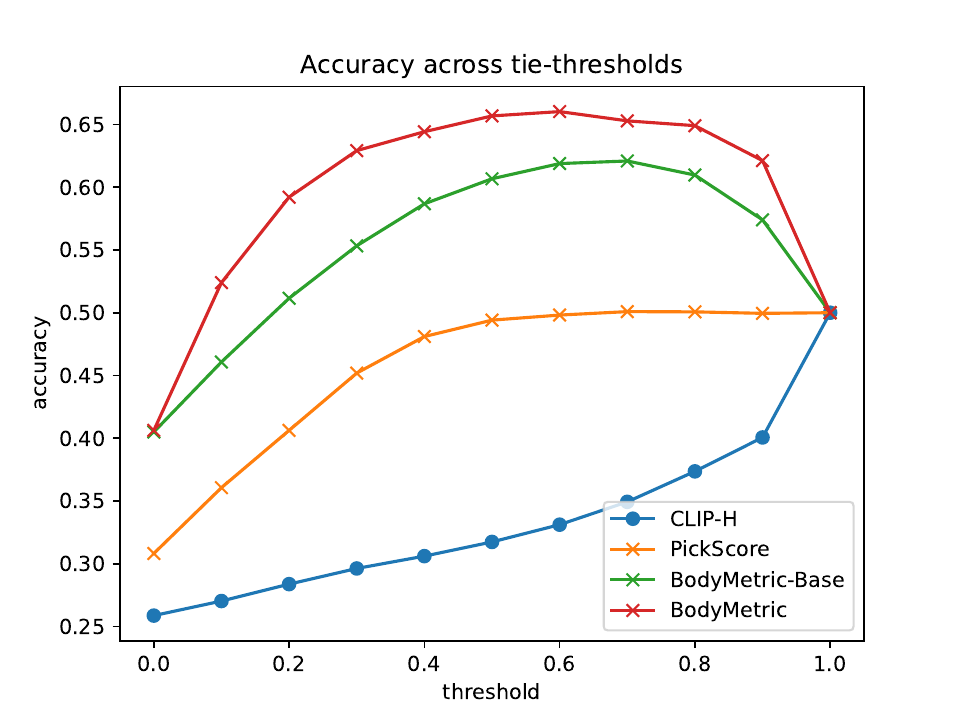}  
    \end{tabular}
    \captionsetup{type=figure,justification=centering} 
    \caption{BodyRealism validation accuracy.}
    \label{fig:val_thresholds}
  \end{minipage}\hfill
  \begin{minipage}[b]{0.25\linewidth}
    \centering
        \caption{Ablations.}
         \vspace{15pt}
        \resizebox{\linewidth}{!}{
    \begin{tabular}{c|c}
        \textbf{Abl. Objective} & \textbf{Acc.}\\
        \hline
        \hline
        {\textcolor{teal}{Base}} & {\textcolor{teal}{0.58}} \\
        Base-Reg & 0.46\\
        Base-Txt & 0.33\\
        \hline
        \hline
        \textbf{Abl. Prior} &  \\
        \hline
        \hline
        Pixel & 0.59\\
        Latent & 0.57\\
        {\textcolor{teal}{Keypoints}} & {\textcolor{teal}{0.61}}
        \end{tabular}}
           \vspace{10pt}
        \label{tab:ablation}
\end{minipage}

% ---------------------------
  \begin{minipage}[b]{0.5\linewidth}
        \begin{tabular}{*{4}{c}}
        &\textbf{ Ex. 1} &\textbf{ Ex. 2} &\textbf{ Ex. 3}  \\
         A &\includegraphics[width=1.8cm]{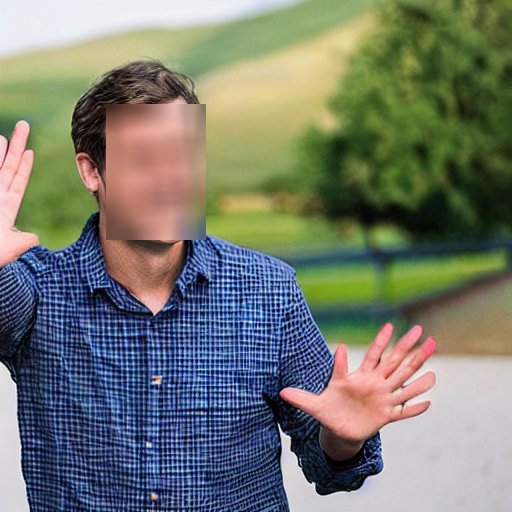} &
        \includegraphics[width=1.8cm]{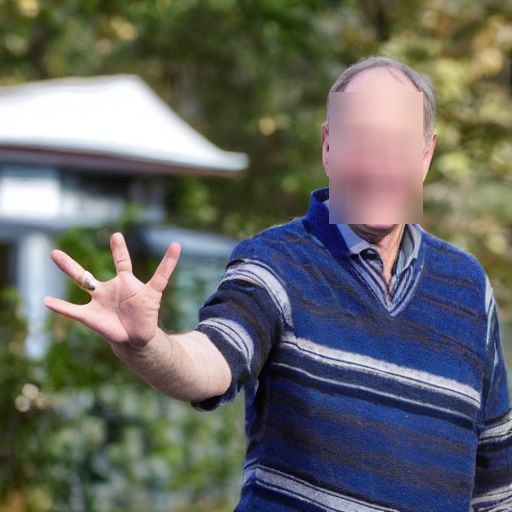} &
        \includegraphics[width=1.8cm]{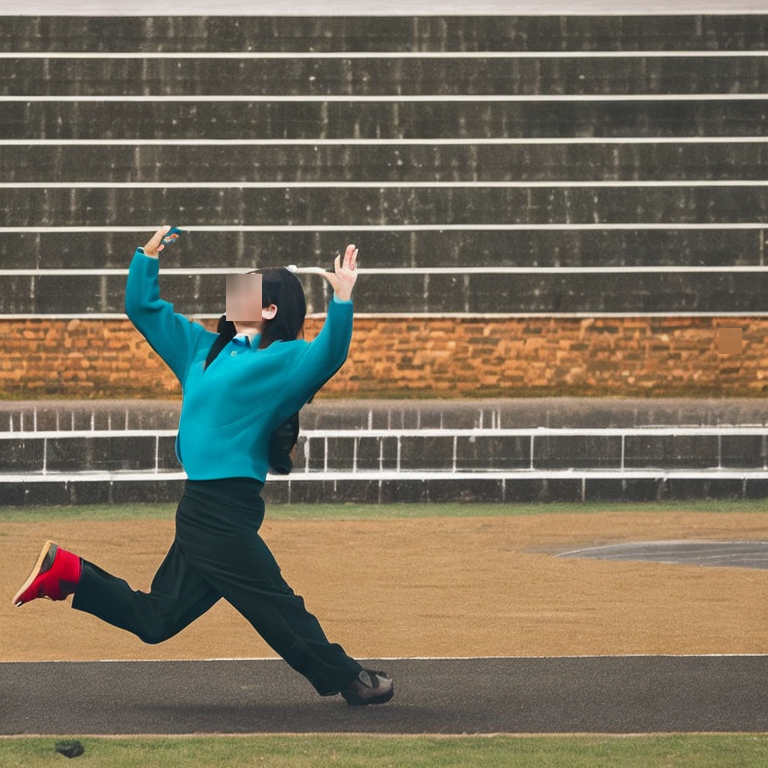} \\
        B & \includegraphics[width=1.8cm]{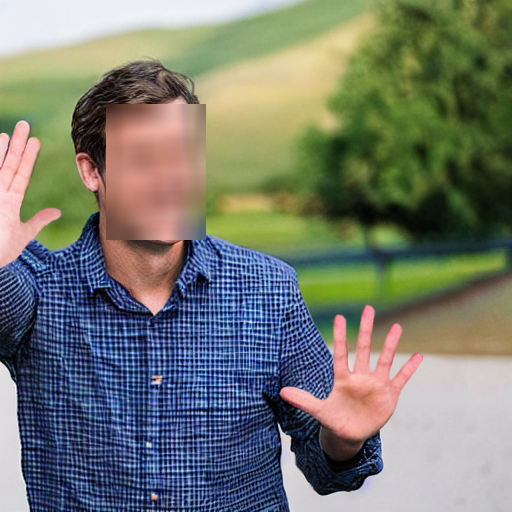} &
         \includegraphics[width=1.8cm]{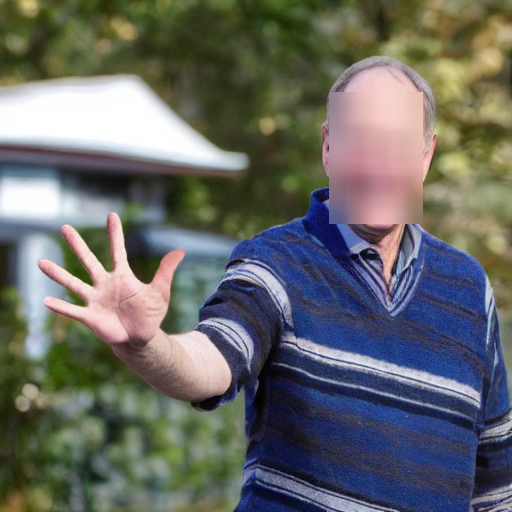} &
        \includegraphics[width=1.8cm]{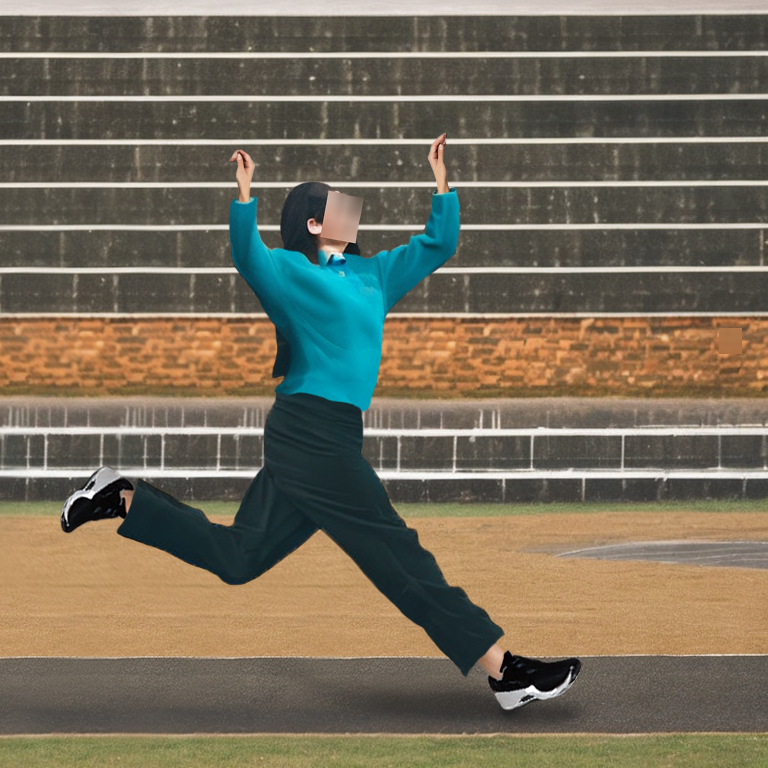} \\
    \end{tabular}
\end{minipage}\hfill
\begin{minipage}[b]{0.5\linewidth}
    \begin{tabular}{cc|*{3}{c}}
          & & \textbf{Ex. 1} & \textbf{Ex. 2} &\textbf{ Ex. 3}  \\
          \hline
          \hline
        \multirow{2}{*}{HPS~\cite{Wu:2023:HPSv2}}& A & 0.50 & 0.50  & 0.50 \\
        & B & 0.50 &0.50& 0.50 \\
        \hline
        \multirow{2}{*}{ImageReward~\cite{Xu:2023:ImageReward}} & A & 0.49 & 0.52 & 0.57 \\
         & B & {\textcolor{teal}{0.51}} & 0.48  & 0.43\\
        \hline
        \multirow{2}{*}{PickScore~\cite{Kirstain:2023:PickScore}}  & A & 0.50 & 0.47 & 0.52 \\
         & B & 0.50 & {\textcolor{teal}{0.53}} & 0.48 \\
        \hline
        % \multirow{2}{*}{\small{BaseMetric}} & A & 0.27 & 0.47 & 0.56 \\
        % & B & \textbf{0.74} &\textbf{0.53} & 0.44 \\
        \hline
        \multirow{2}{*}{\small{BodyMetric}} & A & 0.31 & 0.42 & 0.43 \\
        & B & {\textcolor{teal}{0.70}} &{\textcolor{teal}{0.58}} & {\textcolor{teal}{0.57}} \\
    \end{tabular}
      \end{minipage}
    \captionsetup{type=figure} 
    \caption{Performance of baselines on pair-wise comparison between original images with body artifacts (A) and identical images with corrected artifacts (B).}
    \label{tab:photoshop_images}
\end{table}

\subsection{Ablations}
We ablate on the objective function used to train \modelname as well as our choice of representation for the body prior.
\begin{table}[h]
\begin{minipage}[b]{1\linewidth}
\centering
\begin{tabular}{ccccc}
     & \multicolumn{1}{p{2.65cm}}{\scriptsize{a dancer rehearsing in an empty studio.}} &     \multicolumn{1}{p{2.65cm}}{\scriptsize{a woman that is standing on a snowboard in the snow.}} & \multicolumn{1}{p{2.65cm}}{\scriptsize{an Indian dancer performing hand gestures.}} &  \multicolumn{1}{p{2.65cm}}{\scriptsize{a man sitting down on a bench beside a water bottle.}} \\ 
    \multirow{2}{*}{A}& \includegraphics[width=2.65cm, cfbox=teal 1pt 1pt]{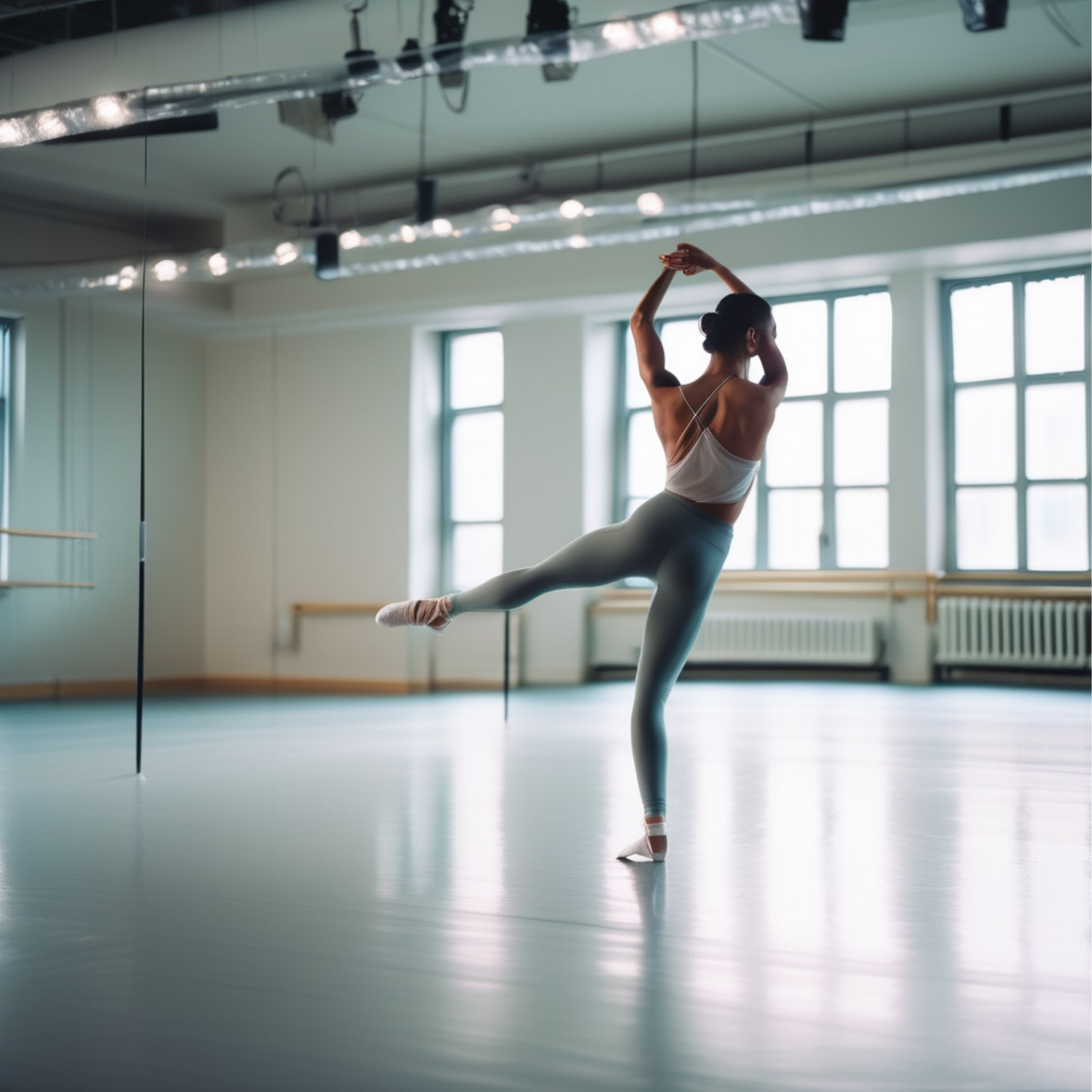} &  \includegraphics[width=2.65cm]{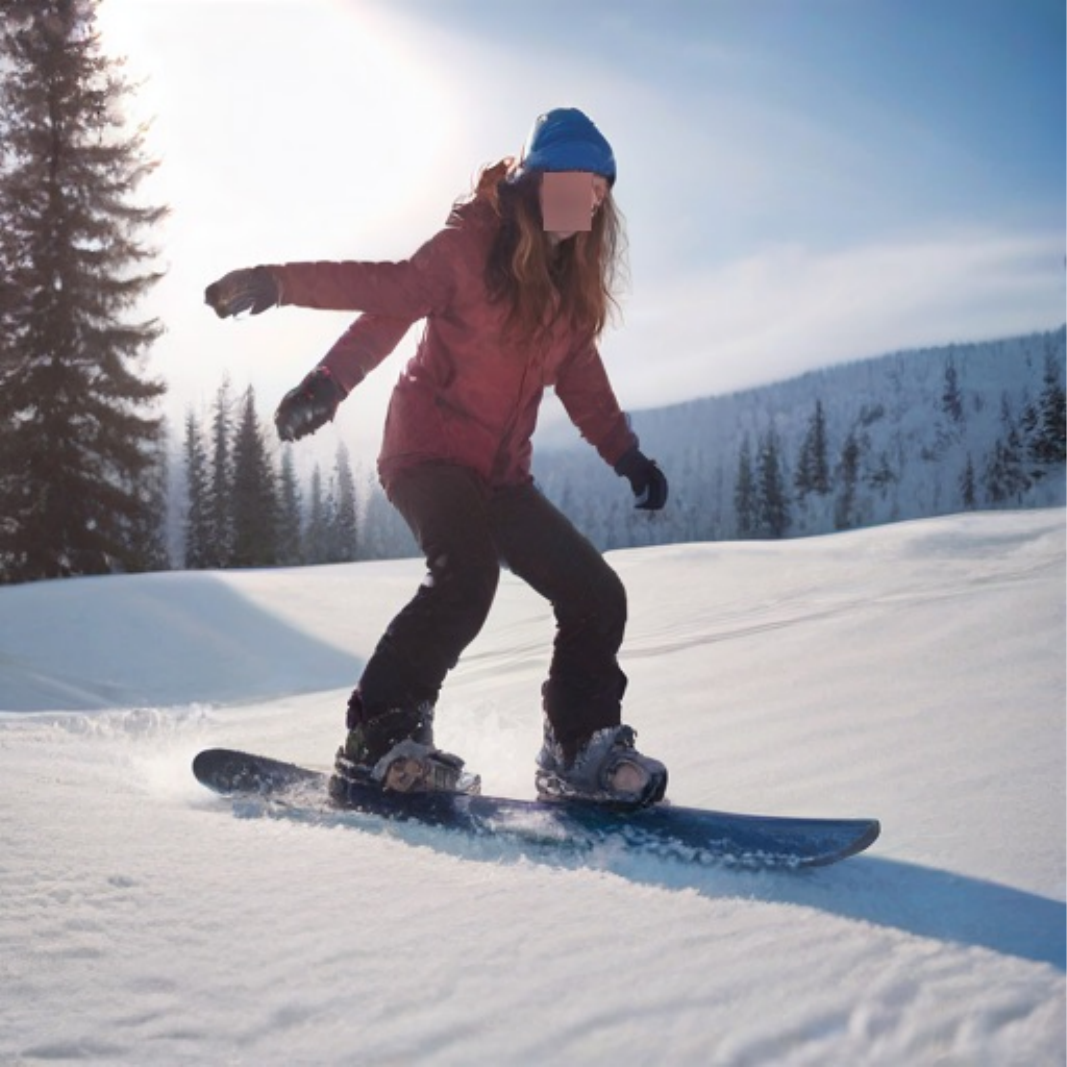}  & \includegraphics[width=2.65cm]{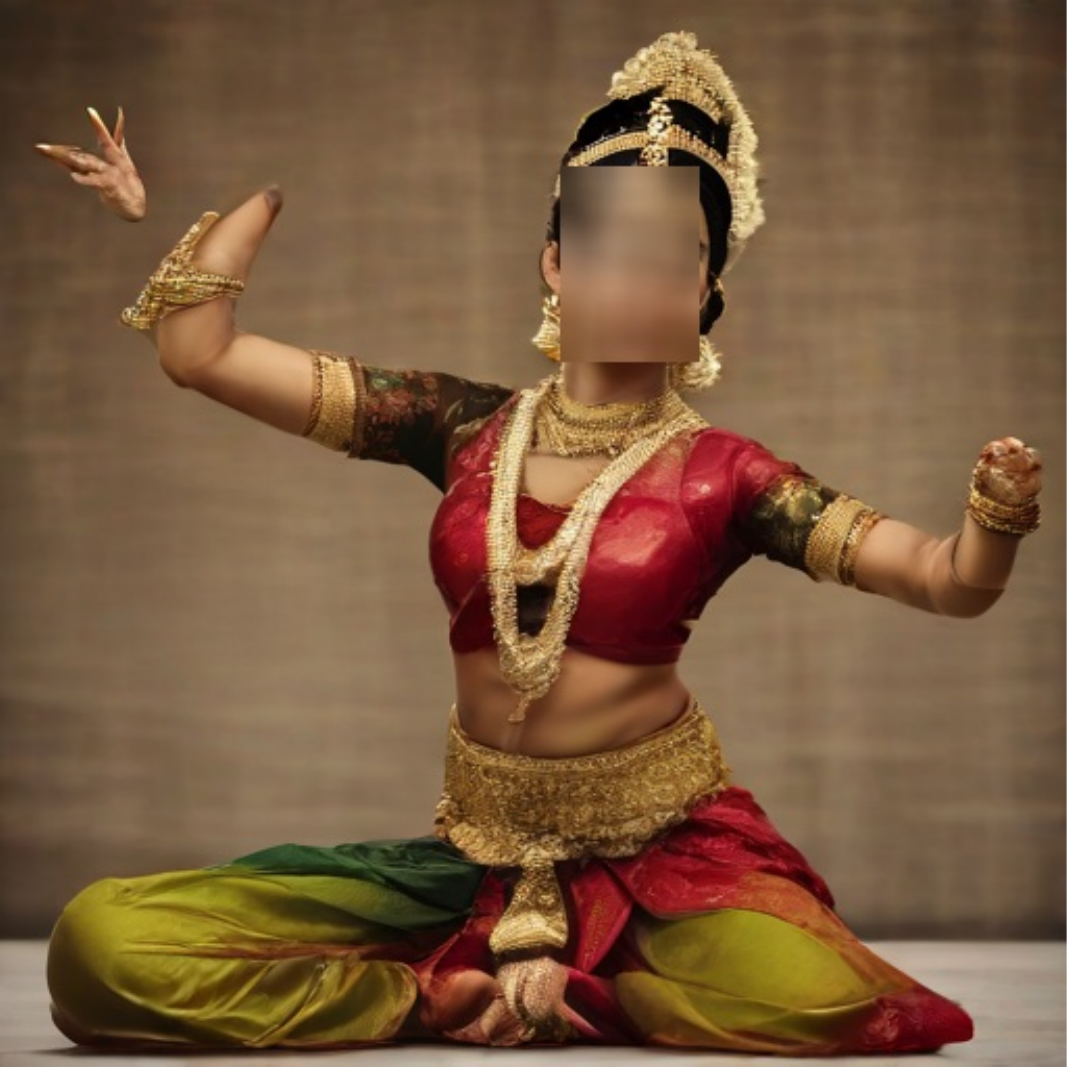} & \includegraphics[width=2.65cm]{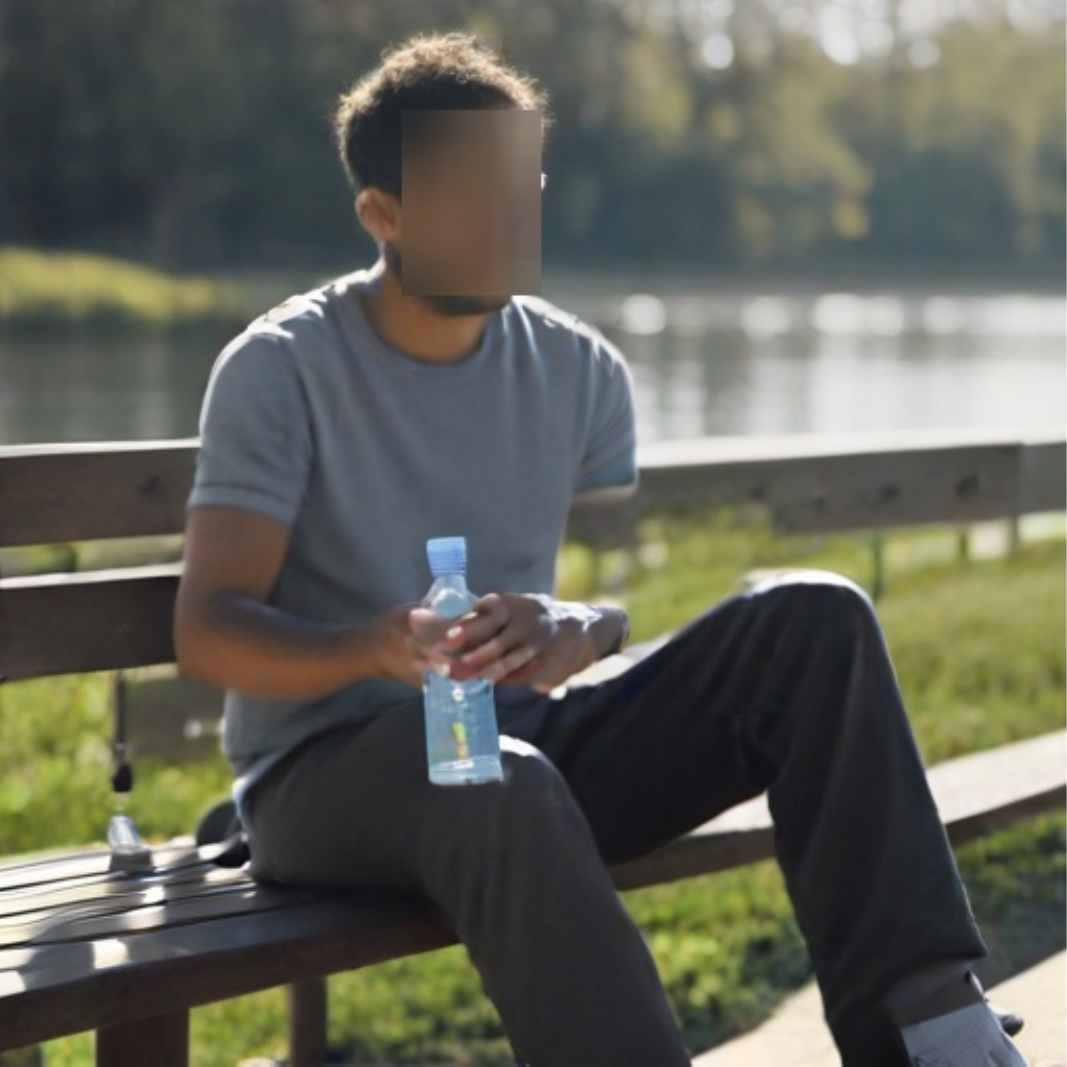} \\
    &\includegraphics[width=2.65cm, cfbox=teal 1pt 1pt]{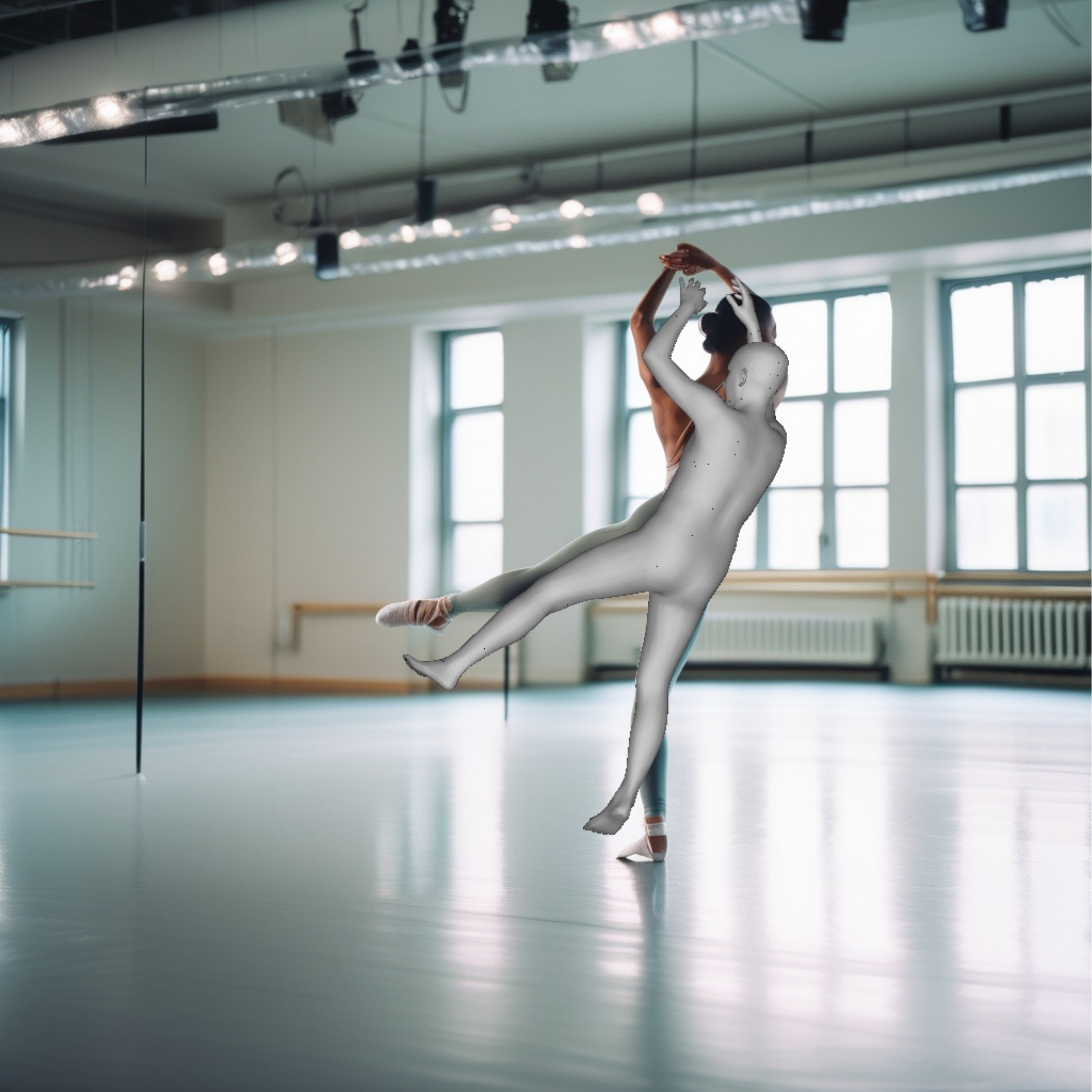} &  \includegraphics[width=2.65cm]{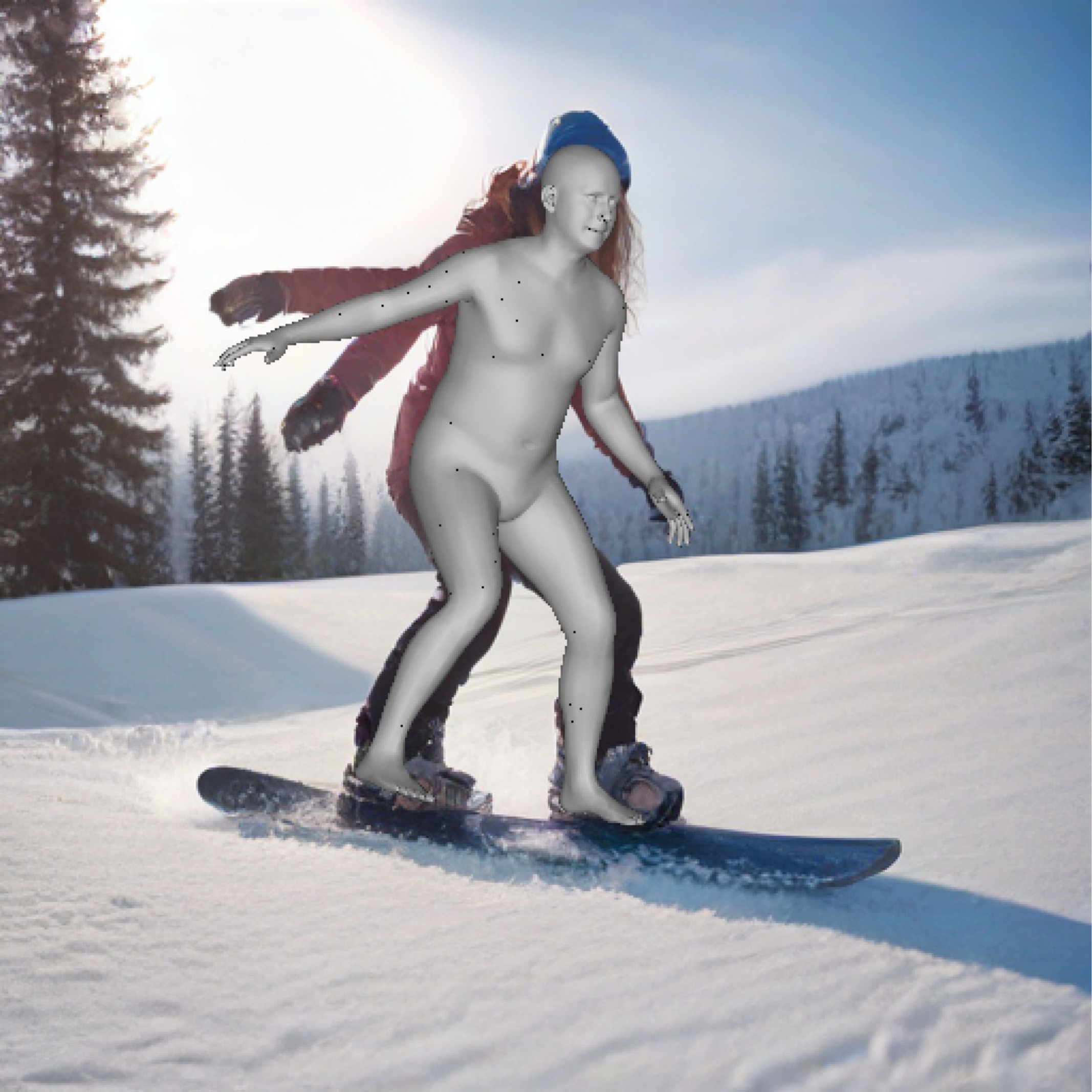}  & \includegraphics[width=2.65cm]{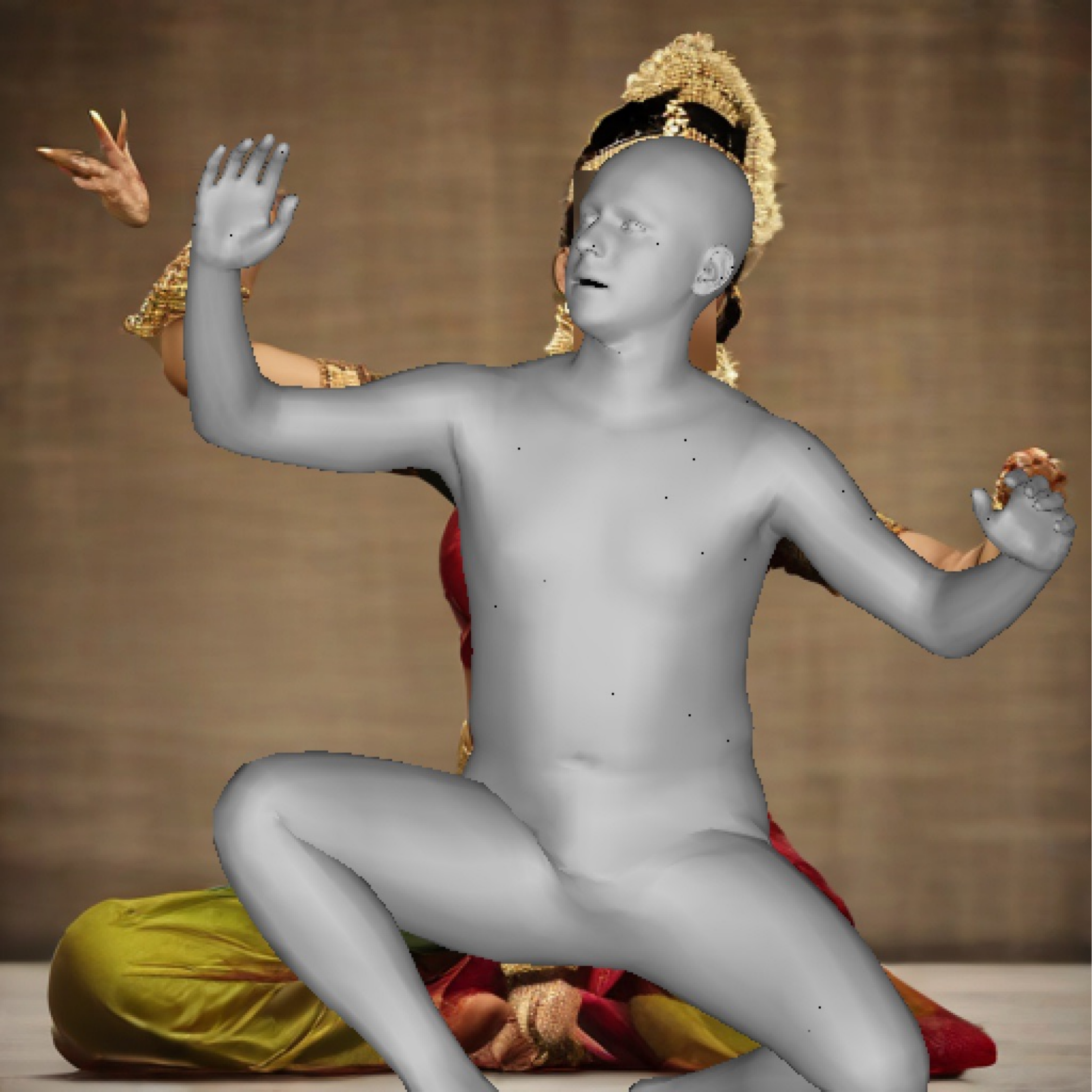} & \includegraphics[width=2.65cm]{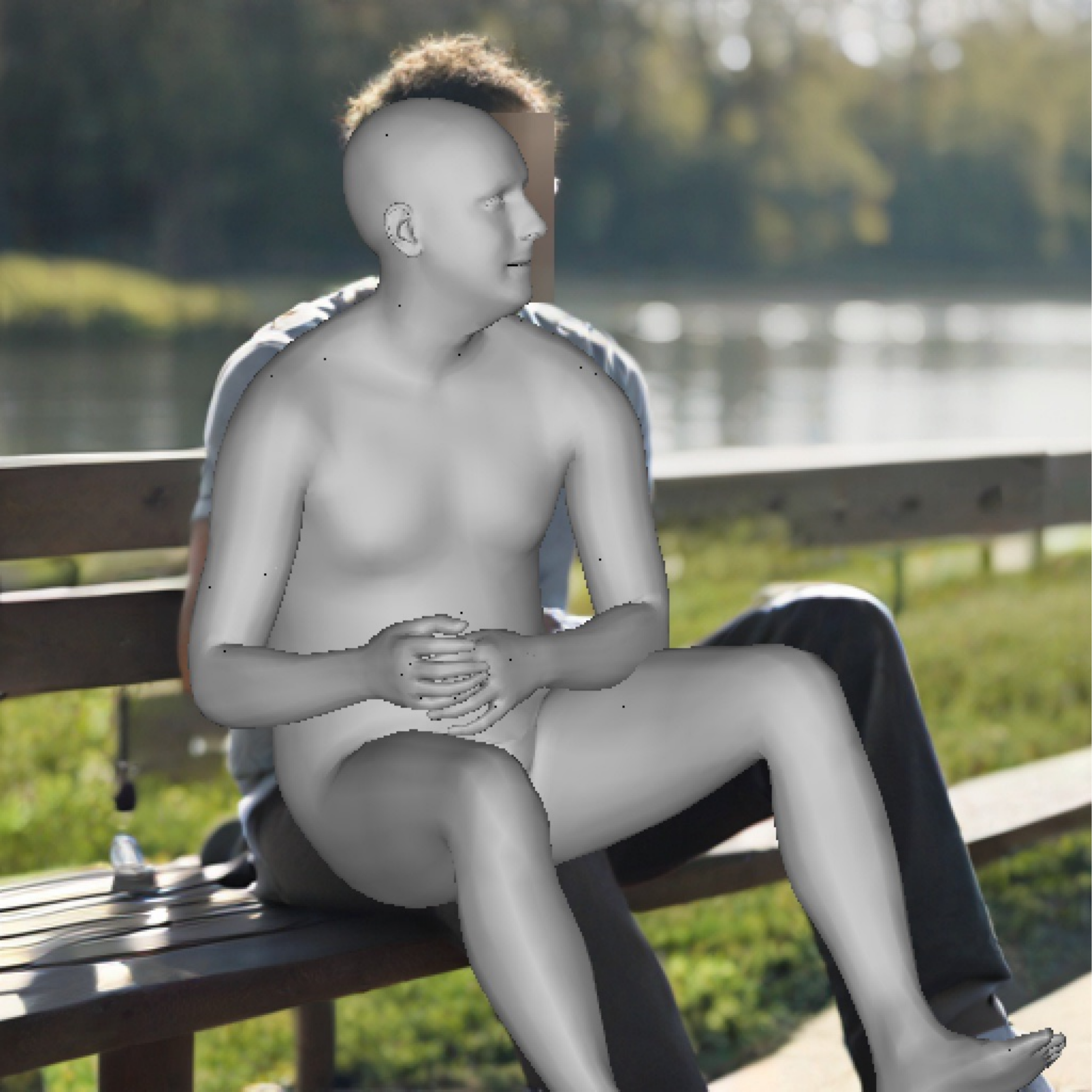} \\
    \multirow{2}{*}{B} & \includegraphics[width=2.65cm]{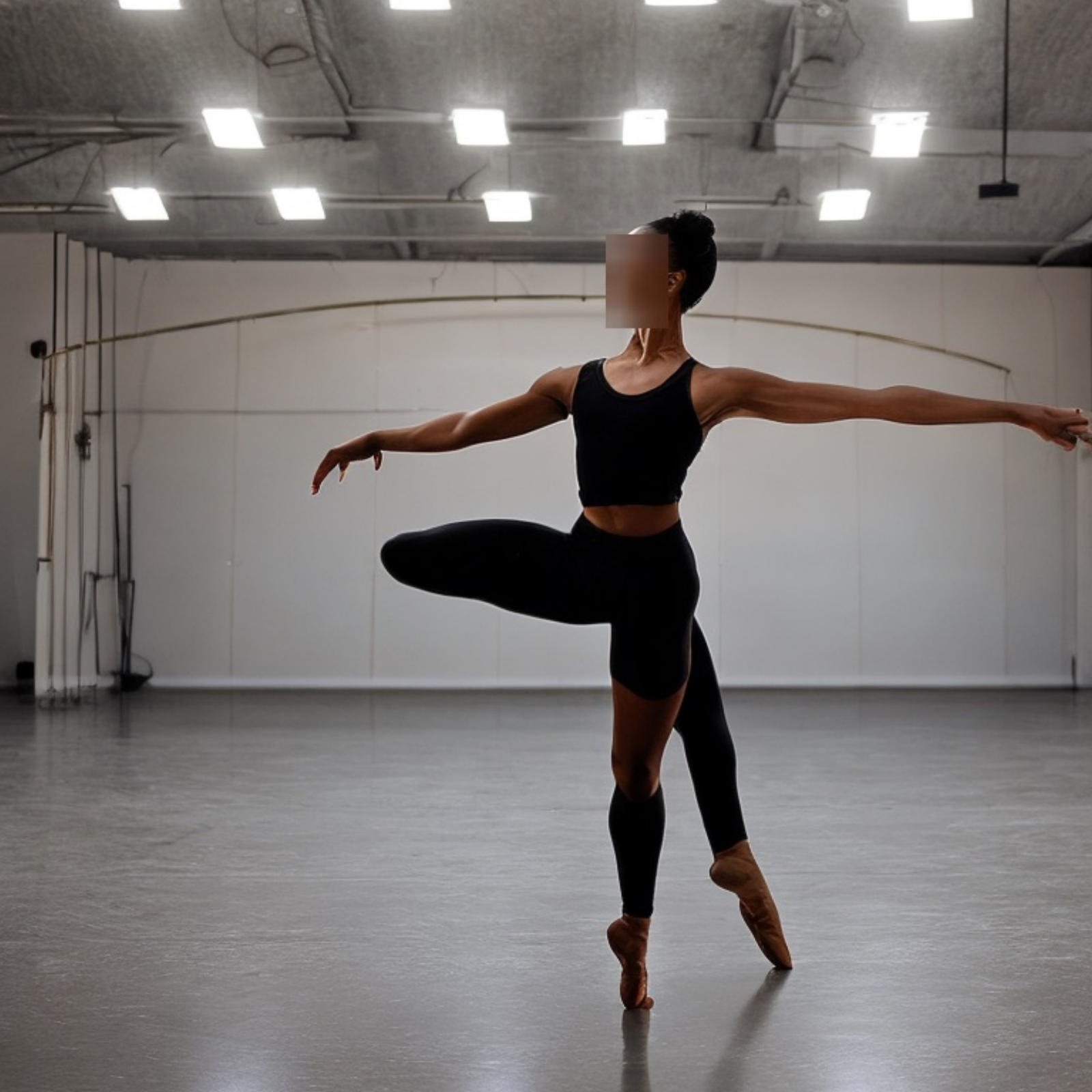}&  \includegraphics[width=2.65cm,cfbox=teal 1pt 1pt]{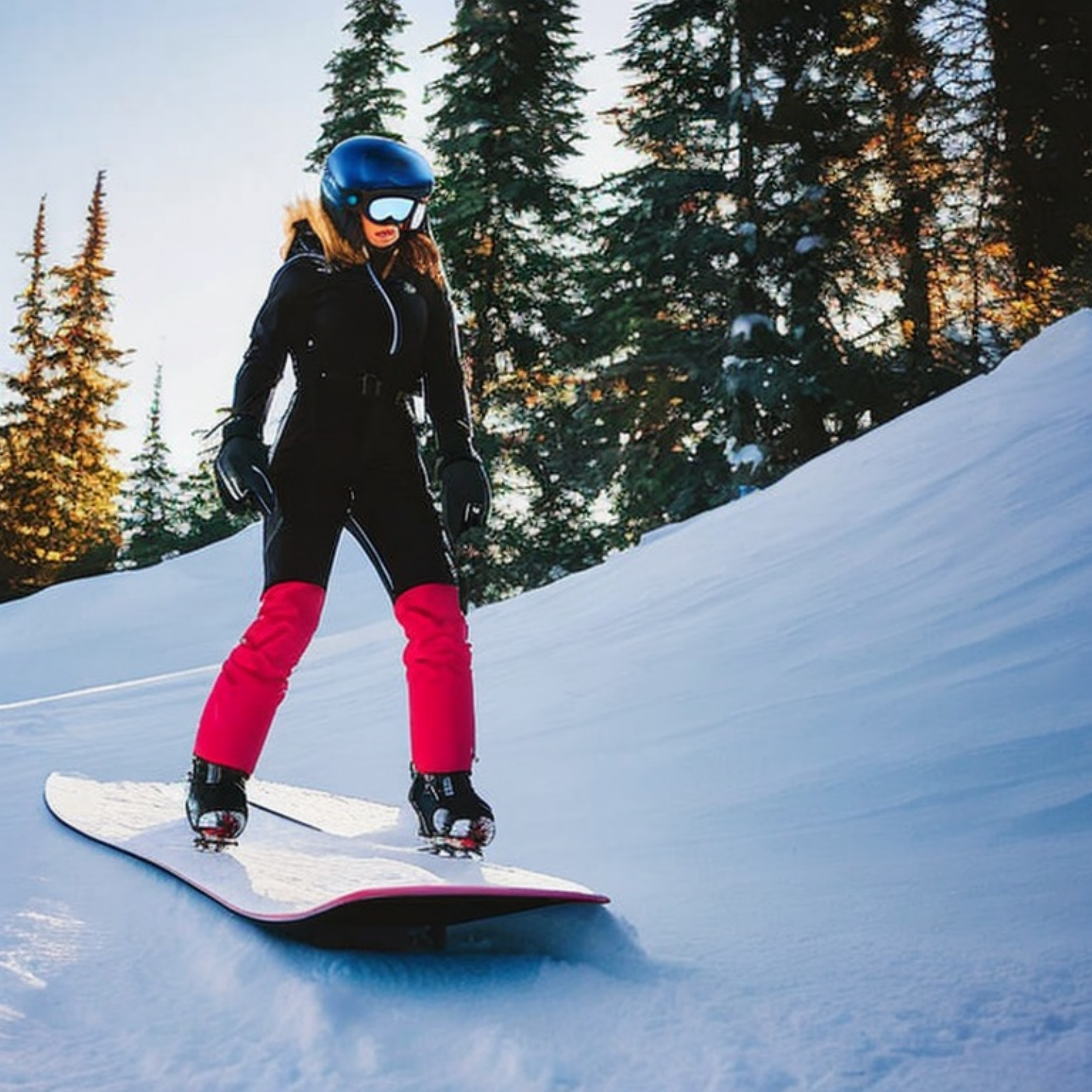}  & \includegraphics[width=2.65cm,cfbox=teal 1pt 1pt]{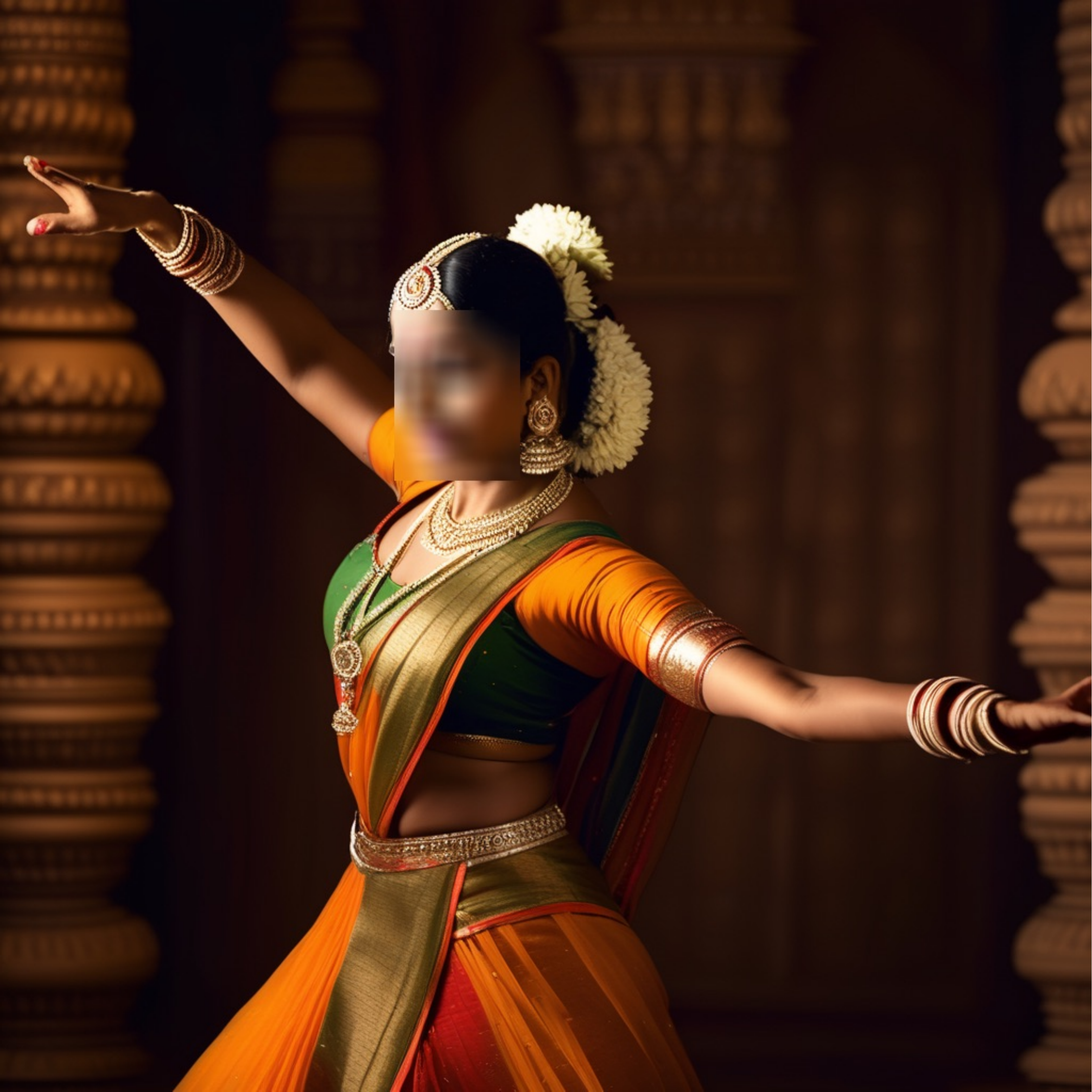} & \includegraphics[width=2.65cm,cfbox=teal 1pt 1pt]{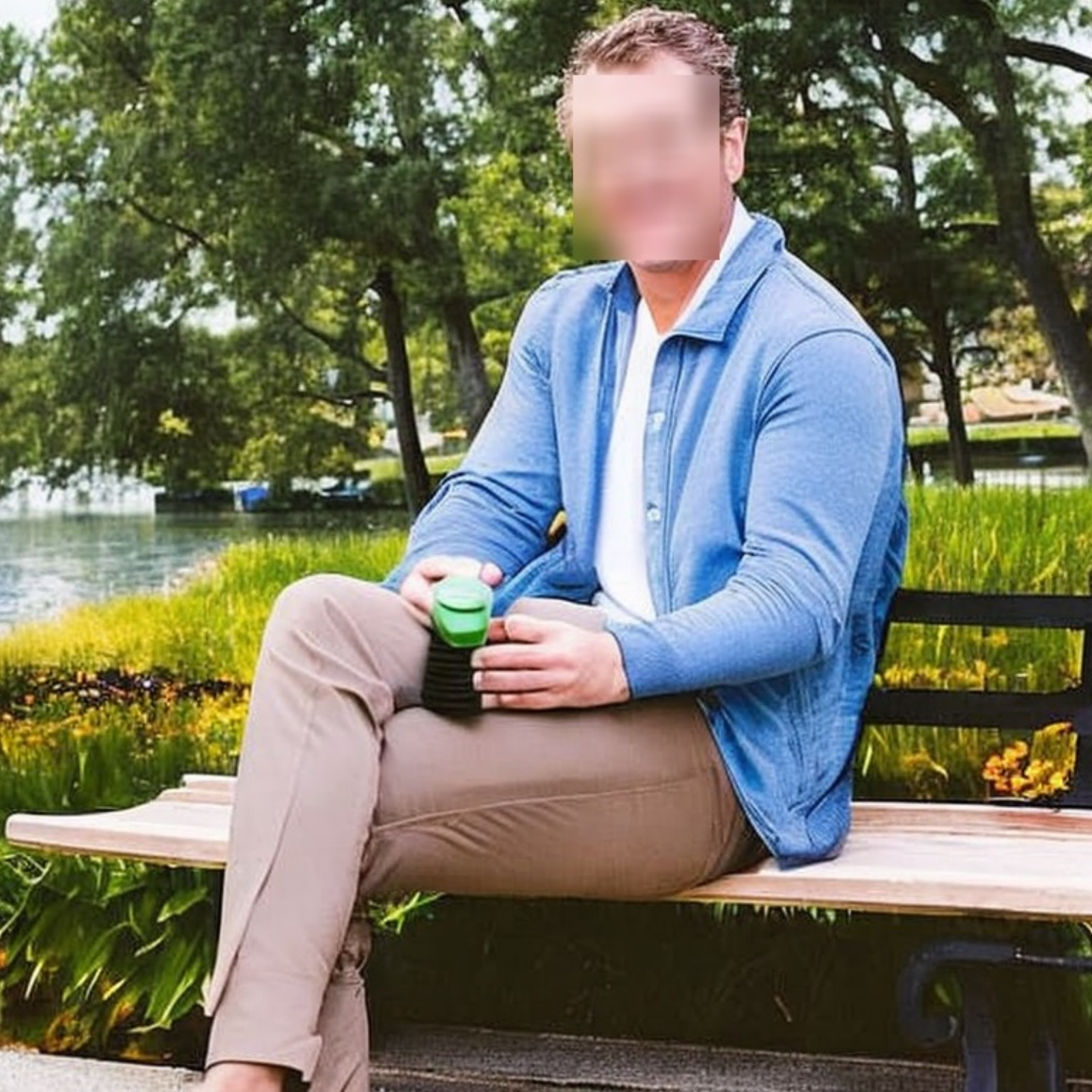} \\
    & \includegraphics[width=2.65cm]{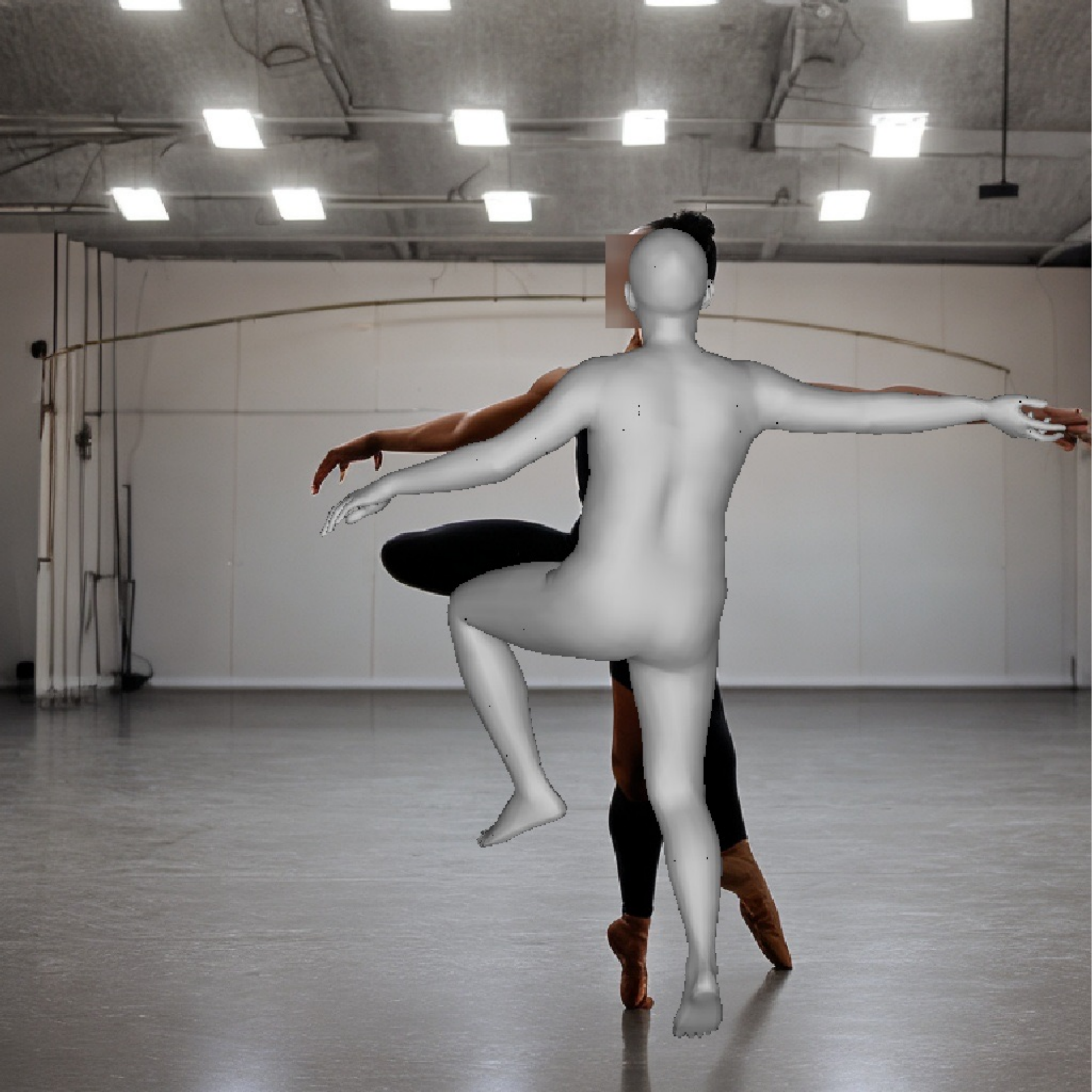}&  \includegraphics[width=2.65cm,cfbox=teal 1pt 1pt]{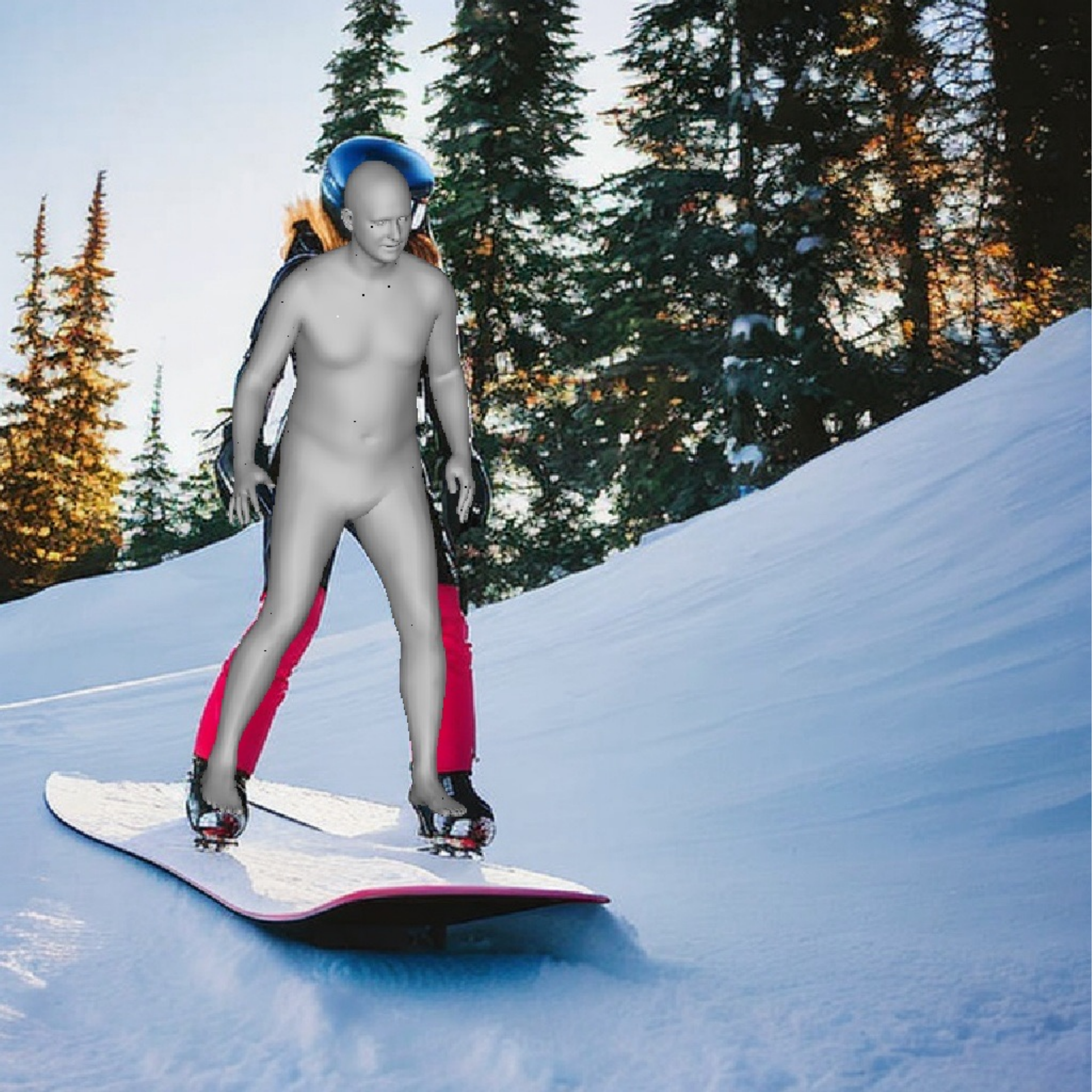}  & \includegraphics[width=2.65cm,cfbox=teal 1pt 1pt]{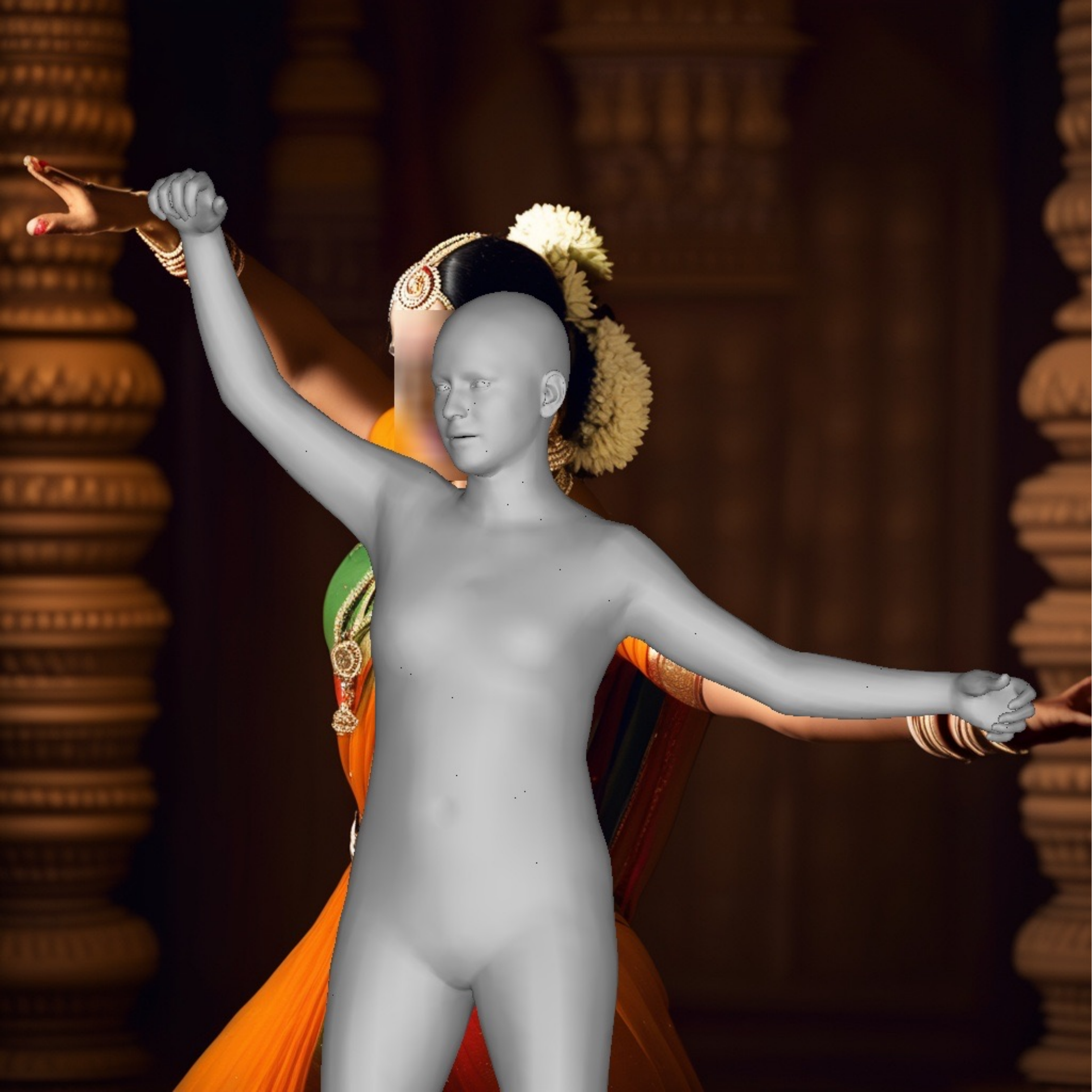} &  \includegraphics[width=2.65cm,cfbox=teal 1pt 1pt]{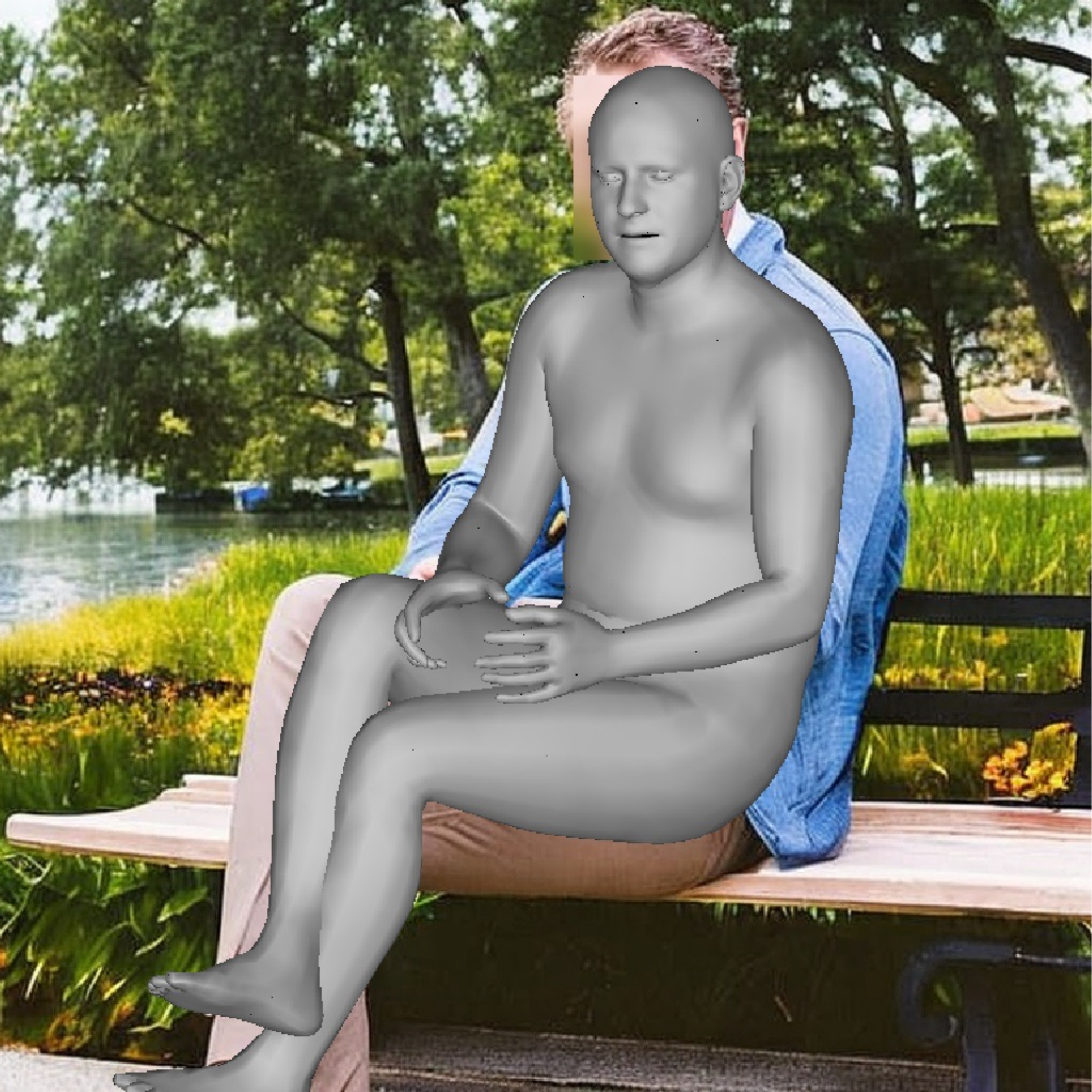} \\
        % A &  0.8475 & 0.0609 & 0.0207 \\
        %  B & 0.1525 & 0.9391 & 0.9793 \\
    \end{tabular}
    \captionsetup{type=figure} 
      \caption{BodyMetric qualitative results. Between image pairs A and B we mark with green border the images preferred by \modelname.}
      \label{fig:qual}
\end{minipage}
\end{table}
\

\noindent \textbf{Training Objective} For the sake of focusing solely on the effect of objective functions, we utilize Base as the starting point. Leveraging the multi-modal information of BodyRealism, we introduce Base-\textit{Txt} which reformulates the objective function around the text modality. In particular, given an image $y$ we formulate two text prompts $x_1, x_2$ reflecting the degree of body realism. For example, if $x$=``a person dancing'', $x_1$=``a person dancing, realistic body'' and $x_2$=``a person dancing, unrealistic body''. The preference distribution is defined based on the  aggregated annotation scores; $p=[1,0]$ if the realism score is less than 3, $p=[0,1]$ if the score is higher than 7 and $p=[0.5, 0.5]$ otherwise. Additionally, we experiment with a regression objective. Given a pair of image $y$ and realism score $r$ we train Base-\textit{Reg} using $L_{reg} = (\mathop{F}(\mathop{E}_{img}({y})) - r)^2,$ where $\mathop{F}$ is a simple MLP that predicts a single score given the image features. We measure the effectiveness of different objectives based on the accuracy. For a fair comparison, we transform the regression results to discrete buckets, similarly to the definition of preference distribution in Base-\textit{Txt}. The results reported in Table ~\ref{tab:ablation} highlight the model trained using the preference objective (Base) as the most effective.

\vspace{6pt}
\noindent \textbf{Body Representation} We consider different ways of representing and injecting the prior. For the representations, we consider: (a) the original image with an overlay of the reconstructed mesh, (b) the 3D body keypoints  (BodyMetric-\textit{Keypoints}). For (a) we use $\mathop{E}_{img}$ to obtain the latent embedding of body features. We then consider two possibilities: 1. merging the two embeddings using an MLP and using \ref{eq:intructGPT} as the final scoring function (BodyMetric-\textit{Pixel}) or, 2. using a latent embedding cosine similarity between $\mathop{E}_{img}(y), \mathop{E}_{body}(b)$ during training to bring the body and image embeddings closer in the latent space (BodyMetric-\textit{Latent}). For the latter, the scoring function reduces to  $s(x,y)=\mathop{E}_{txt}(x).\mathop{E}_{img}(y)$. Tab.~\ref{tab:ablation} shows the accuracy for the different prior formats, highlighting the 3D keypoints as the most effective choice.
\section{Applications}
\label{sec:results}
We empirically validate the utility of BodyMetric within the emerging domain of text-to-image generation, and introduce the BodyRealism benchmark to spur advances towards the generation of images with increased body realism. The BodyRealism benchmark consists of 100 synthetic prompts describing diverse and challenging poses.

\subsection{Benchmarking Text-to-Image Models}
We systematically apply BodyMetric (Eq.~\ref{eq:intructGPT}) to calculate body realism scores of SOTA text-to-image models on the BodyRealism benchmark. For each T2I model, we repeat the generation 20 times and report the mean across all samples. We use a common set of randomly negative prompts across all models. We follow the process described in Sec.~\ref{subsec:dataset} to blur the faces, filter out images and moderate NSFW samples. The results, as shown in Tab.~\ref{tab:BodyRealism_benchmark}, coincide with our observations; BodyMetric identifies SD-XL~\cite{Podel:2023:SDXL} as the model that generates more realistic humans, with Wuerstchen~\cite{pernias2023wuerstchen}, following and SD-1.4~\cite{Roombach:2022} as the model which generates the least realistic humans. 

\vspace{-6mm}
\begin{table}[h]
\begin{minipage}[b]{1\linewidth}
\centering
\caption{Benchmarking T2I on BodyRealism. Left-to-right: least to most realistic generated bodies.} 
\begin{tabular}{c|c|c|c|c|c}
\multicolumn{1}{c}{} & \multicolumn{5}{c}{\textbf{Text-to-Image Models}} \\
\hline
\hline
\tstrut
 & SD-1.4~\cite{Roombach:2022} & SD-XL-T~\cite{Podel:2023:SDXL} & SD-2.1~\cite{Roombach:2022} & Wuerst.~\cite{pernias2023wuerstchen} & SD-XL~\cite{Sauer:2023:SDXLTurbo} \\ 
\hline
\tstrut
\tstrut
BodyMetric & $-0.45 $ & $-0.26$ & $-0.25 $ & $0.69 $ & $0.92$  \\ 
\end{tabular}
\tstrut
\label{tab:BodyRealism_benchmark} 
\end{minipage}
\end{table}

\subsection{Image Ranking}
BodyMetric can be used to rank sets of generated images by sorting the predicted scores from highest to lowest; a higher score corresponds to higher body realism. Fig.~\ref{fig:ranking} demonstrates how BodyMetric successfully ranks sets of 3 images. We carefully design the image sets so that the samples cover the three-tiered severity scale as described in Sec.~\ref{sec:realism_annotation}. 
\begin{table}
 \begin{minipage}[b]{0.5\linewidth}
    \begin{tabular}{cc}
    \multicolumn{2}{p{5.5cm}}{a girl in a dress standing in a field.} \\
    \includegraphics[width=2.5cm, cfbox=brown 1pt  1pt]{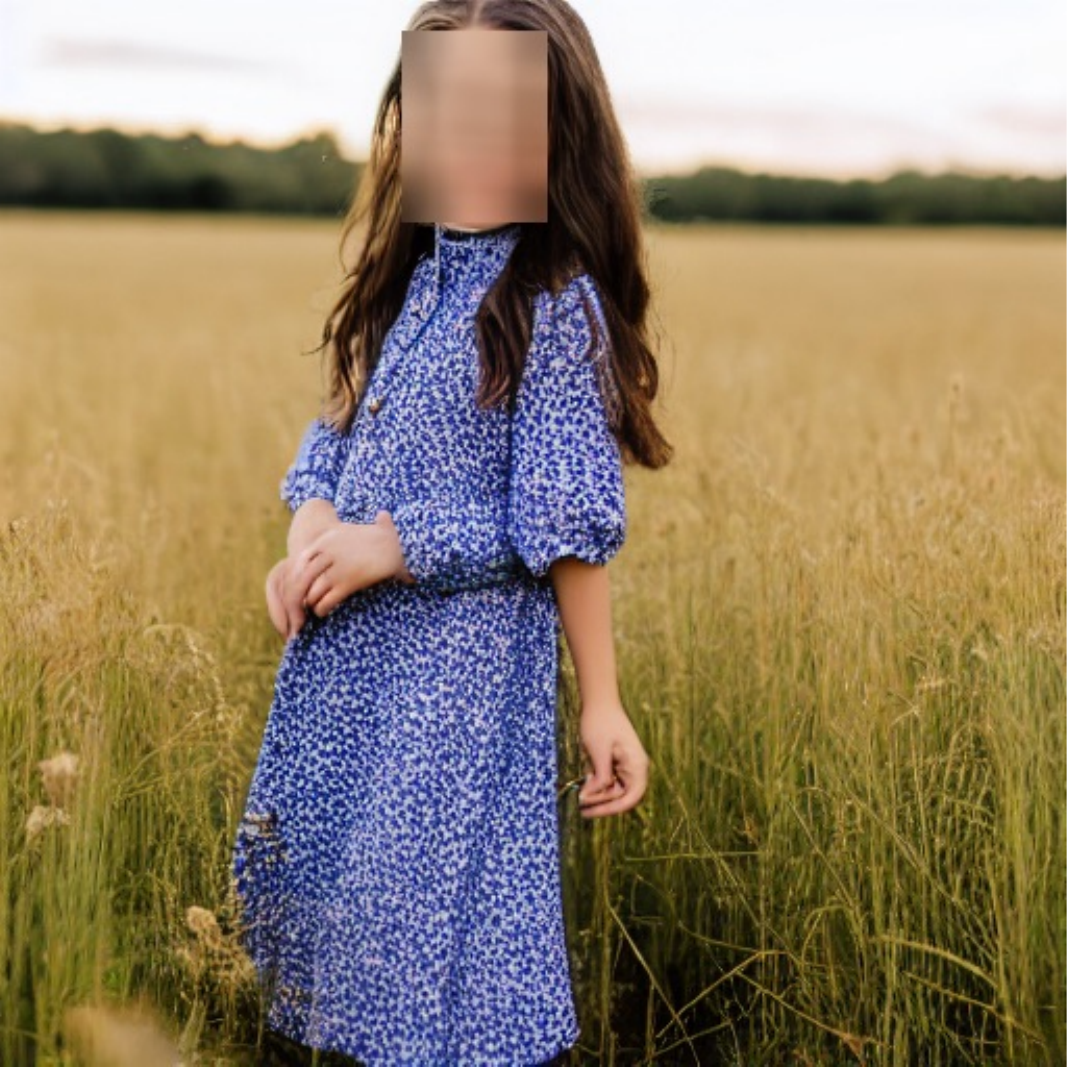} & \includegraphics[width=2.5cm, height=2.5cm]{bodymetric/figures/failure_cases/eg4/image1.pdf} 
    \end{tabular}
\end{minipage}\hfill%
    \begin{minipage}[b]{0.5\linewidth}
    \begin{tabular}{cc}
        \multicolumn{2}{p{5.5cm}}{a woman leading a horse.} \\
        \includegraphics[width=2.5cm,cfbox=brown 1pt  1pt]{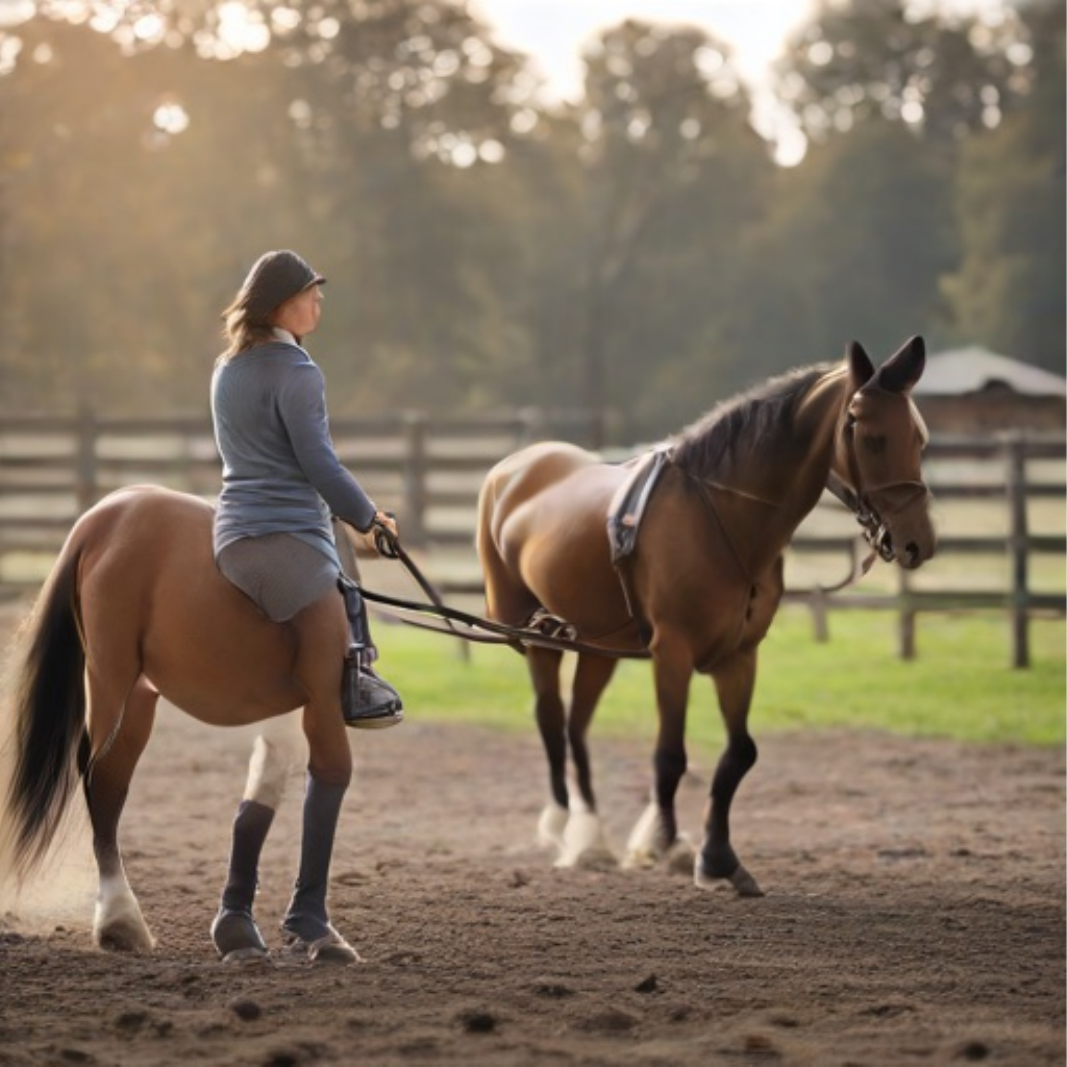} & \includegraphics[width=2.5cm]{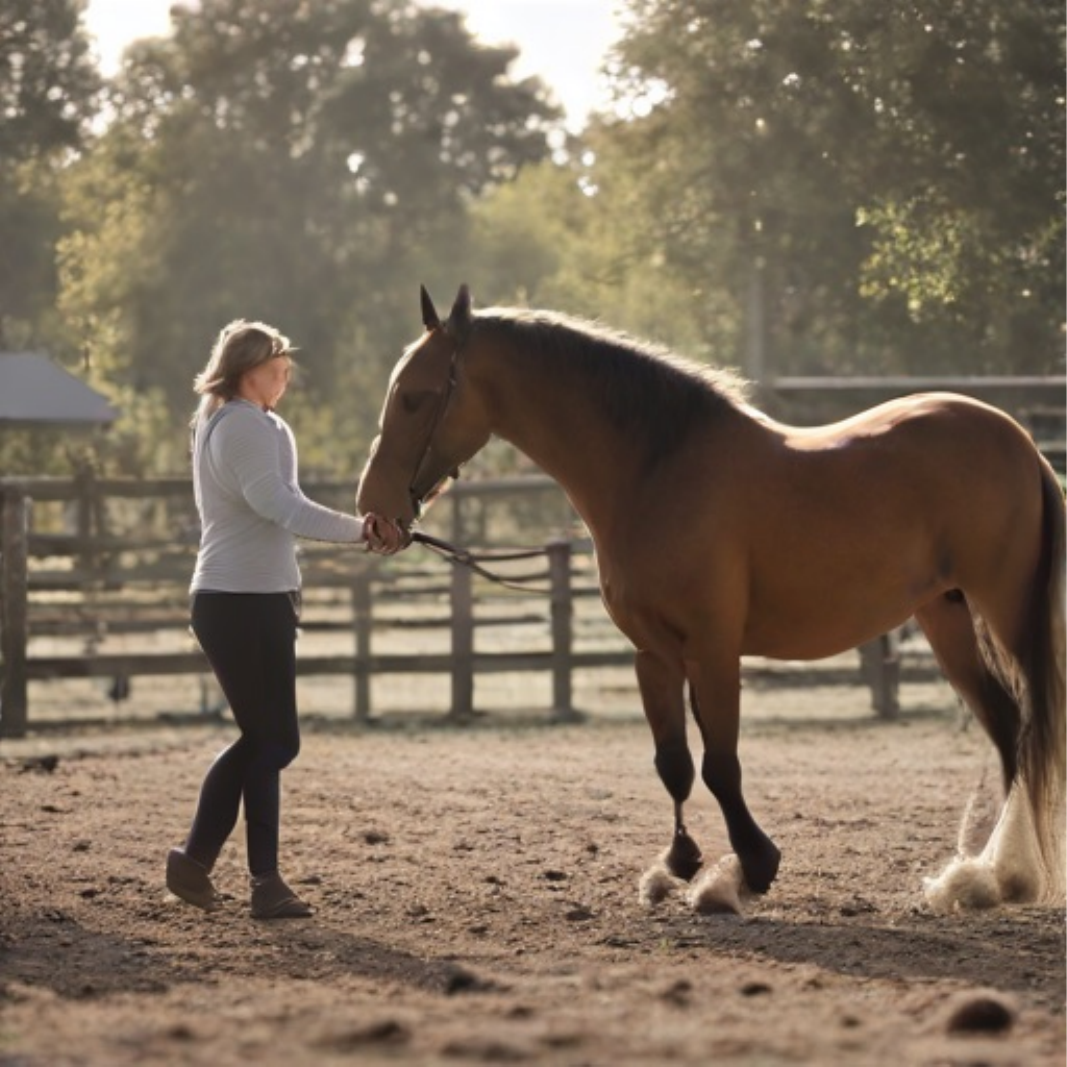} 
    \end{tabular}
    \end{minipage}\hfill%
\captionsetup{type=figure} 
\caption{Failure cases. BodyMetric incorrectly predicts the images highlighted with red border as the most realistic in each pair.}
\label{fig:failure_cases}
\end{table}
 
\begin{table}[h]
\begin{minipage}[b]{0.5\linewidth}
\begin{tabular}{ p{1.8cm}p{1.8cm}p{1.8cm}}
    \multicolumn{3}{p{5.5cm}}{\scriptsize{a ballet dancer en pointe, their form poised and graceful. }} \\
     \includegraphics[width=1.8cm]{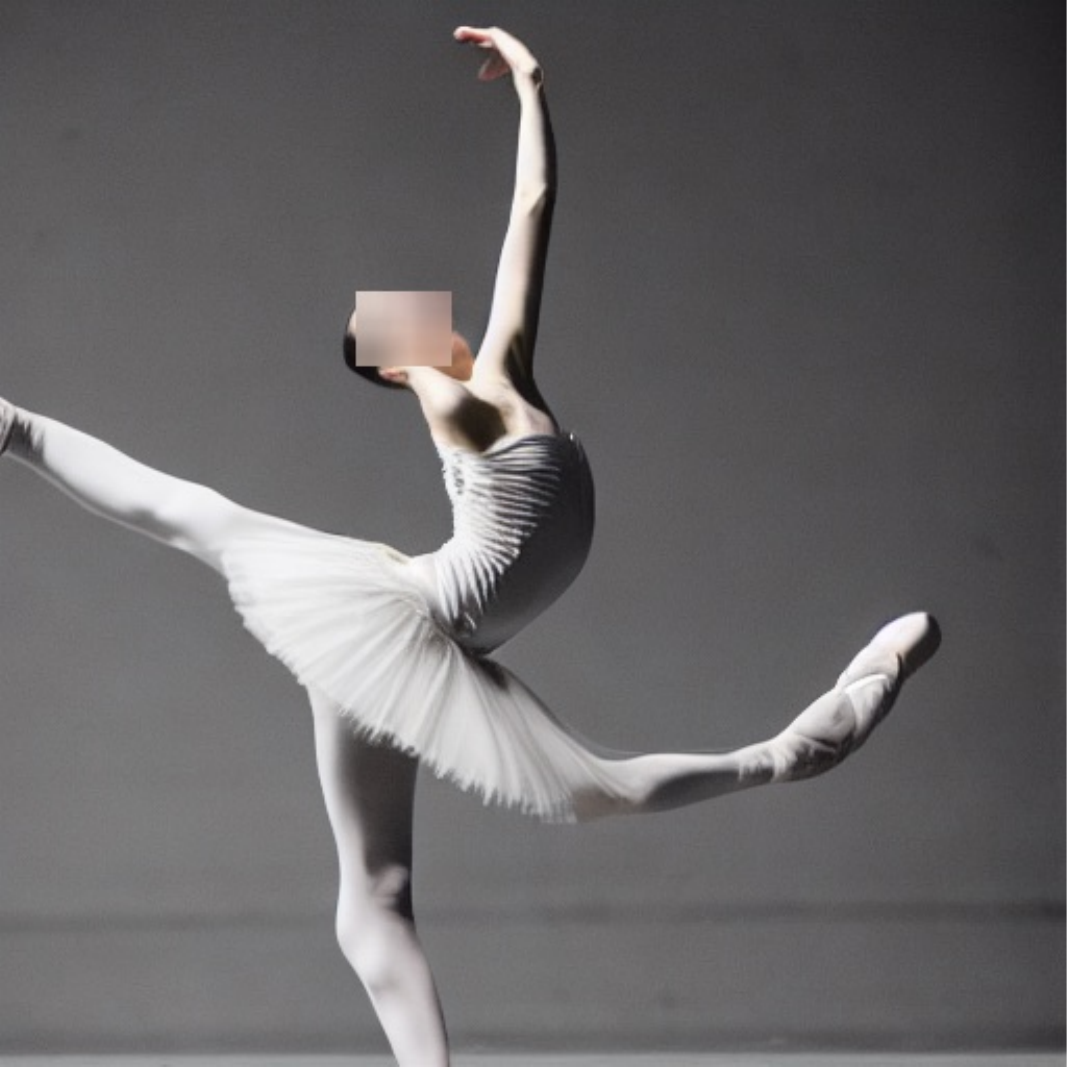} &   \includegraphics[width=1.8cm]{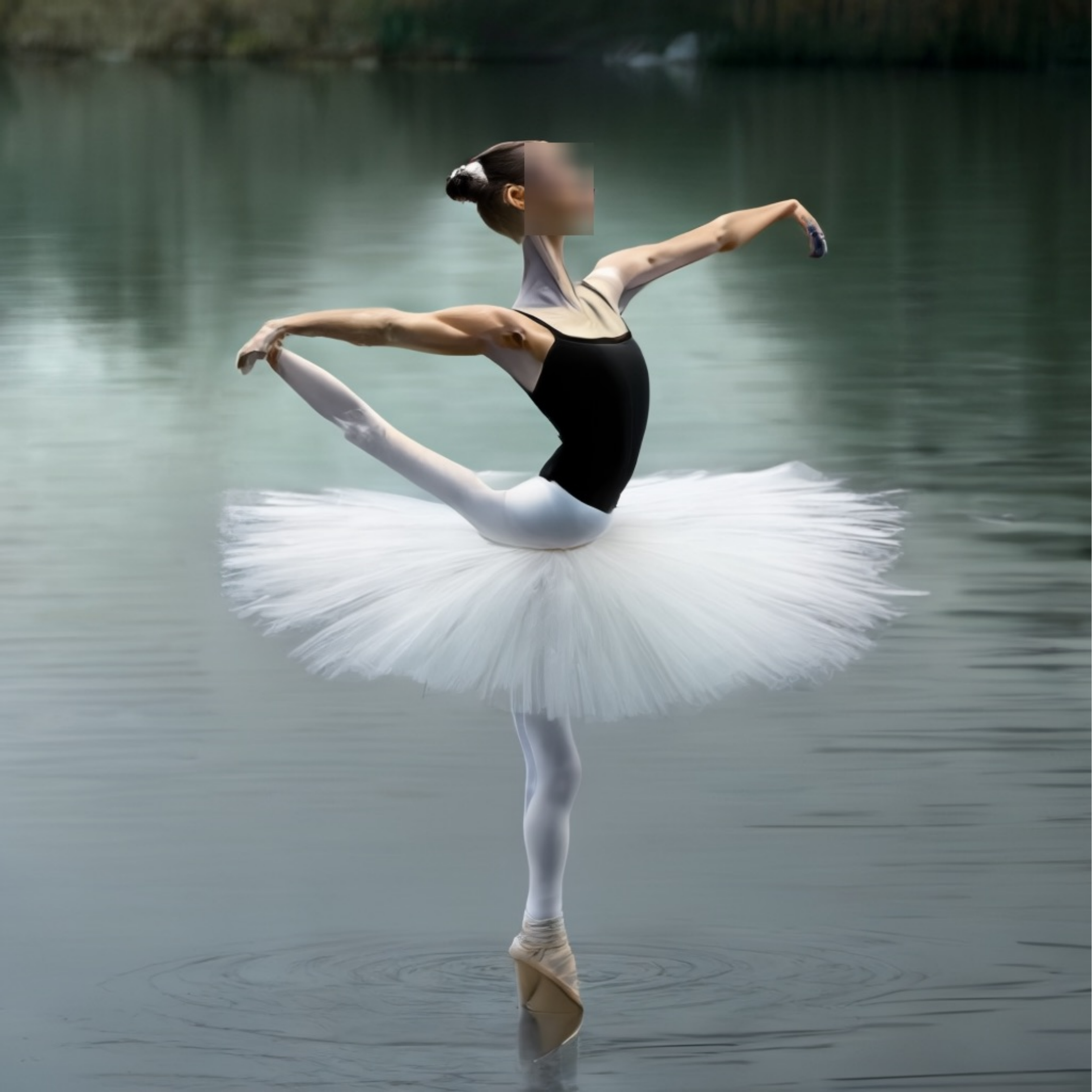} &   \includegraphics[width=1.8cm]{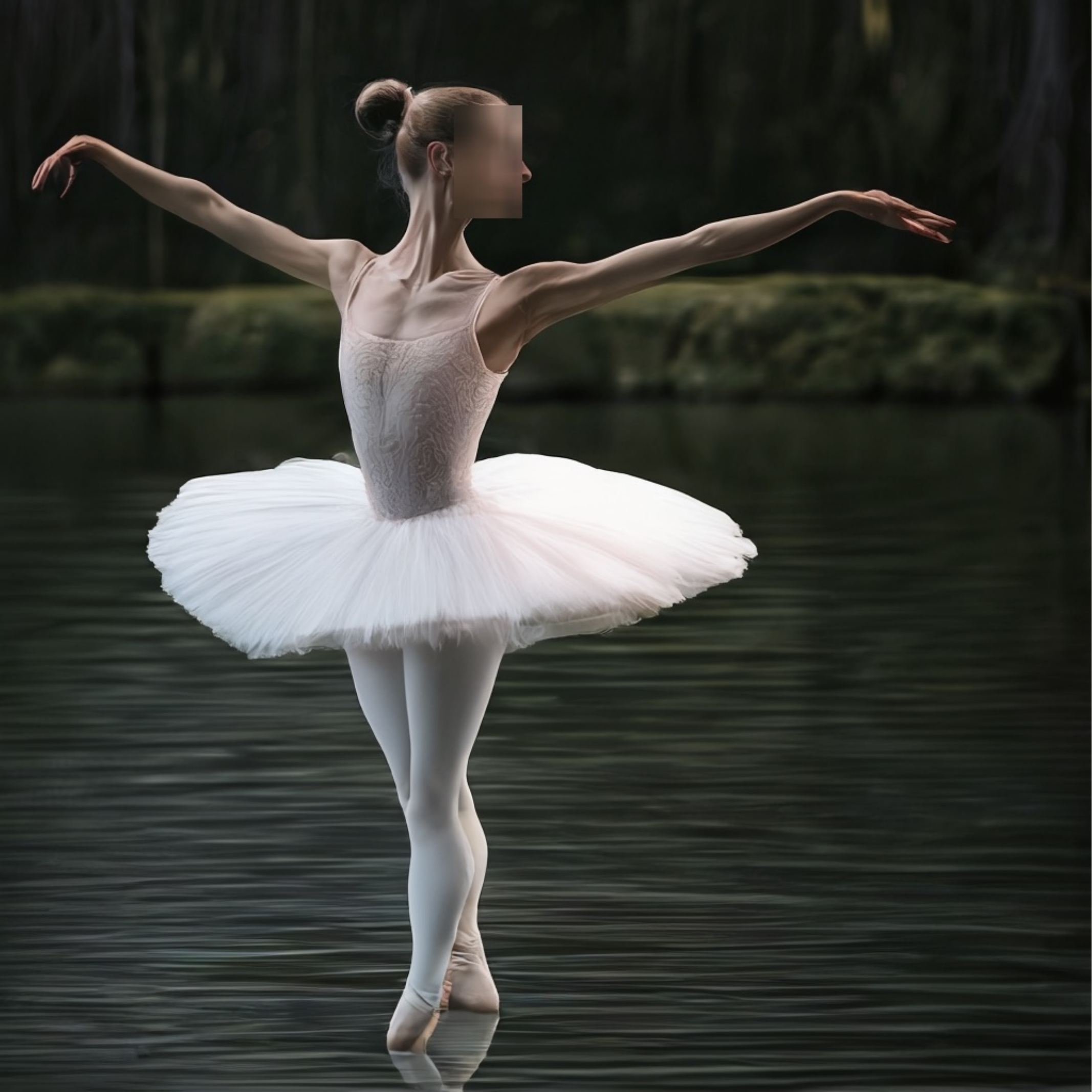} \\
         \multicolumn{3}{p{5.5cm}}{\scriptsize{a model with their weight evenly distributed on both feet.}} \\
     \includegraphics[width=1.8cm]{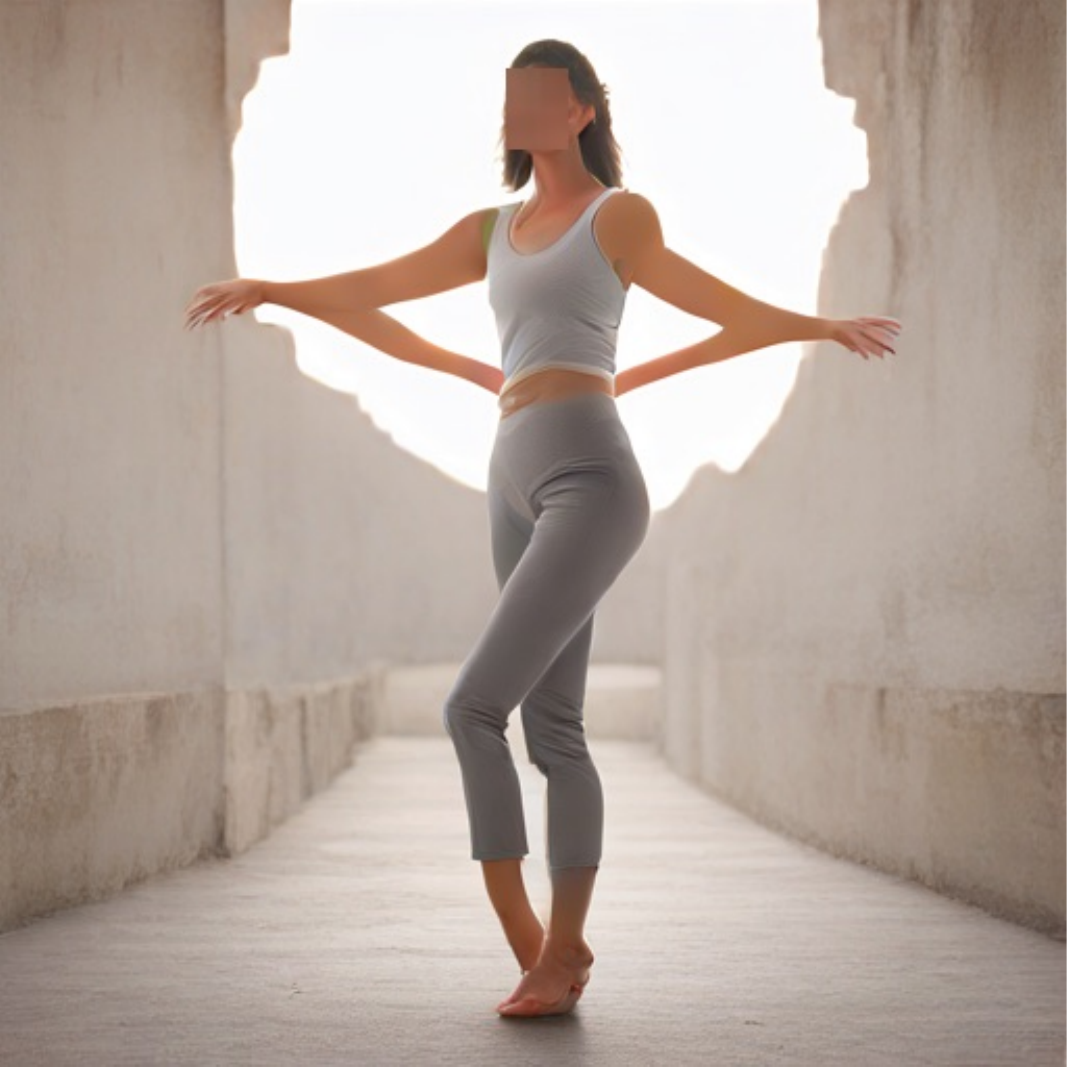} &   \includegraphics[width=1.8cm]{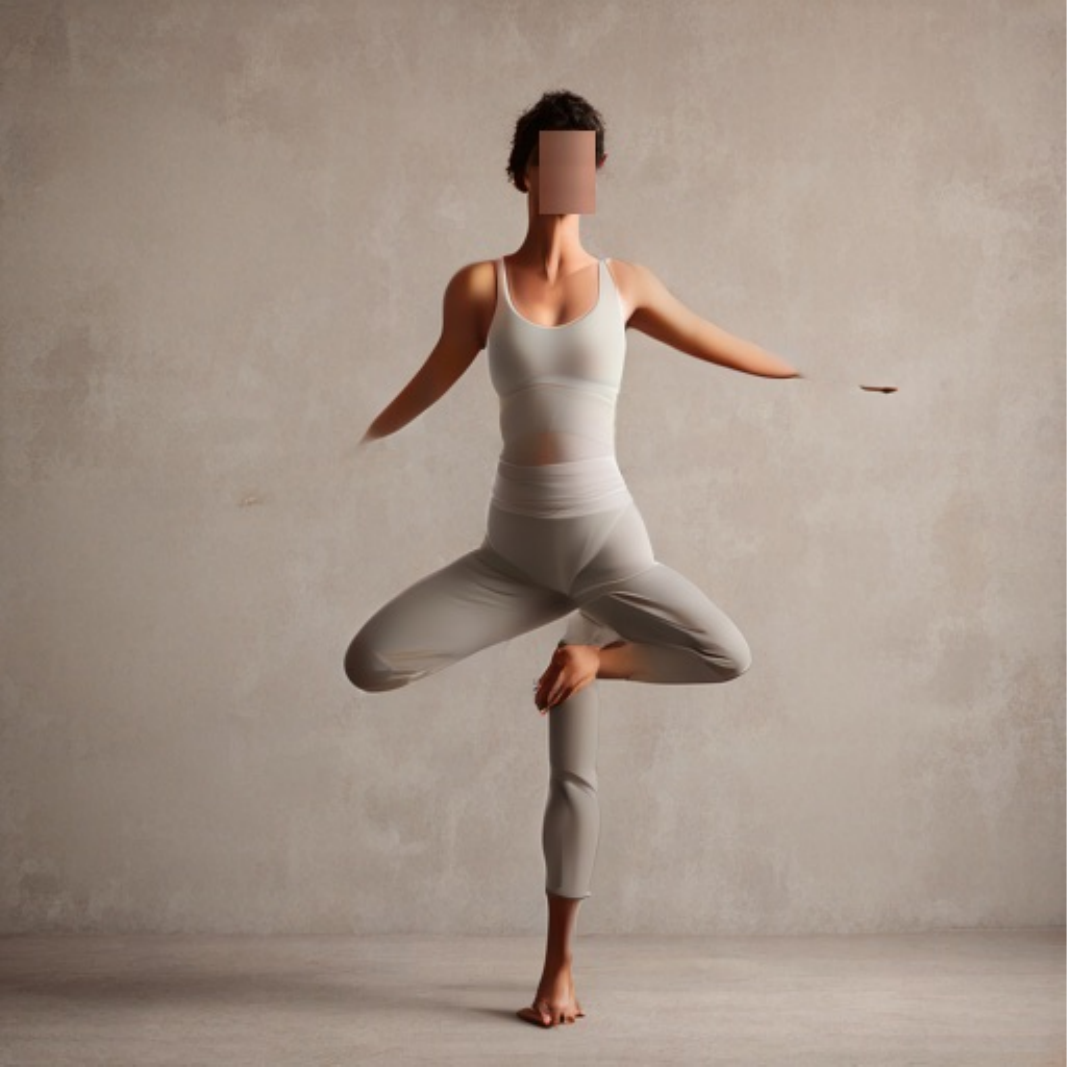} &   \includegraphics[width=1.8cm]{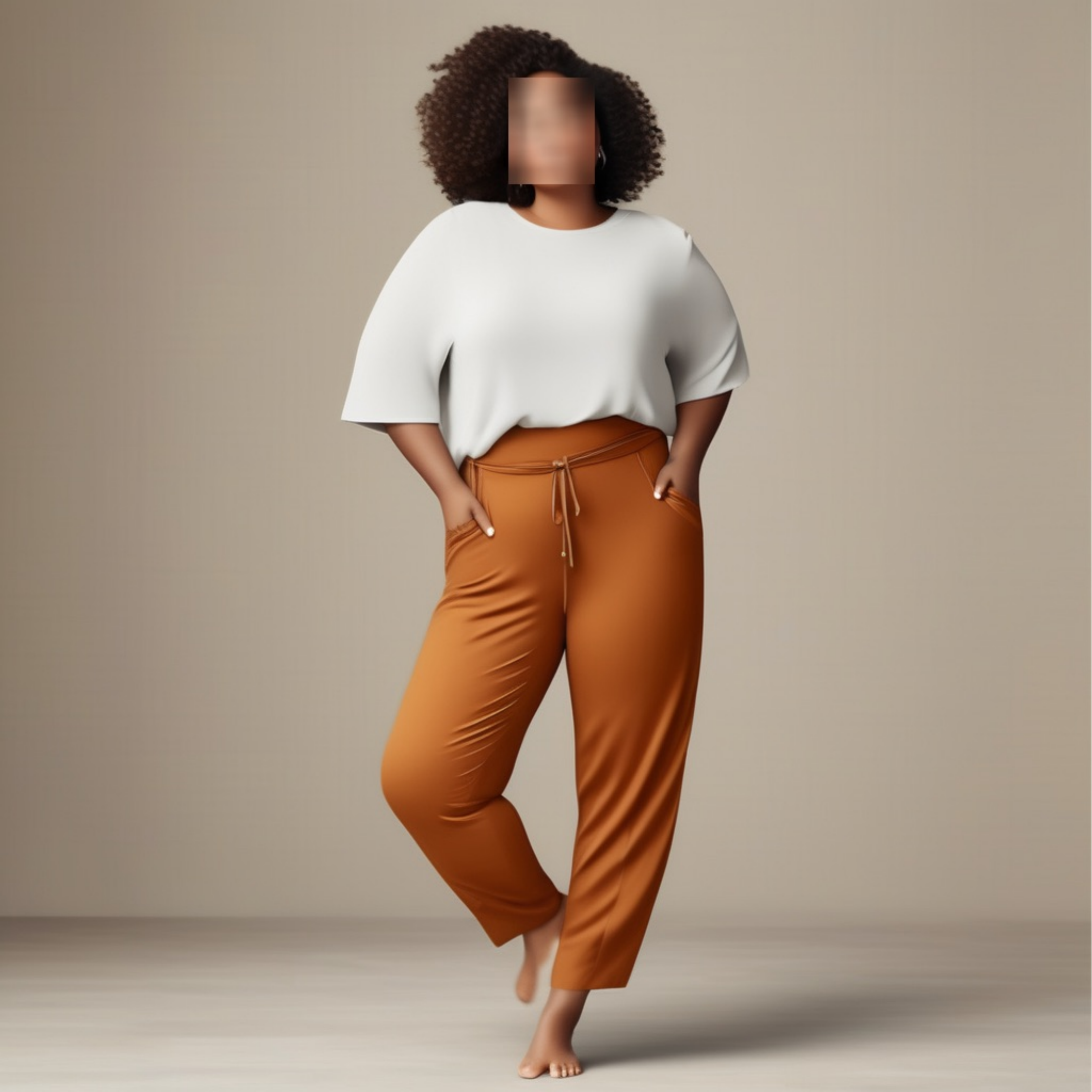} \\
\end{tabular}
\end{minipage}\hfill
\begin{minipage}[b]{0.5\linewidth}
\begin{tabular}{ p{1.8cm}p{1.8cm}p{1.8cm}}
       \multicolumn{3}{p{5.5cm}}{\scriptsize{a male model with a cane stands tall and proud.}} \\
    \includegraphics[width=1.8cm]{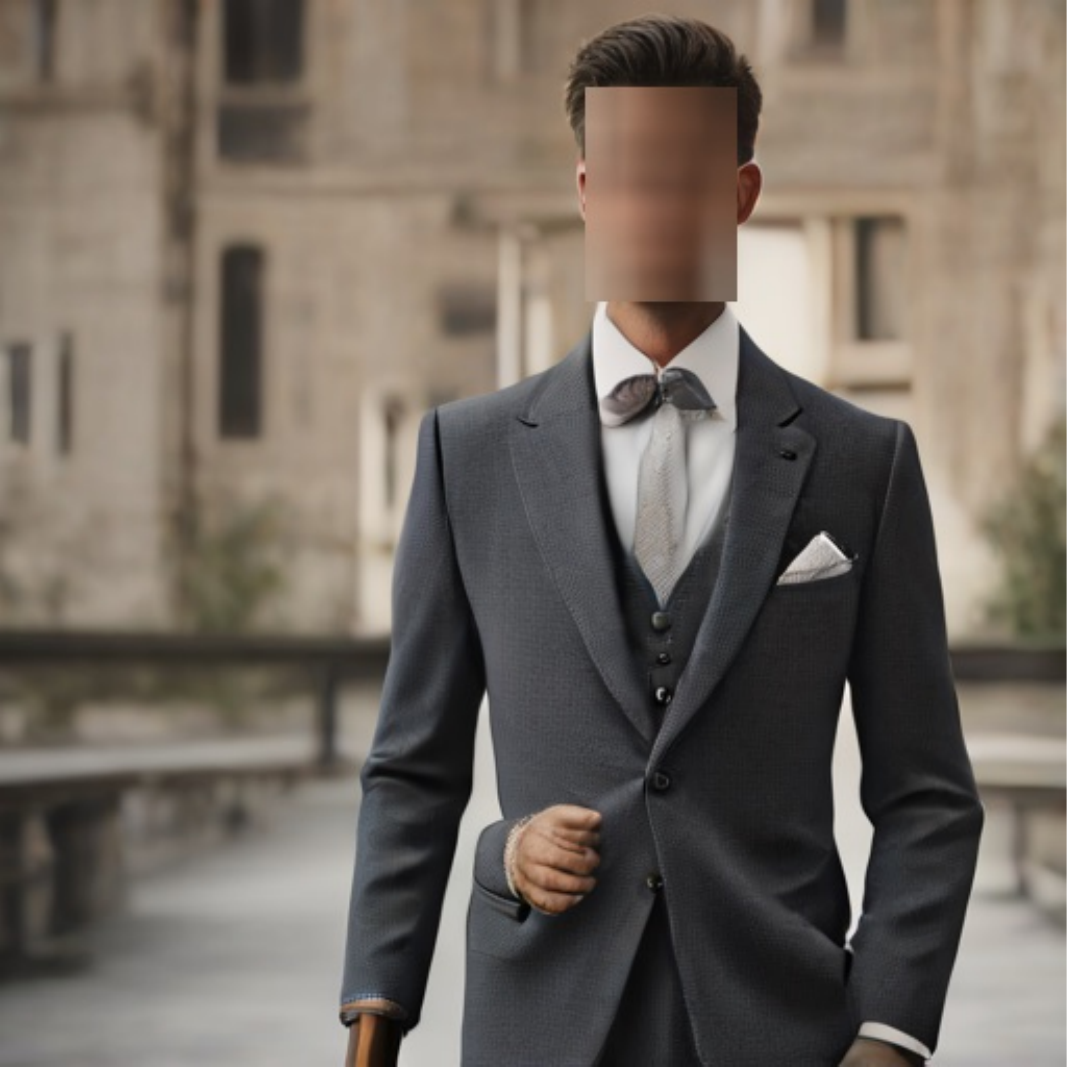} &   \includegraphics[width=1.8cm]{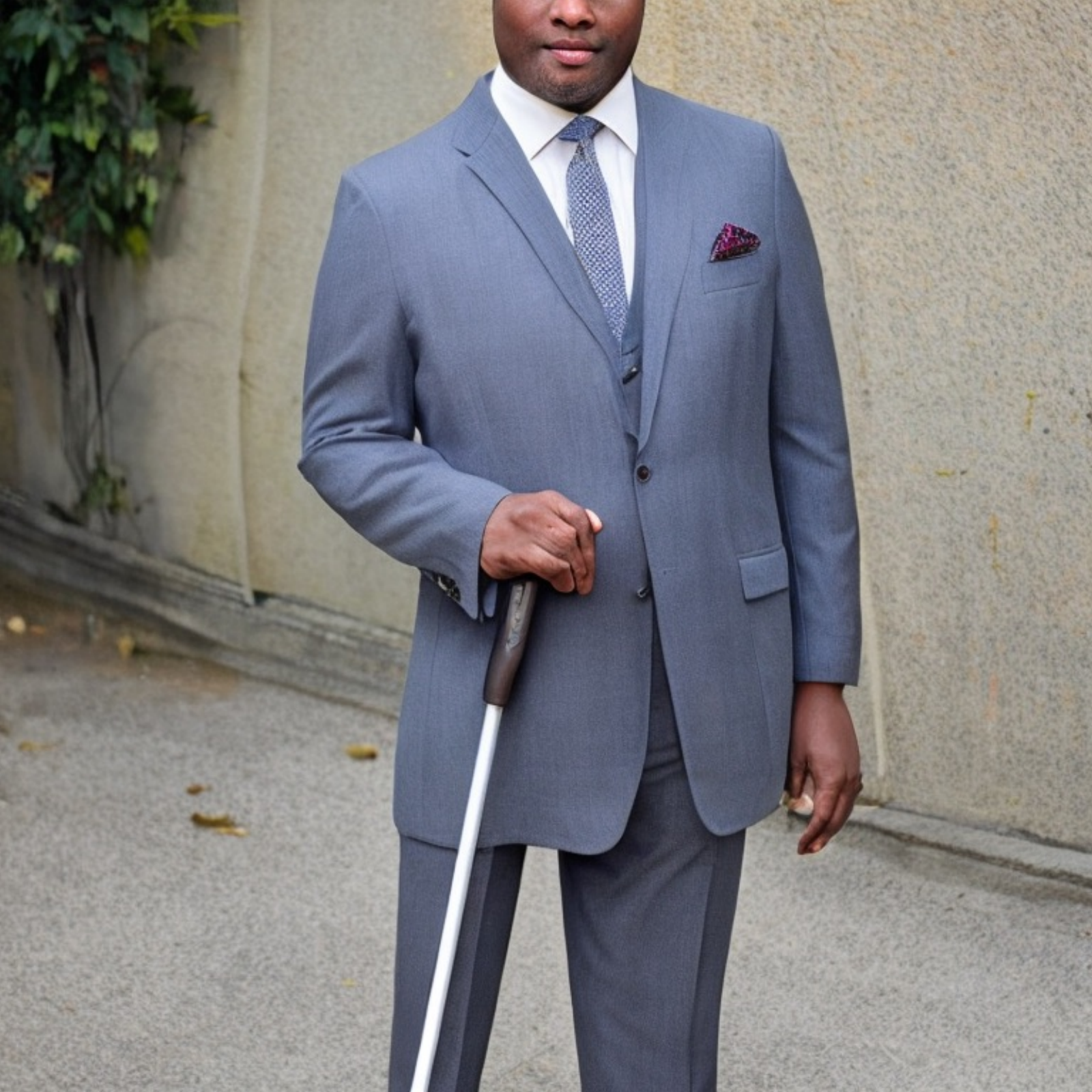} &   \includegraphics[width=1.8cm]{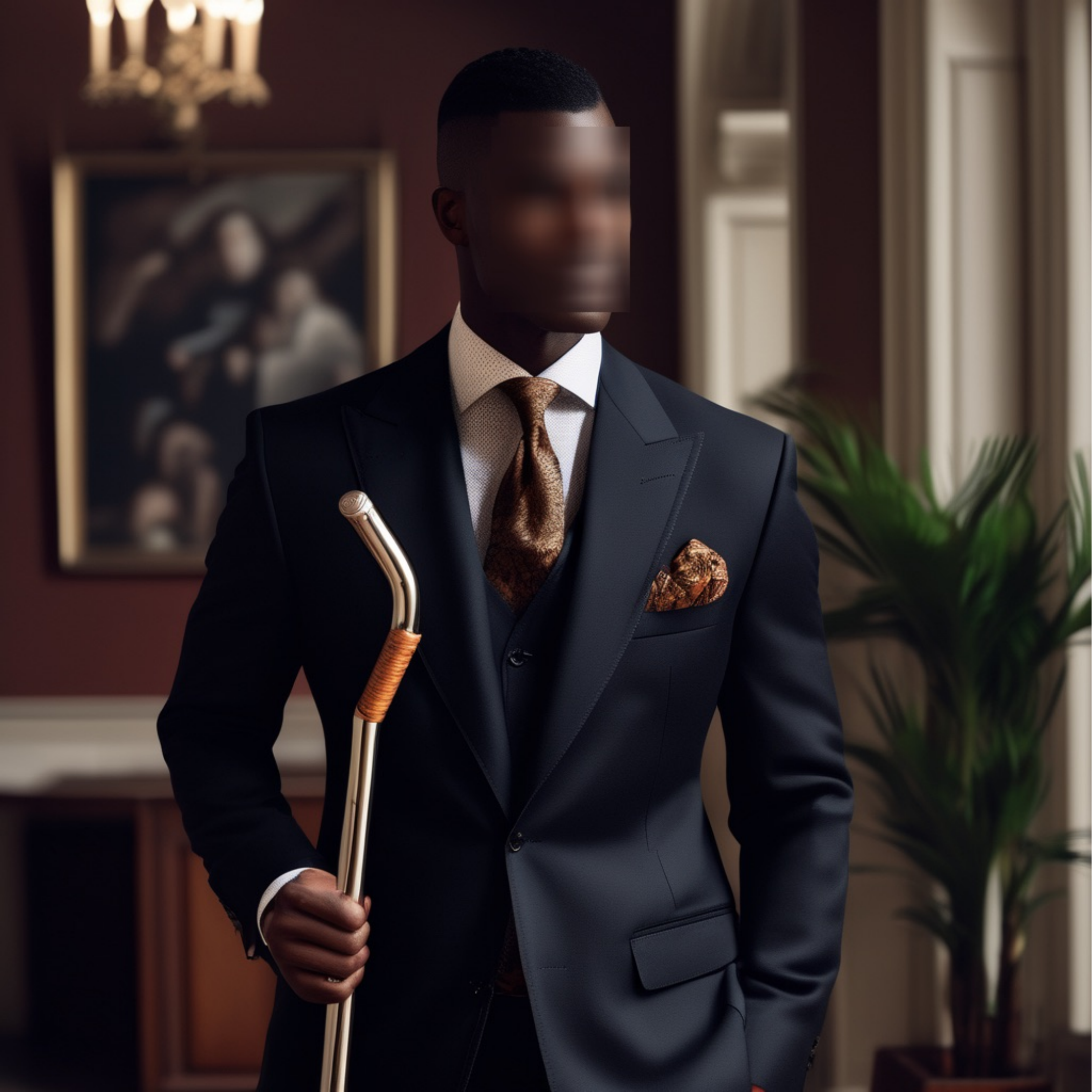} \\
           \multicolumn{3}{p{5.5cm}}{\scriptsize{a skateboarder grinding down a handrail, fearless and exhilarated.}} \\
    \includegraphics[width=1.8cm]{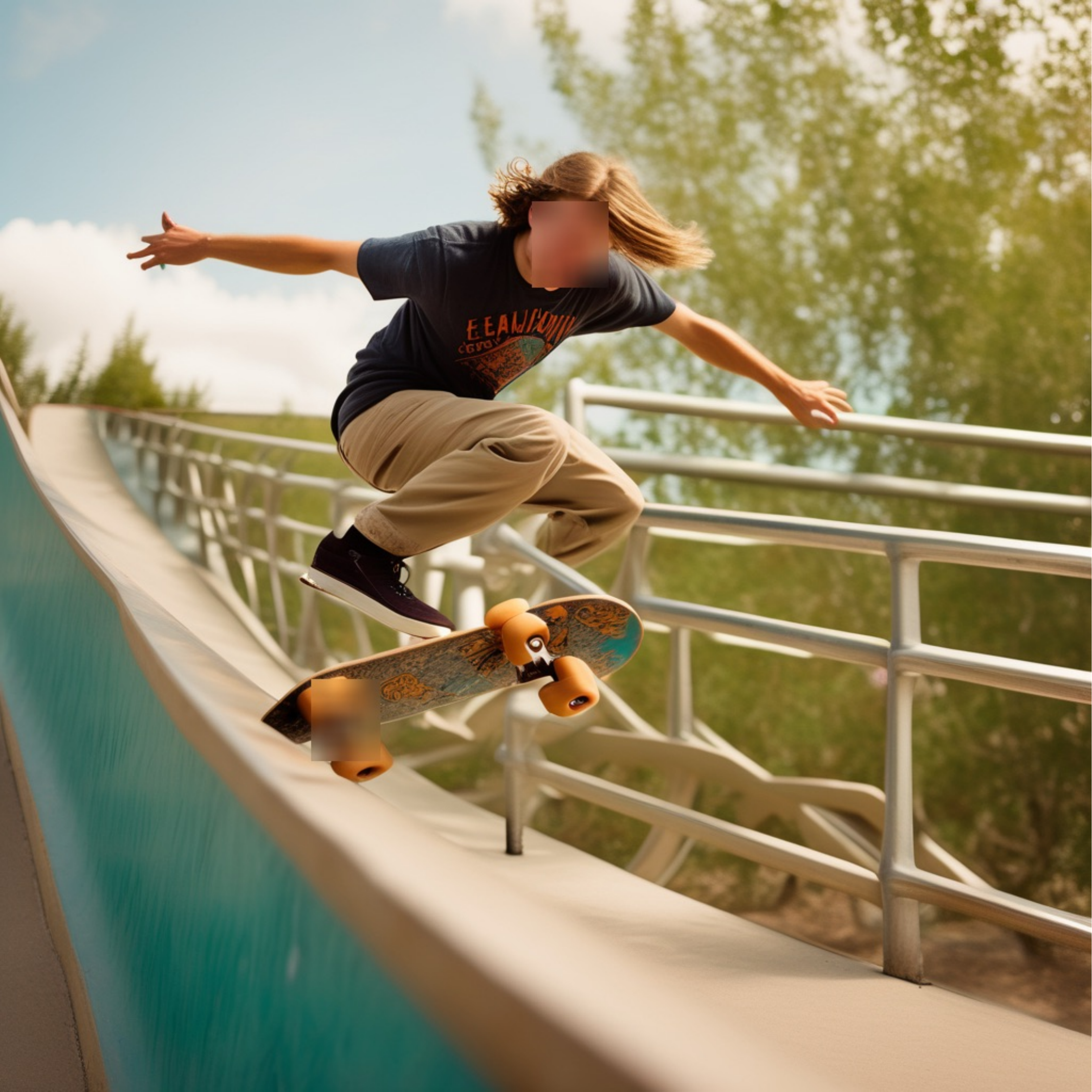} &   \includegraphics[width=1.8cm]{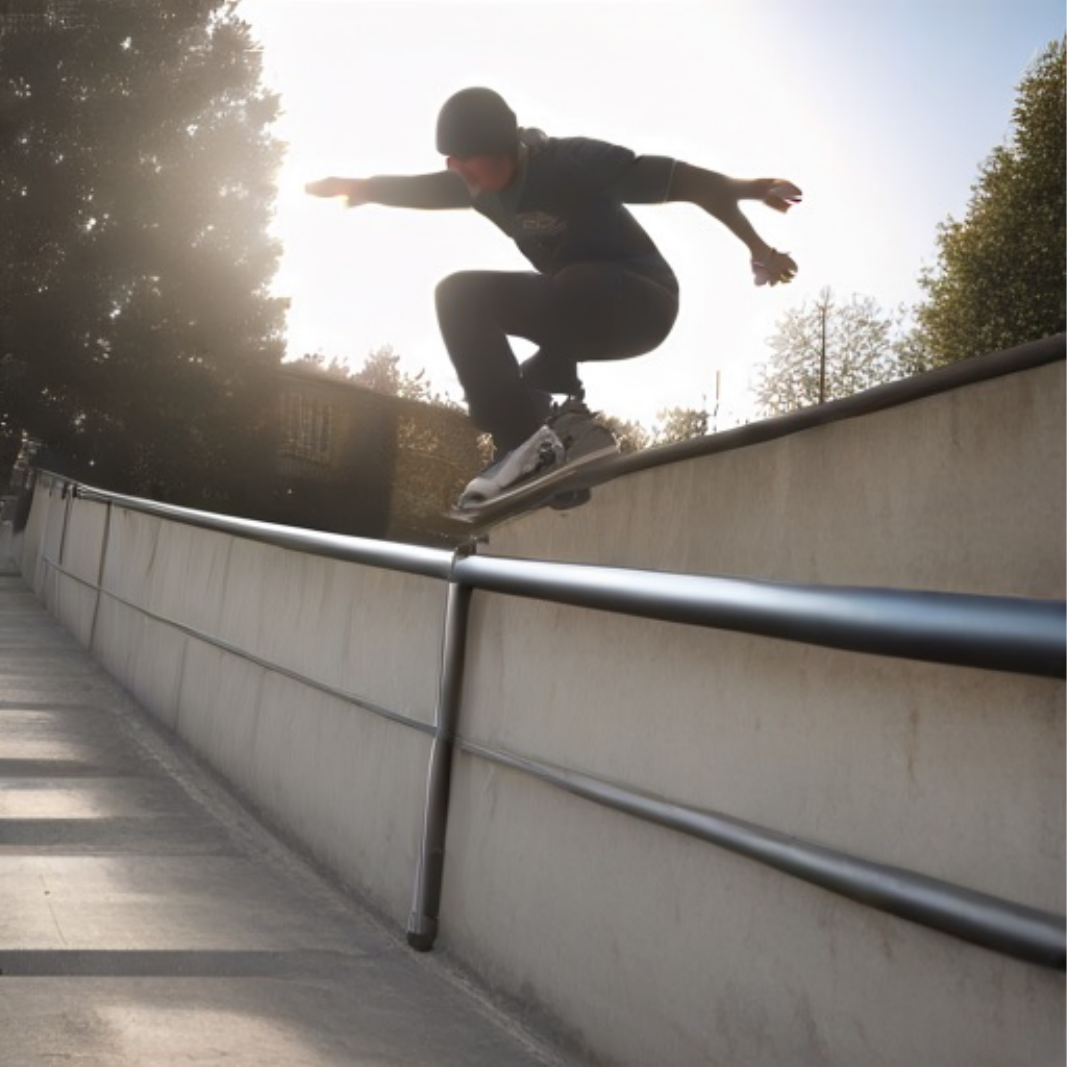} &   \includegraphics[width=1.8cm]{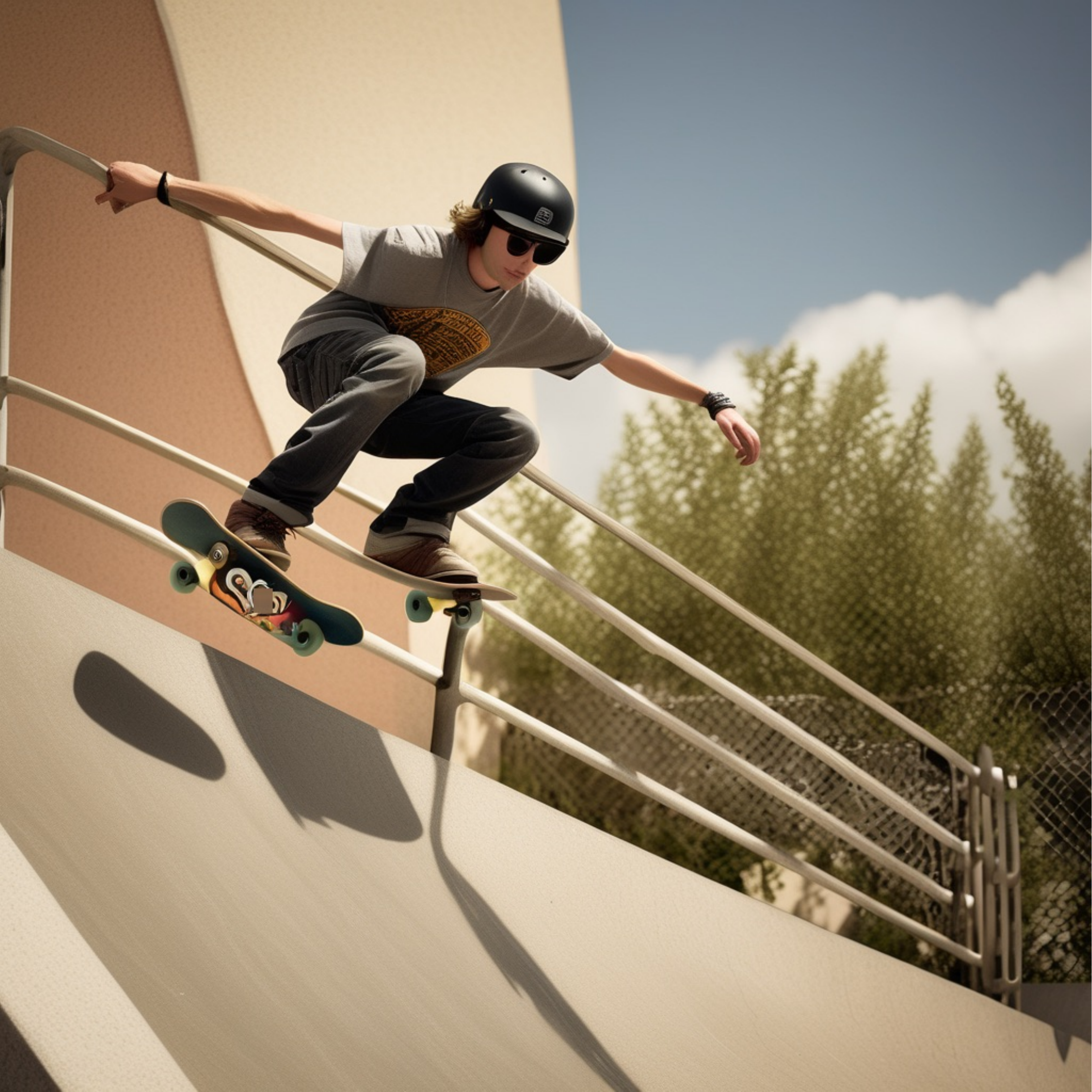} \\
\end{tabular}
\end{minipage}\hfill
\captionsetup{type=figure} 
\caption{Ranking generated images. From left to right, unrealistic to realistic bodies as ranked by BodyMetric.}
 \label{fig:ranking}
\end{table}
\vspace{-5mm}

\section{Conclusion}
\label{sec:Conclusion}
We design a learnable metric to quantify human body realism related artifacts in images, a key challenge in text-to-image generation. Our metric is trained using the BodyRealism dataset, a new  curated multi-modal dataset of images, text descriptions, expert annotations on body realism, and 3D SMPL-X joint keypoints. BodyMetric uses a carefully designed architecture to leverage the multi-modal signal, making it the first body-aware image evaluation metric. We demonstrate both qualitatively and quantitatively that BodyMetric better reflects the quality of generated humans, compared to existing evaluation metrics. In addition, we define a challenging benchmark for text-to-image generation utilising BodyRealism and BodyMetric. We demonstrate how BodyMetric can be used to improve ranking of generated images, leading to the selection of images depicting realistic bodies.

\noindent \textbf{Failure Cases}
As seen in Fig.~\ref{fig:failure_cases} BodyMetric struggles to accurately capture body realism in cases where the body parts appear merged with other objects such as clothing or animals. 
\vspace{6pt} \\
\noindent \textbf{Future Work}
The framework introduced in BodyMetric is adaptable and extensible for various envisioned scenarios such as multi-humans, different 3D body models, and different image generators. Keeping future expansion possibilities in mind, BodyMetric leverages the text modality to serve for diversity and inclusivity; although a typical human figure follows the SMPL-x structure, this is not the case for amputees or humans with other anatomical variations. Having the text enables techniques such as weighting SMPL-x reconstructions based on the textual context.
  
In the current version of the dataset, body realism annotations are defined on a 1-10 scale. In future versions of the dataset, we plan to collect fine-grained annotations on the specific body-parts with unrealistic characteristics, enabling more targeted analysis. Similarly to other existing metrics, despite including images of varying body realism and resolution, additional fine-tuning may be needed for vastly different image generators. To support this, we provide the current text space for generating new images, an HTML annotation template for obtaining body realism scores, and standardized instructions to ensure consistency with existing scores.

BodyMetric currently supports images depicting single humans. An interesting expansion would be for images depicting multiple humans, in which case body realism can be measured by considering BodyMetric scores across single-human crops of the image. Inter-human interactions (such as hand-shakes, hugs, occlusions of humans by humans) introduce additional interesting challenges that would be valuable to capture in an extended BodyRealism dataset.

\newpage
\clearpage
\setcounter{page}{1}
\newpage
\Large

{\centering{\textbf{BodyMetric: Evaluating the Realism of Human Bodies in Text-to-Image Generation}}}\\
\vspace{0.5em}
{\centering{Supplementary Material \\}}
\vspace{-3mm}
\normalsize
\renewcommand{\thesection}{\Alph{section}}
\setcounter{section}{0}
\counterwithin{figure}{section}

\section{Dataset}

\subsection{Images}
\noindent \textbf{Blurring faces} We use Multi-Task Cascaded Convolutional Neural Networks (MTCNN~\cite{MTCNN}) from Facenet-Pytorch to detect the faces in each image. Then we use the bounding box indices to identify a rectangular region around each face and apply Gaussian blur.
\vspace{10pt}

\noindent \textbf{Filtering} 
 In order to minimize the occurrence of invalid images, we use COCO - InstanceSegmentation from Detectron2 to filter out images under the following conditions: (a) number of detected humans is more than 3 (eliminate case of multiple humans), (b) number of unique detected classes is more than 3 and (eliminated non-human classes/occluded humans), (c) confidence score for detected humans is lower than 98\%. 
\vspace{10pt}

\noindent \textbf{Moderation} We perform moderation to remove all NSFW images. To do so, we use Amazon Rekognition to obtain the moderation labels with a minimum confidence threshold of 0.9. 

\subsection{Prompts}
We formulate part of our text space using CLIP ImageNet templates. For that, we use particular classes for the subject (e.g. person, woman, man, girl, boy, child) and specific actions which are likely to result in generated humans with body artifacts (e.g. standing, waving, sitting, walking, jogging, dancing). Part of our text space consists of prompts from DiffusionDB, TiFa, MS COCO, Pick-a-pic and openPARTI. We utilise prompts containing keywords such as ``person, woman, man, girl, boy, child''; we have also experimented with a second level filtering using a \textit{Llama-2-7b-chat-hf} but have found this to not be so effective.

\subsection{Standard Operating Procedure (SOP) for Body Realism Annotations}
We carefully design an instructional template which is used to collect the body-realism annotations (see Fig.~\ref{fig:BodyRealism}). We use a 1-10 scale for body photo-realism scores. Each score correlates with the degree of body related errors in the image. Low scores (3 or less) correspond to major artifacts, such as extra/missing legs or arms. High-scores (8 or more) correspond to human bodies that do not have obvious artifacts. Bodies with less major artifacts (blurred bodies or parts, extra/missing fingers) are scored between 4-7. We instruct annotators to follow a mental three-tiered process and assign scores of 1-3 when body related errors are immediately obvious when looking at the image and scores higher than 7 to images with no obvious body errors. Scores 4-7 correspond to a ``grey-area'' with moderate errors. 

Images are labelled as invalid when more than one person is visible in the image. In addition, we consider invalid images for which less than 3 body parts are visible (excluding the head) and images which are non-photorealistic, i.e. paintings, cartoons or photographs.

By understanding how artifacts relate inversely to realism scores, readers gain useful context for interpreting evaluation results. Fig.~\ref{tab:instruct} provides representative image examples for different realism scores.

\subsection{Consolidating Annotations}
We provide details on the consolidation algorithm used to distil the scores assigned by all 5 annotators to a unique indicative body realism score for each image.
\begin{table}[h]
    \centering
    \begin{tabular}{c|c||c|c||c}
        Score & Image & Score & Image & invalid\\
        \hline
            1 & \includegraphics[width=3cm]{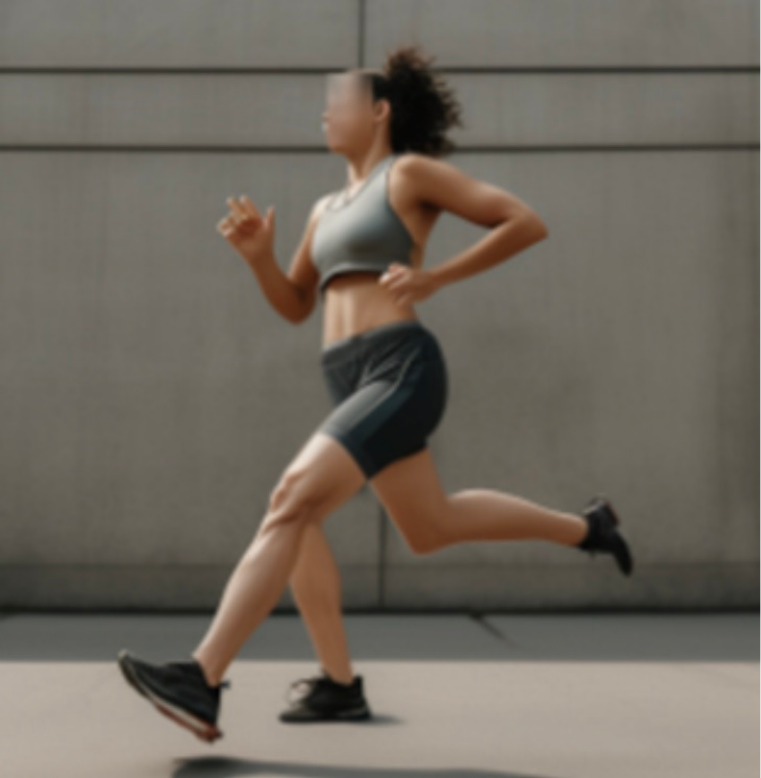} & 6 & \includegraphics[width=3cm]{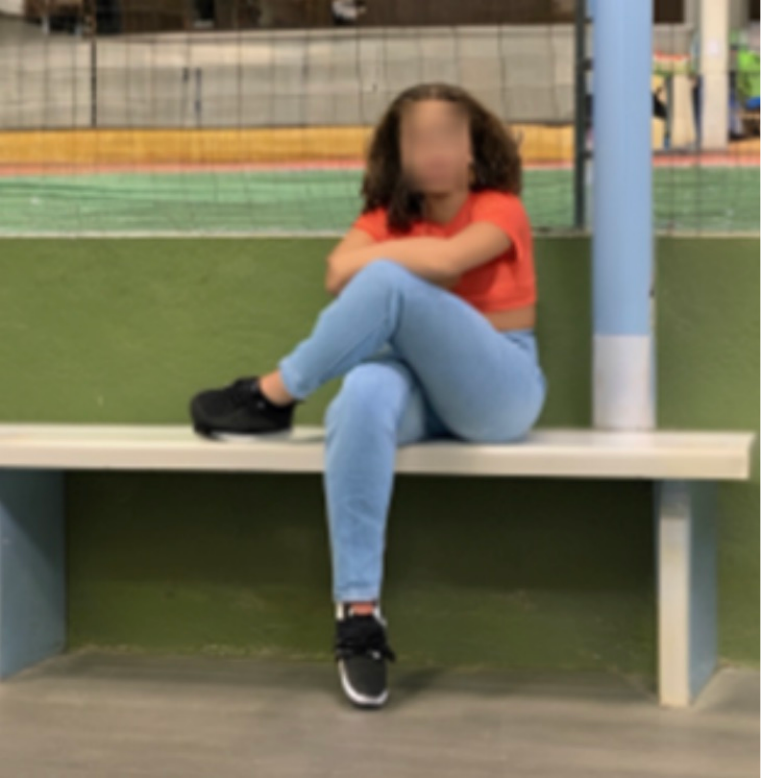}  & \includegraphics[width=3cm]{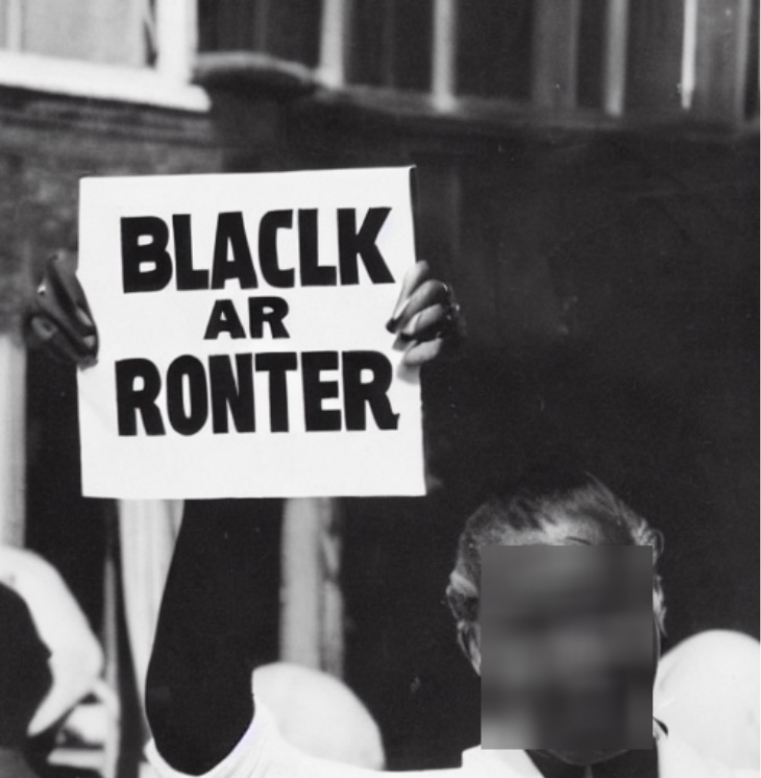} \\
        2 & \includegraphics[width=3cm]{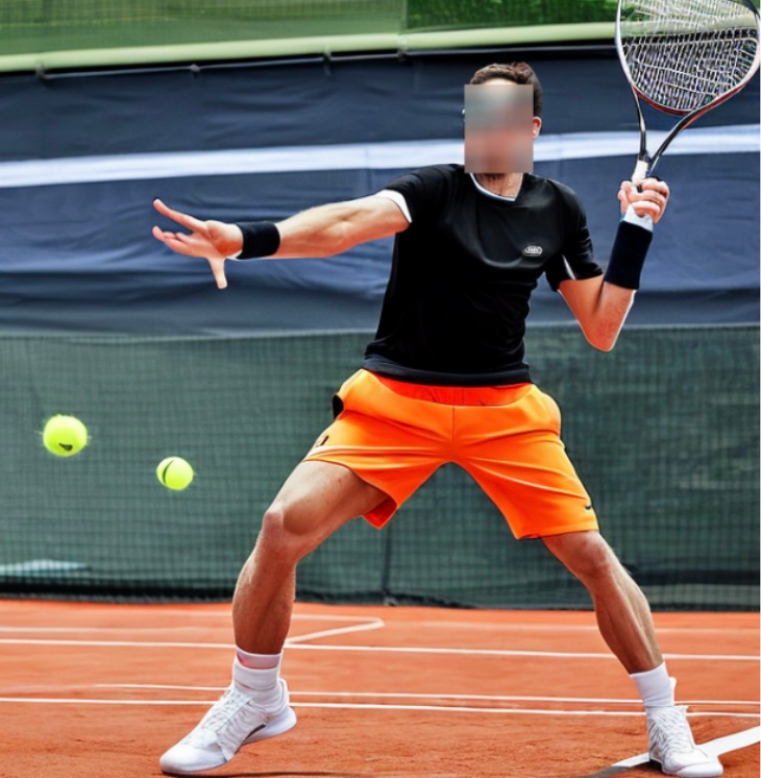} & 7 & \includegraphics[width=3cm]{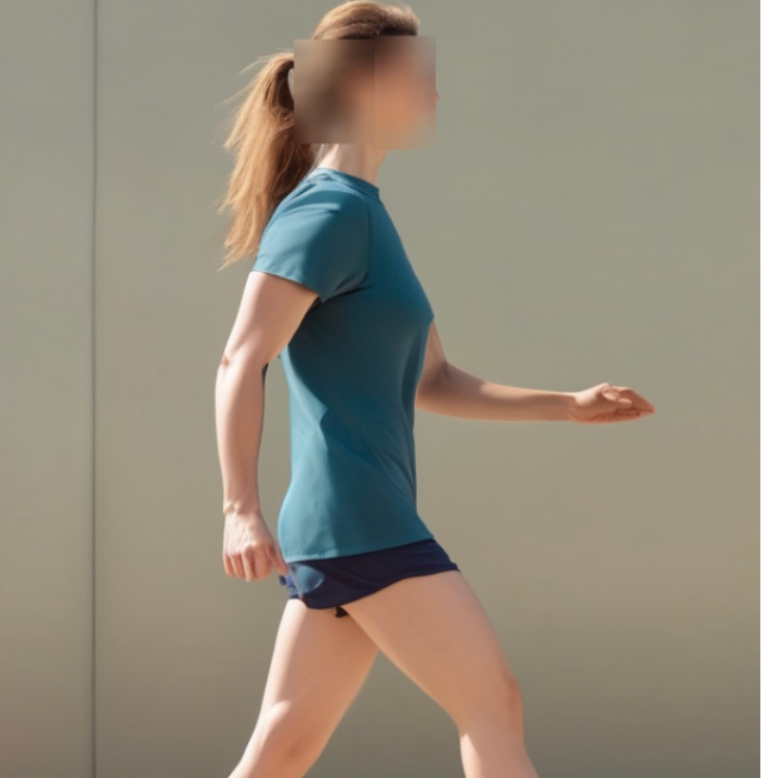} & \includegraphics[width=3cm]{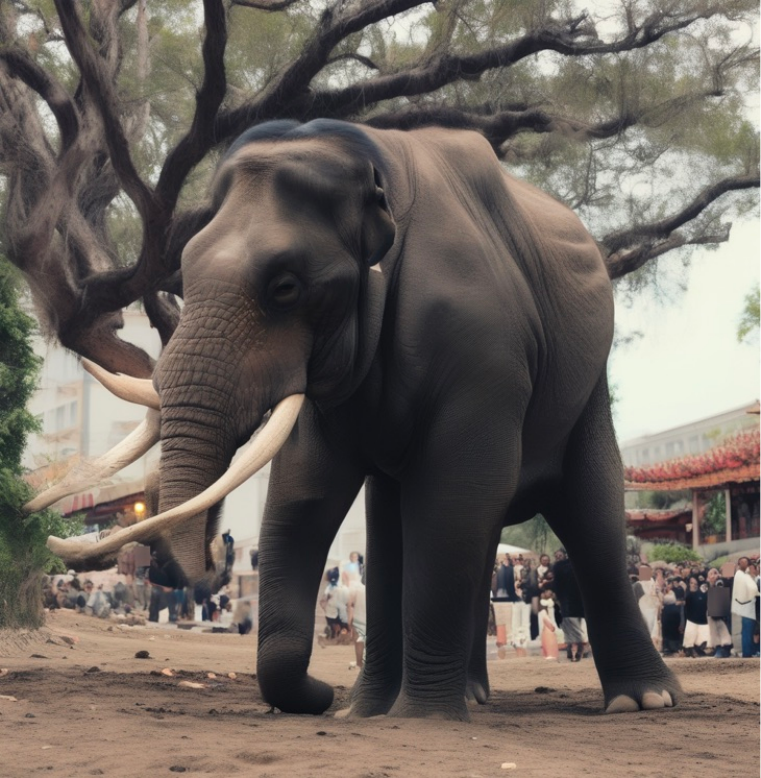}\\
        3 & \includegraphics[width=3cm]{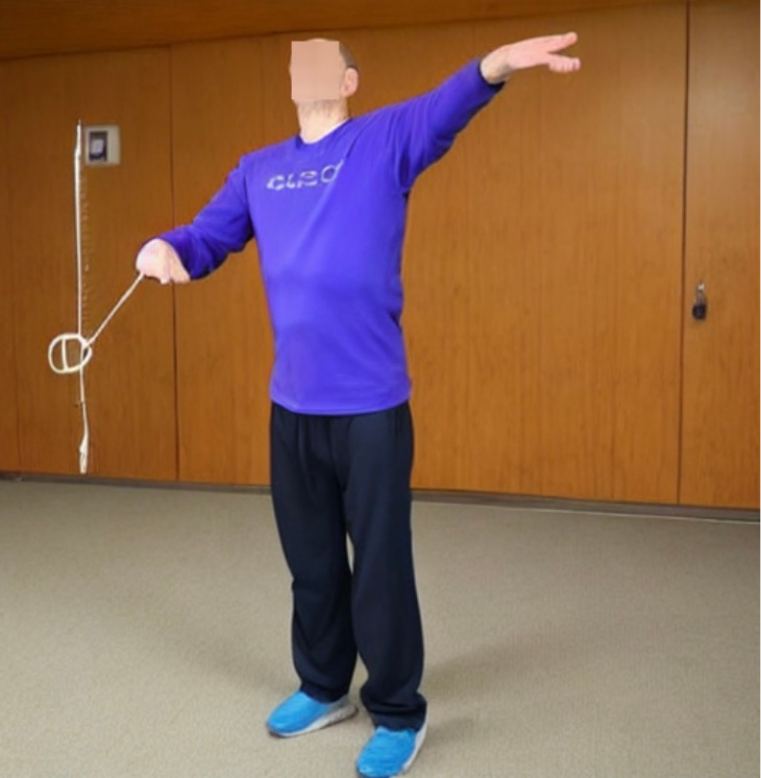} & 8 & \includegraphics[width=3cm]{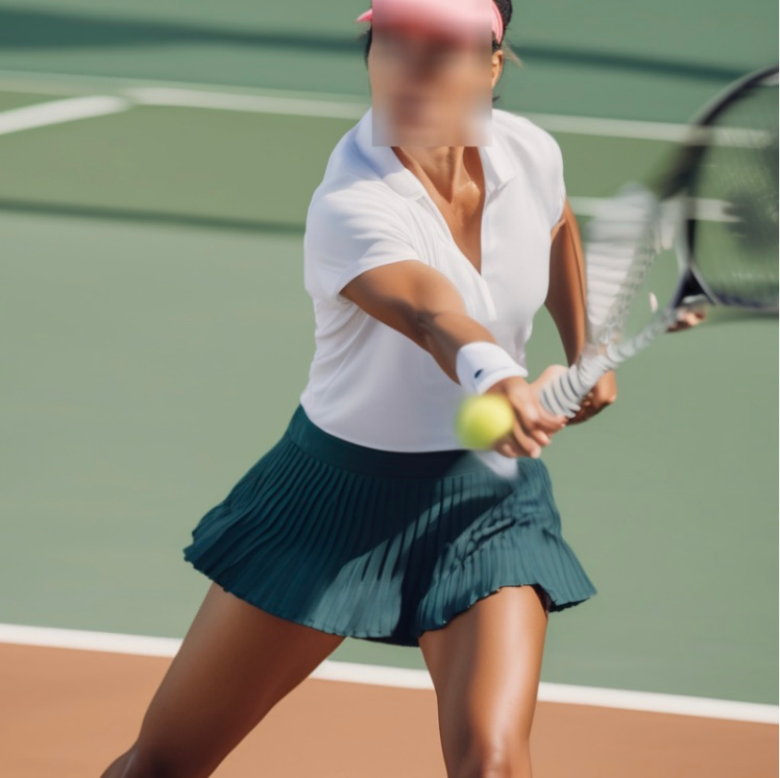} & \includegraphics[width=3cm]{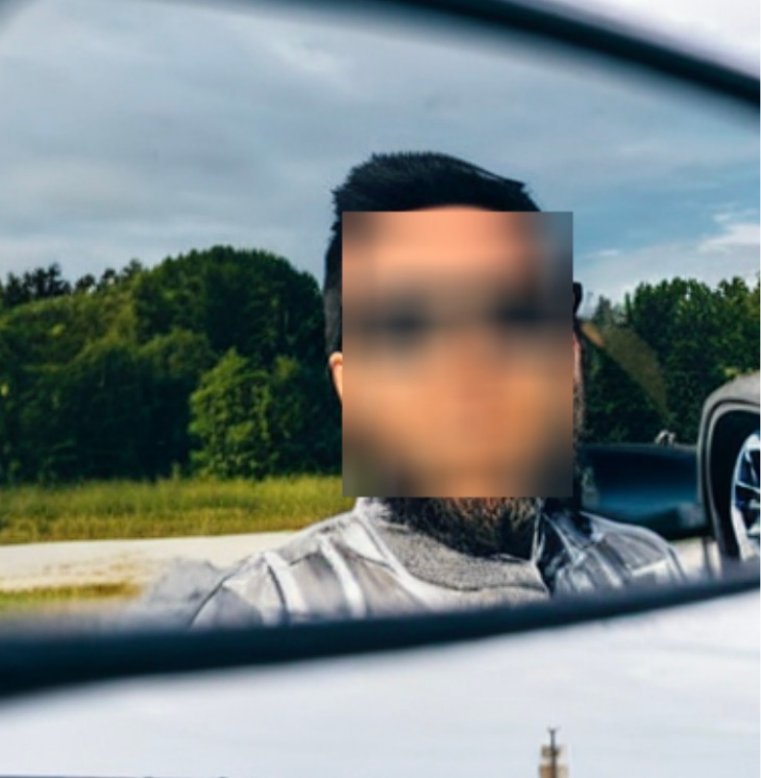}\\
        4 & \includegraphics[width=3cm]{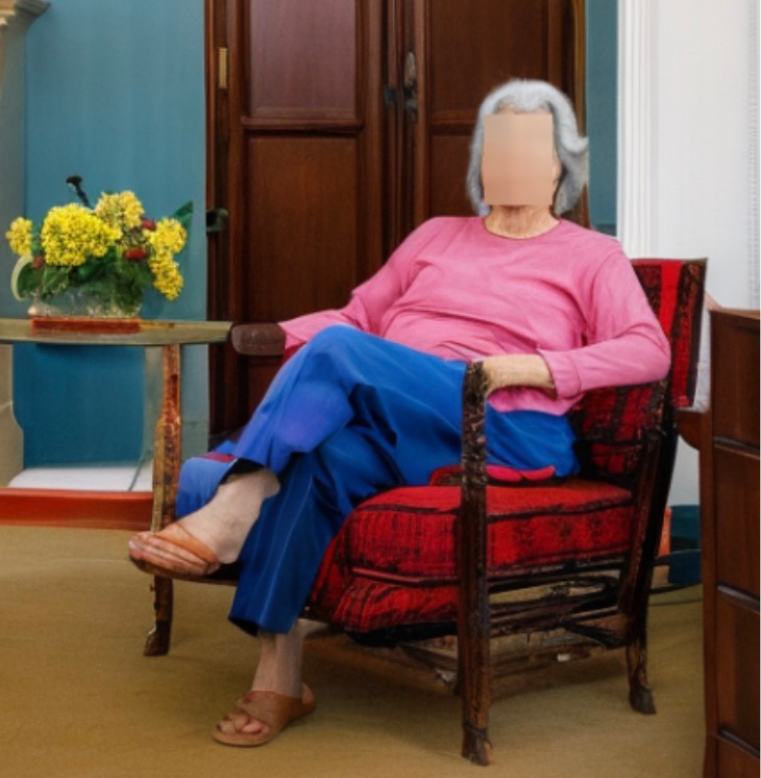} & 9 & \includegraphics[width=3cm]{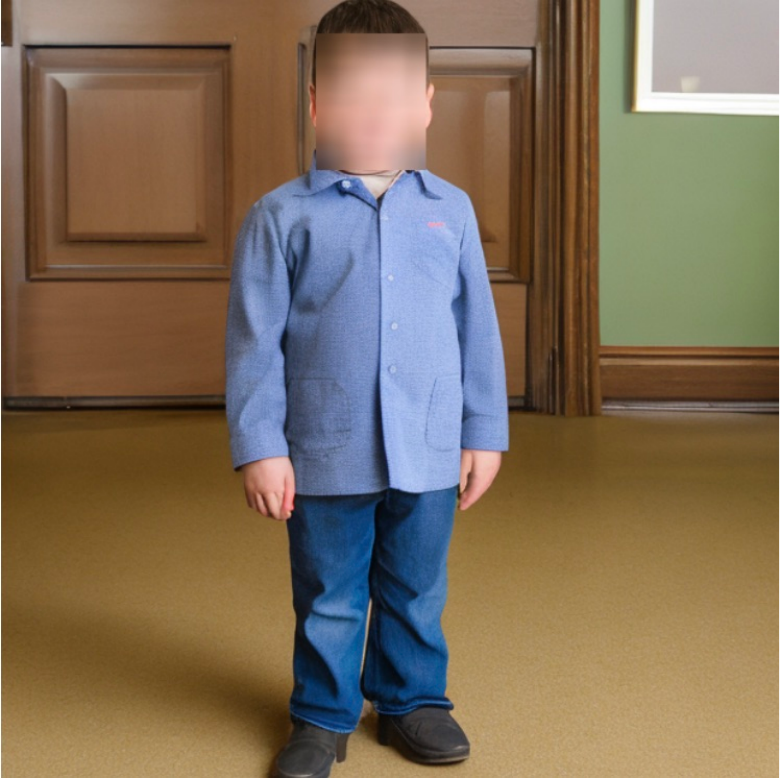} & \includegraphics[width=3cm]{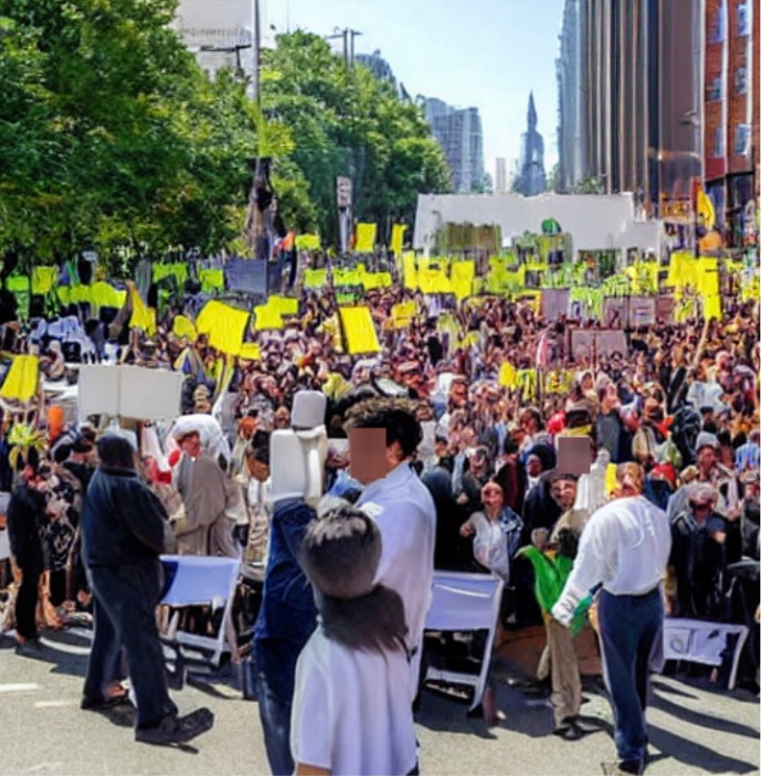}\\
        5 & \includegraphics[width=3cm]{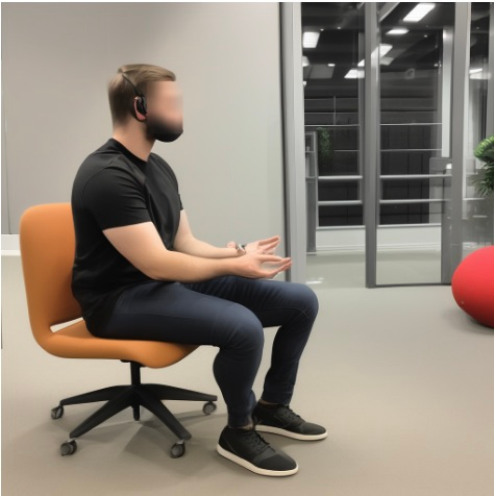} & 10 & \includegraphics[width=3cm]{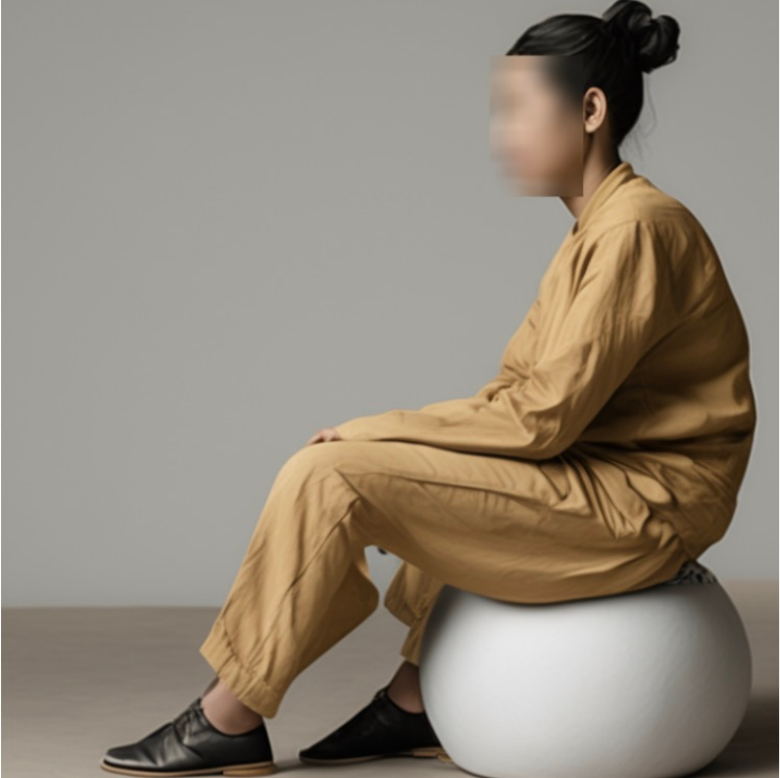} & \includegraphics[width=3cm]{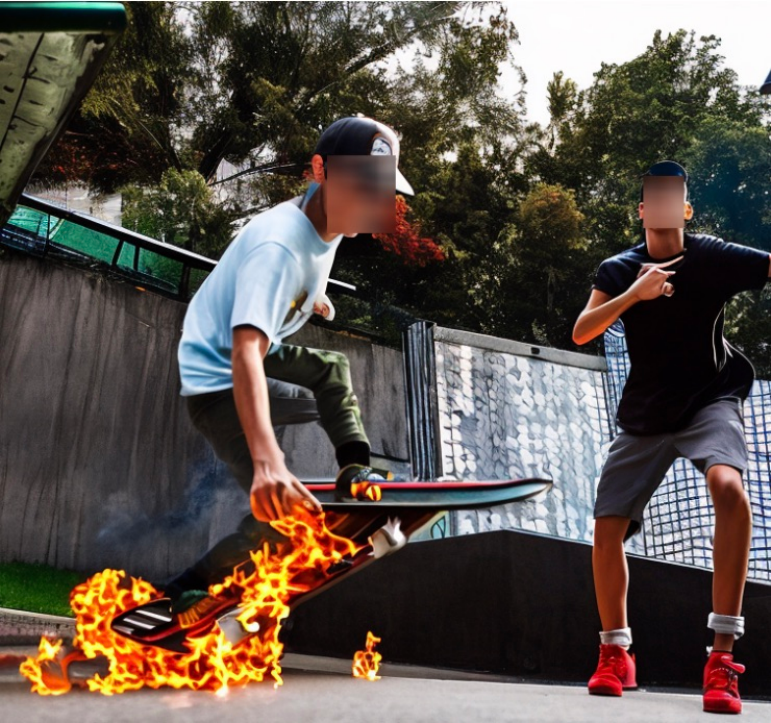} \\
    \end{tabular}
    \vspace{2mm}
    \captionsetup{type=figure}
    \caption{Annotation SOP - Representative images corresponding to body realism scores.}
    \label{tab:instruct}
\end{table}

\begin{table}[h]
\scalebox{0.8}{
\begin{minipage}{0.5\linewidth}
\begin{algorithm}[H]
\caption{Annotation consolidation}\label{alg:cap}
% \textbf{Input:} $r \in \mathbb{R}^{1 \times N}$\\
% \textbf{Output:} $s \in \mathbb{R}$ \\
\begin{algorithmic}
\State $\hat{r} = \left[ \text{ }\right]$
\State $r \gets Sorted(r)$
\If{max(r) >= 8}
\State $s \gets Mean(r[-2:])$
\Else
\State $m \gets Median(r)$
\State $L_{low}, L_{high} \gets (m \pm 0.5 * IQR(r))$ 
\For{$r_j$ in $r$}
\If{$ L_{low}< r_j < L_{high}$}
\State $\hat{r}$.append($r_j$)
\EndIf
\EndFor
\If{$len(\hat{r})>0$}
\State $s \gets Mean(\hat{r})$
\Else
\State $s \gets Mean(r)$
\EndIf
\EndIf
\end{algorithmic}
\end{algorithm}
\end{minipage}\hspace{4mm}
}
\begin{minipage}{0.5\linewidth}
\centering
\vspace{8mm}
\includegraphics[width=4.8cm]{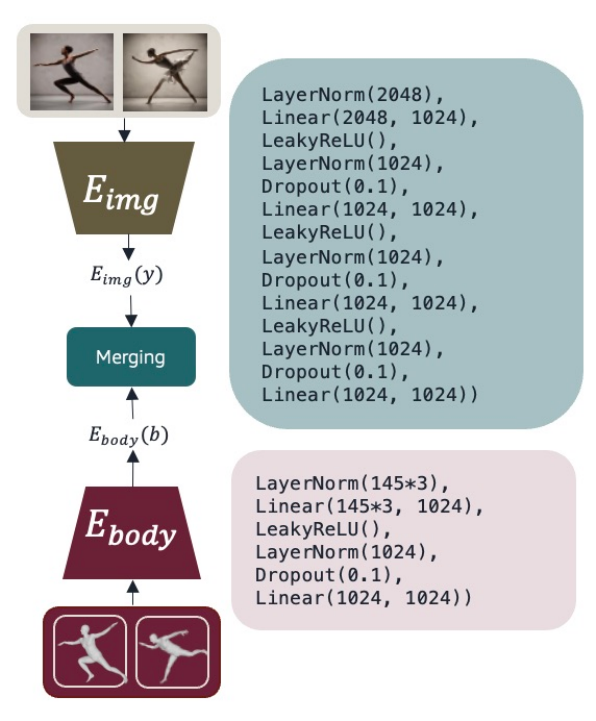}
% \vspace{10mm}
\captionsetup{type=figure}
\caption{Architectural details on Body Encoder and Merging module.}
\end{minipage}
\end{table}

\subsection{Correcting Artifacts in Generated Images}
In Sec.~\ref{tab:photoshop_images} we showcase the performance of several image quality score functions on pairs of generated and corrected images. We instruct experts to edit the given images so as to eliminate errors relating to the bodies. Errors within scope are: missing, extra, deformed limbs (arms, legs, hands, feet, fingers, toes). Faces, background, colours, image resolution are out of the scope of this effort.

\section{Comparison to State-of-the-Art}
Fig.~\ref{fig:body_vs_ir_hps} includes additional comparisons between BodyMetric and ImageReward~\cite{Xu:2023:ImageReward}. For a fair comparison we convert the logits obtained with ImageReward to probabilities using softmax. 
\begin{table}[h]
\begin{minipage}[b]{0.32\linewidth}
\begin{tabular}{c|cccccc}
 \multicolumn{1}{c}{}&\multicolumn{1}{p{1.5cm}}{\tiny{a man standing next to a stack of red luggage.}
}&\multicolumn{1}{p{1.5cm}}{\tiny{a person playing frisbee on a field in sport wear.}}
&\multicolumn{1}{p{1.5cm}}{\tiny{an Indian dancer performing hand gestures.}} 
&\multicolumn{1}{p{1.5cm}}{\tiny{a man standing on a hillside next to a lake holding frisbee.}} & \multicolumn{1}{p{1.5cm}}{ \tiny{a man holding a fuchsia umbrella.}} & \multicolumn{1}{p{1.5cm}}{\tiny{a woman running across a tennis court.}} \\
 \multicolumn{1}{c}{A} & \includegraphics[width=1.8cm]{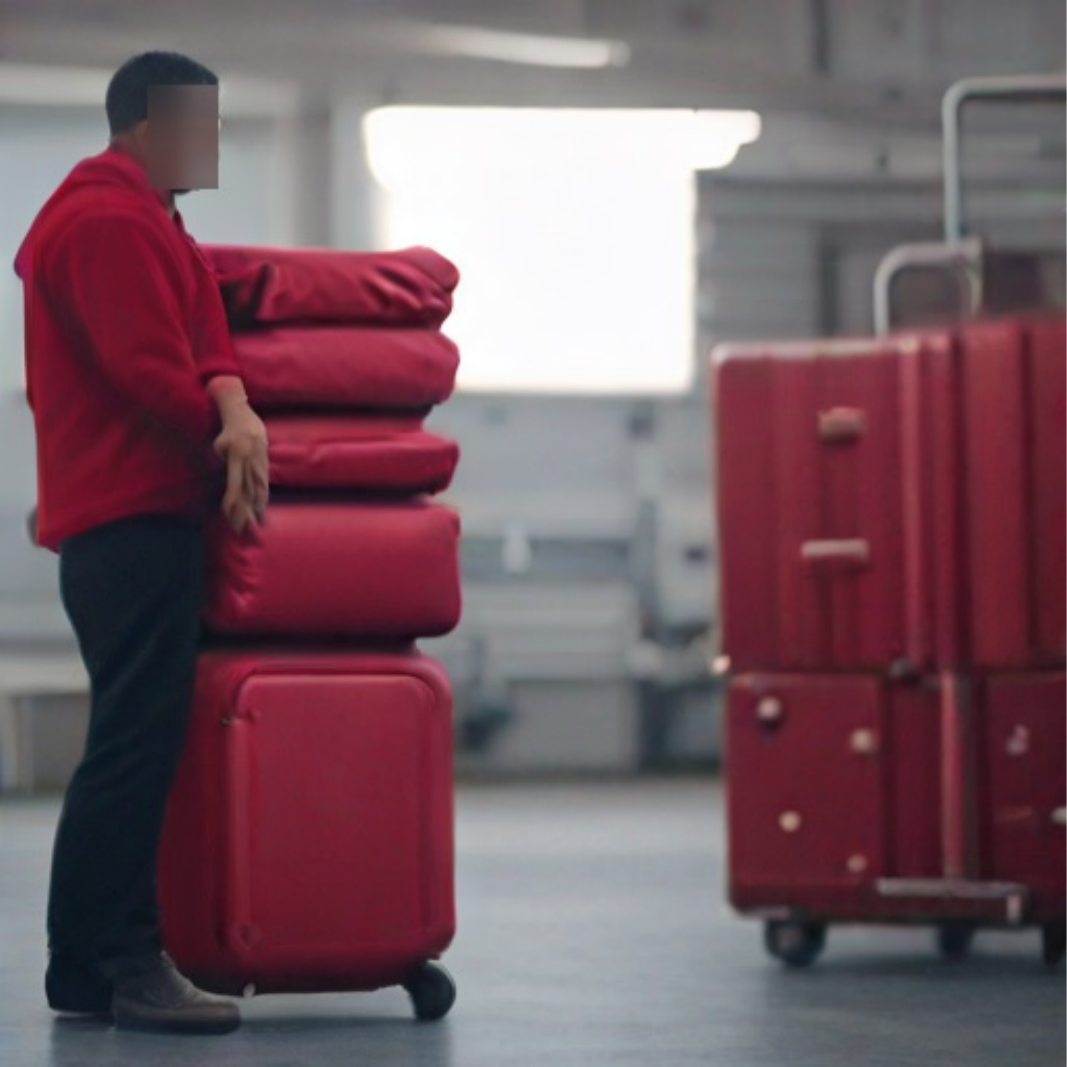} &  \includegraphics[width=1.8cm]{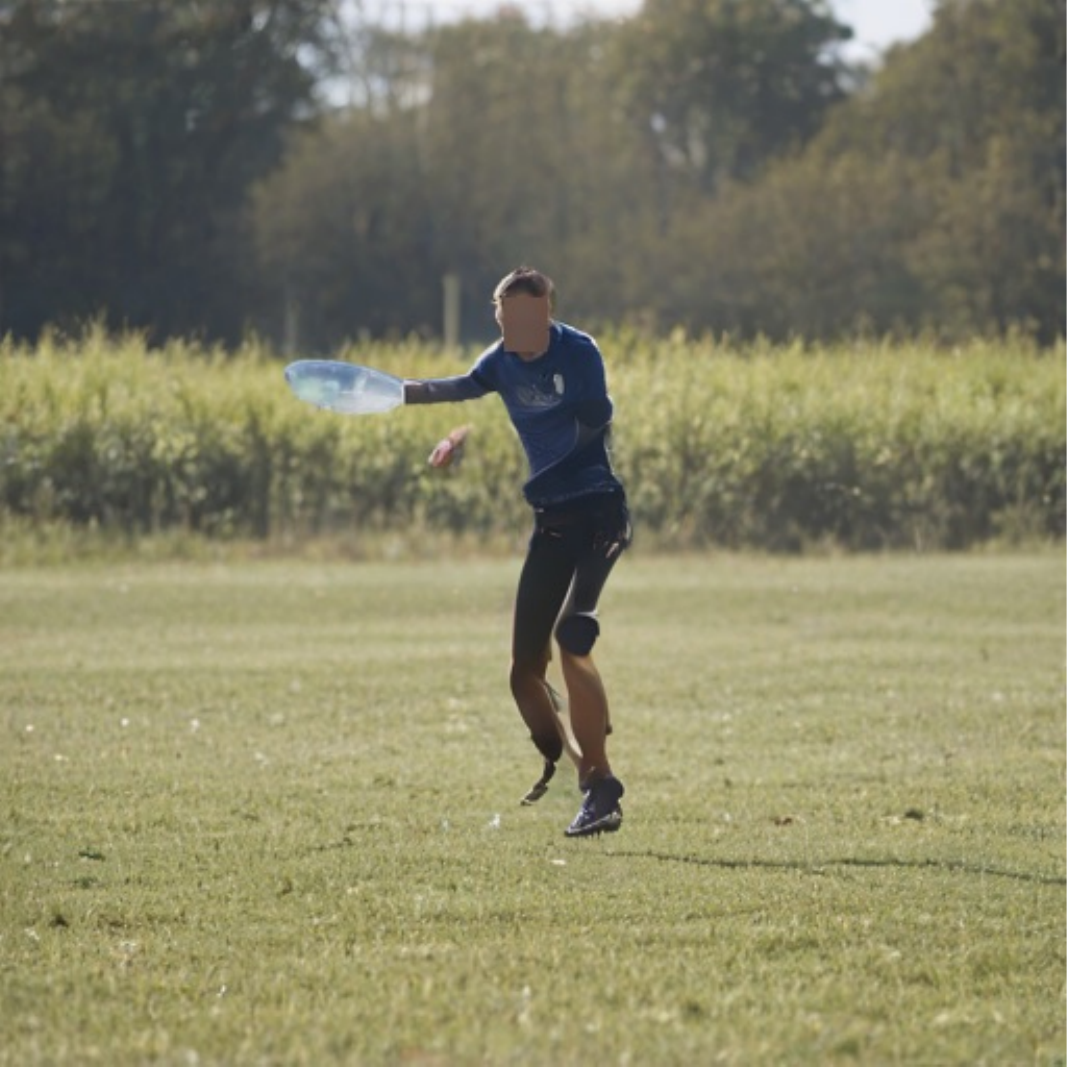} & \includegraphics[width=1.8cm]{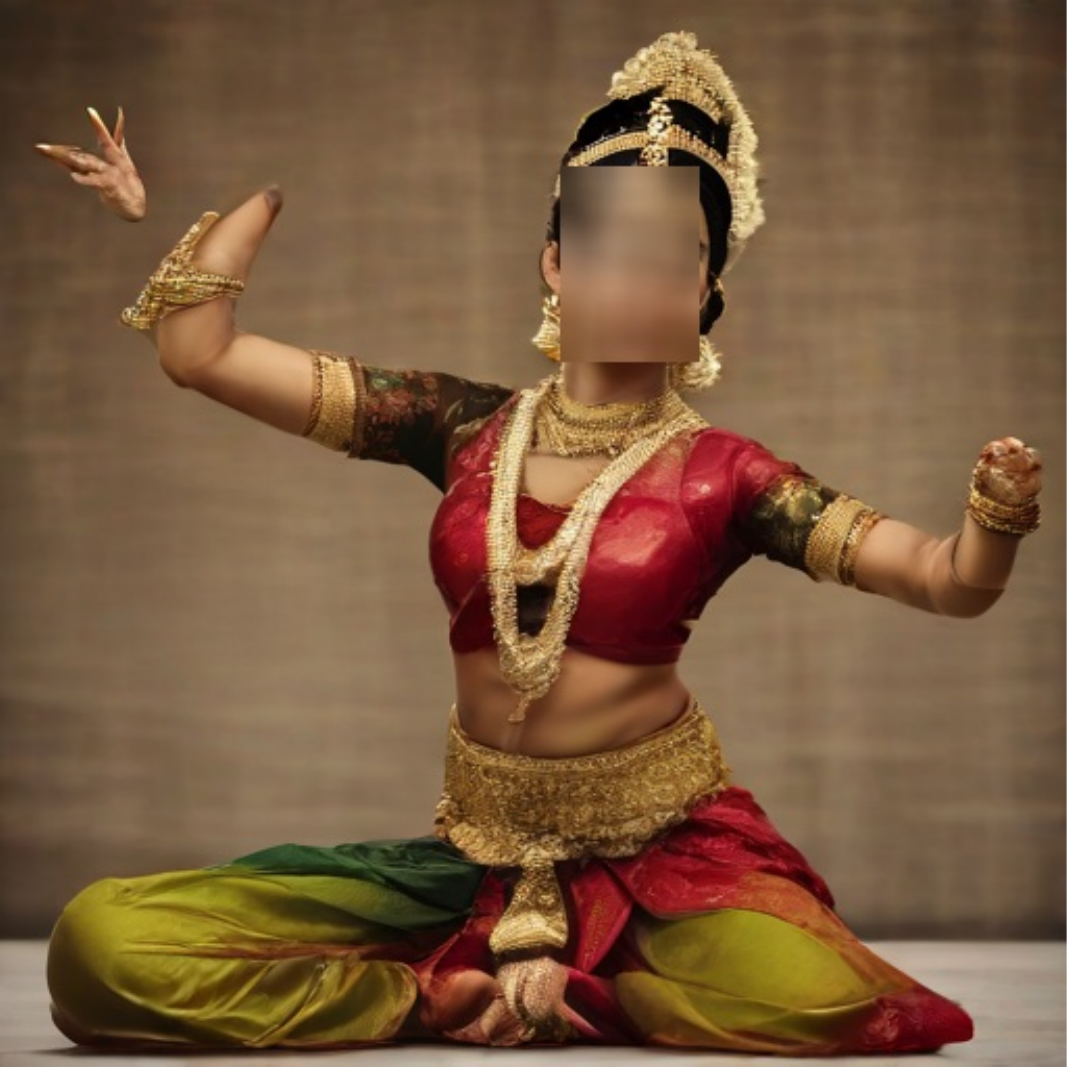} & \includegraphics[width=1.8cm]{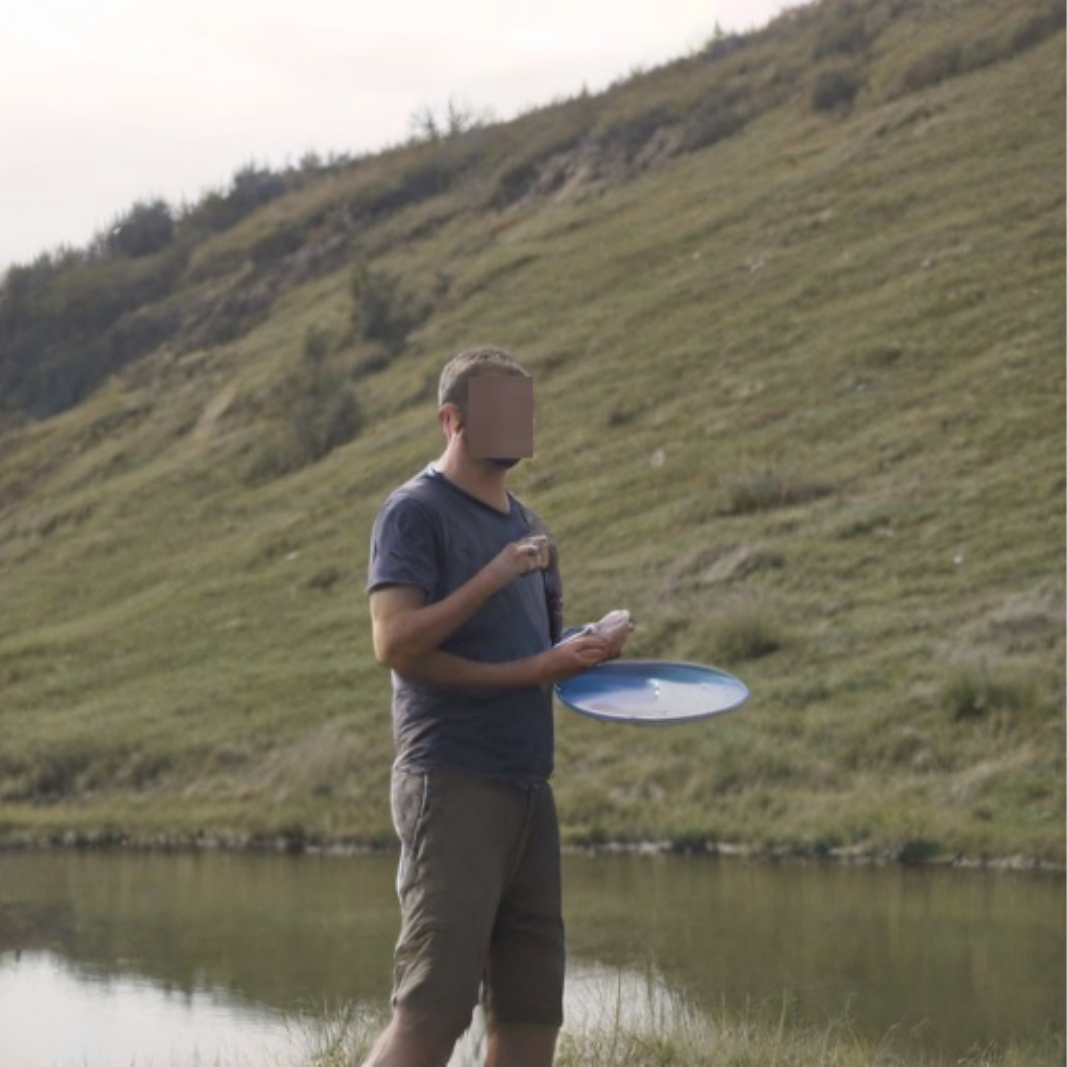} & \includegraphics[width=1.8cm]{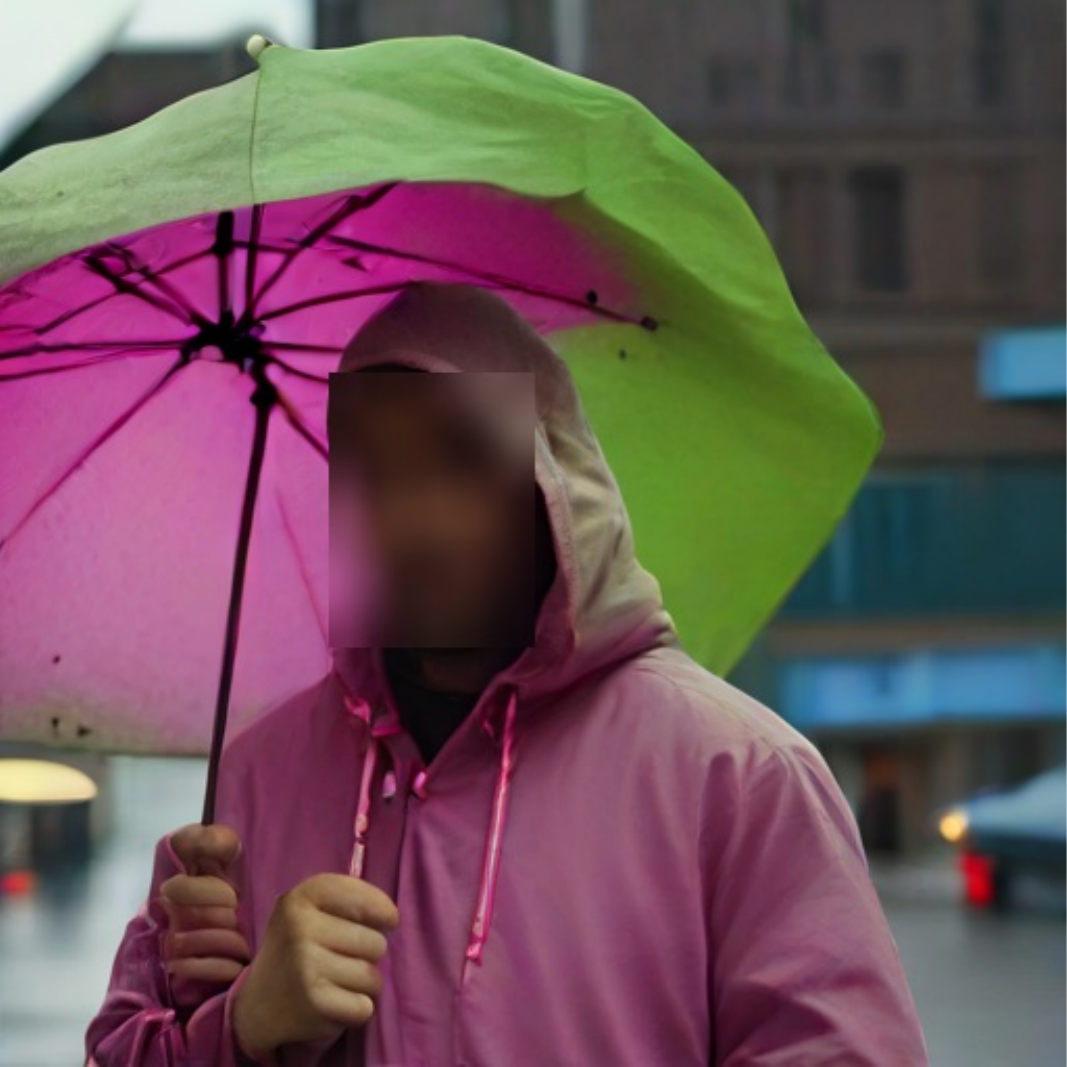} & \includegraphics[width=1.8cm]{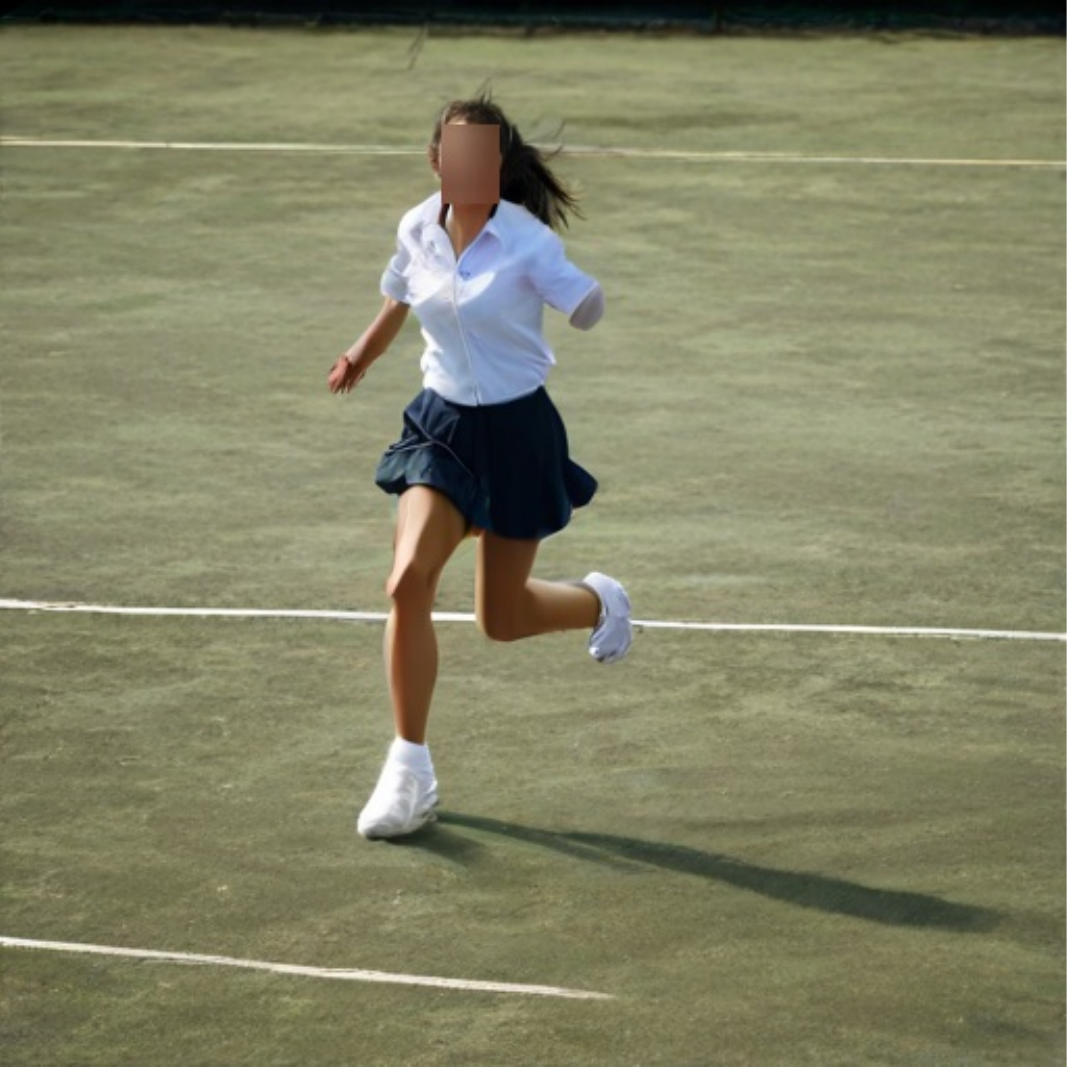}   \\
   \multicolumn{1}{c}{B} & \includegraphics[width=1.8cm]{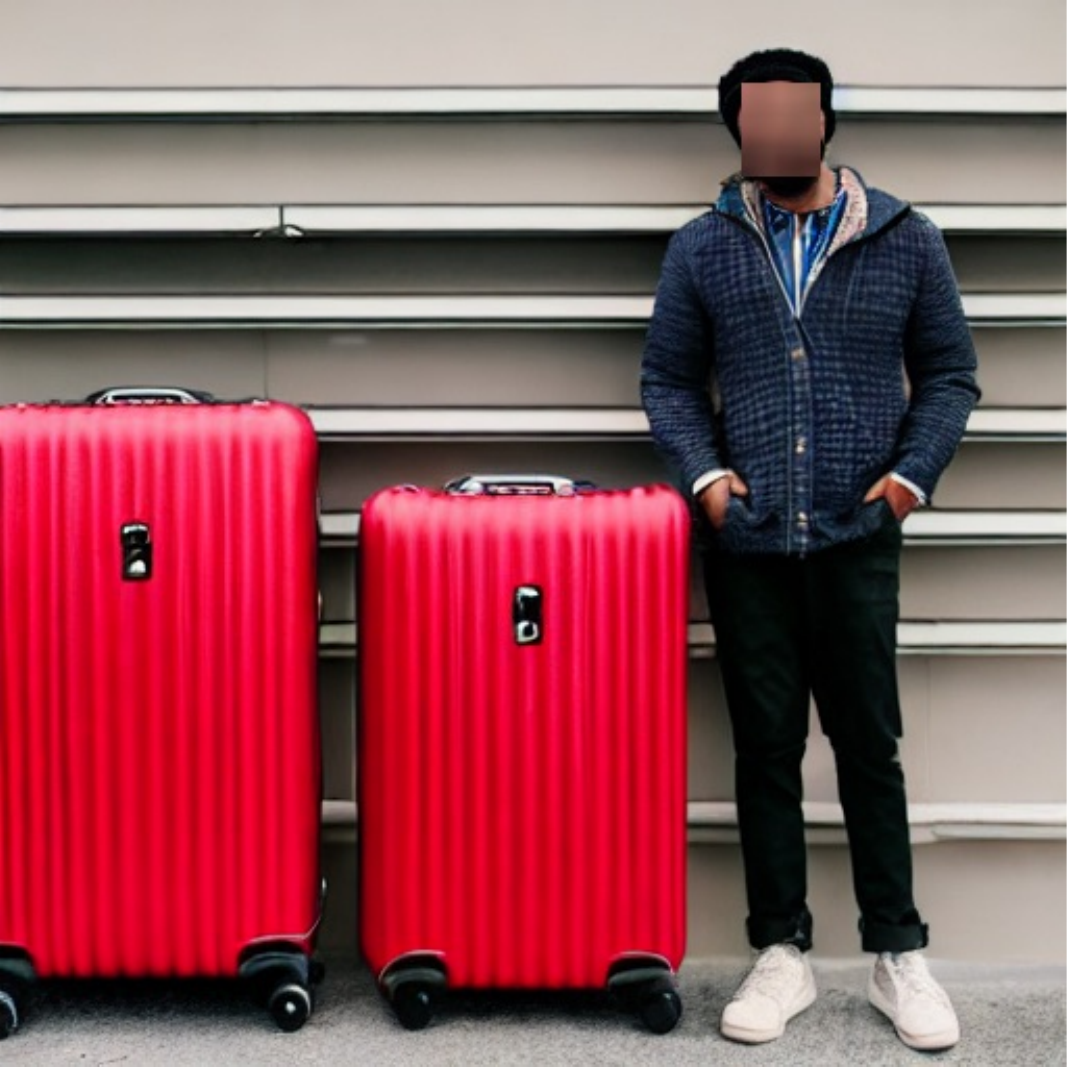}& \includegraphics[width=1.8cm]{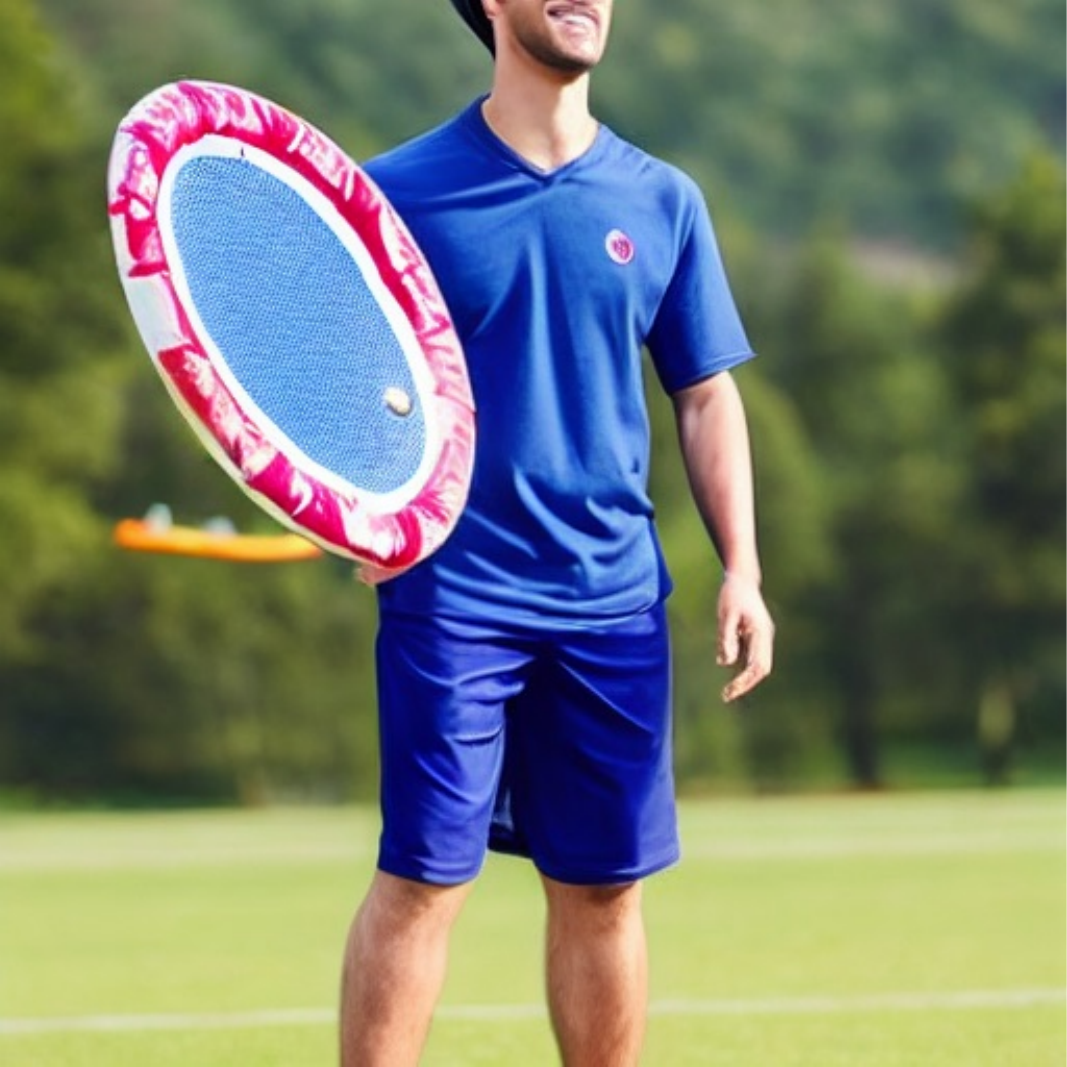} & \includegraphics[width=1.8cm]{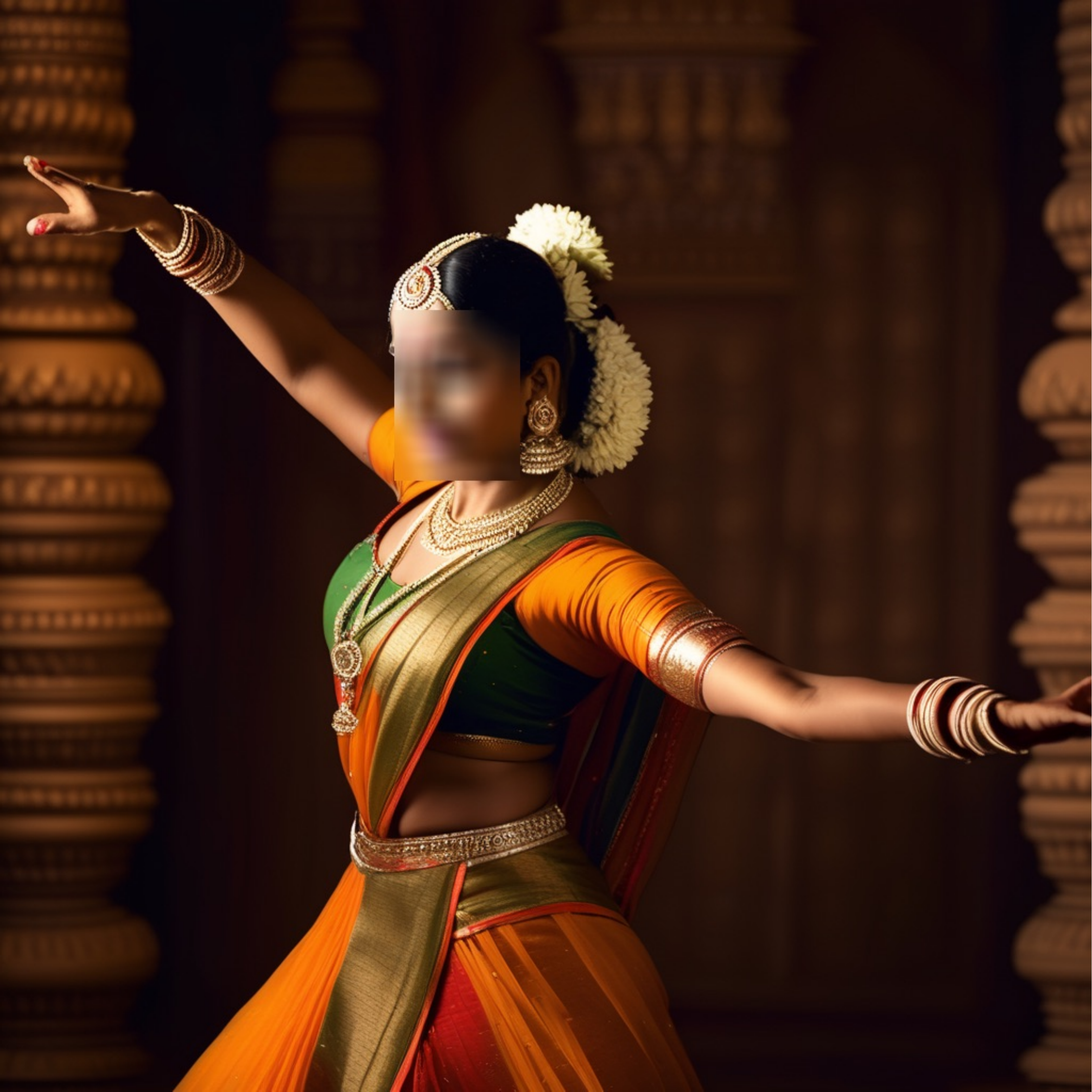} & \includegraphics[width=1.8cm, height=1.8cm]{bodymetric/figures/ir_hps_body/eg7/image1.pdf} & \includegraphics[width=1.8cm]{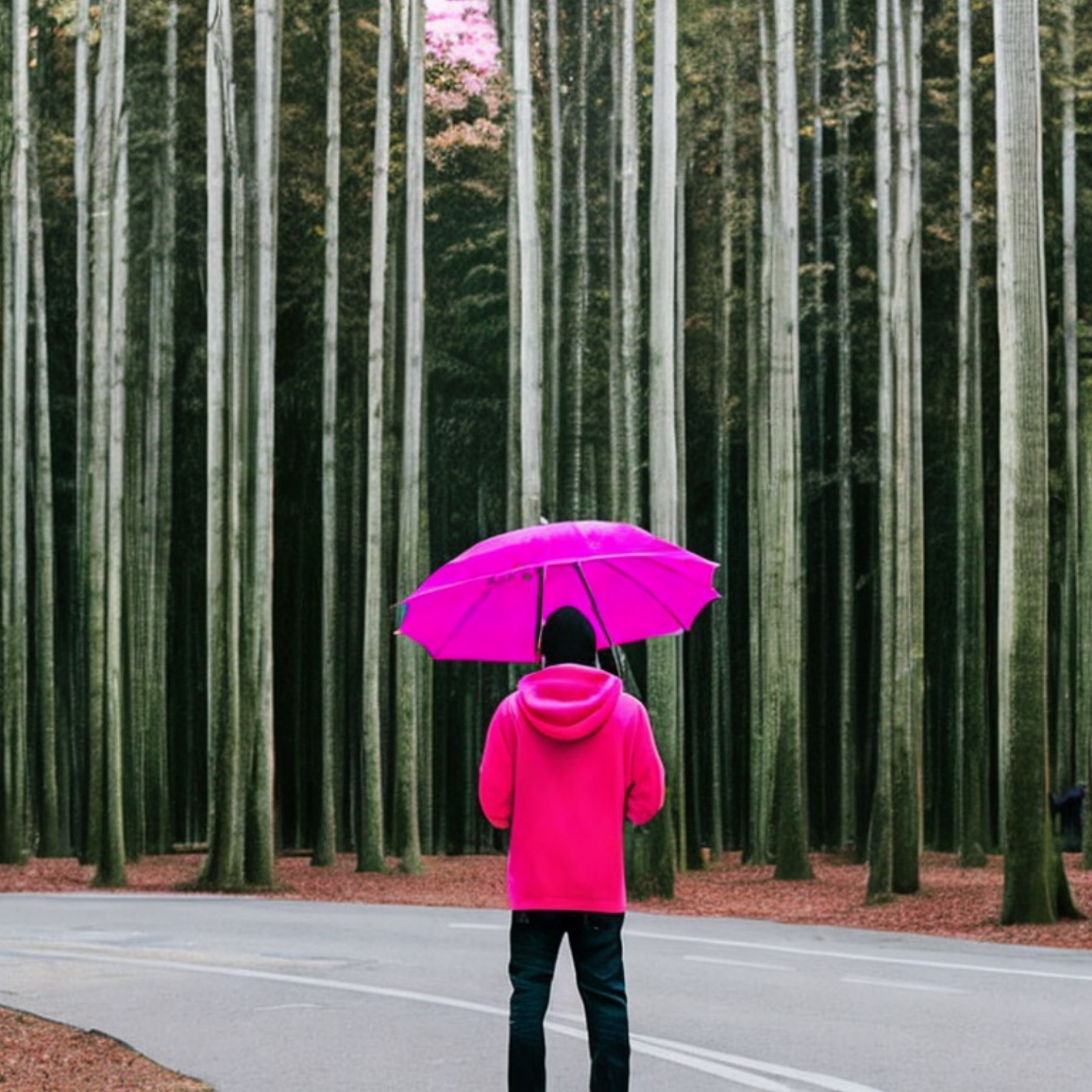} & \includegraphics[width=1.8cm]{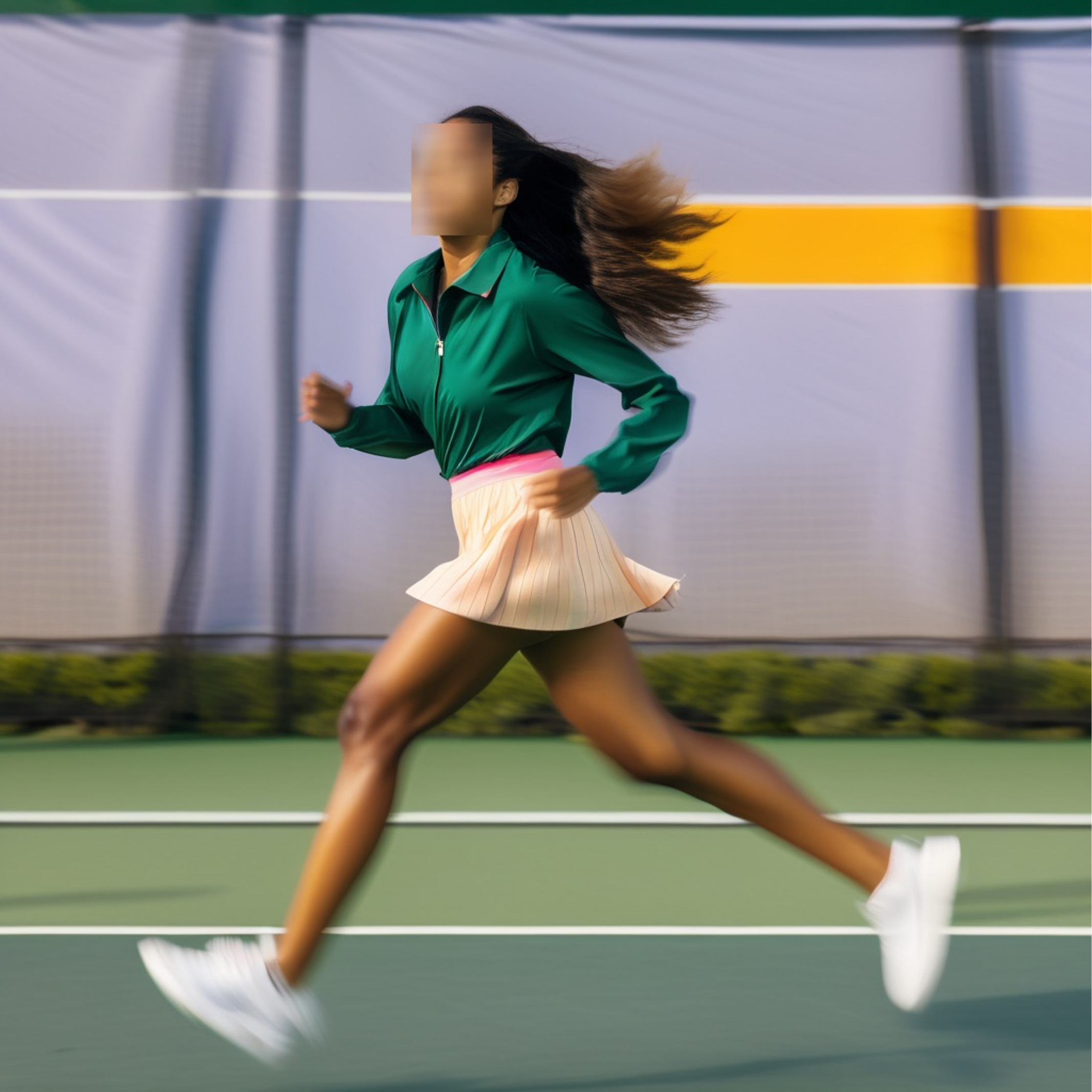}   \\
  \multicolumn{7}{c}{{ImageReward~\cite{Xu:2023:ImageReward}}}  \\
  \hline
  \hline
  A & \textcolor{brown}{0.60} & \textcolor{brown}{0.74} & \textcolor{brown}{0.56} & \textcolor{brown}{0.56} & \textcolor{brown}{0.80} & \textcolor{brown}{0.54}\\
  B & \textcolor{brown}{0.40} & \textcolor{brown}{0.26} & \textcolor{brown}{0.44} & \textcolor{brown}{0.44} & \textcolor{brown}{0.20} & \textcolor{brown}{0.46} \\
  % \multicolumn{7}{c}{{HPSv2}}  \\
  % \hline
  % \hline
  %   A &  \textcolor{brown}{0.50} &\textcolor{brown}{0.50} &\textcolor{brown}{0.50} &\textcolor{brown}{0.50} &\textcolor{brown}{0.50} &  \textcolor{brown}{0.50} \\
  %   B & \textcolor{brown}{0.50} & \textcolor{brown}{0.50} & \textcolor{brown}{0.50} & \textcolor{brown}{0.50} & \textcolor{brown}{0.50} & \textcolor{brown}{0.50} \\
  \multicolumn{7}{c}{{BodyMetric}}  \\
  \hline
  \hline
  A & \textcolor{teal}{0.05} & \textcolor{teal}{0.36} & \textcolor{teal}{0.02} & \textcolor{teal}{0.01} & \textcolor{teal}{0.08} & \textcolor{teal}{0.01} \\
  B & \textcolor{teal}{0.95} & \textcolor{teal}{0.64} & \textcolor{teal}{0.98} & \textcolor{teal}{0.99} & \textcolor{teal}{0.92} & \textcolor{teal}{0.99} \\
\end{tabular}
\end{minipage}
\captionsetup{type=figure}
\caption{Pair-wise image preference using ImageReward~\cite{Xu:2023:ImageReward} and BodyMetric. Images in row A are annotated by human experts as less realistic than B. We display the scores per pair for each model and highlight the \textbf{\textcolor{teal}{correct}} and \textbf{\textcolor{brown}{incorrect}} predictions.}
\label{fig:body_vs_ir_hps}
\end{table}

\section{Qualitative Results}
Fig.~\ref{tab:success2} showcases pair-wise preference prediction with BodyMetric. Among pairs generated using the same prompt, BodyMetric successfully selects the image with a more realistic body.

\section{Benchmark}
In Fig.~\ref{fig:bench1},~\ref{fig:bench2} we demonstrate images representative of the quality of bodies generated by State-of-the-Art text-to-image models such as SD-1.4, SD-2.1~\cite{Roombach:2022} SD-XL~\cite{Podel:2023:SDXL}, SD-XL-Turbo~\cite{Sauer:2023:SDXLTurbo}, Wuerstchen~\cite{pernias2023wuerstchen}. We observe that the perceived quality in terms of body realism aligns with our findings on T2I benchmarking reported in Tab.~\ref{tab:BodyRealism_benchmark}.

\begin{table}[h]
    \centering
    \begin{tabular}{ccccc}
        \multicolumn{2}{p{5cm}}{a man standing on a hillside next to a lake holding frisbee.} & &  \multicolumn{2}{p{5cm}}{a young man holding a white ball while running through a field.} \\
    \includegraphics[width=3cm]{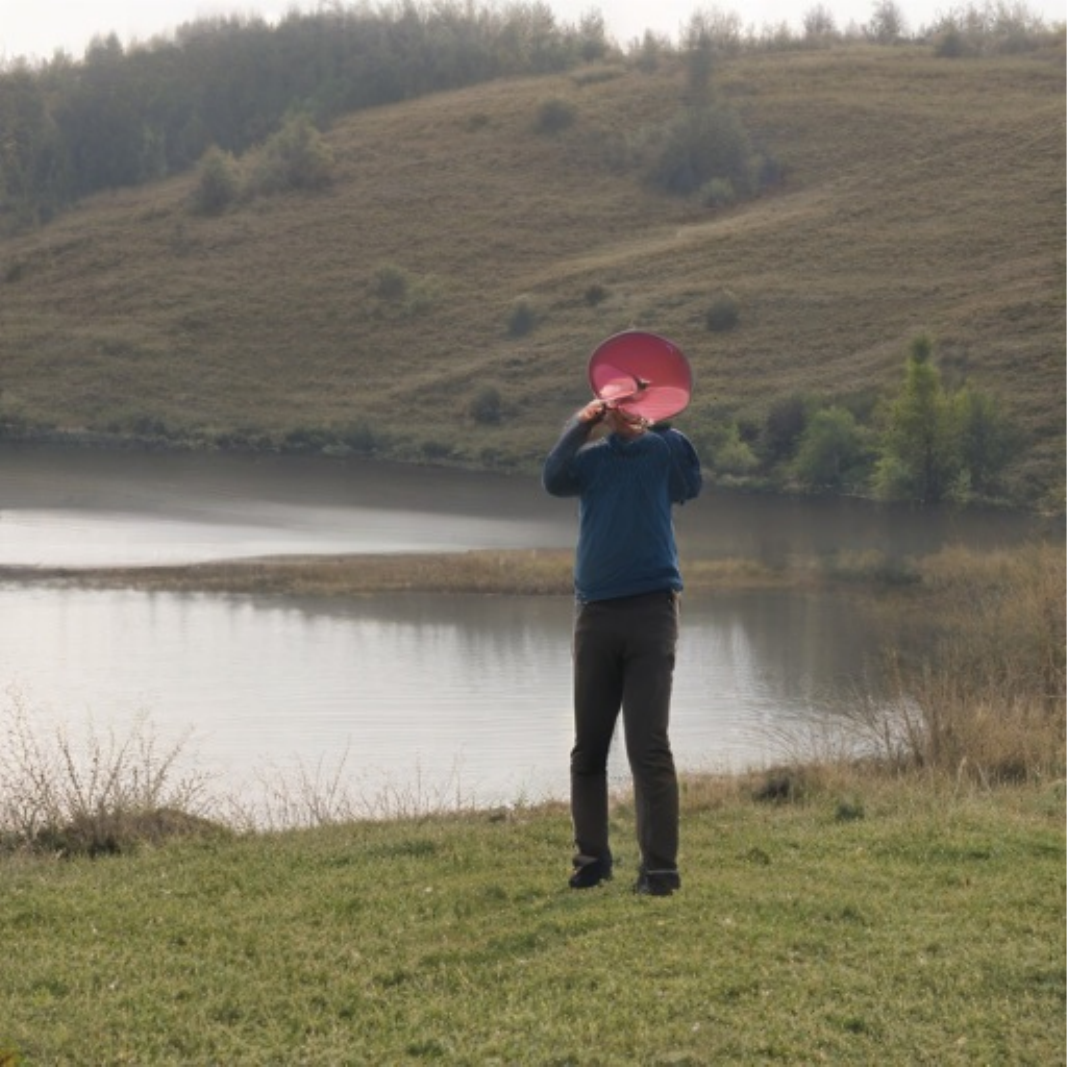}& \includegraphics[width=3cm, cfbox=teal 1pt 1pt]{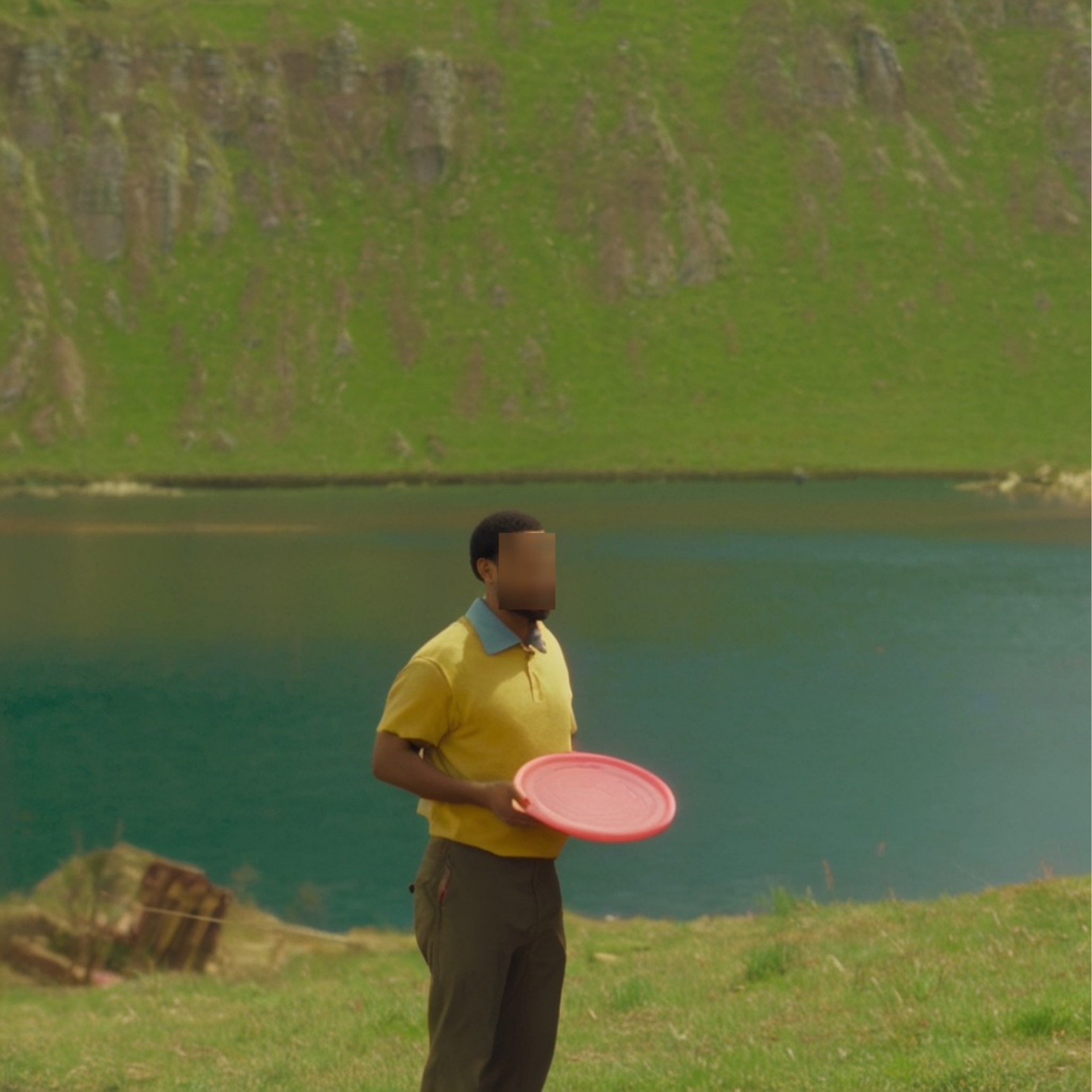}& &\includegraphics[width=3cm]{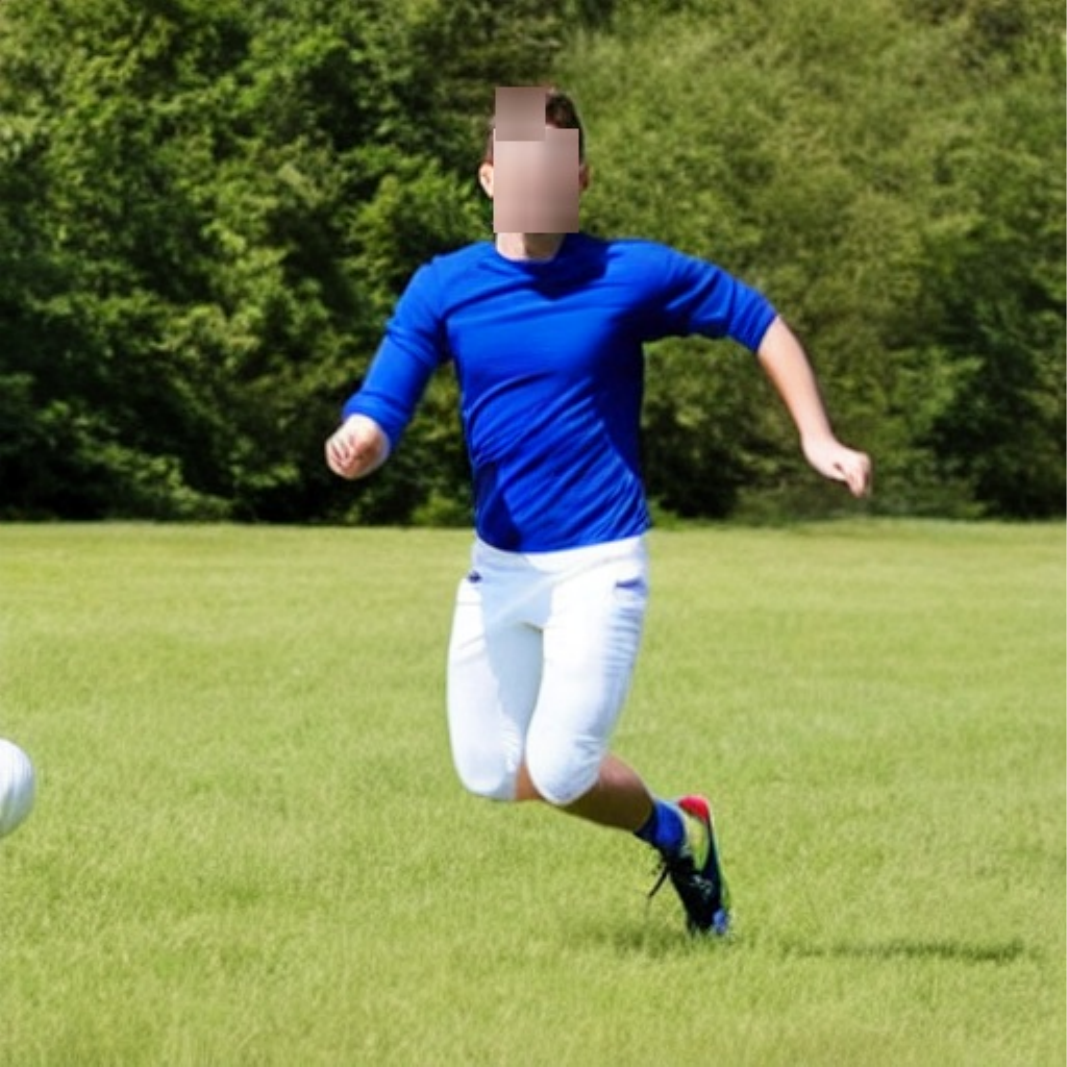}&\includegraphics[width=3cm, height=3cm,  cfbox=teal 1pt 1pt]{bodymetric/figures/success_cases/eg6/image1.pdf}\\
     \multicolumn{2}{p{5cm}}{a person in rubber boots and a rain coat seated on a bench.}& & \multicolumn{2}{p{5cm}}{a woman standing next to a yellow fire hydrant.}  \\
      \includegraphics[width=3cm]{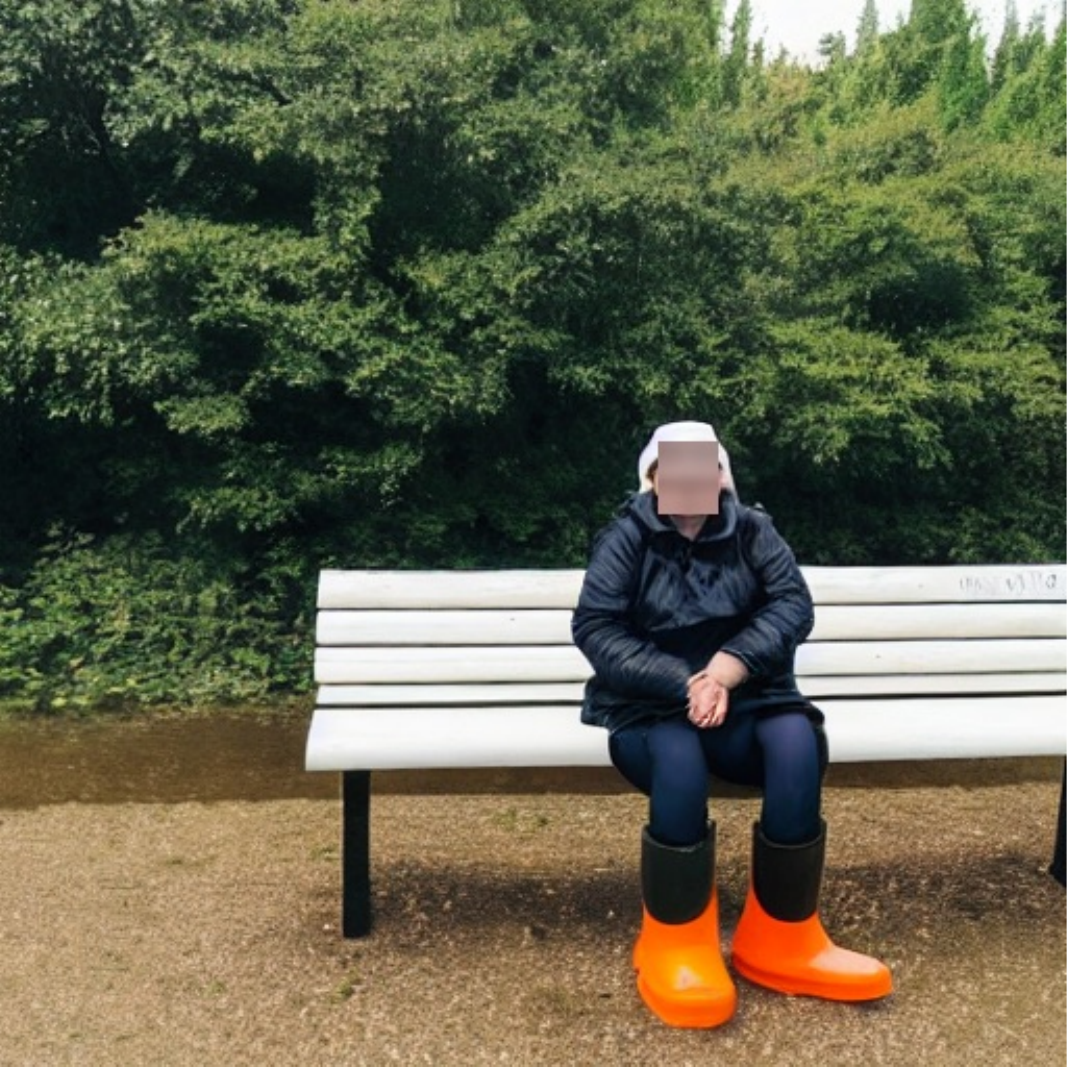}& \includegraphics[width=3cm, cfbox=teal 1pt 1pt]{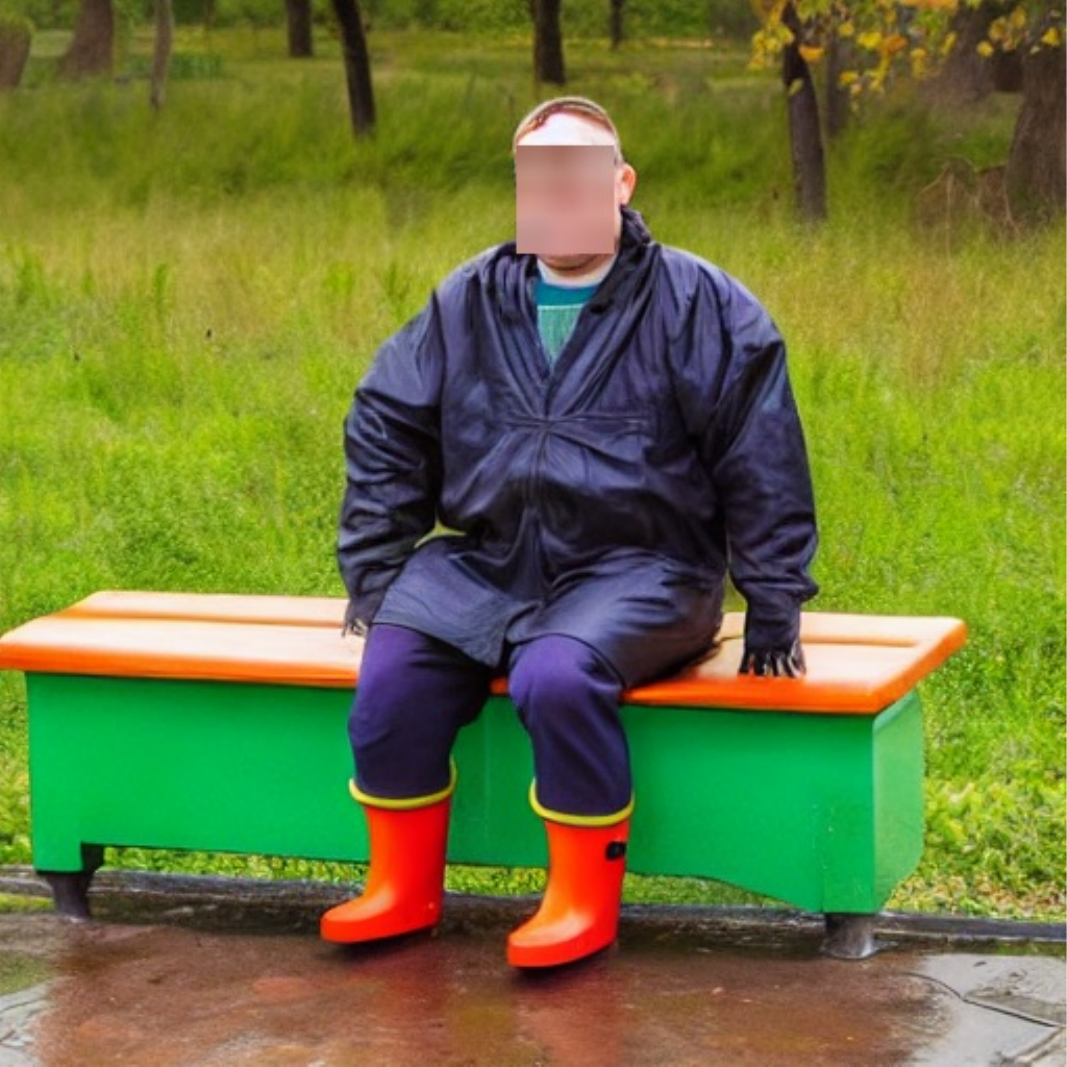}& &\includegraphics[width=3cm]{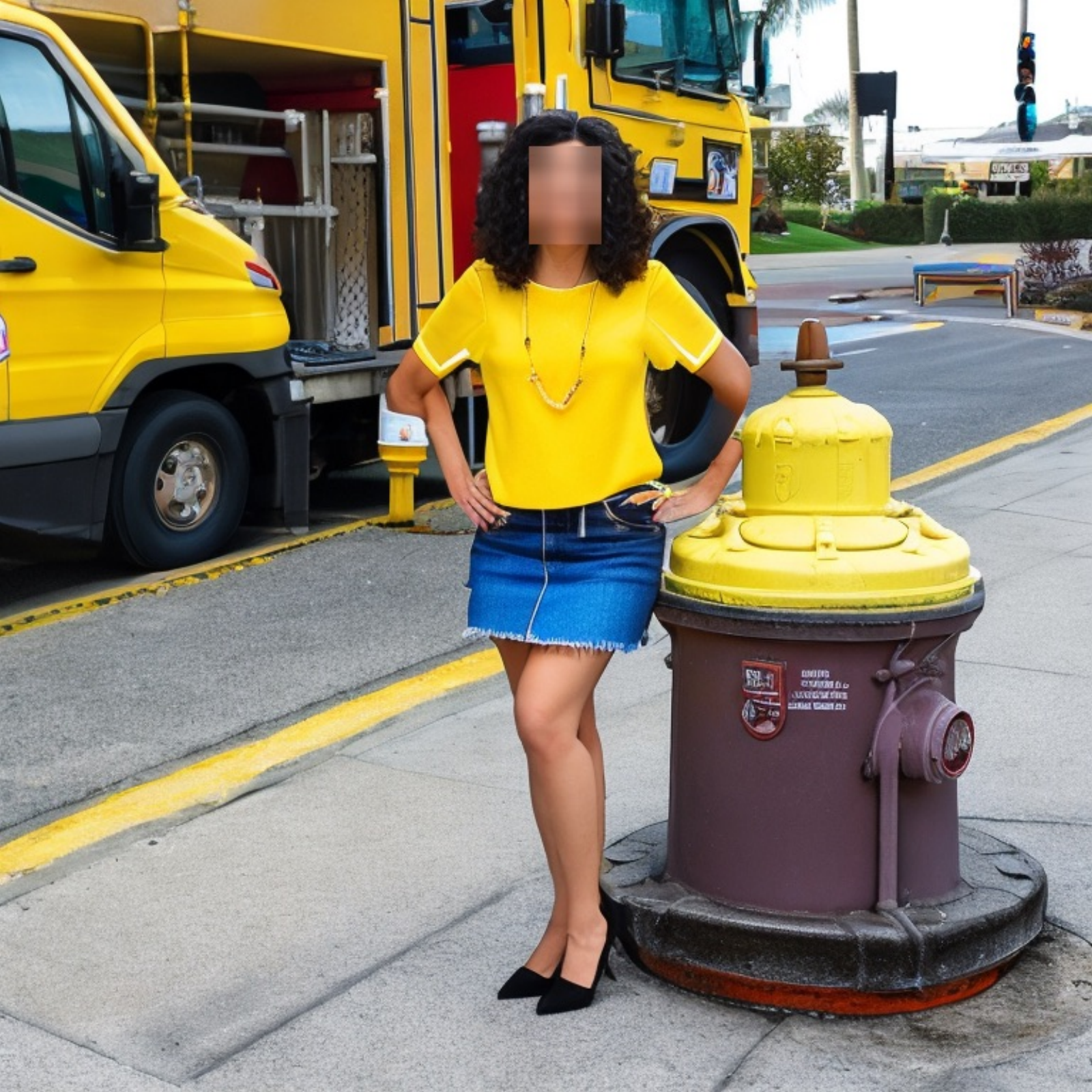}&\includegraphics[width=3cm, height=3cm,  cfbox=teal 1pt 1pt]{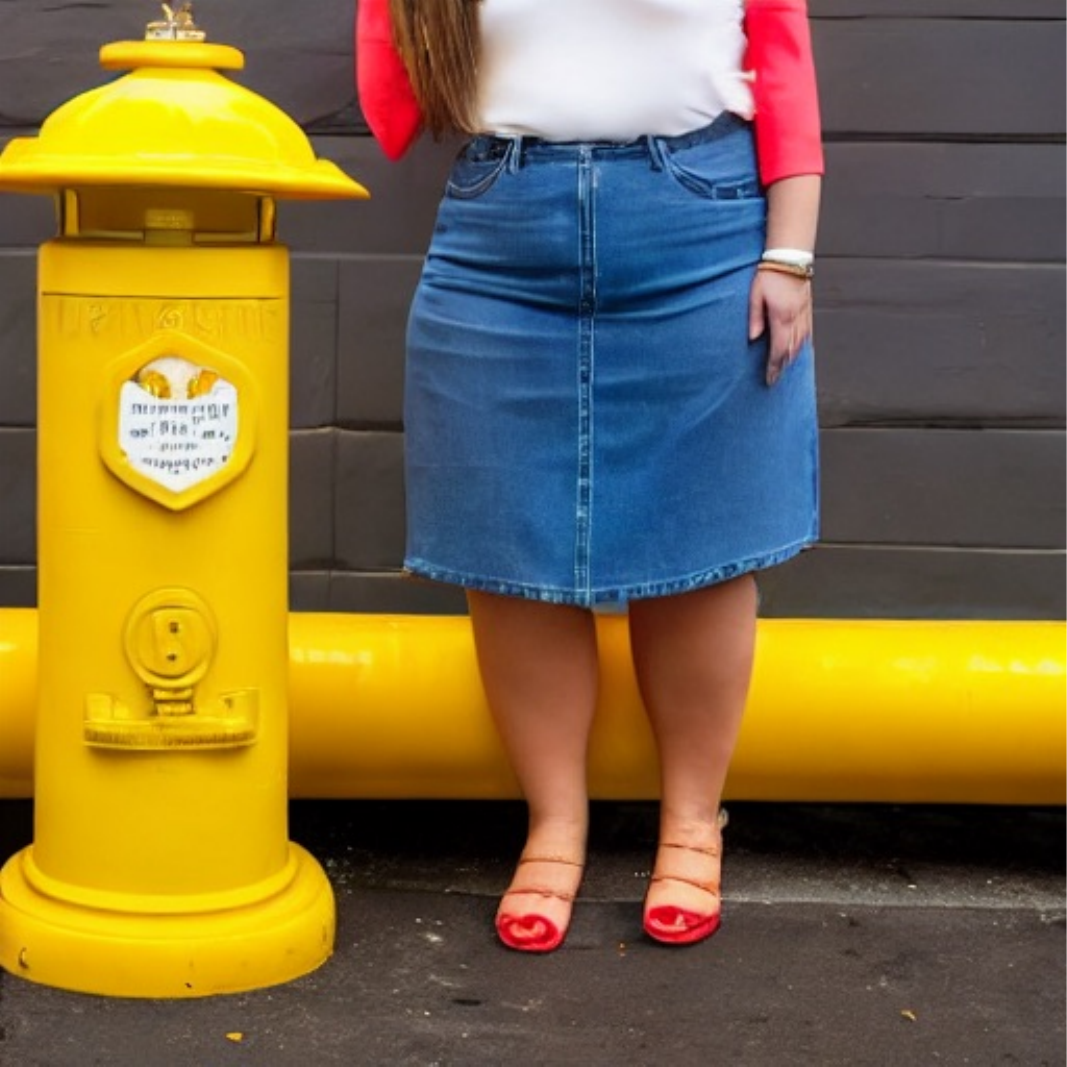}\\
     \multicolumn{2}{p{5cm}}{a man in a shirt and tie motioning with his hand.}& & \multicolumn{2}{p{5cm}}{there is a man standing next to the kitchen counter.}  \\  
    \includegraphics[width=3cm]{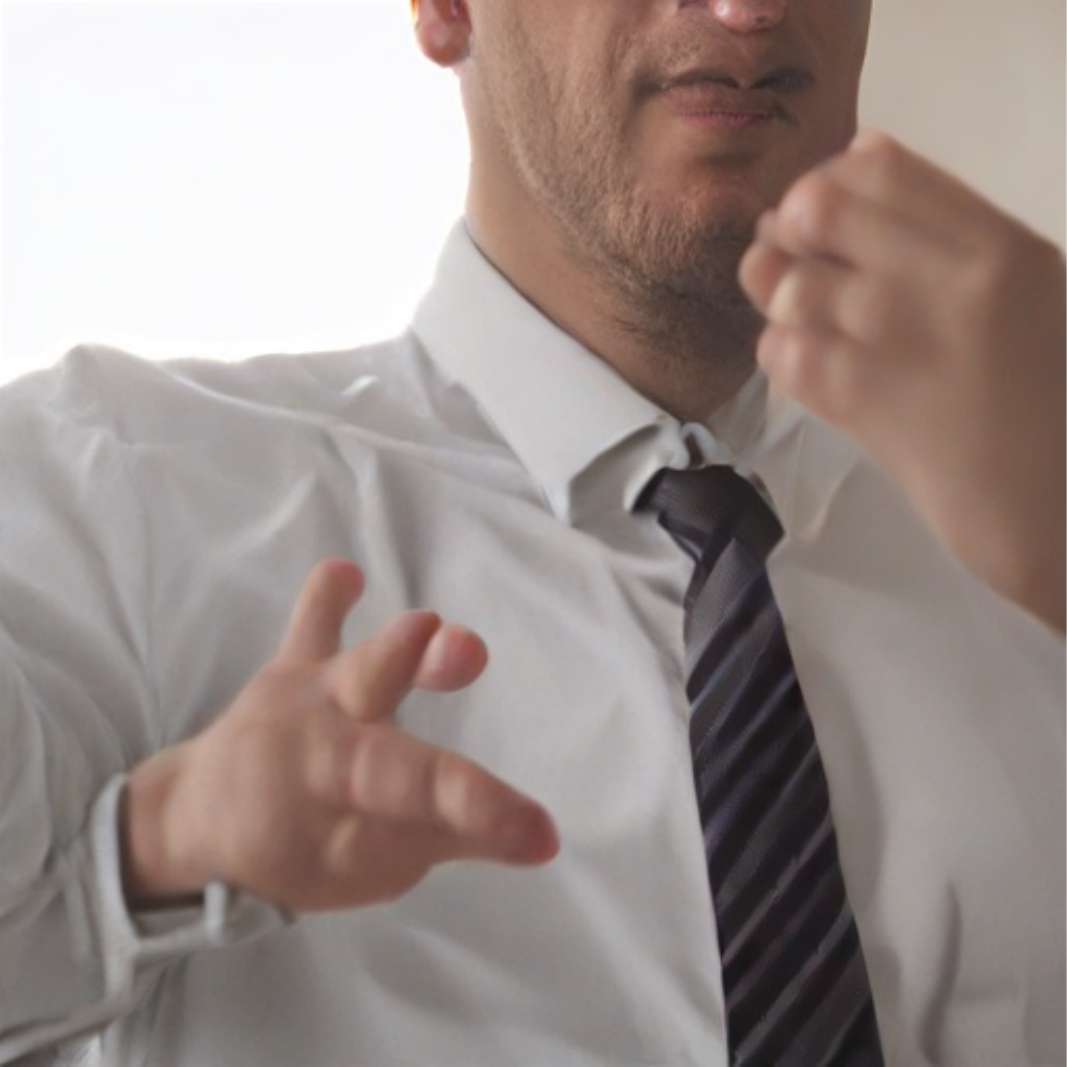}& \includegraphics[width=3cm, cfbox=teal 1pt 1pt]{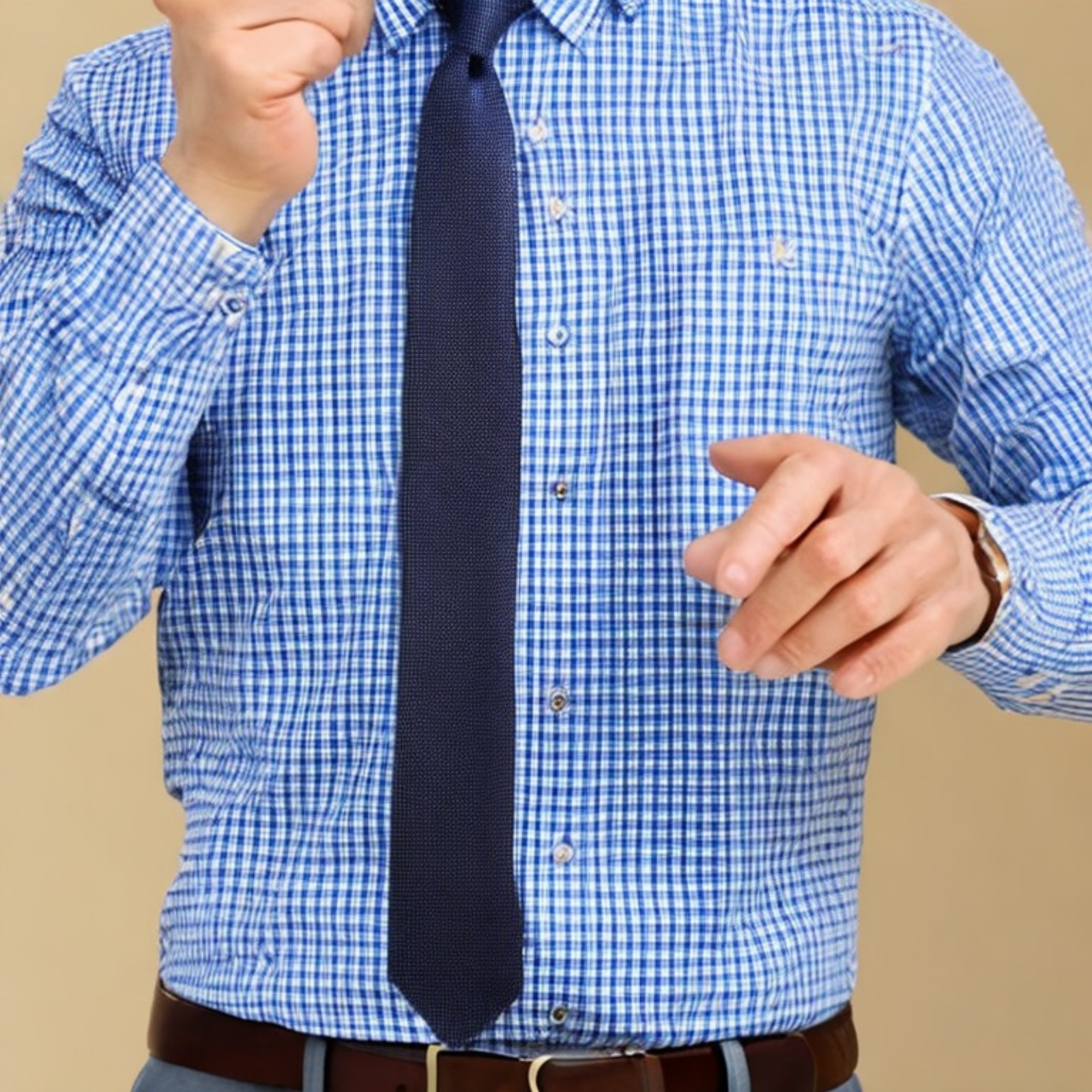}& &\includegraphics[width=3cm]{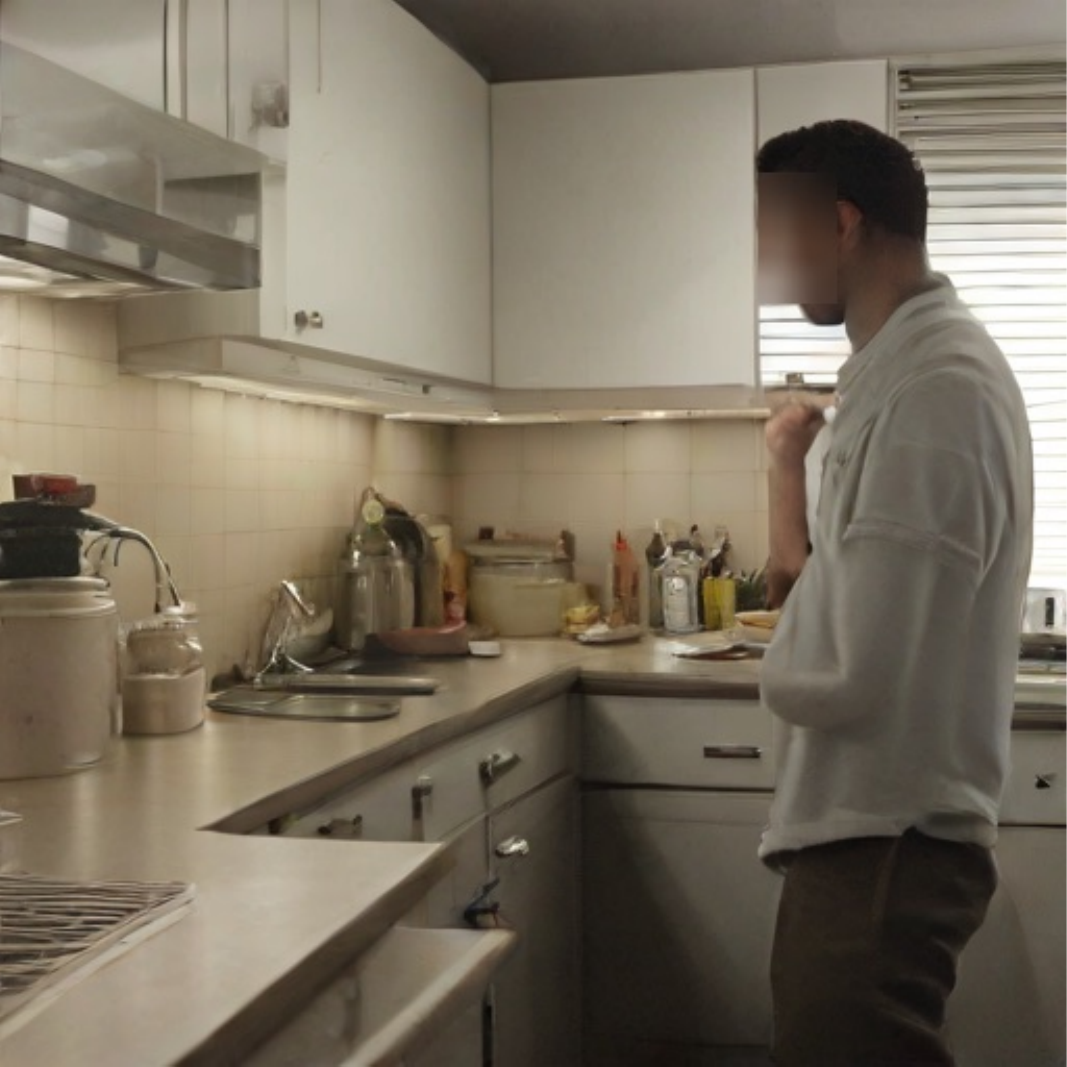}&\includegraphics[width=3cm, height=3cm, cfbox=teal 1pt 1pt]{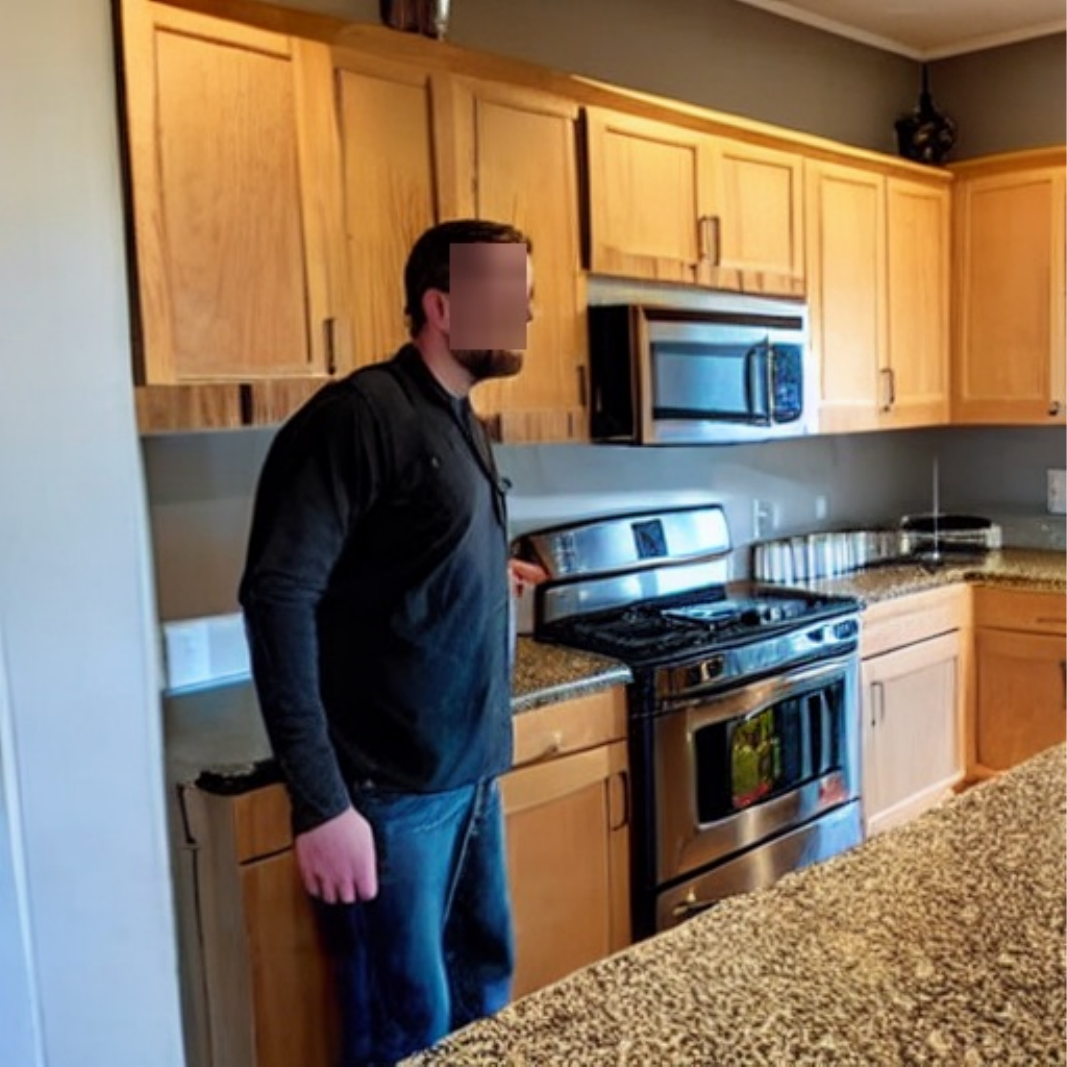} \\
     \multicolumn{2}{p{5cm}}{an attractive young woman leads a grey horse through a paddock.}& & \multicolumn{2}{p{5cm}}{a woman sitting on a unique chair beside a vase.}  \\  
      \includegraphics[width=3cm]{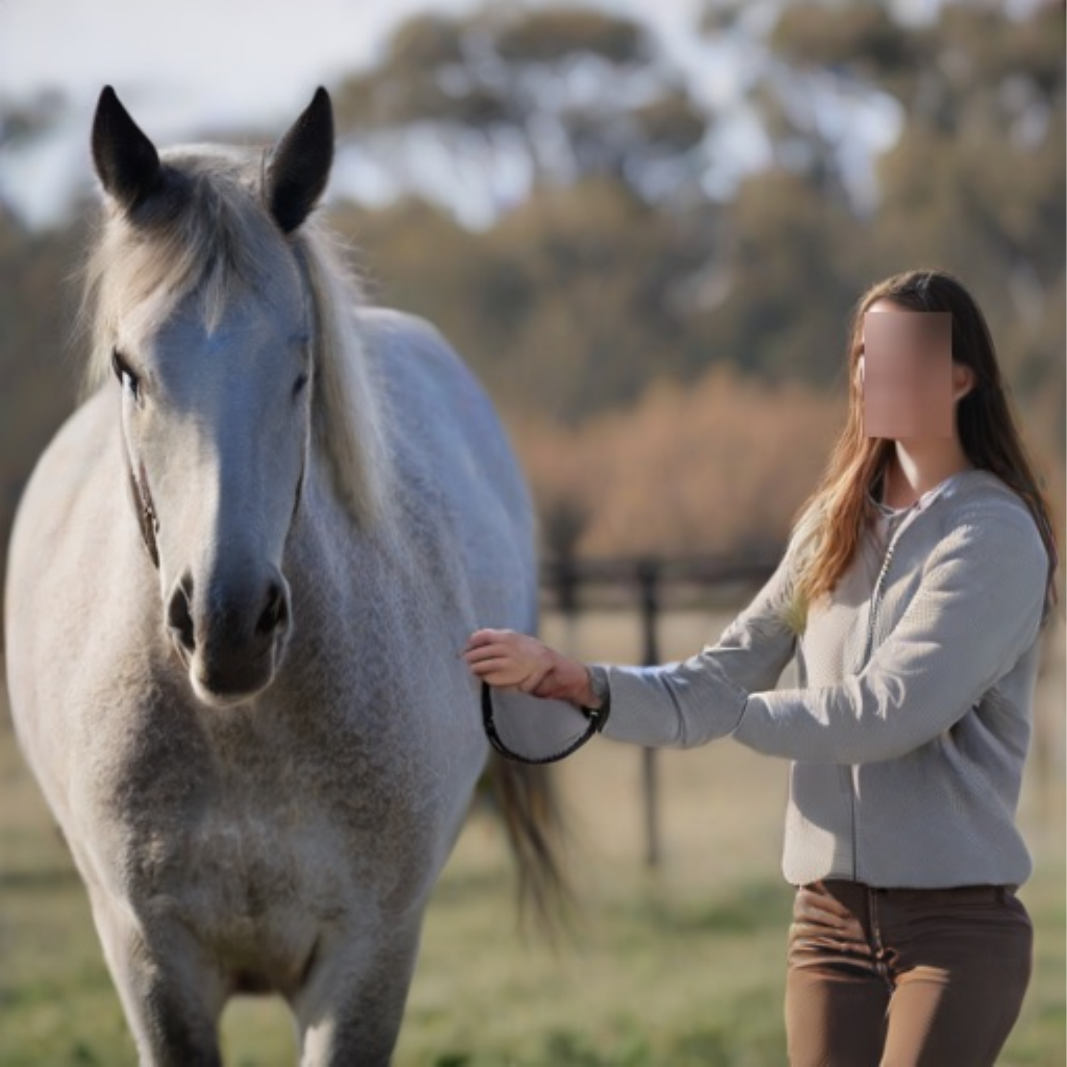}& \includegraphics[width=3cm, cfbox=teal 1pt 1pt]{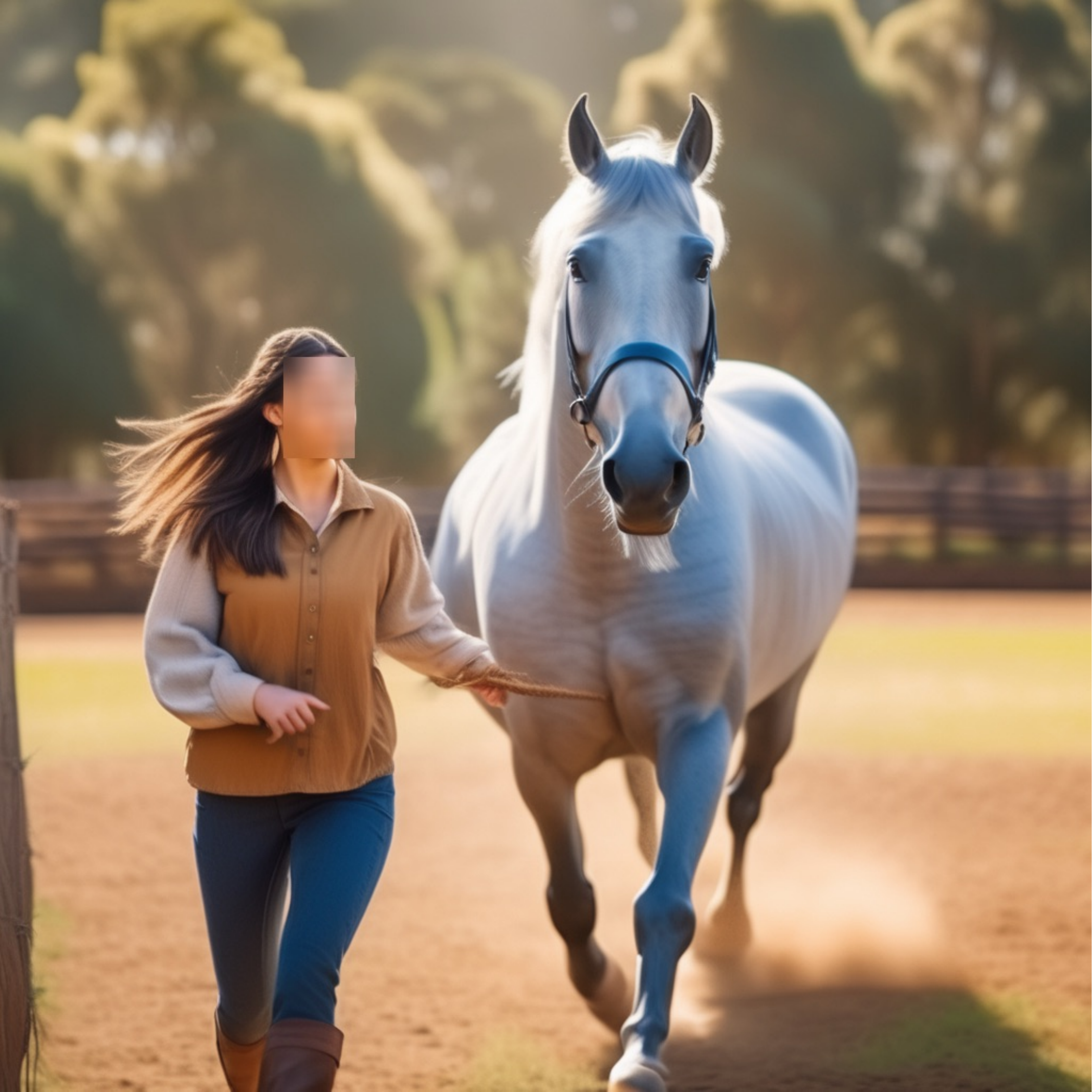}& &\includegraphics[width=3cm, cfbox=teal 1pt 1pt]{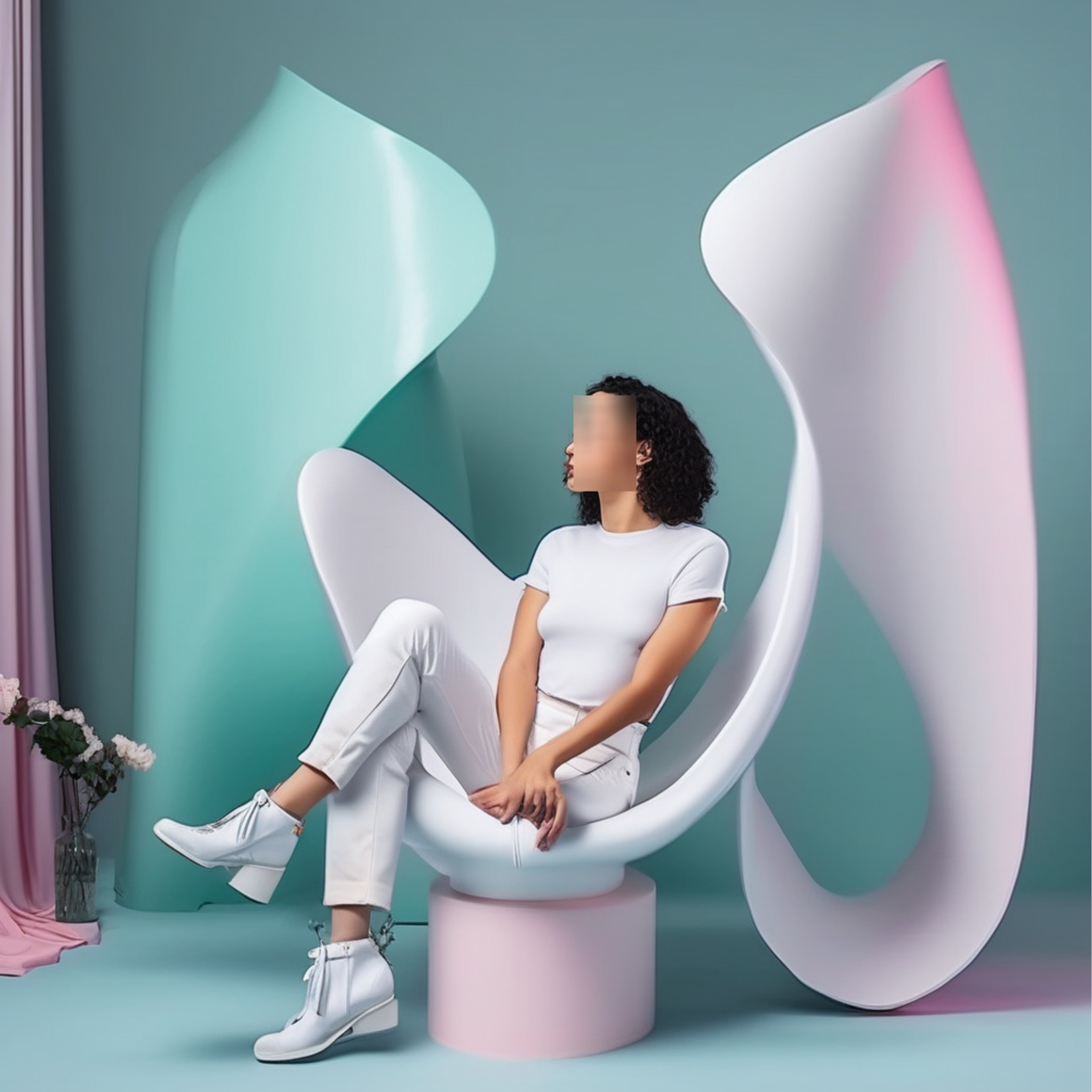}&\includegraphics[width=3cm, height=3cm]{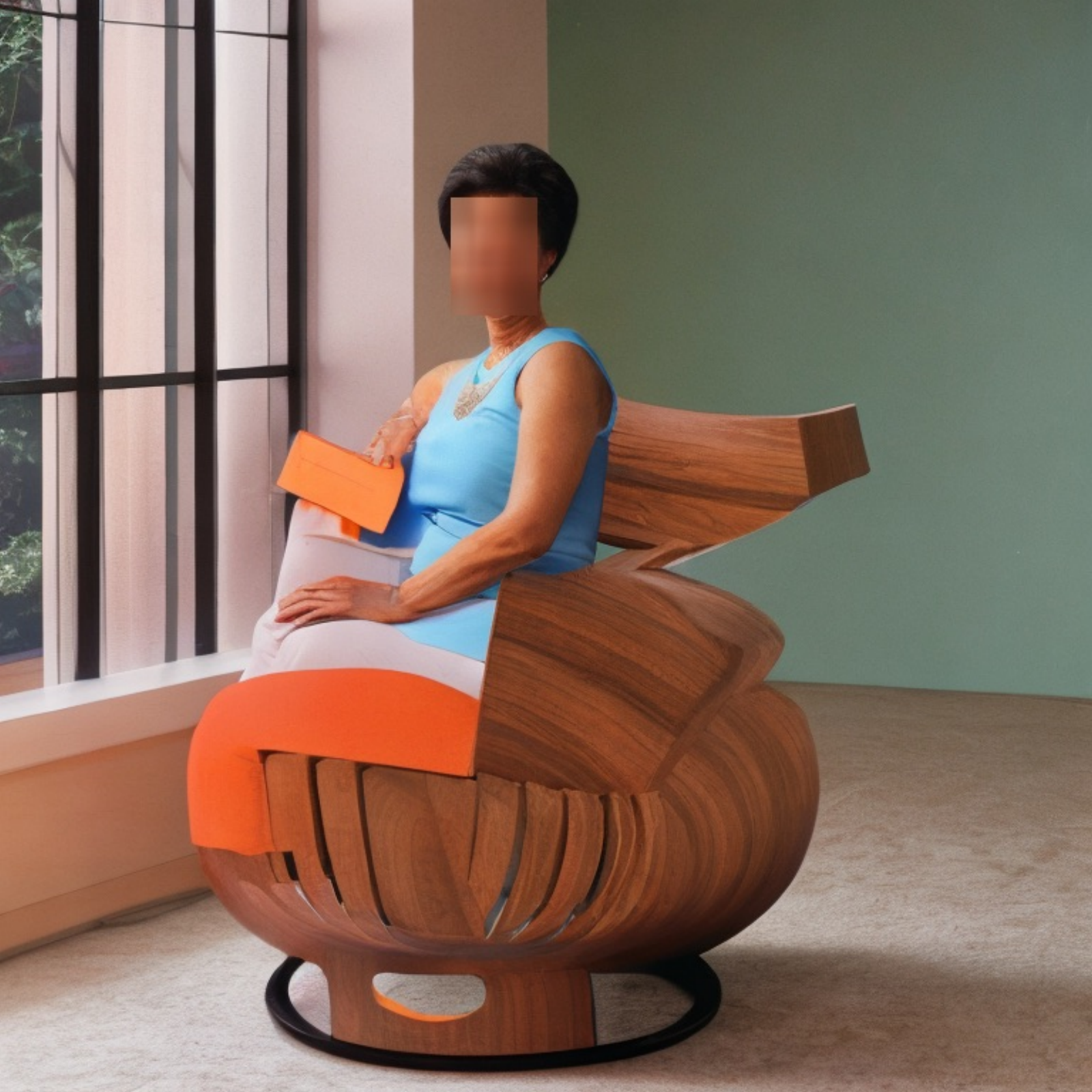}\\
       
    \end{tabular}
    \caption{Pair-wise image preference with BodyMetric. The preferred images are highlighted in green. }
    \label{tab:success2}
\end{table}

\begin{table}[]
    \centering
    \begin{tabular}{ccccc}
    \multicolumn{5}{p{11cm}}{
\centering{{a jazz dancer improvising with soulful style, their movements a tribute to the improvisational spirit of jazz music.}}} \\
    \hline
        \multicolumn{5}{c}{SD-1.4} \\ 
             \hline
                  \hline
        \includegraphics[width=2.3cm]{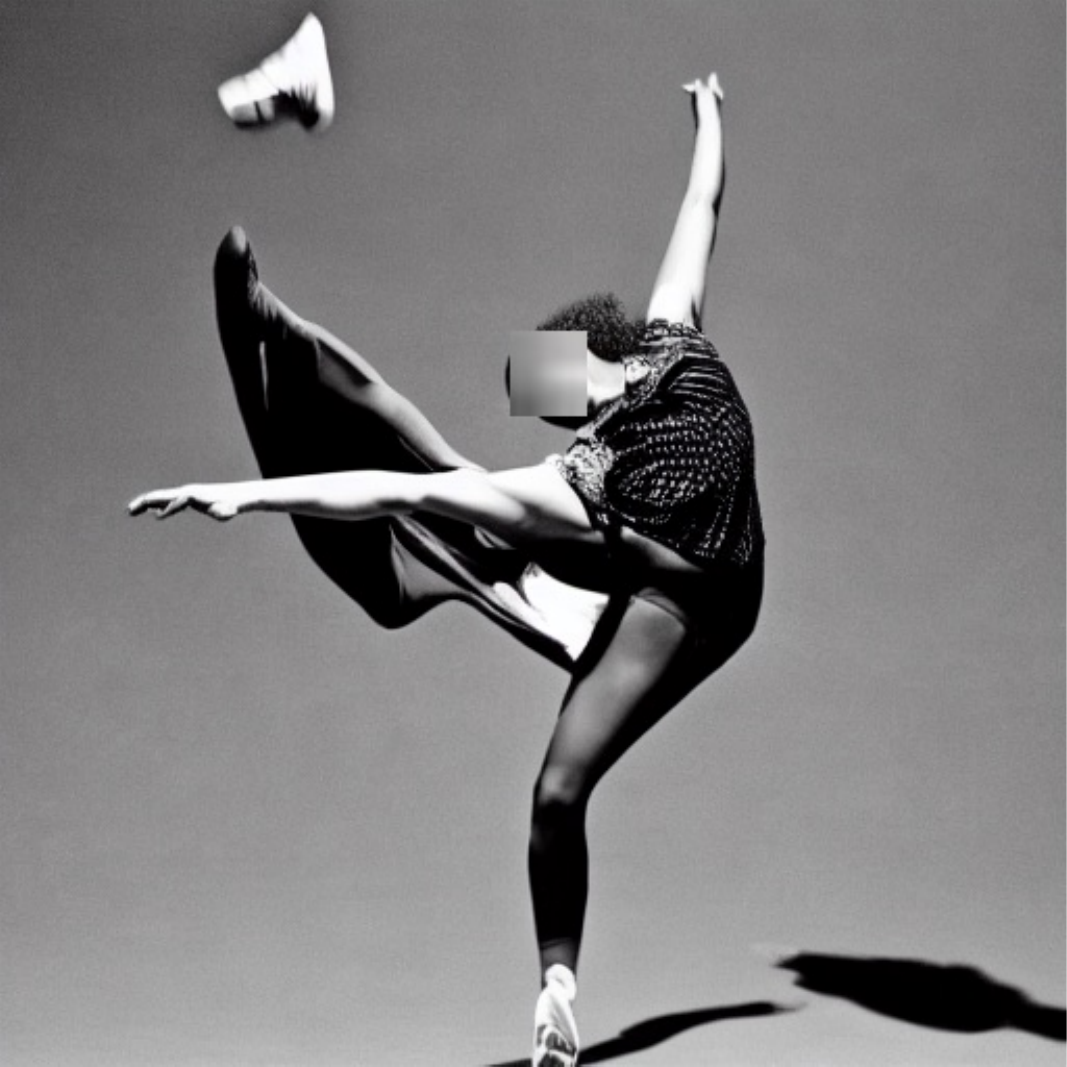} & 
           \includegraphics[width=2.3cm]{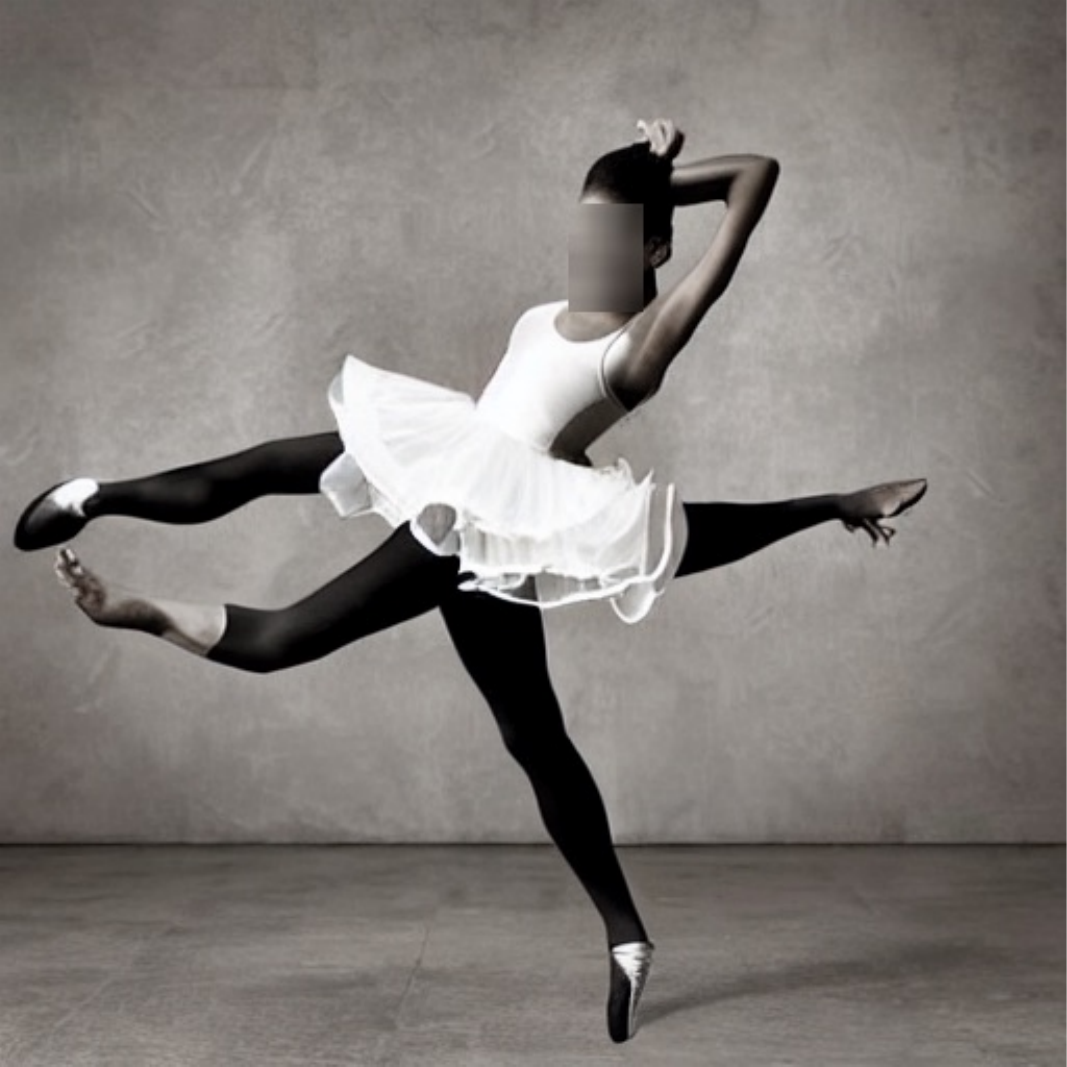} & 
              \includegraphics[width=2.3cm]{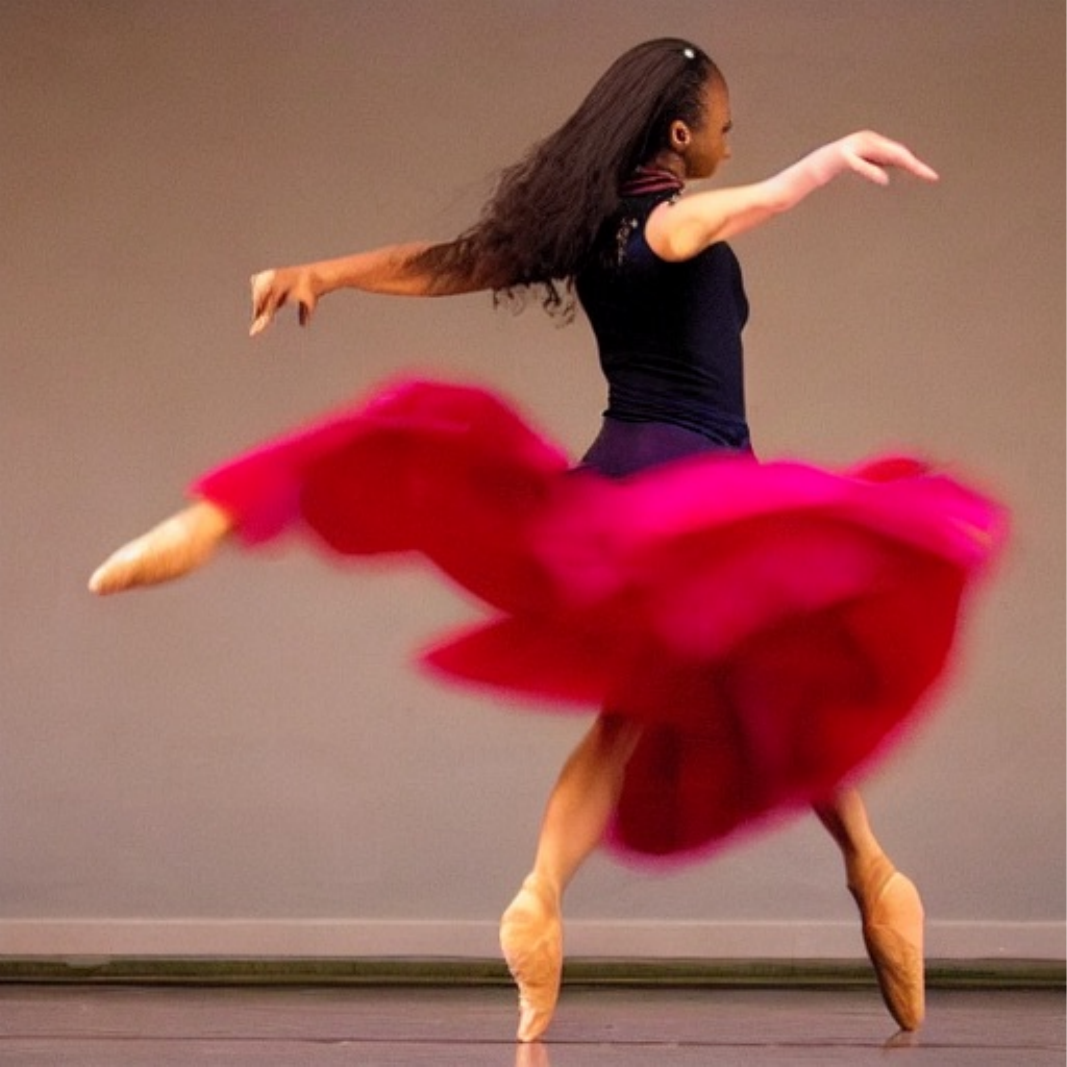} & 
                 \includegraphics[width=2.3cm]{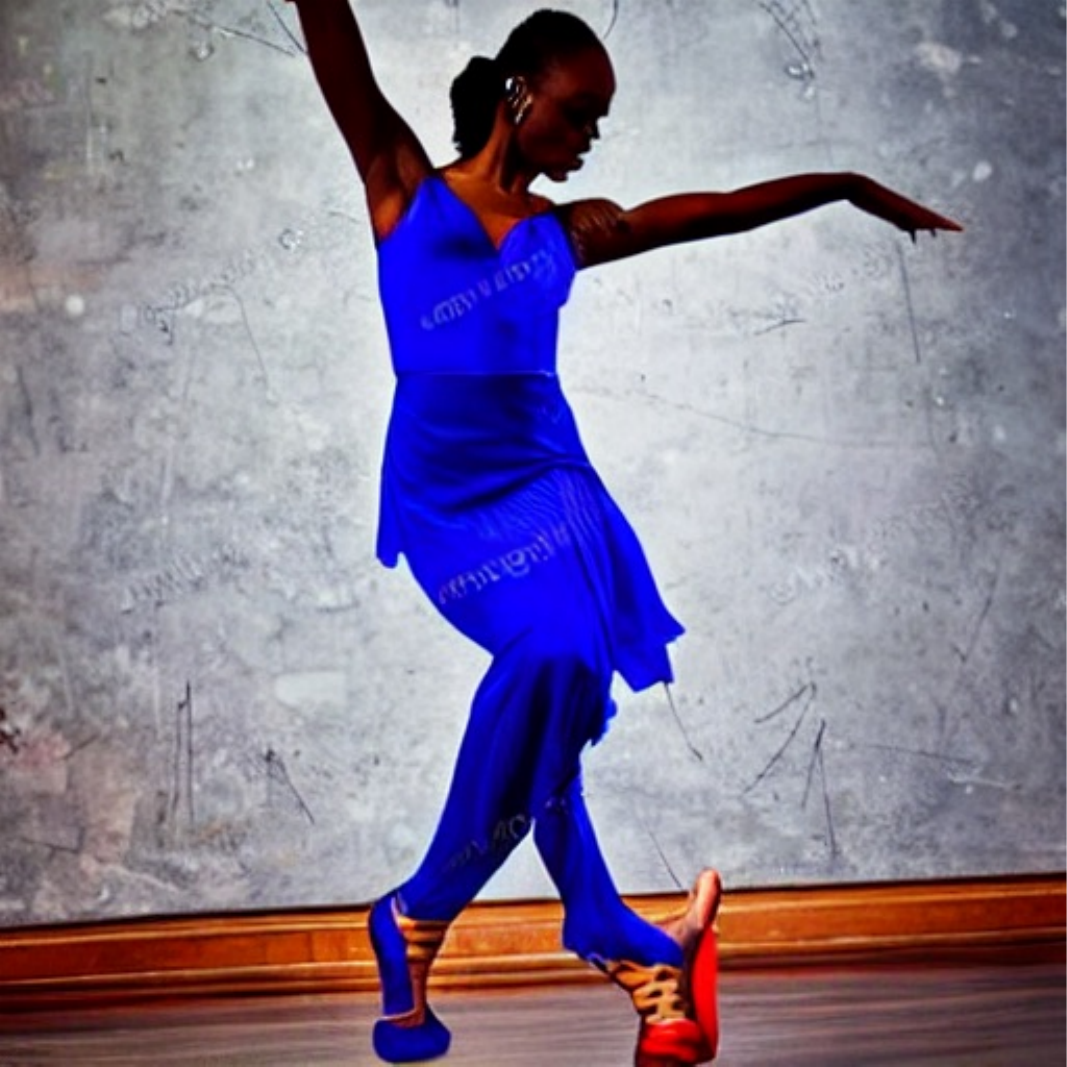} & 
                    \includegraphics[width=2.3cm]{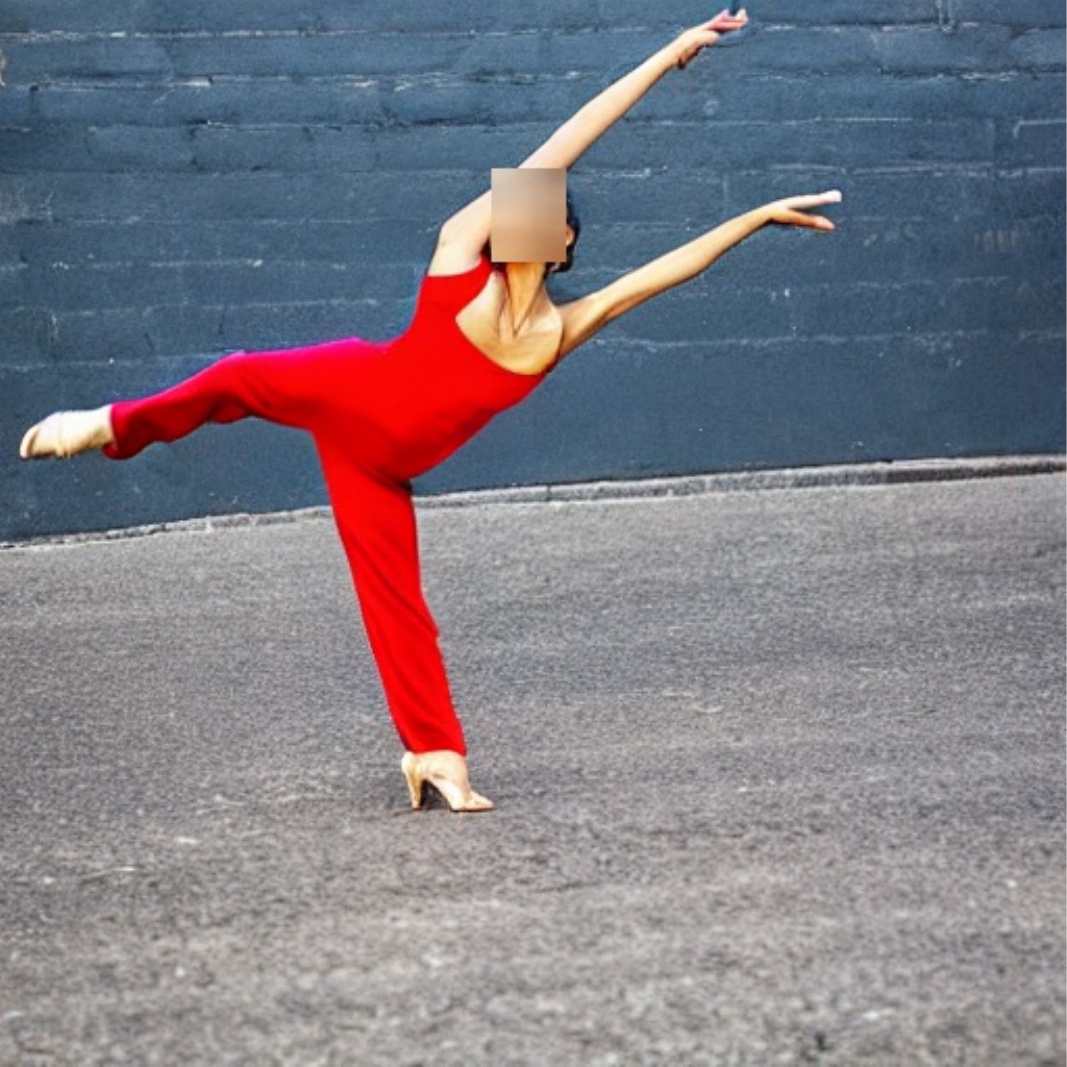} \\
                     \multicolumn{5}{c}{SD-XL-Turbo} \\
         \hline
          \hline
      \includegraphics[width=2.3cm]{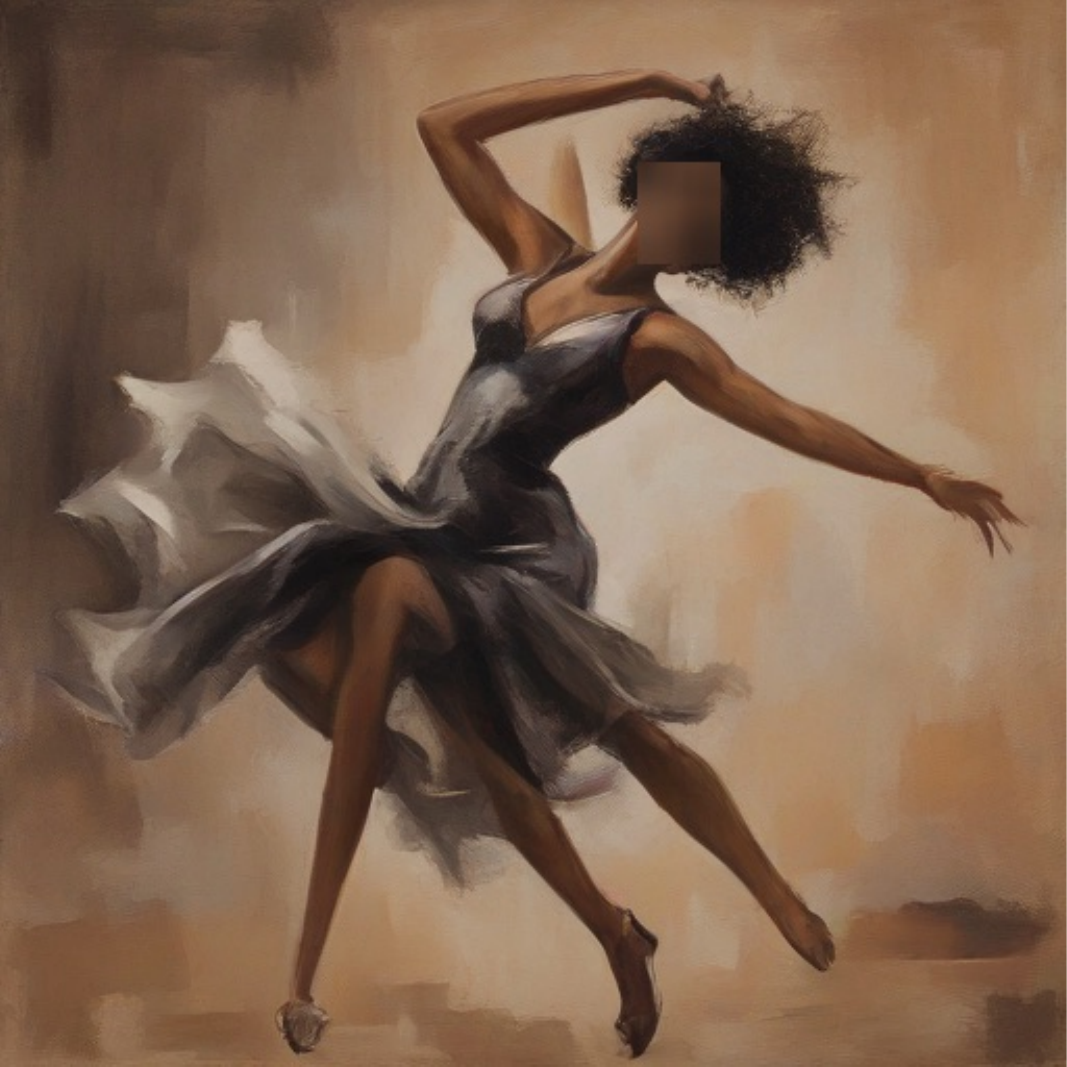} &    
      \includegraphics[width=2.3cm]{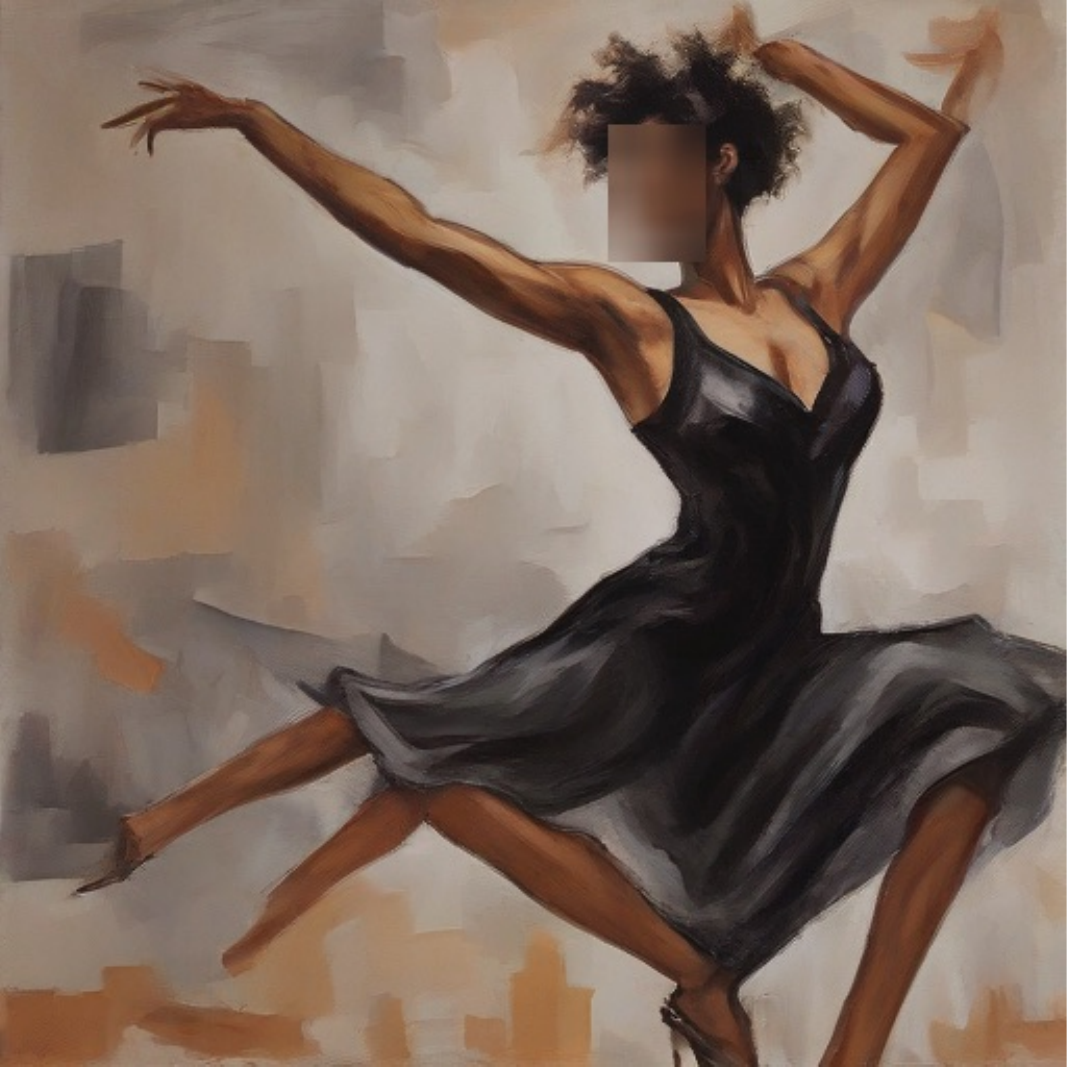} &
       \includegraphics[width=2.3cm]{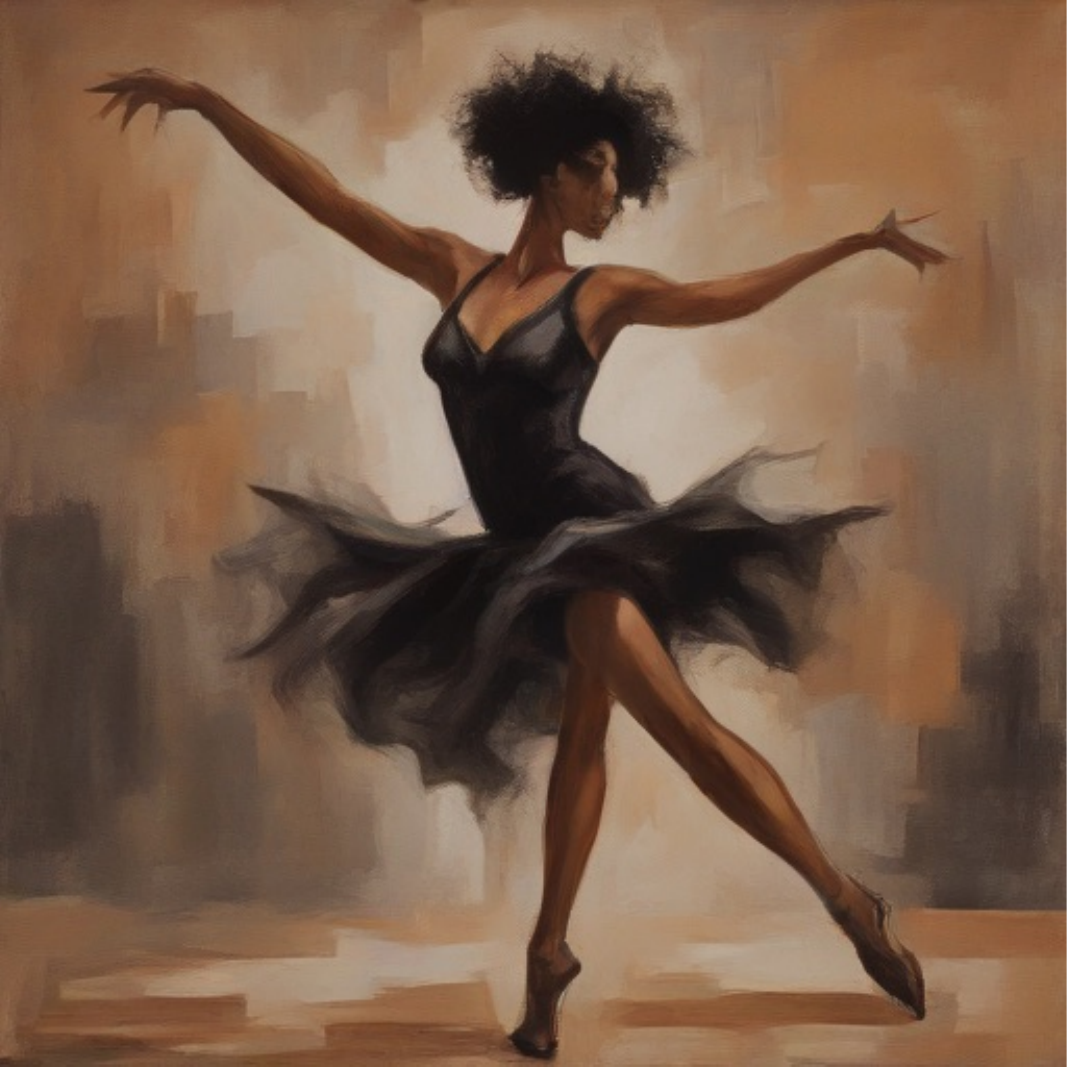} &
        \includegraphics[width=2.3cm]{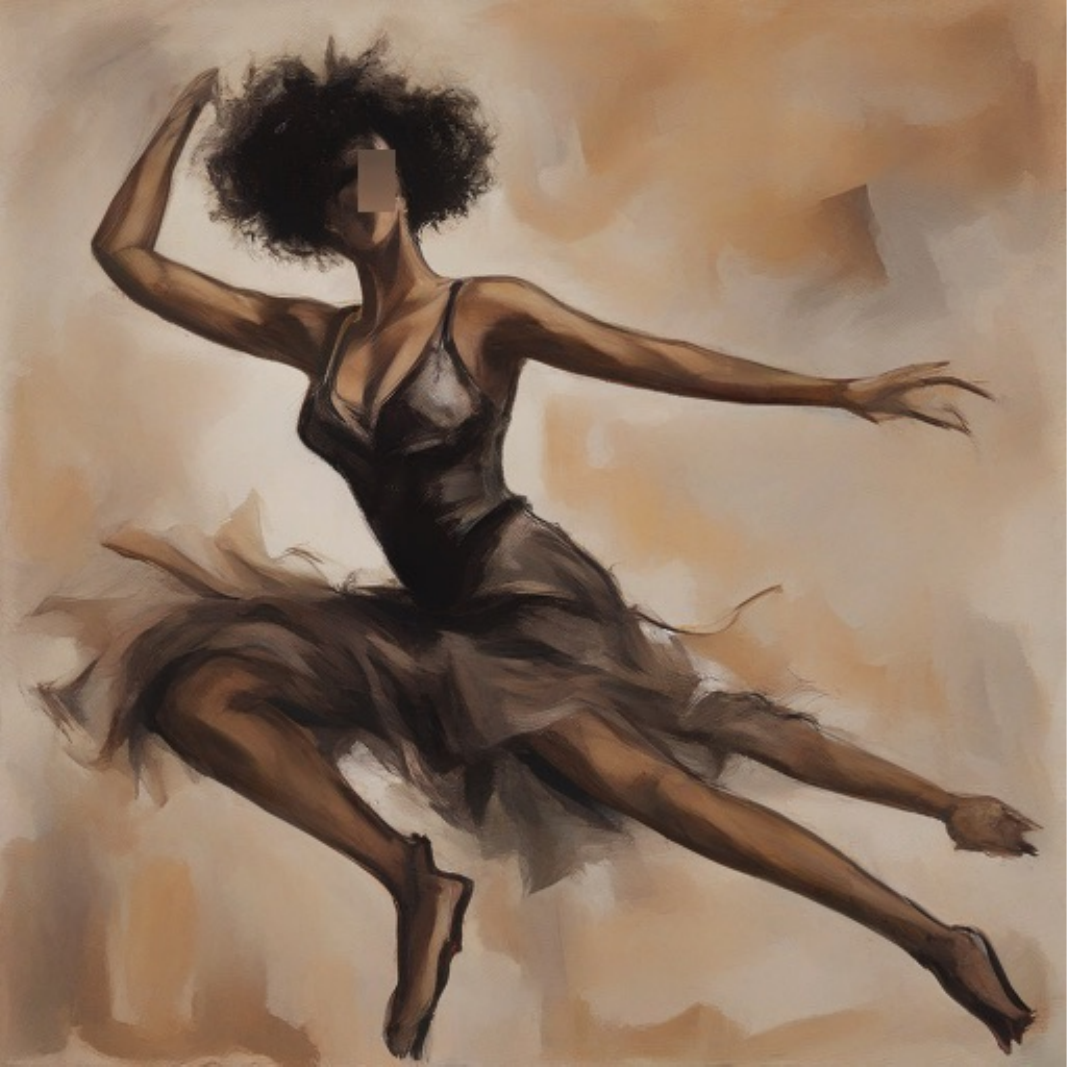} &
     \includegraphics[width=2.3cm]{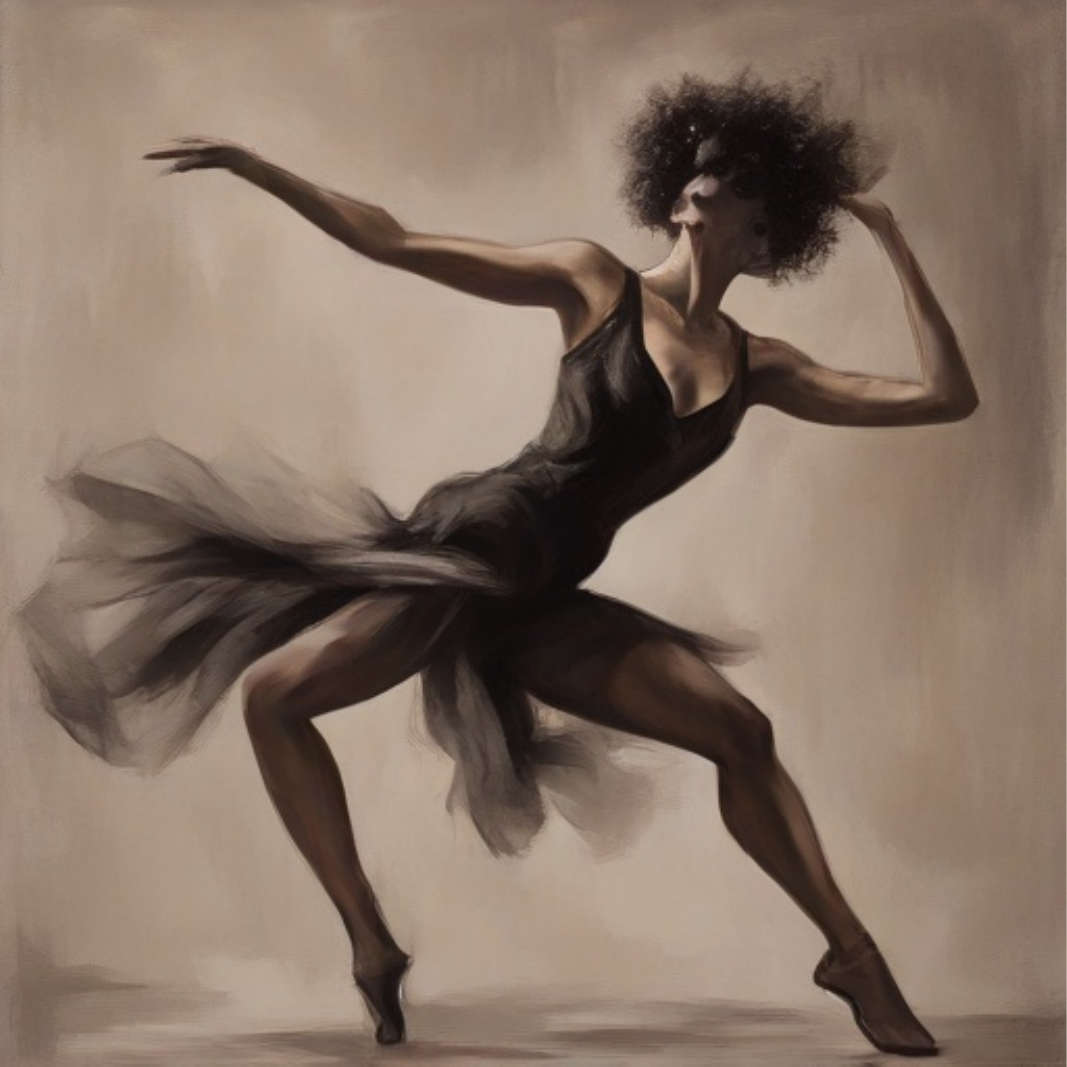} \\
        \multicolumn{5}{c}{SD-2.1} \\ 
             \hline
                  \hline
        \includegraphics[width=2.3cm]{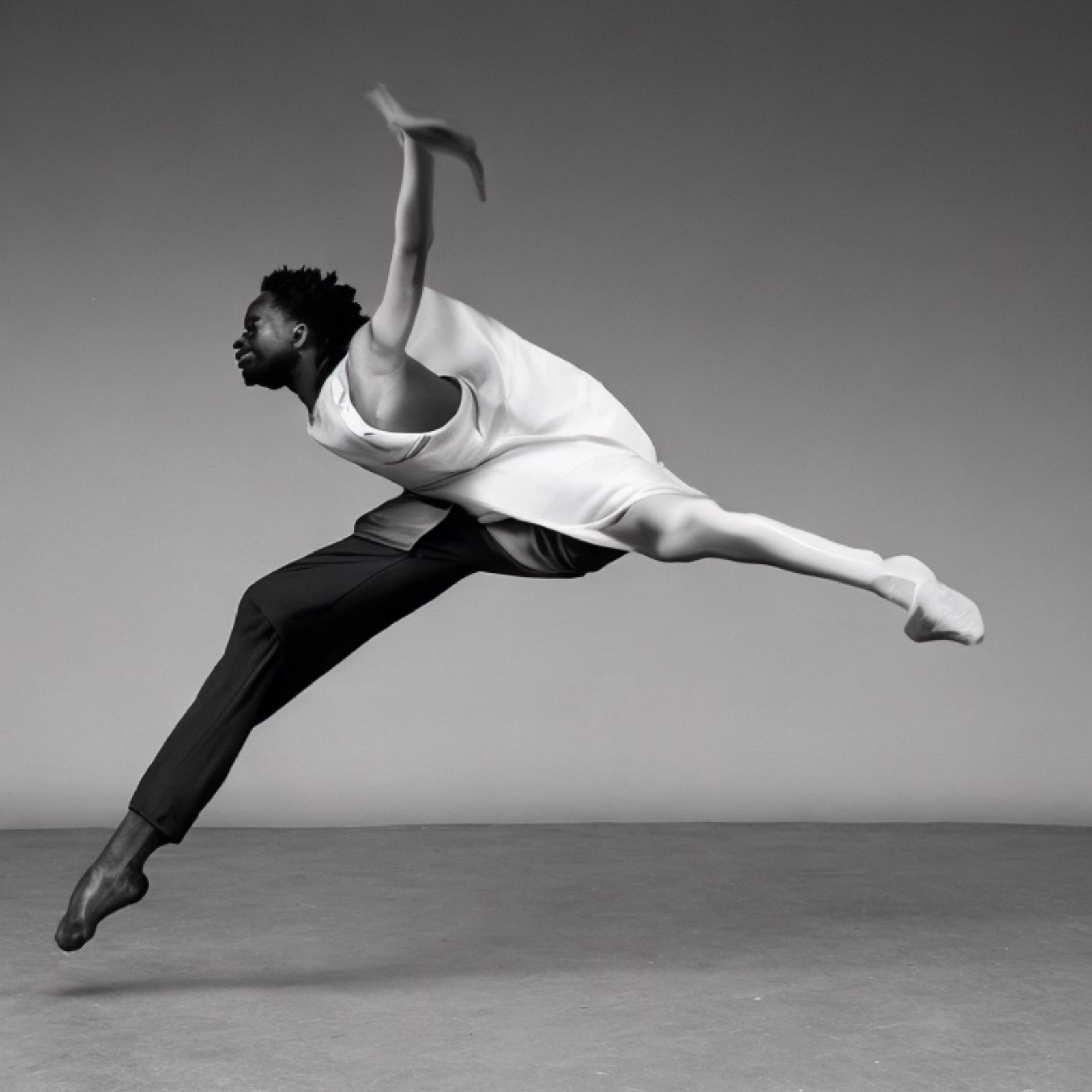} &     \includegraphics[width=2.3cm]{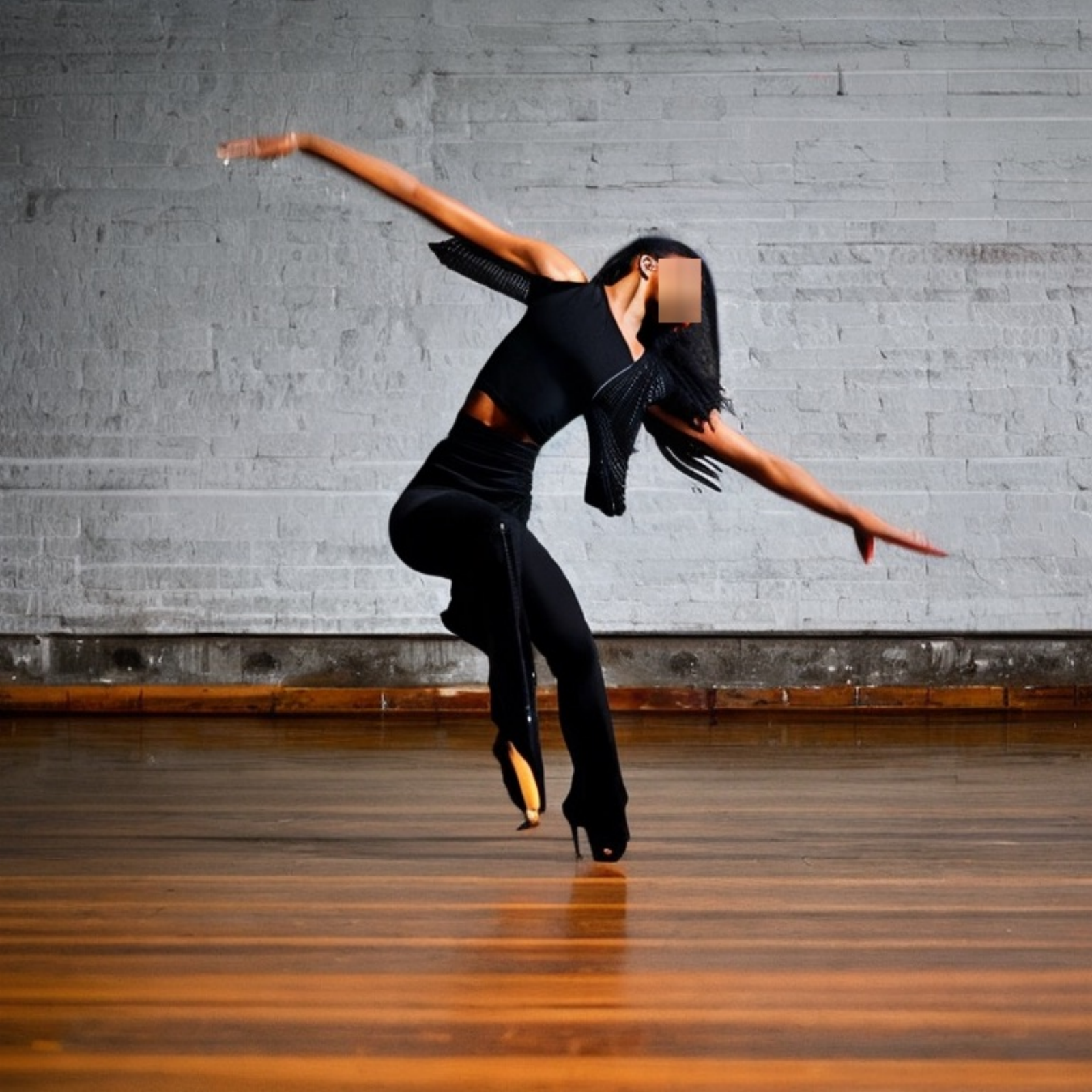} &     \includegraphics[width=2.3cm]{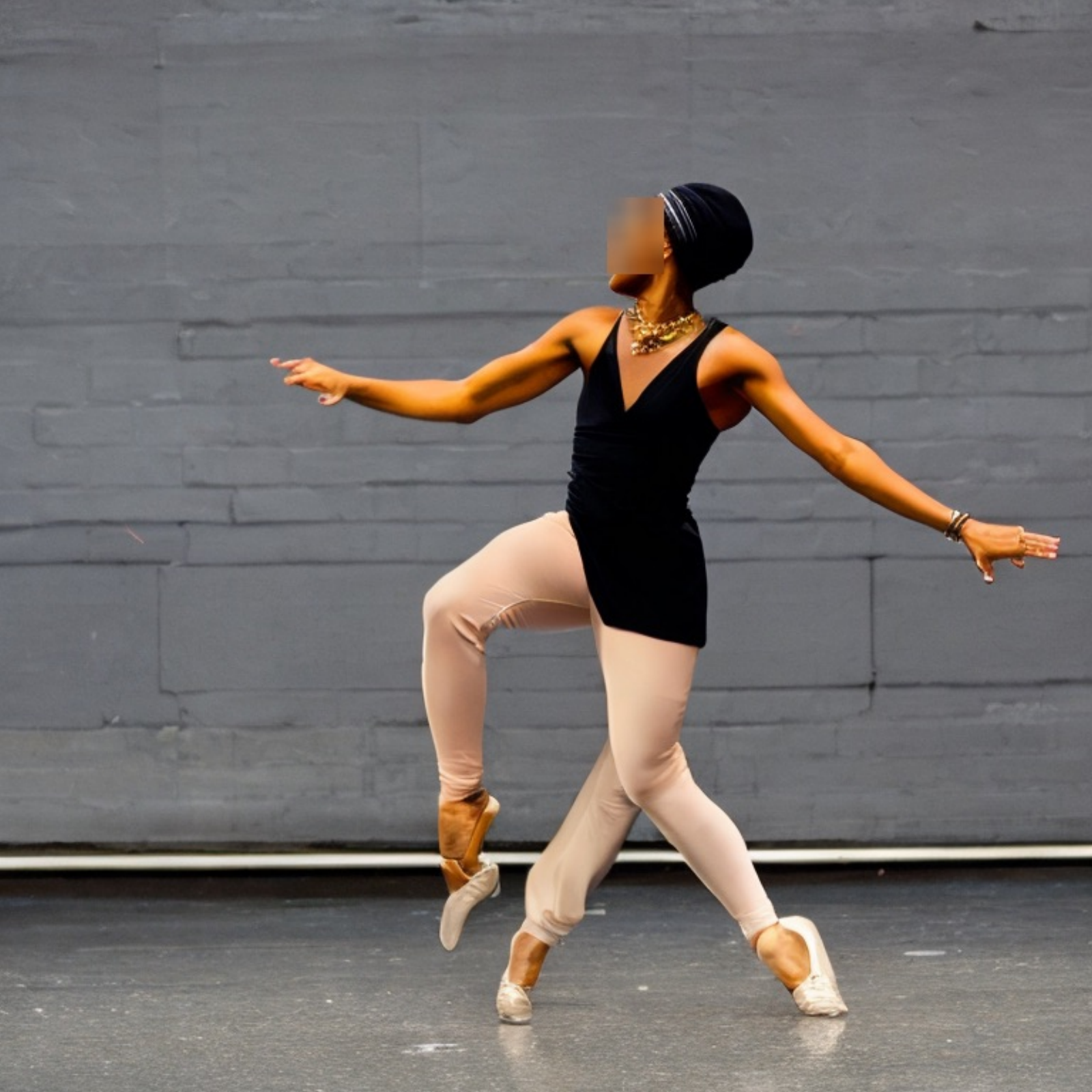} &     \includegraphics[width=2.3cm]{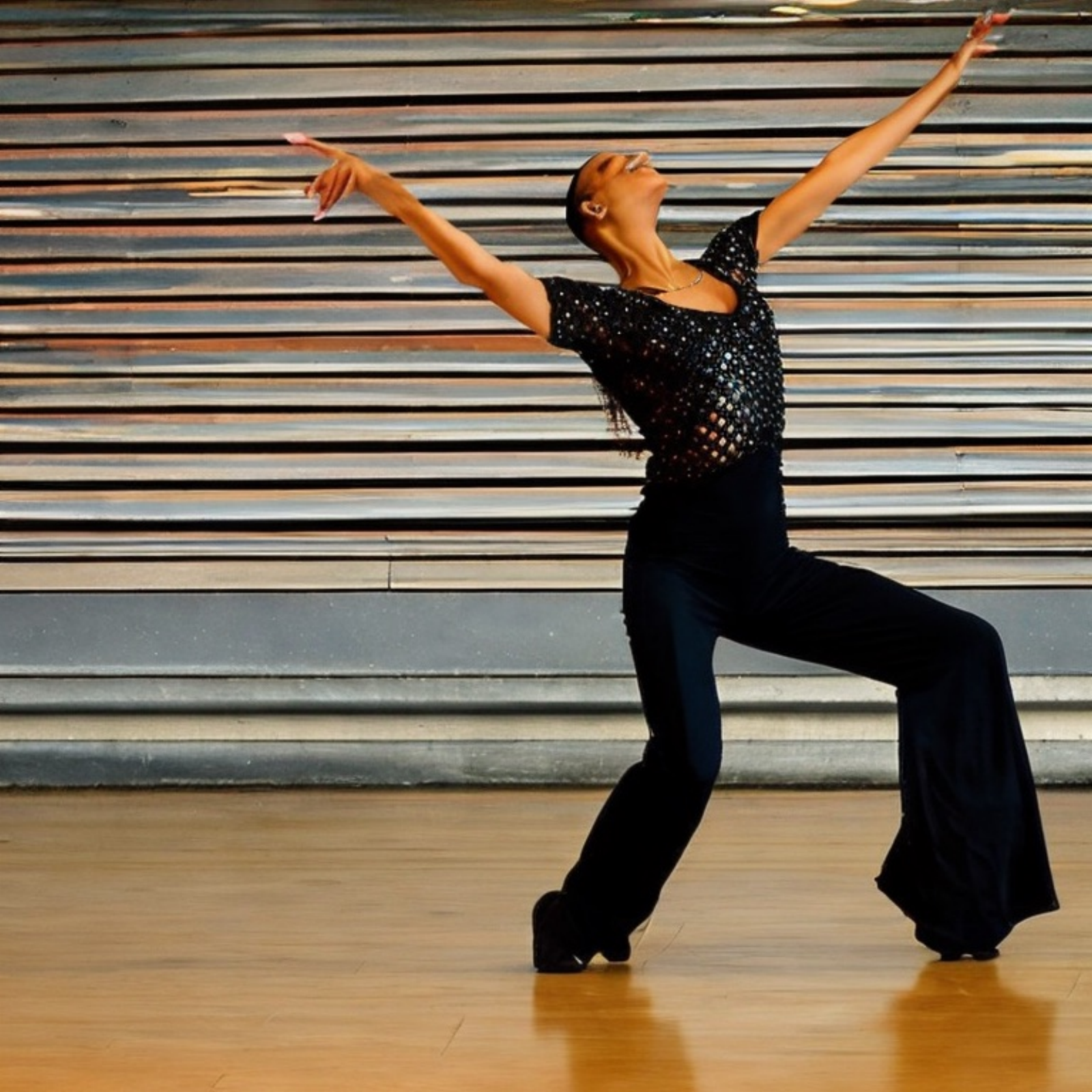} &     \includegraphics[width=2.3cm]{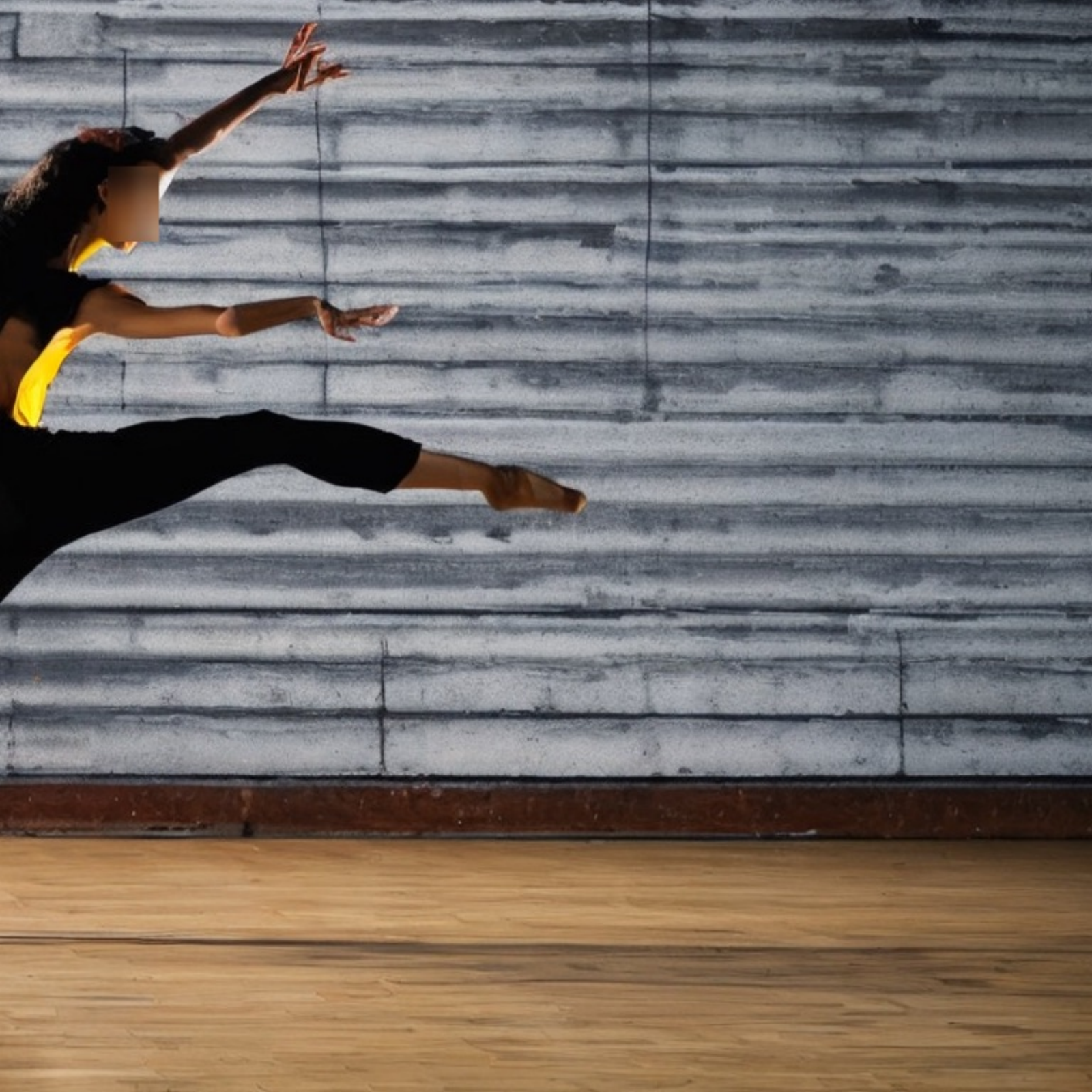} \\

        \multicolumn{5}{c}{Wuerstchen} \\ 
             \hline
                  \hline
        \includegraphics[width=2.3cm]{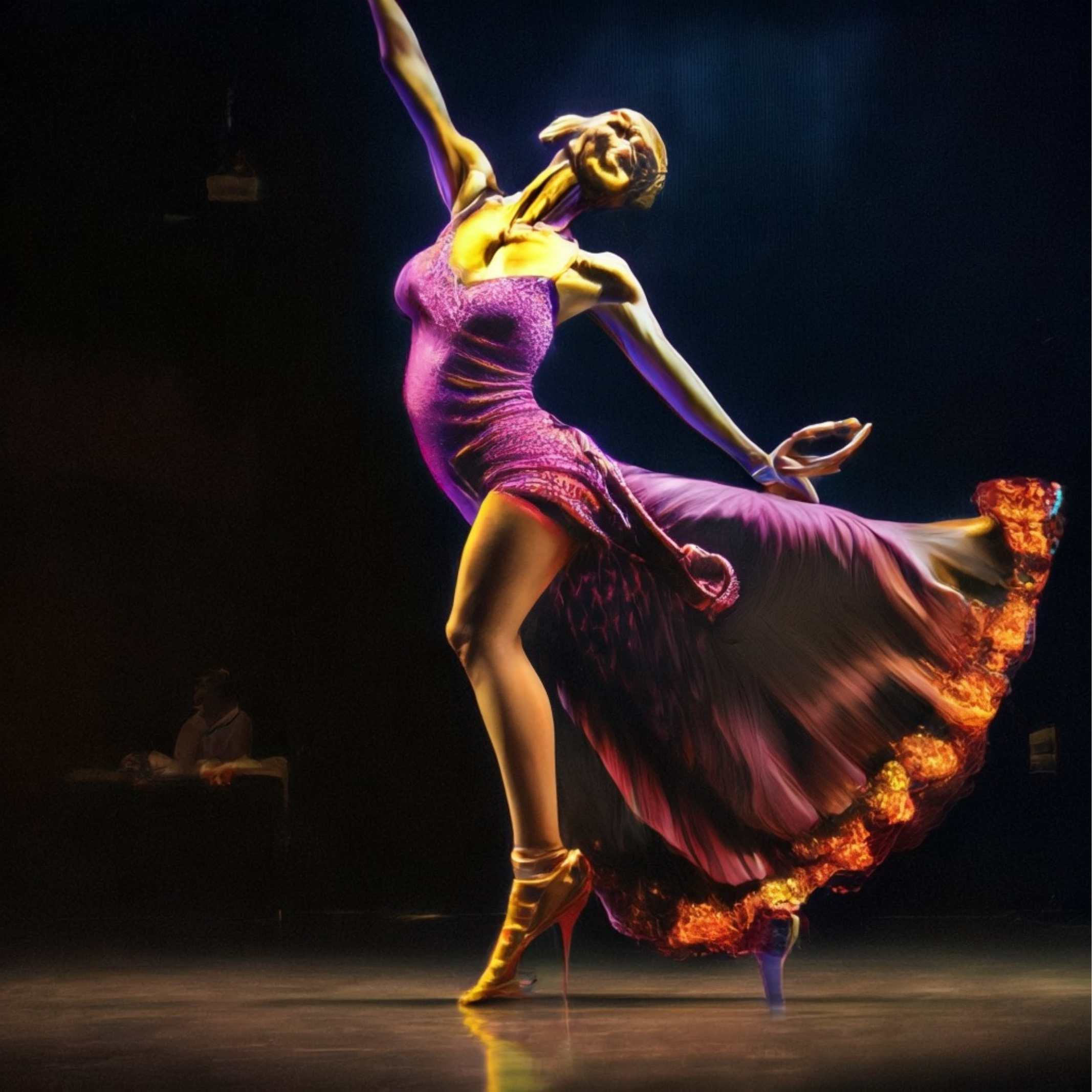} &\includegraphics[width=2.3cm]{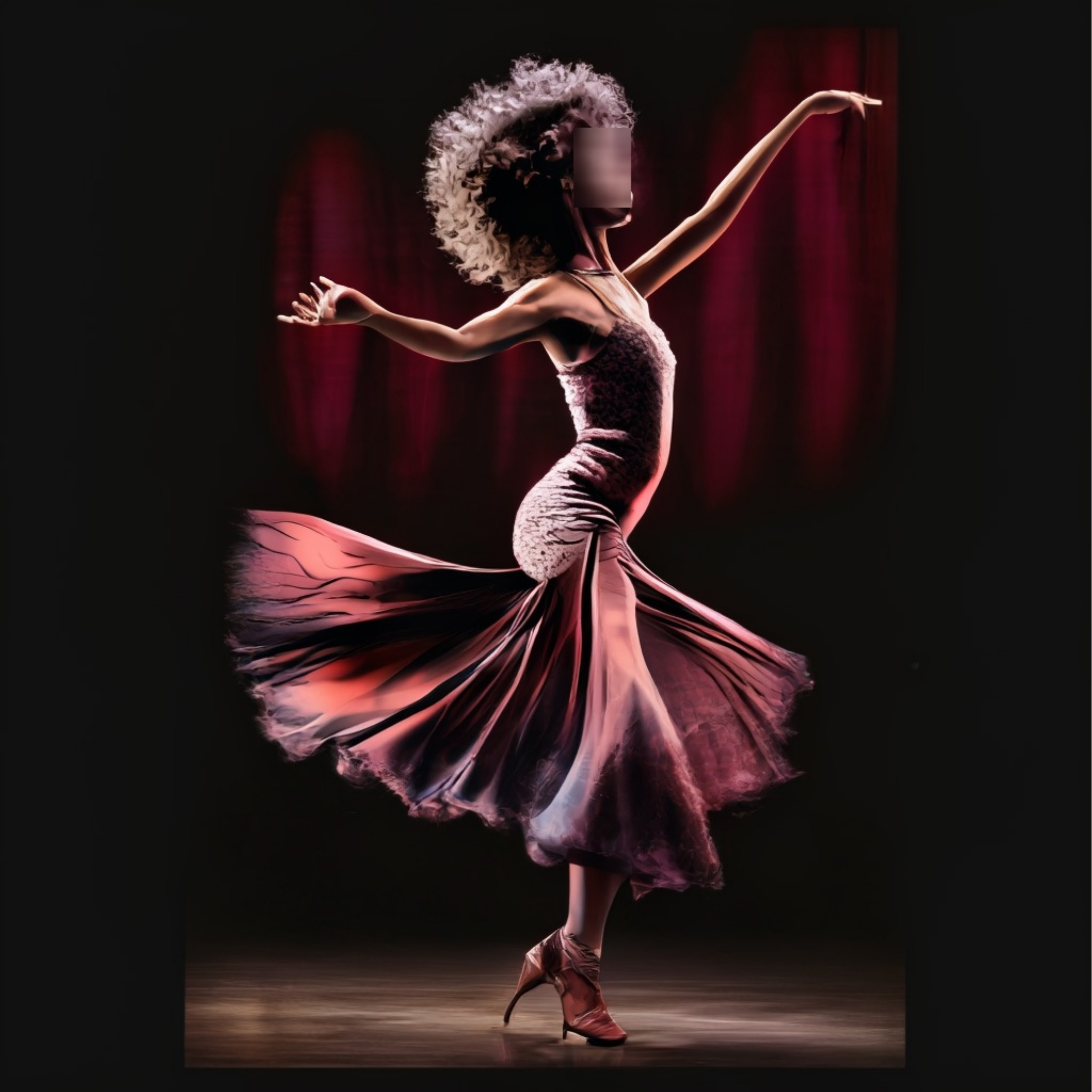} &\includegraphics[width=2.3cm]{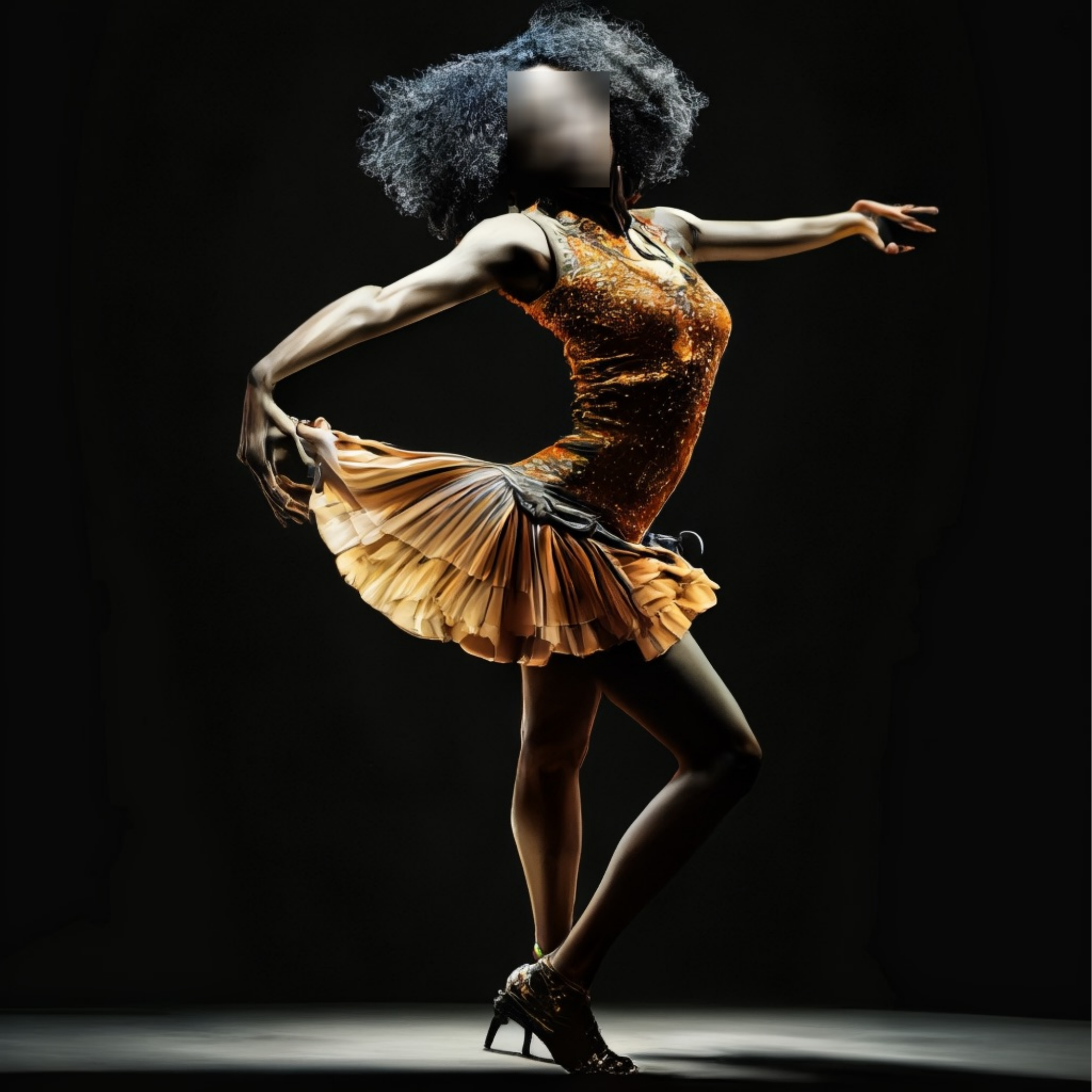} &\includegraphics[width=2.3cm]{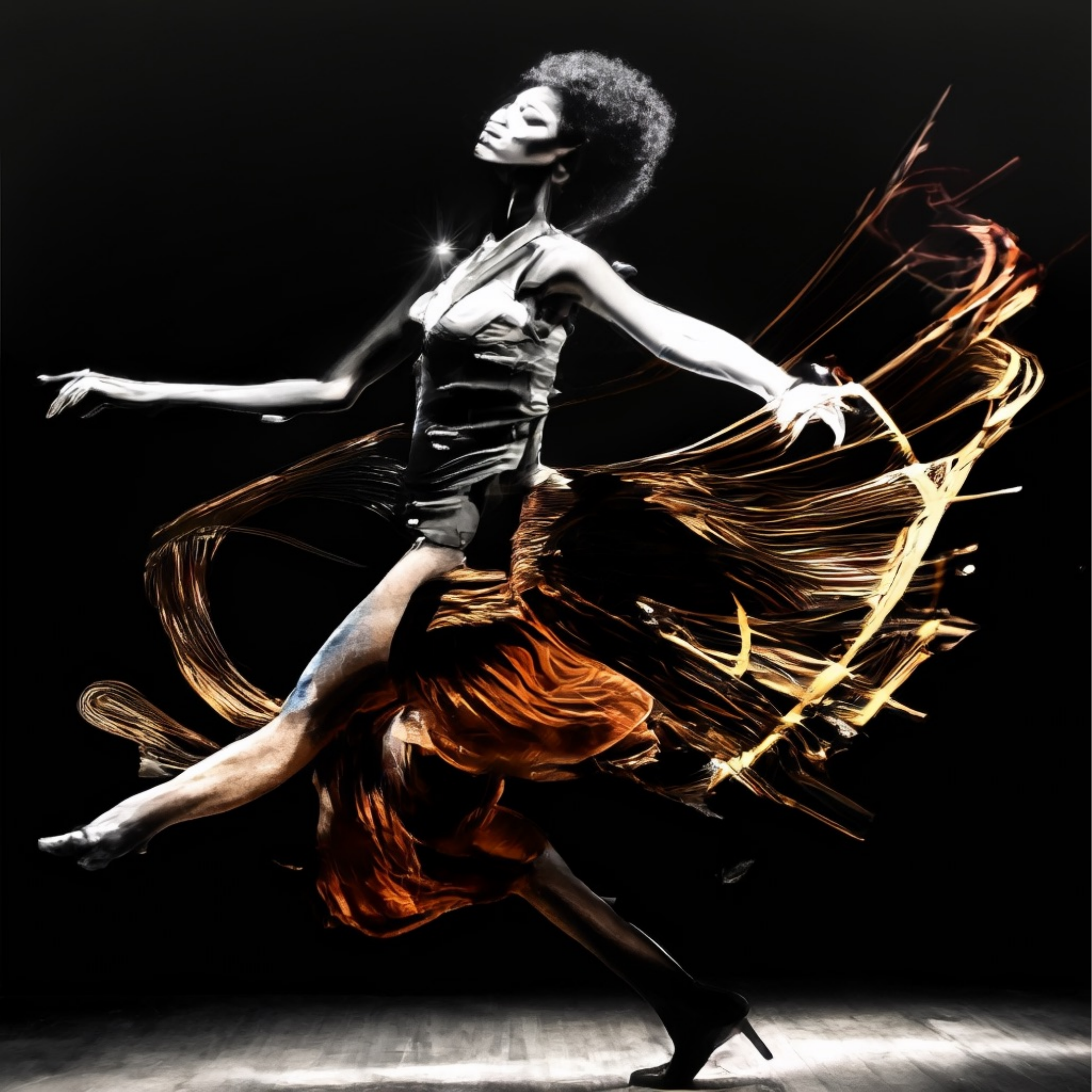} &\includegraphics[width=2.3cm]{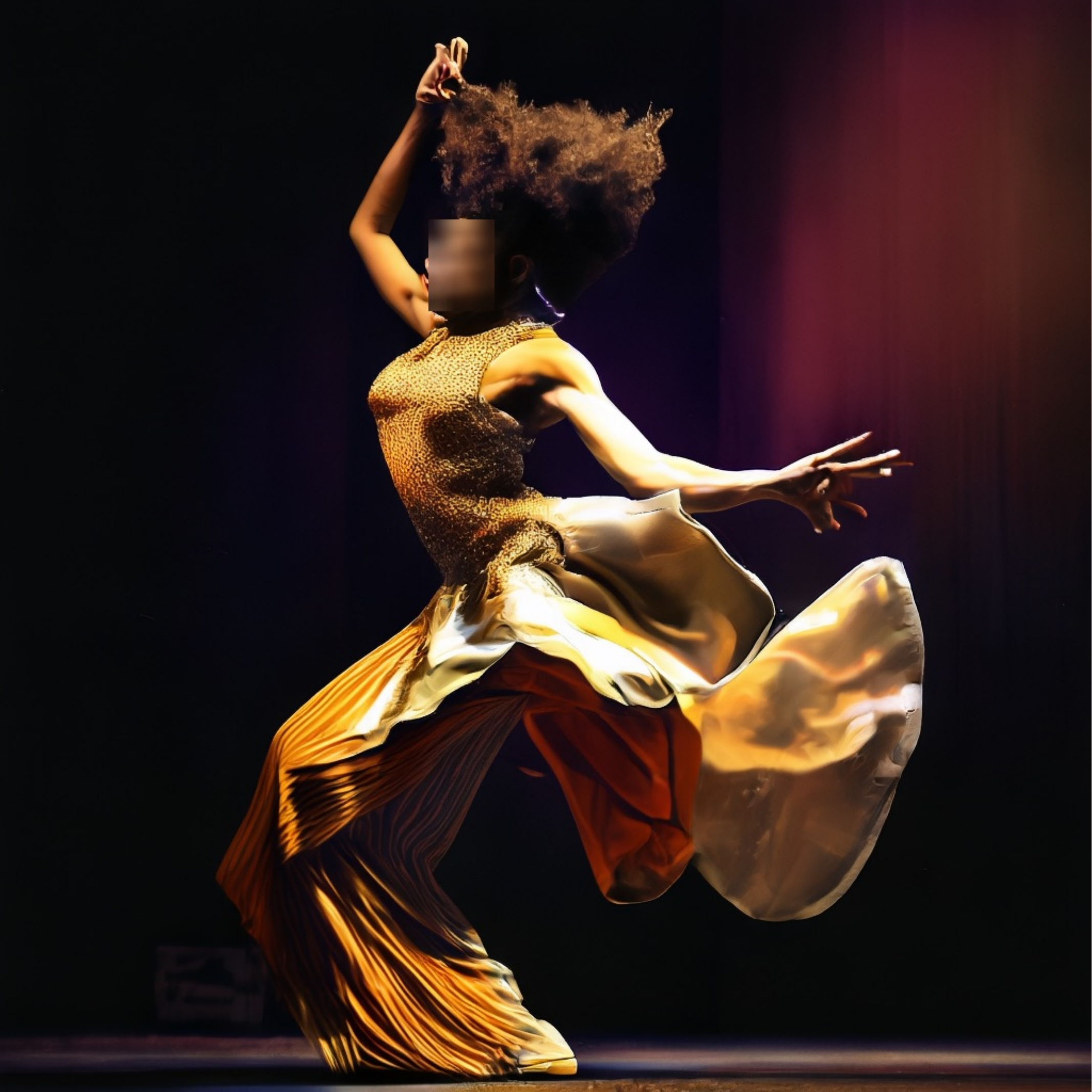} \\
        \multicolumn{5}{c}{SD-XL} \\ 
        \hline
        \hline
        \includegraphics[width=2.3cm]{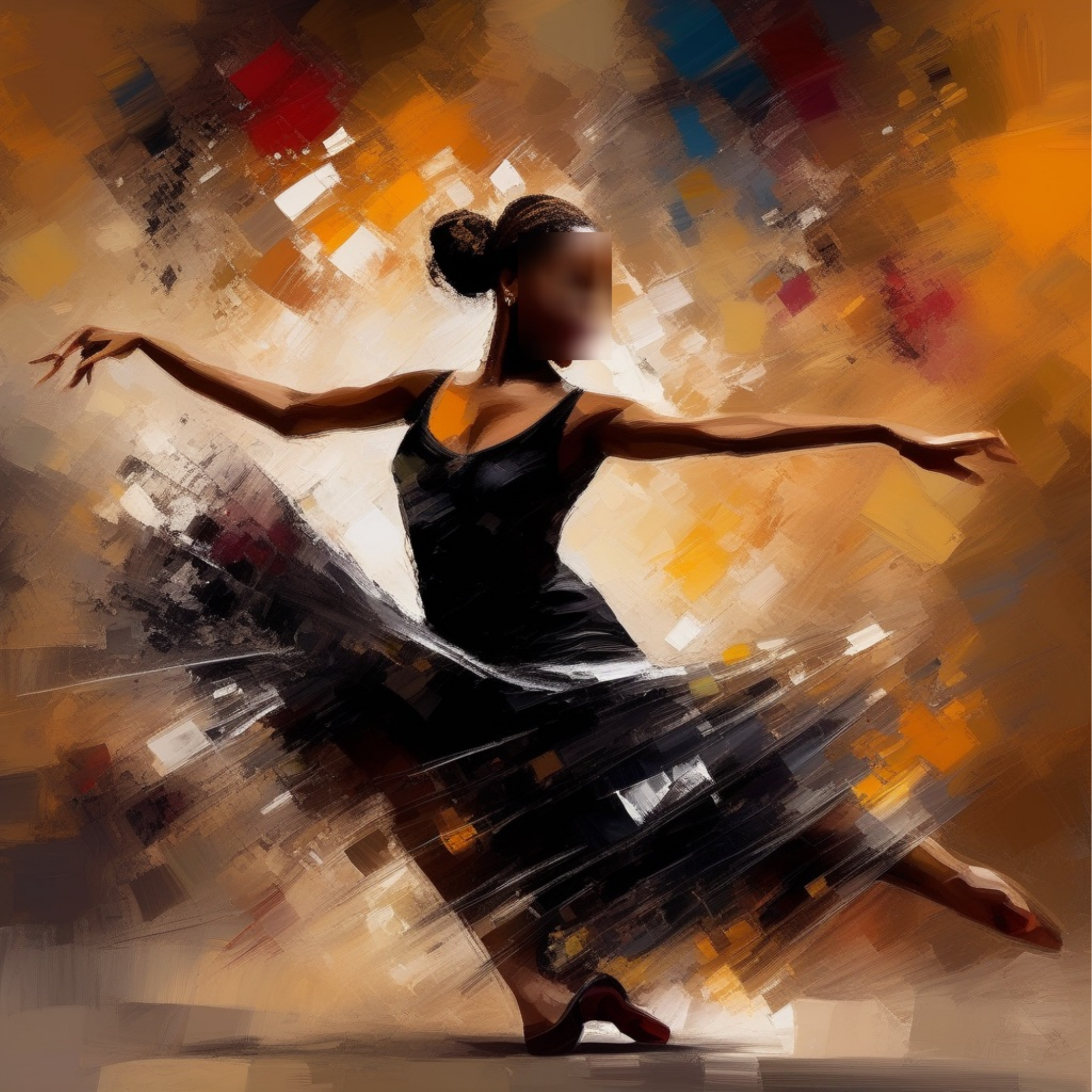} & \includegraphics[width=2.3cm]{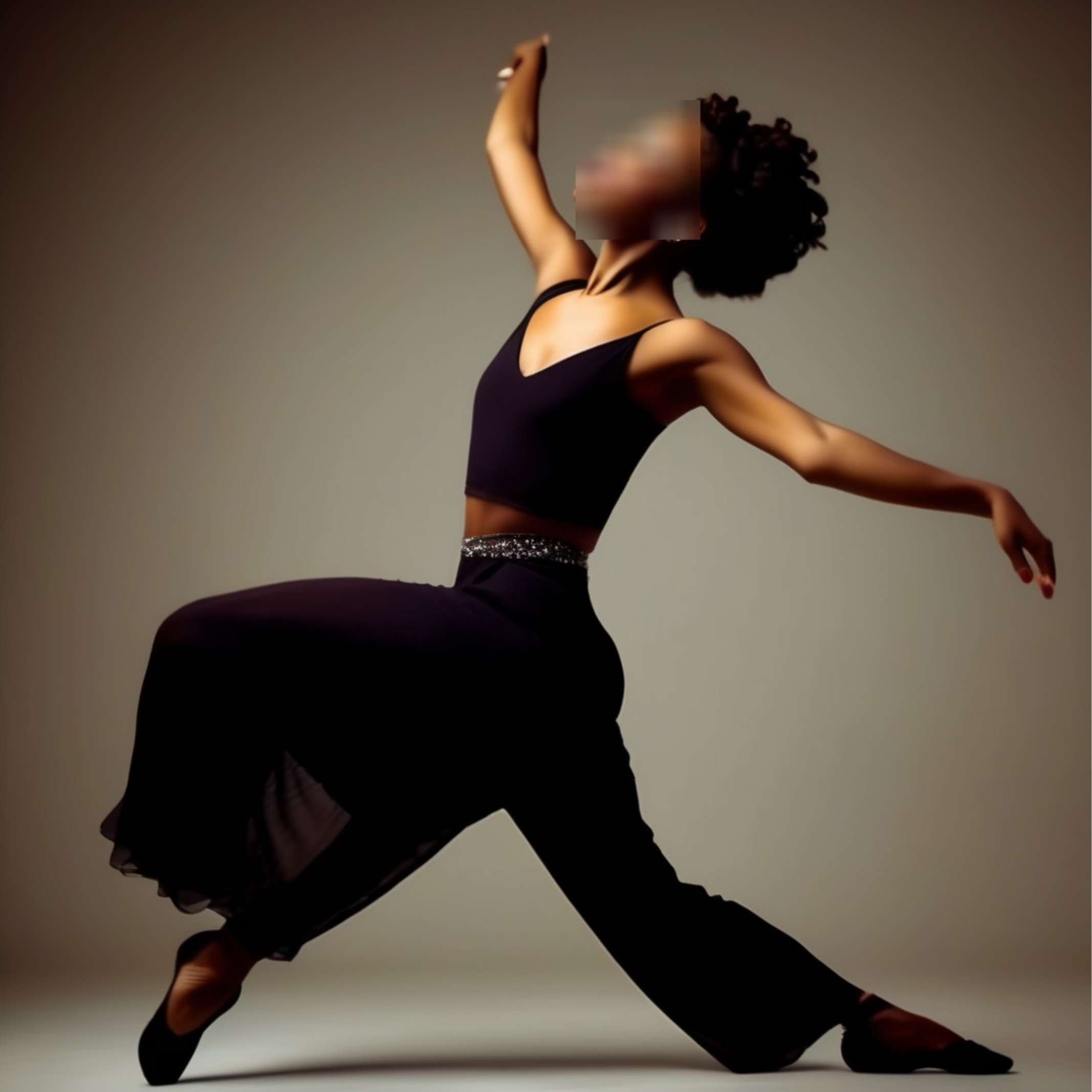} & \includegraphics[width=2.3cm]{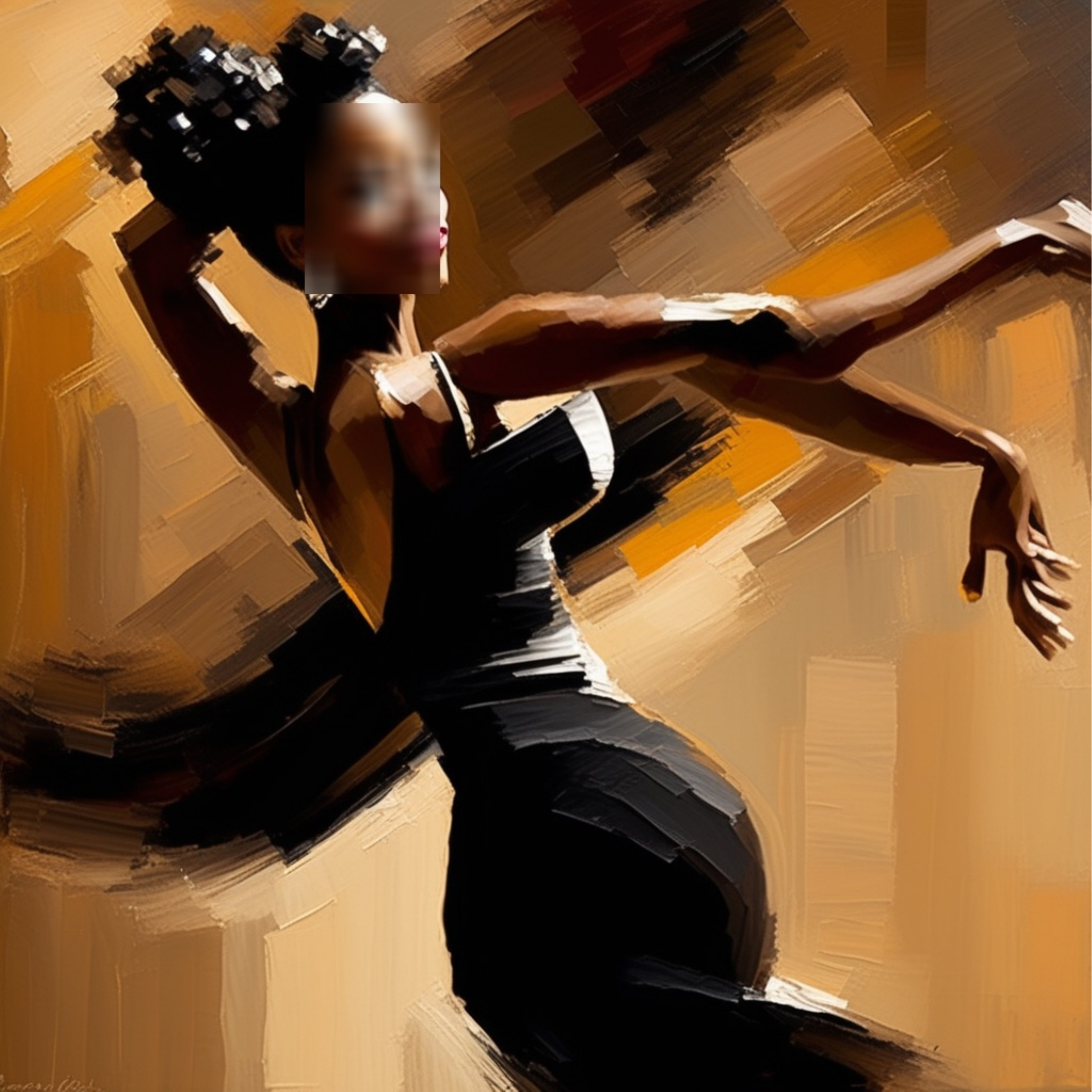} & \includegraphics[width=2.3cm]{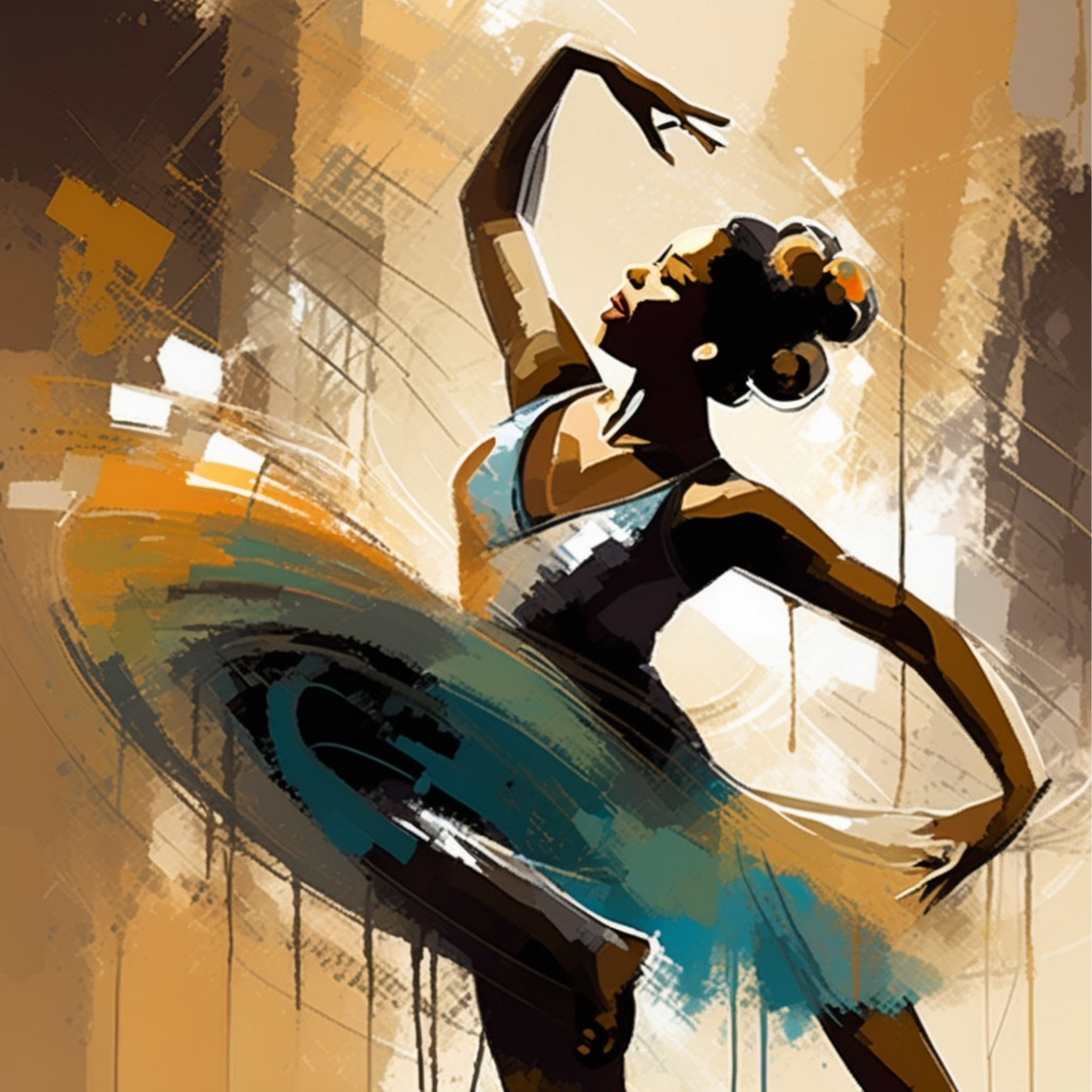} & \includegraphics[width=2.3cm]{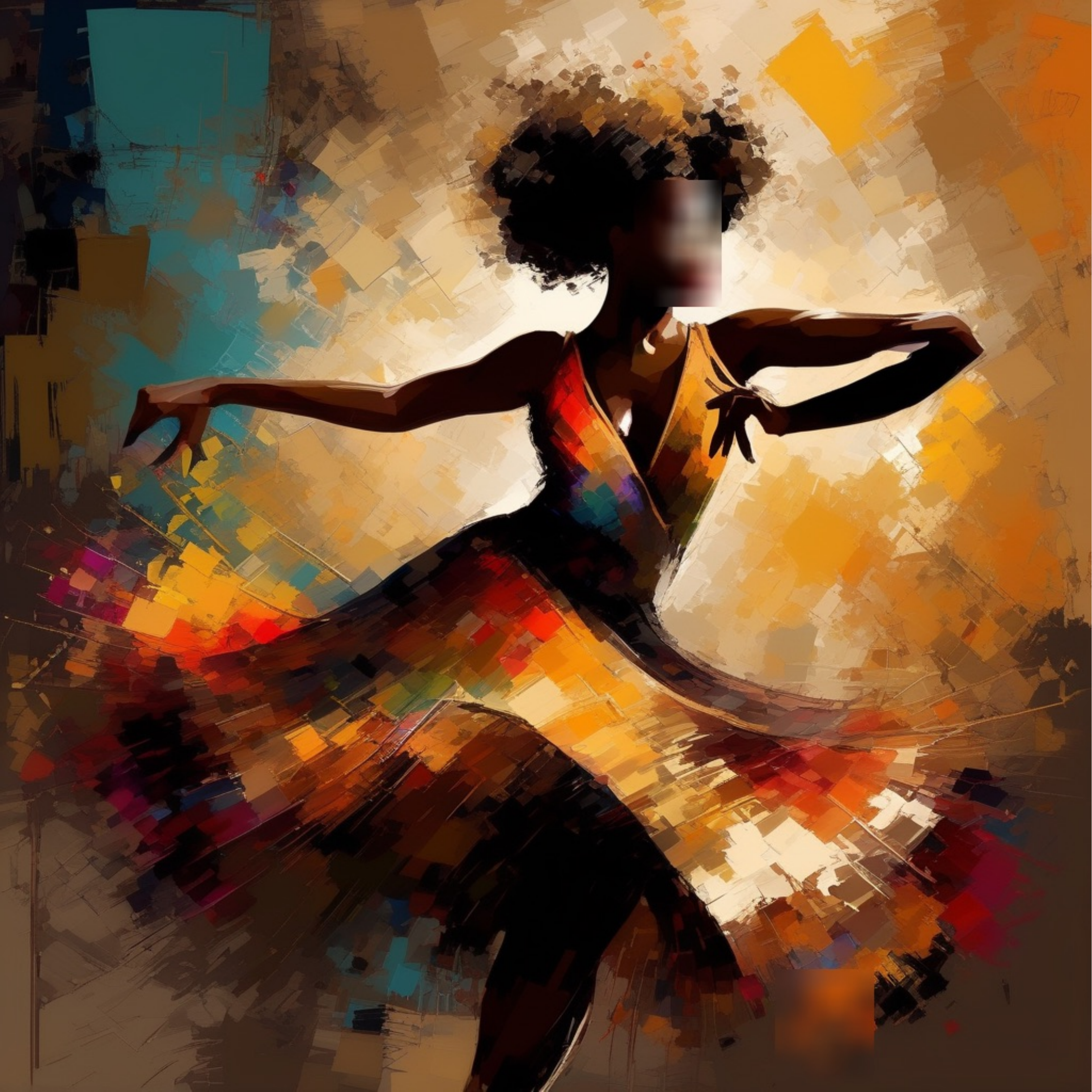} \\ 
    \end{tabular}
    
    \captionsetup{type=figure}
    \caption{Randomly chosen samples generated by SOTA text-to-image models.}
    \label{fig:bench1}
\end{table}

\begin{table}[]
    \centering
    \begin{tabular}{ccccc}
     \multicolumn{5}{p{11cm}}{
\centering{{a contemporary dancer exploring themes of identity and self-expression through movement, their performance a testament to personal liberation.}}} \\
    \hline
        \multicolumn{5}{c}{SD-1.4} \\ 
             \hline
                  \hline
        \includegraphics[width=2.3cm]{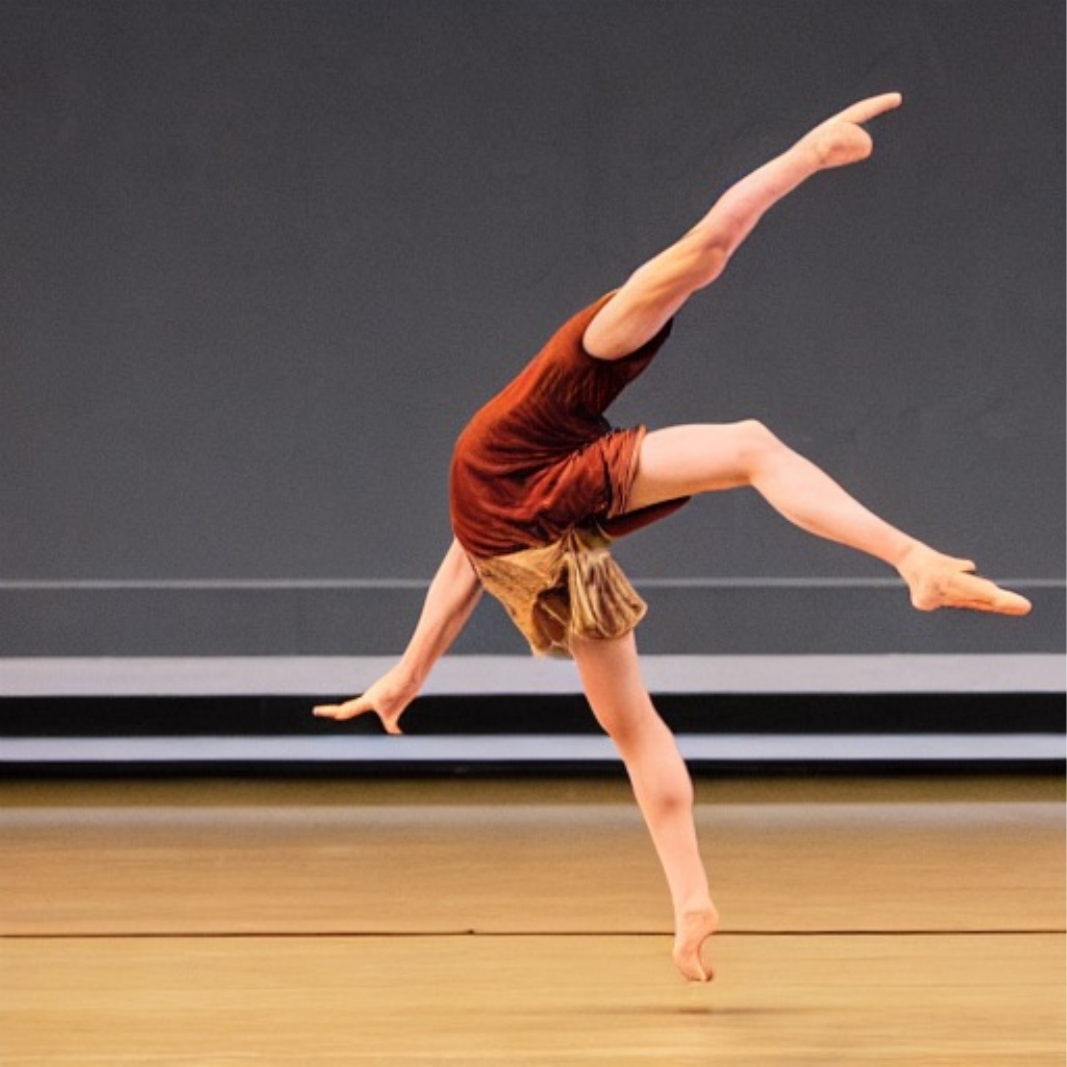} & 
        \includegraphics[width=2.3cm]{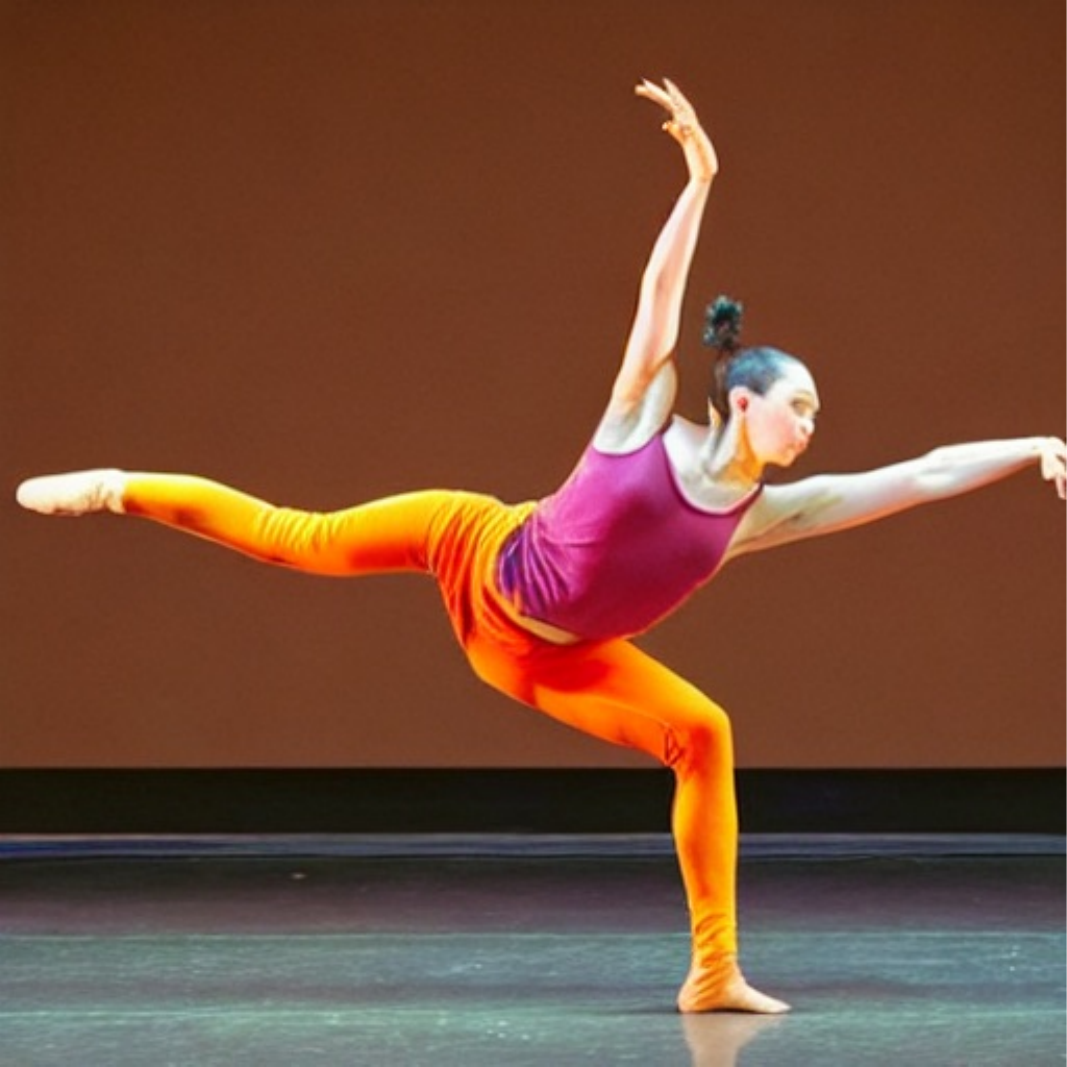} & 
        \includegraphics[width=2.3cm]{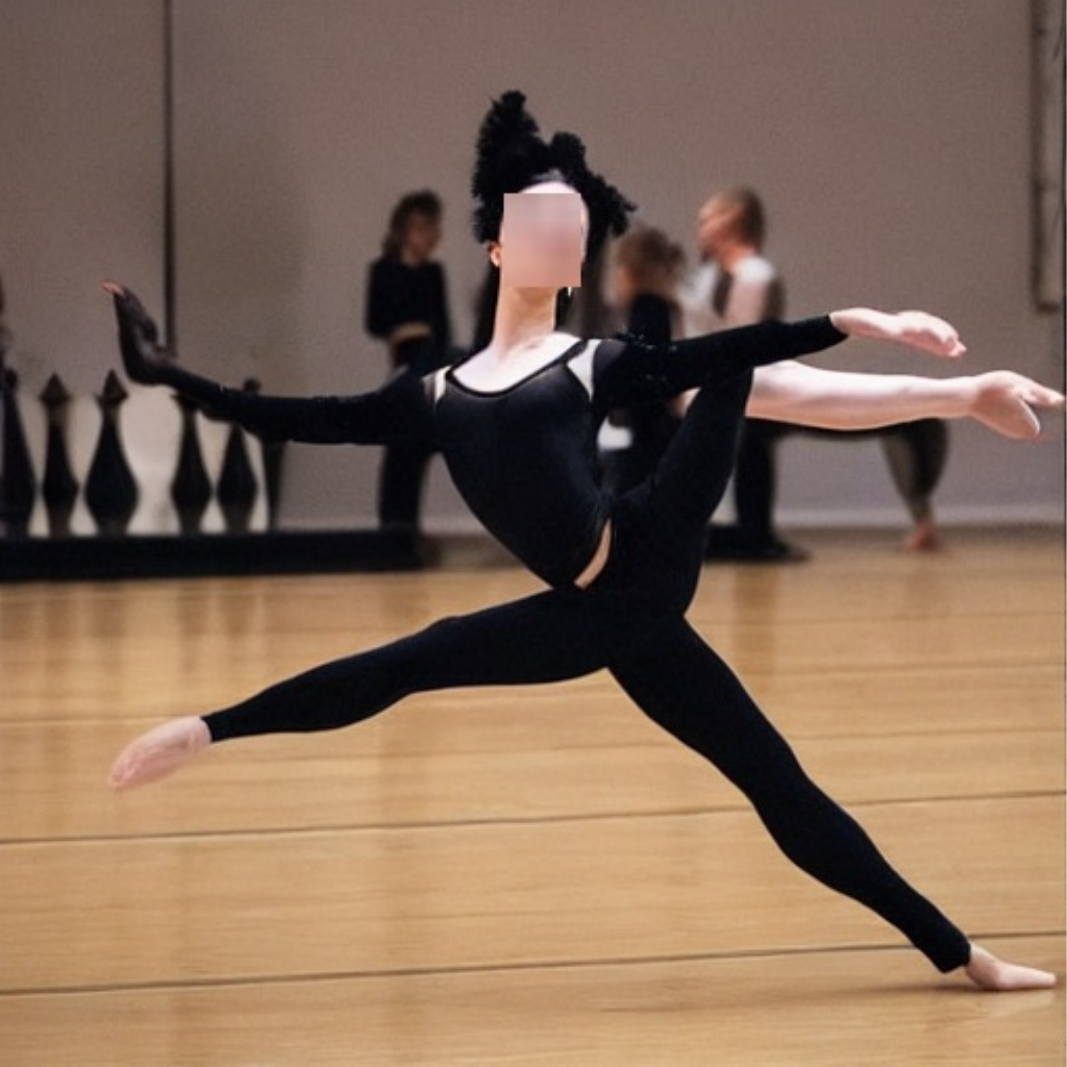} & 
        \includegraphics[width=2.3cm]{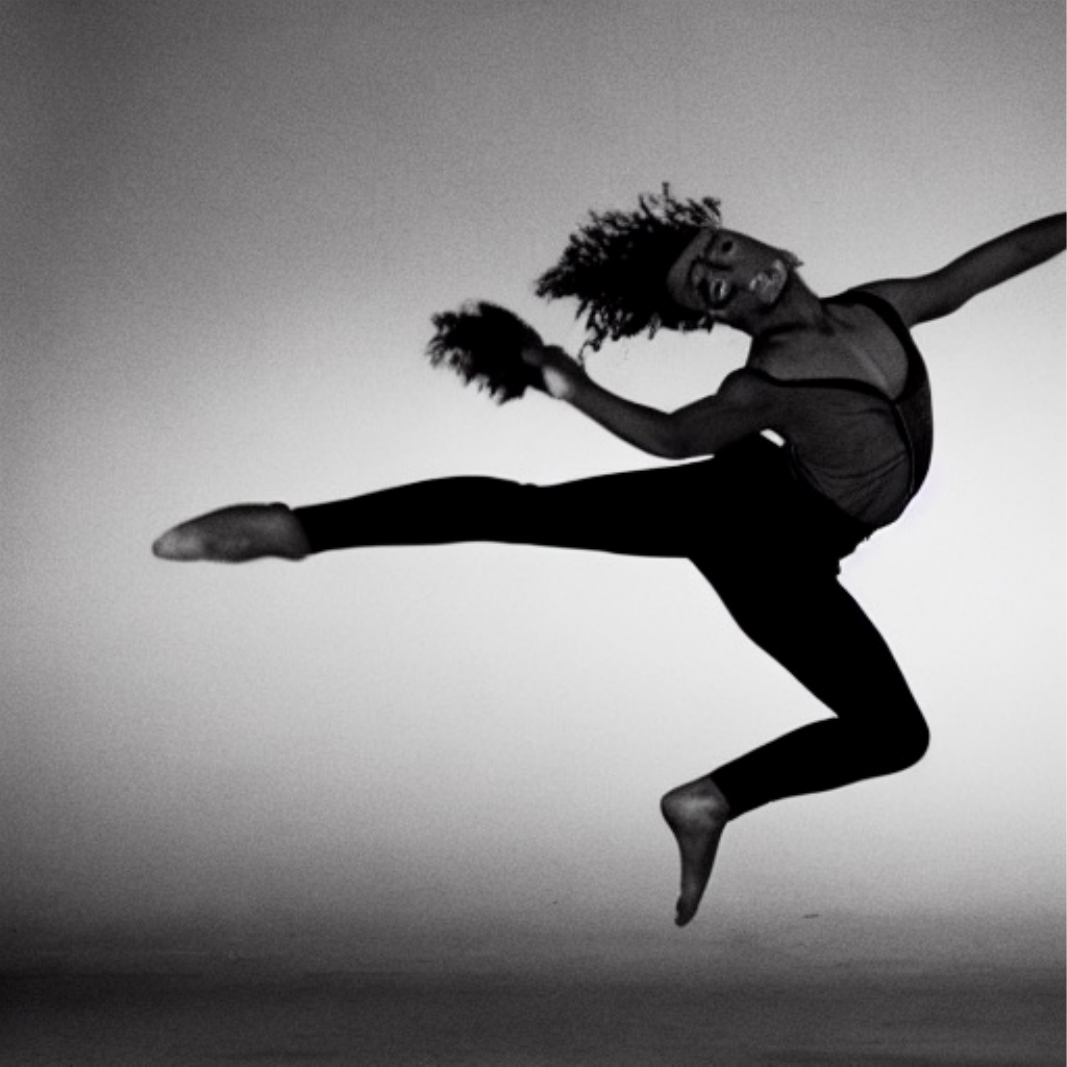} & 
        \includegraphics[width=2.3cm]{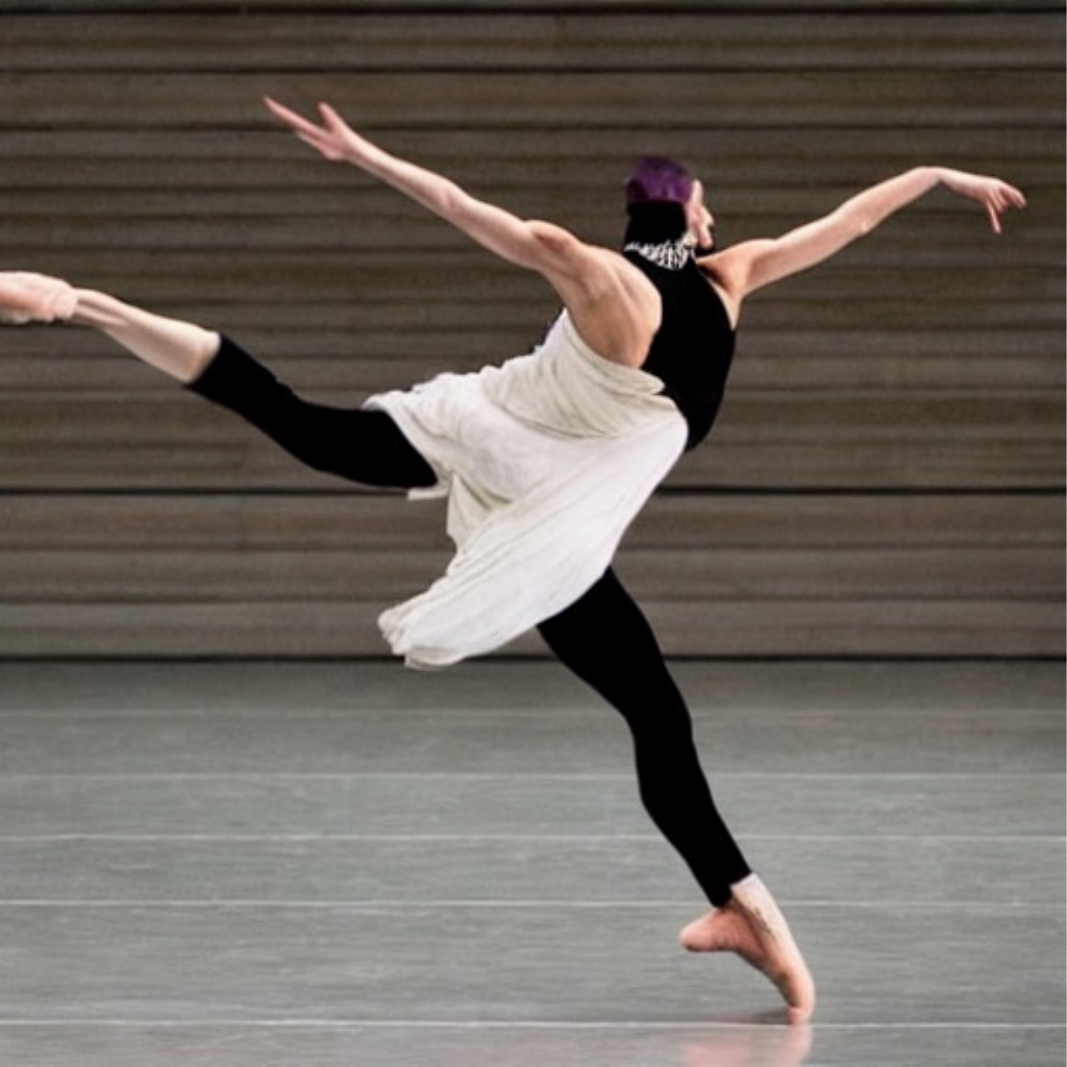} \\
        \multicolumn{5}{c}{SD-XL-Turbo} \\
        \hline
        \hline
        \includegraphics[width=2.3cm]{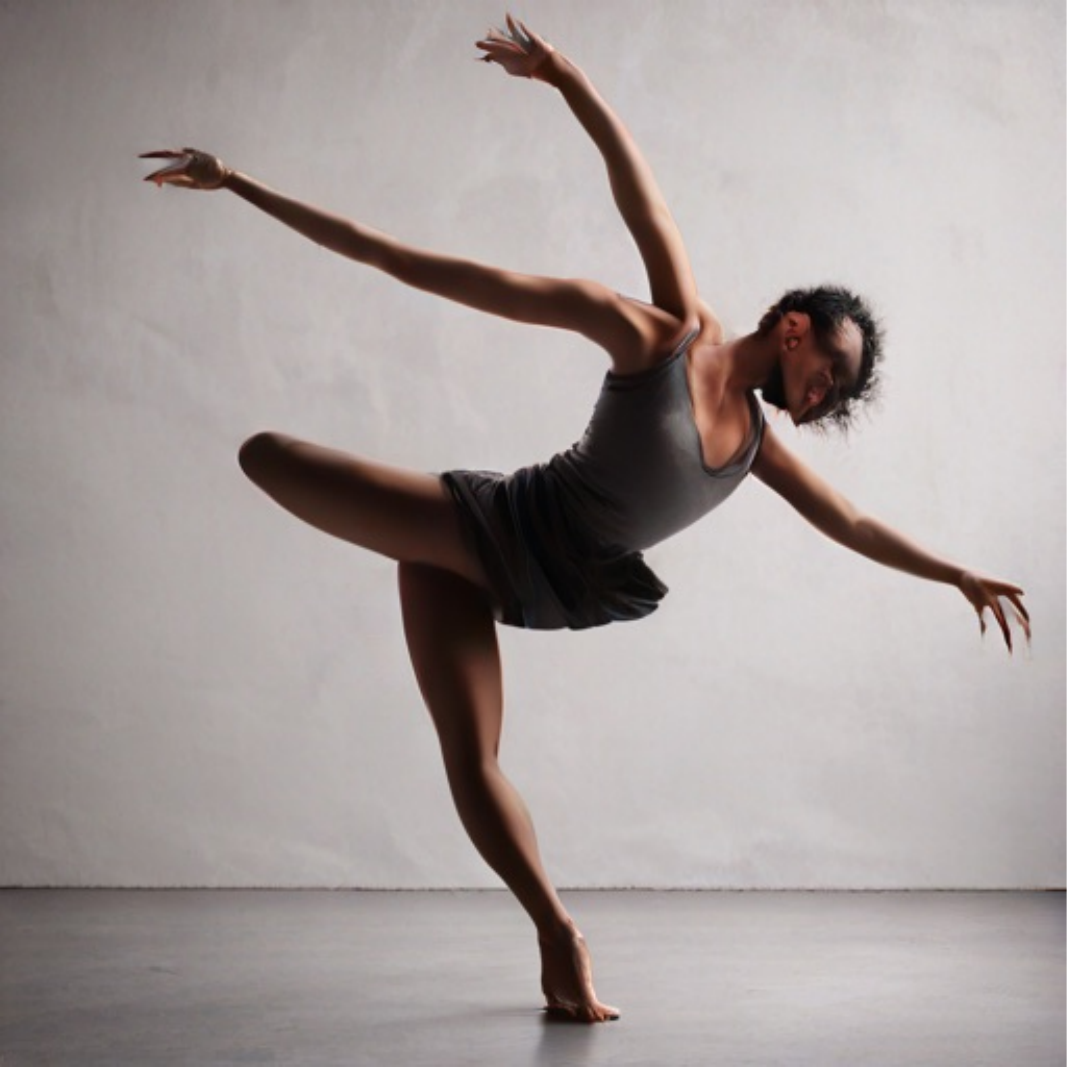} &    
        \includegraphics[width=2.3cm]{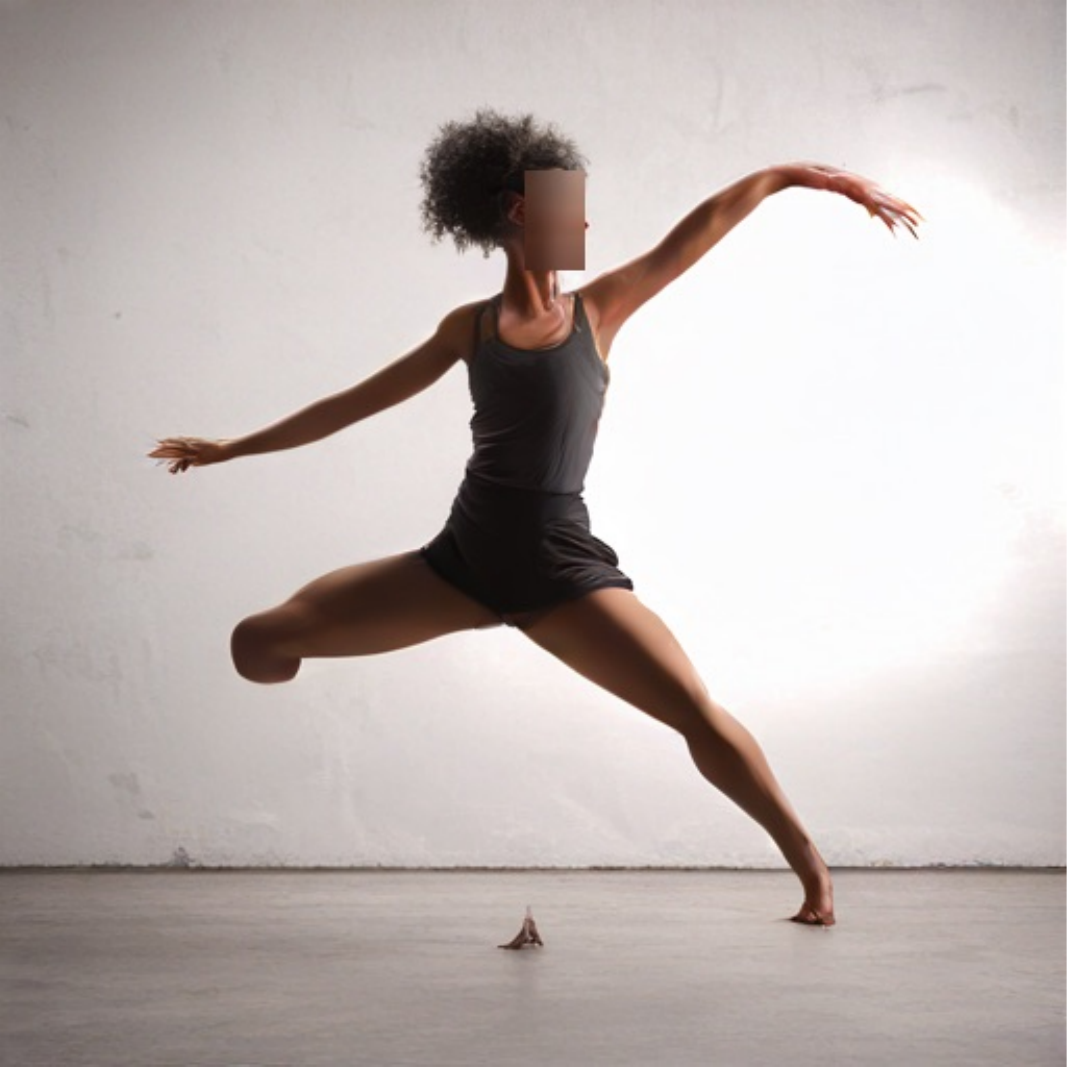} &
        \includegraphics[width=2.3cm]{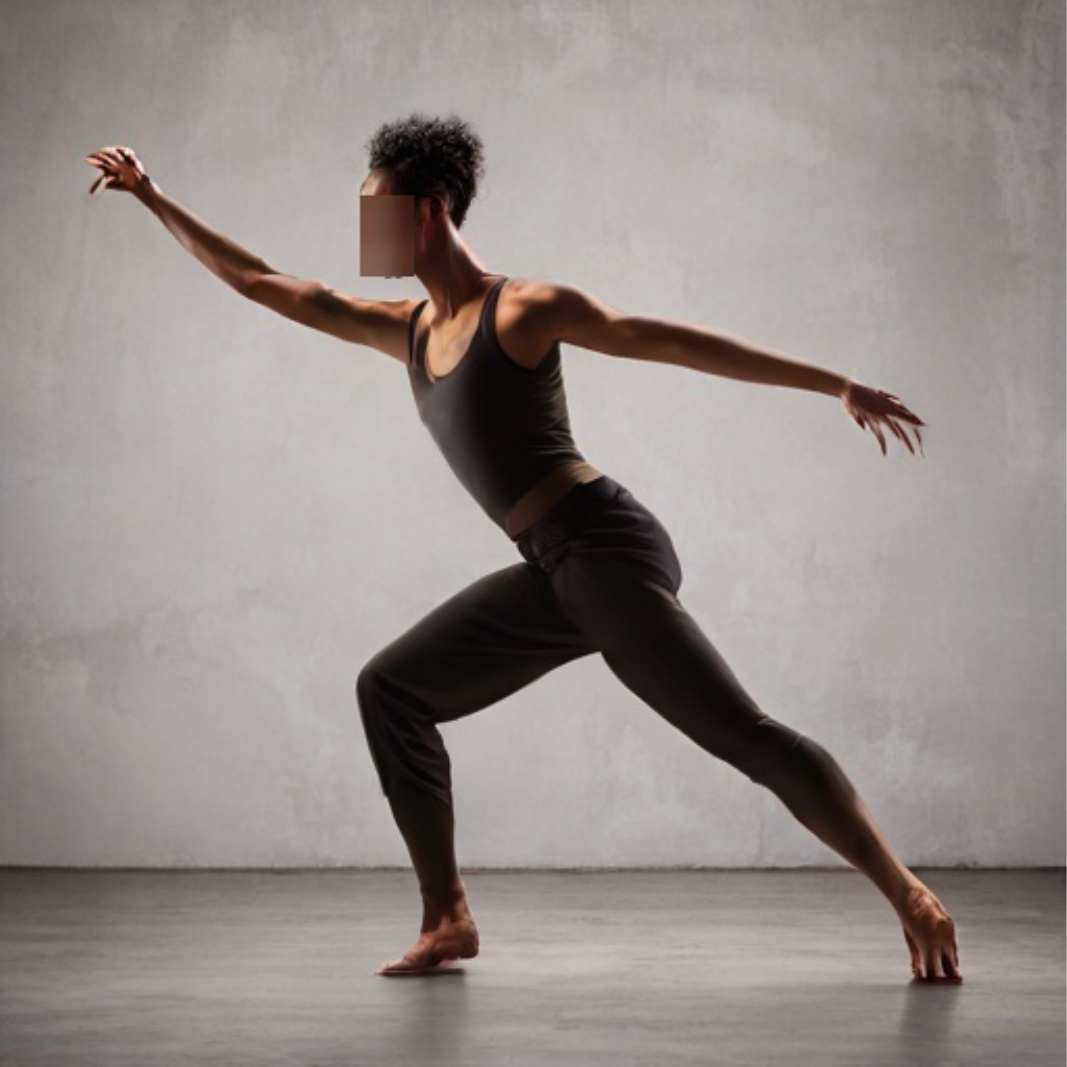} &
        \includegraphics[width=2.3cm]{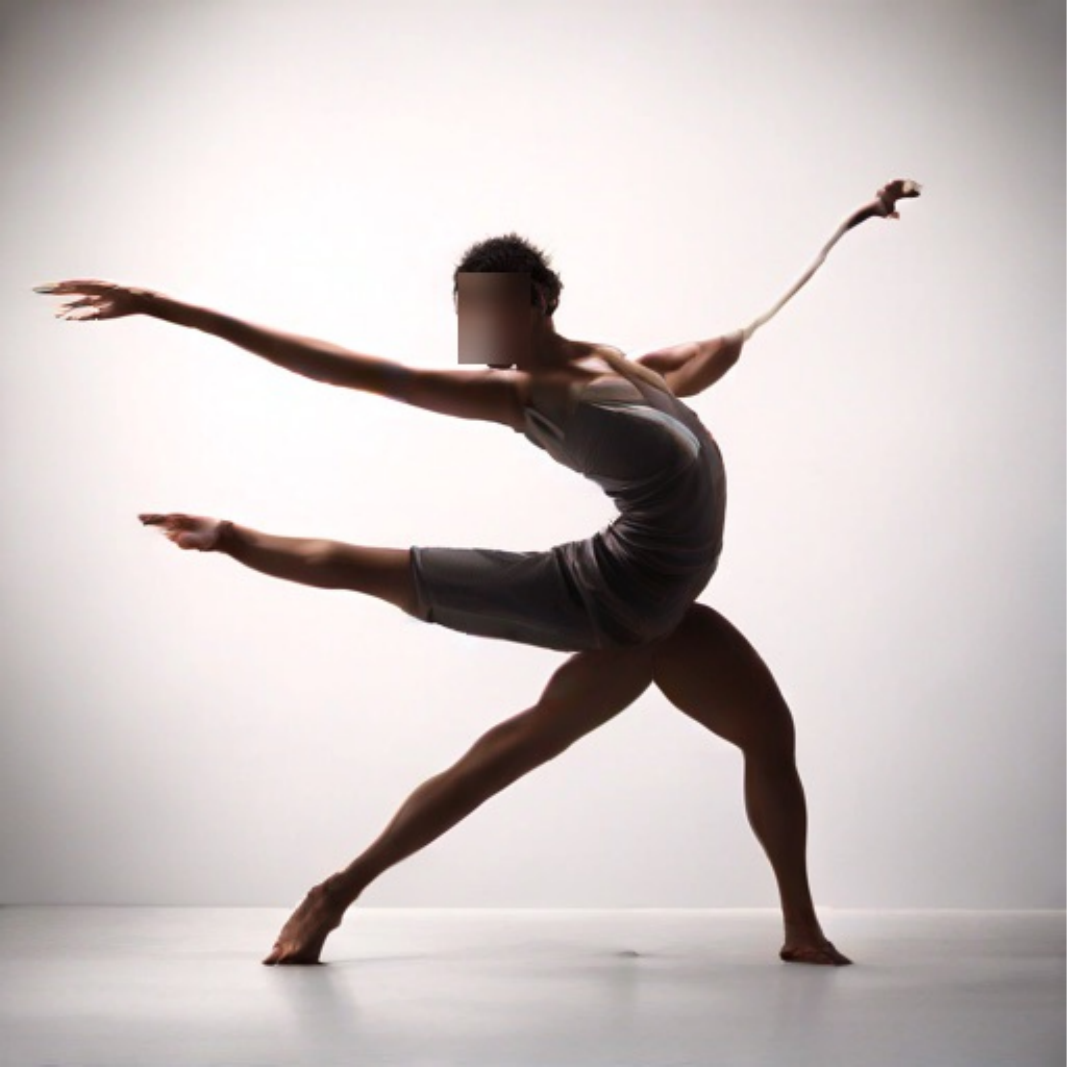} &
        \includegraphics[width=2.3cm]{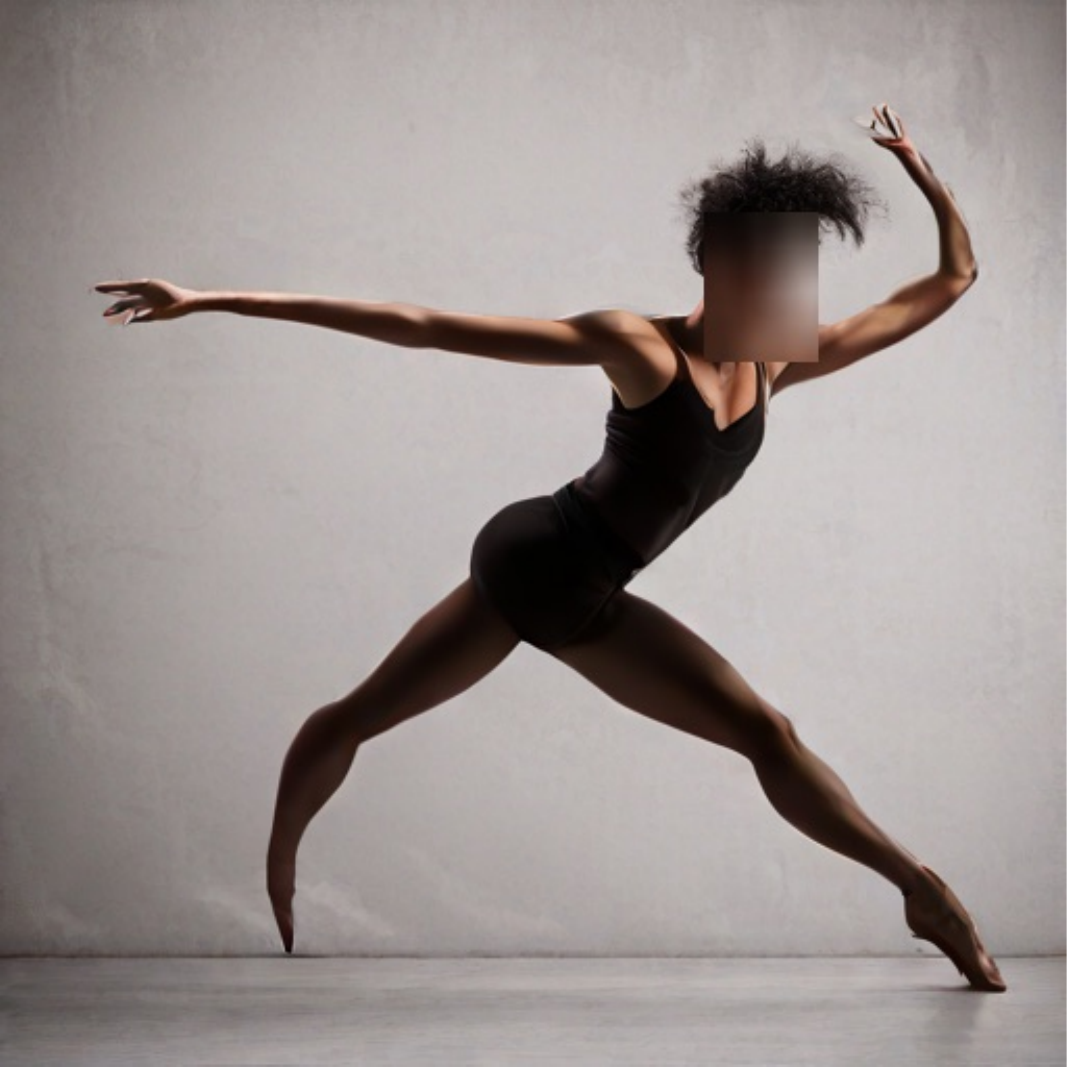} \\
        \multicolumn{5}{c}{SD-2.1} \\ 
        \hline
        \hline
        \includegraphics[width=2.3cm]{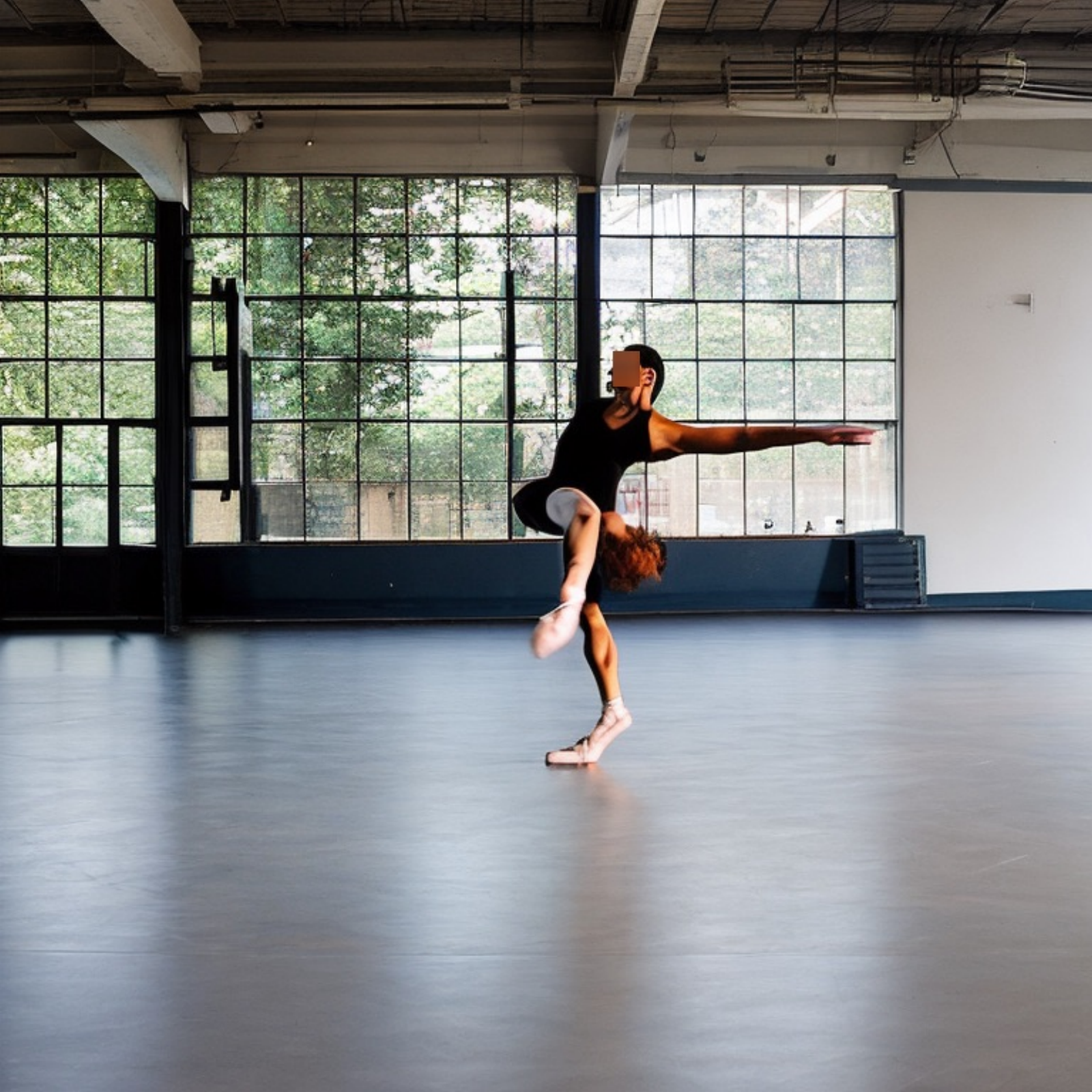}& \includegraphics[width=2.3cm]{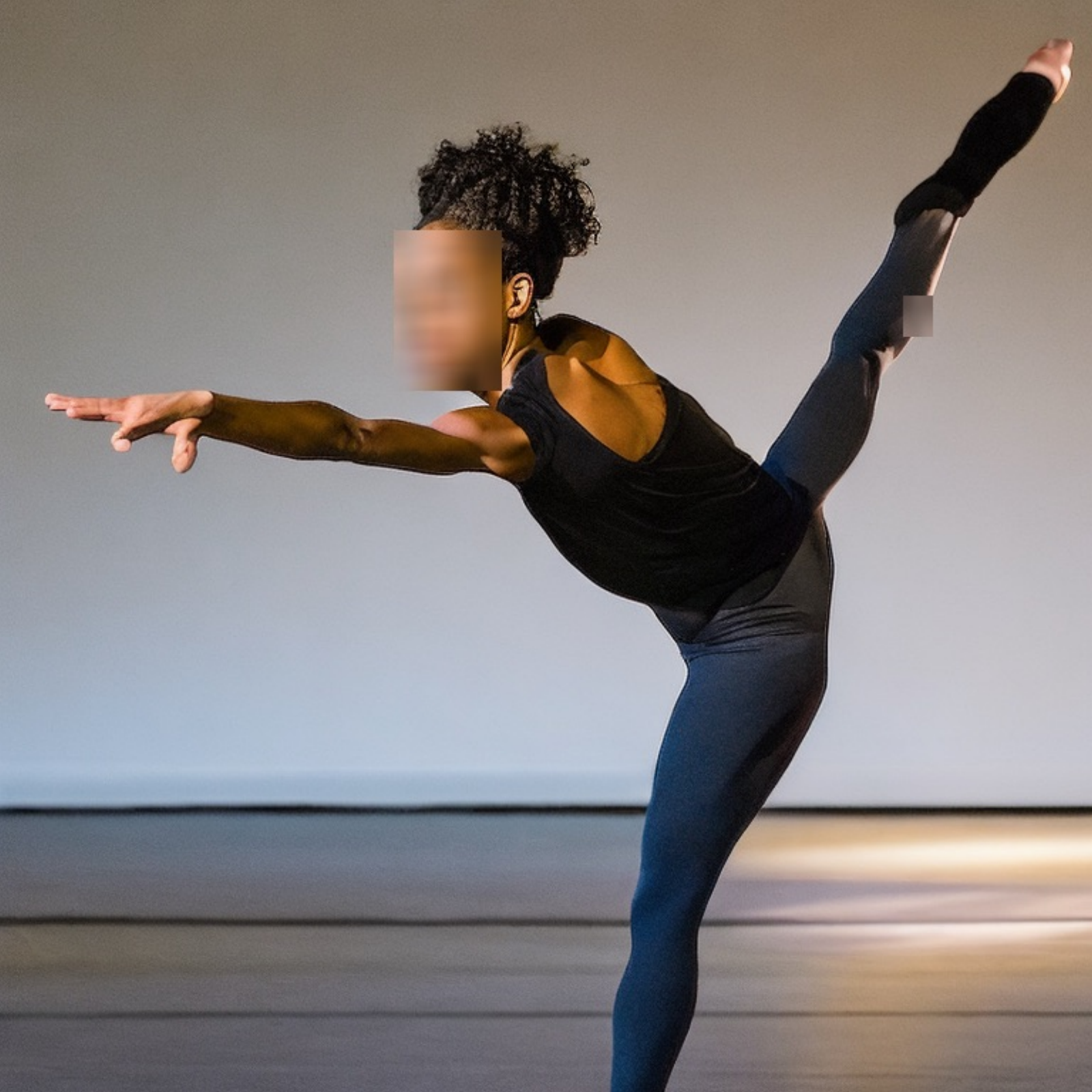} &     \includegraphics[width=2.3cm]{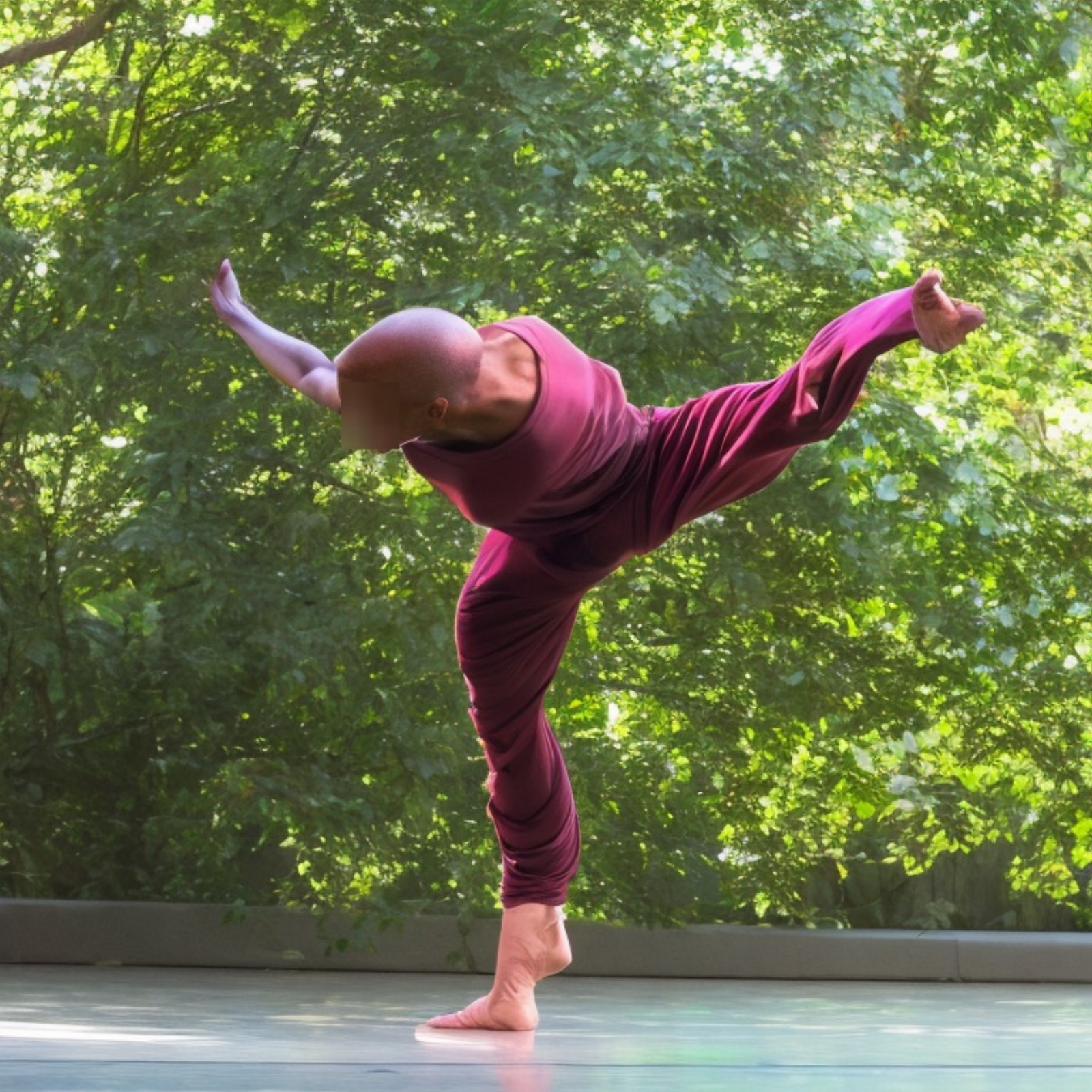} &     \includegraphics[width=2.3cm]{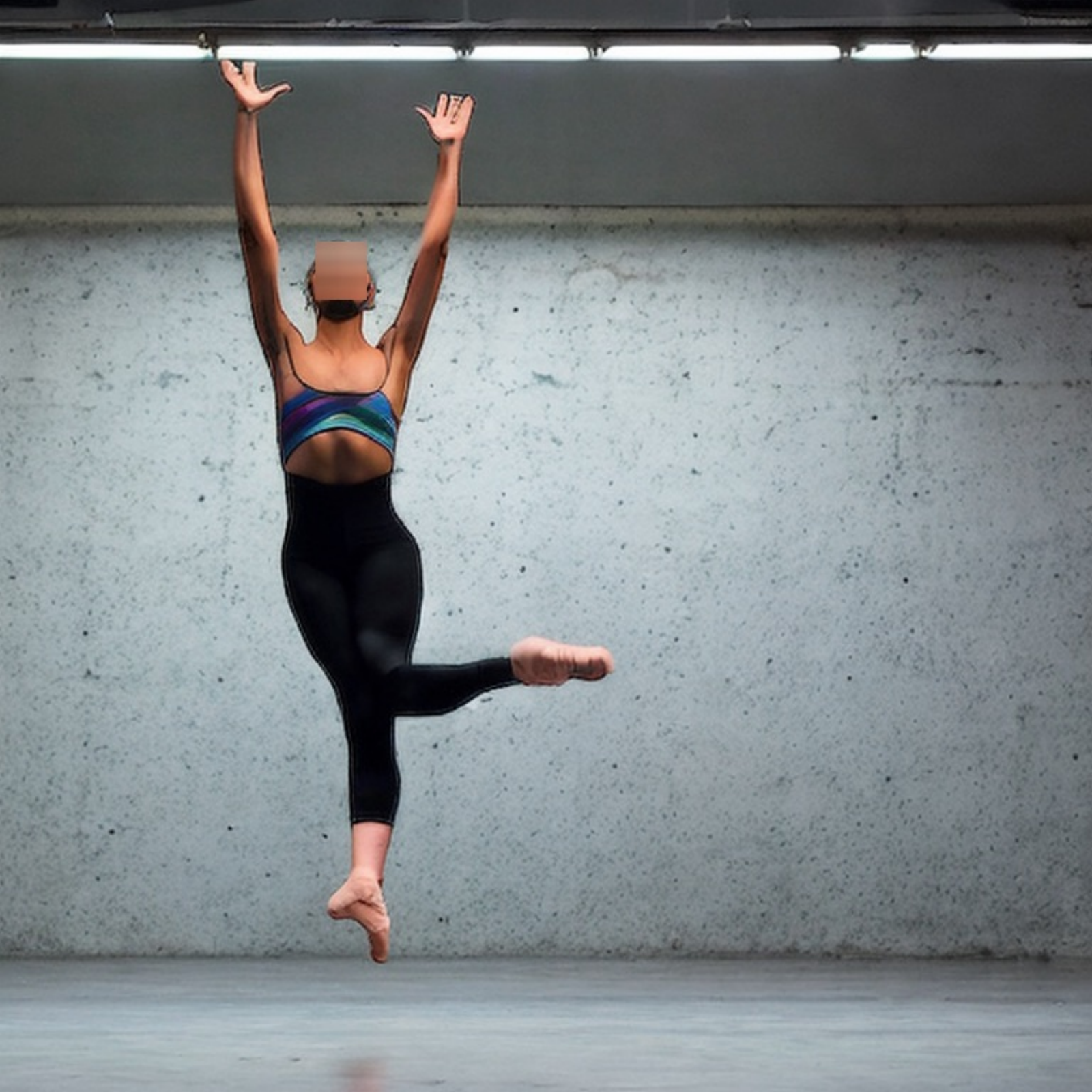} &     \includegraphics[width=2.3cm]{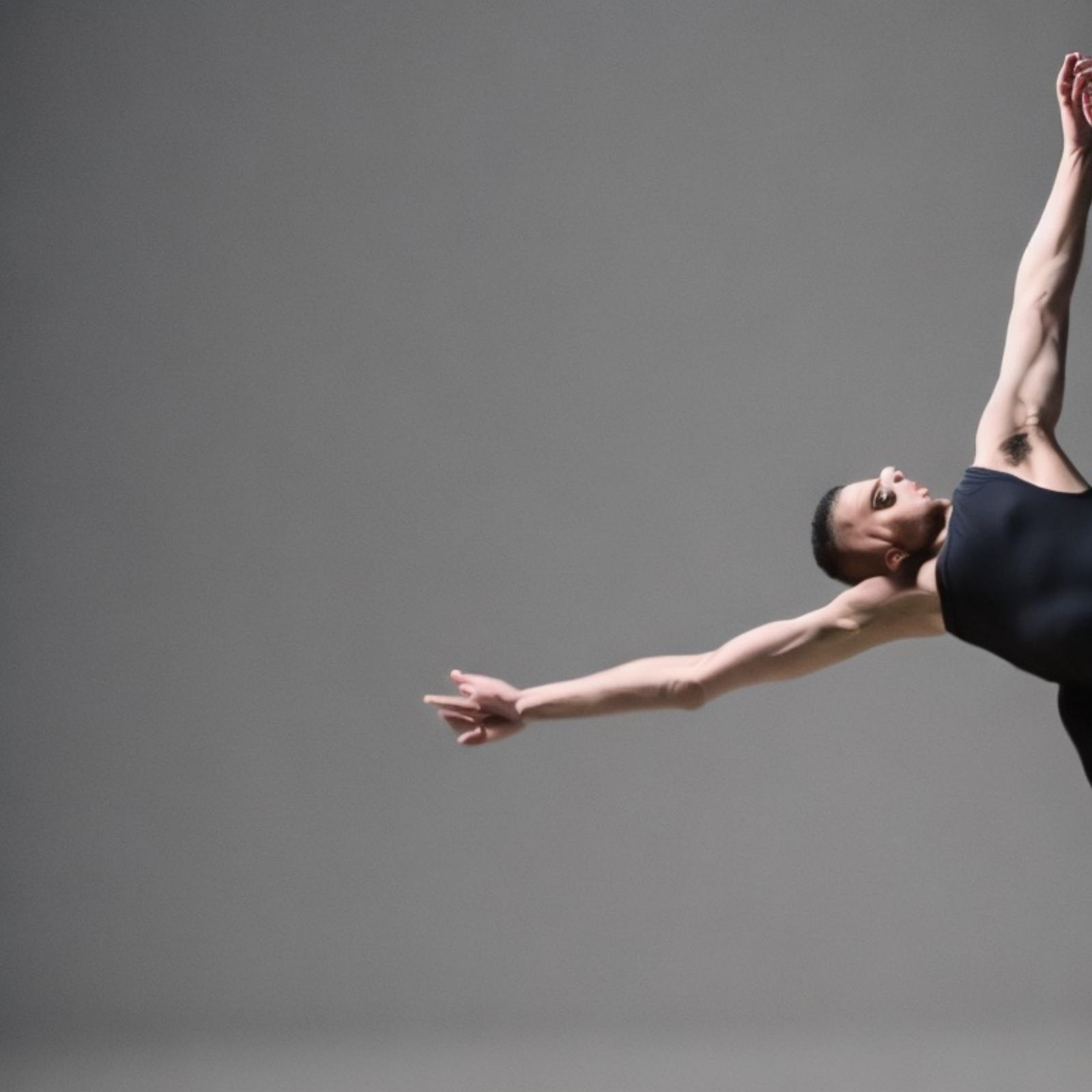} \\
        \multicolumn{5}{c}{Wuerstchen} \\ 
        \hline
        \hline
        \includegraphics[width=2.3cm]{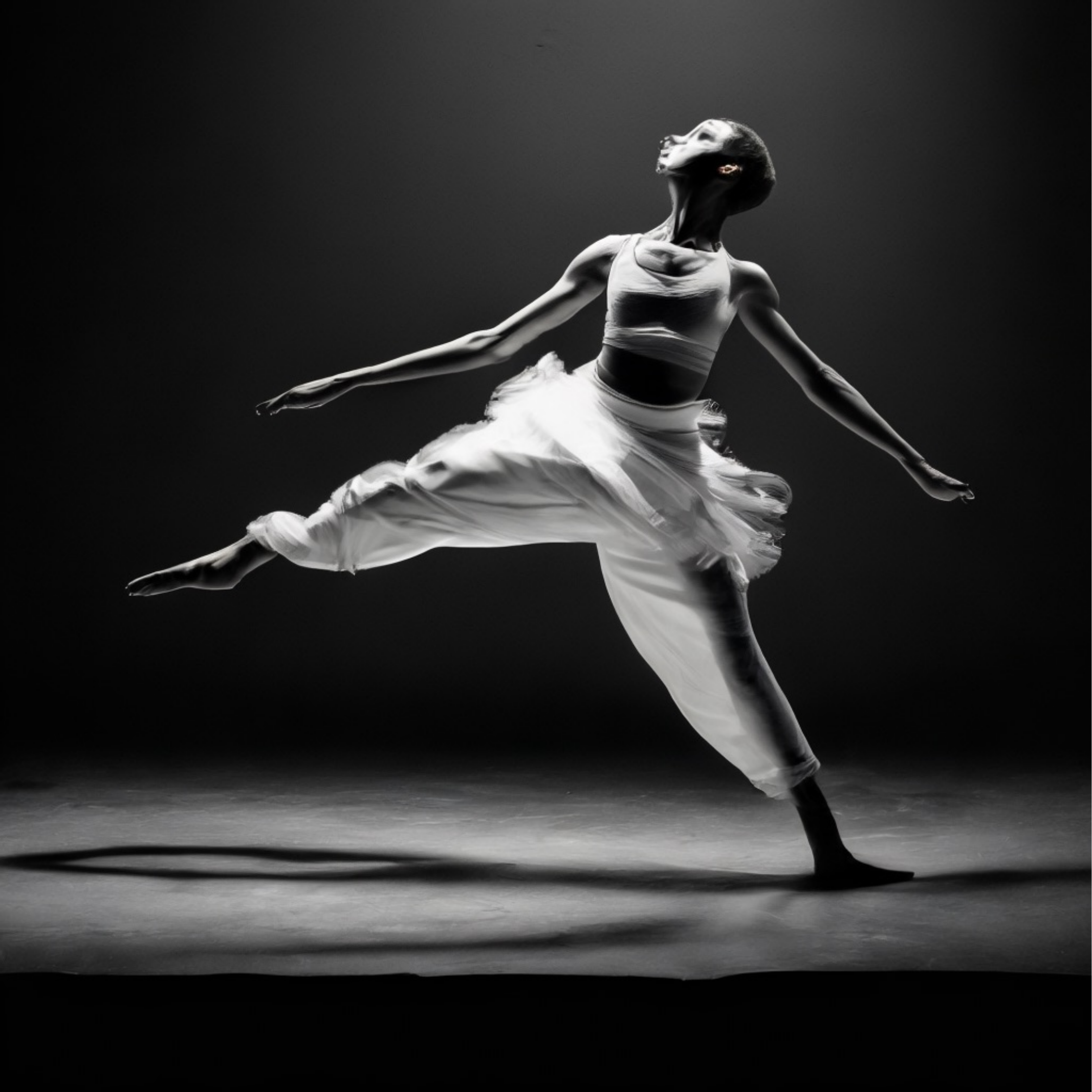} &\includegraphics[width=2.3cm]{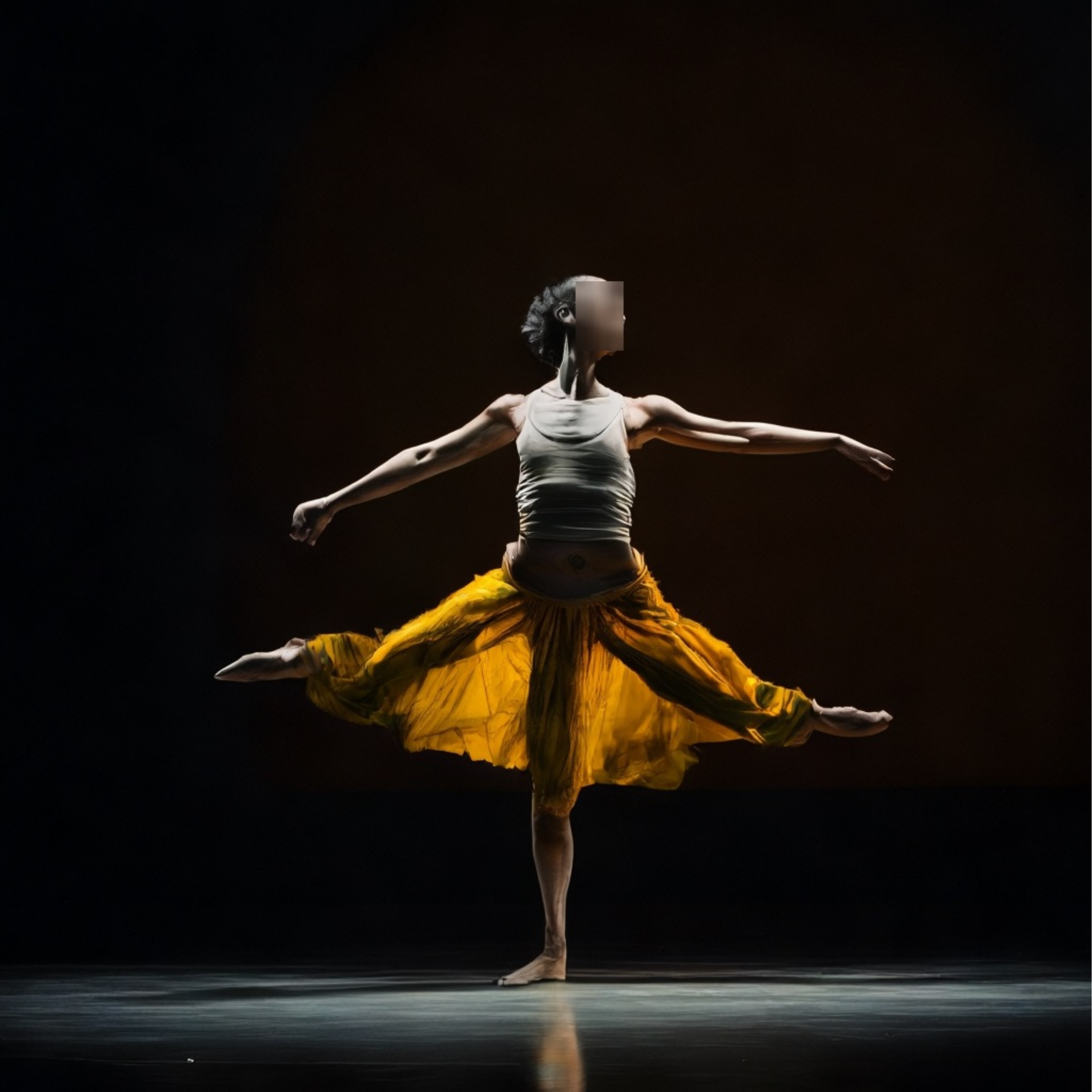} &\includegraphics[width=2.3cm]{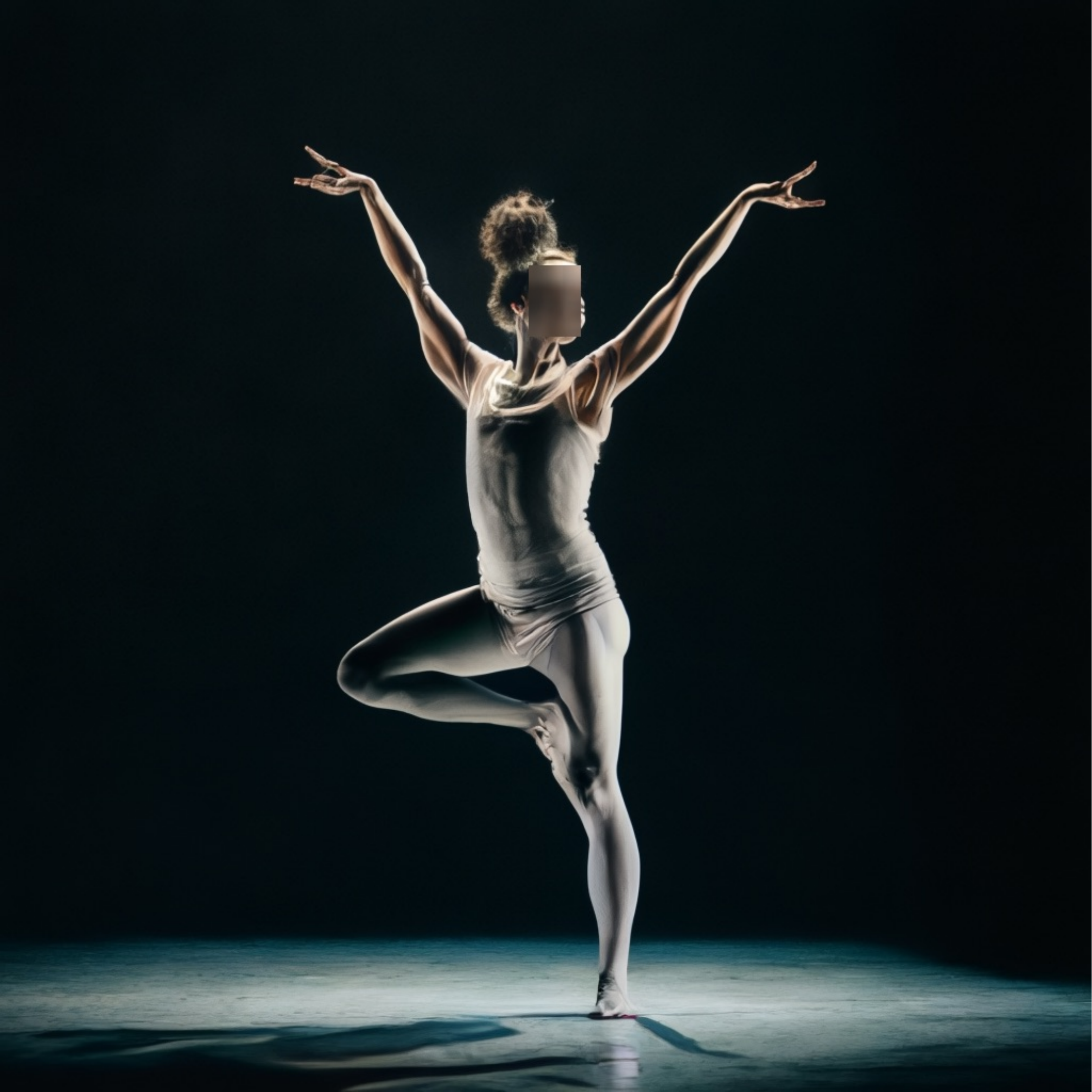} &\includegraphics[width=2.3cm]{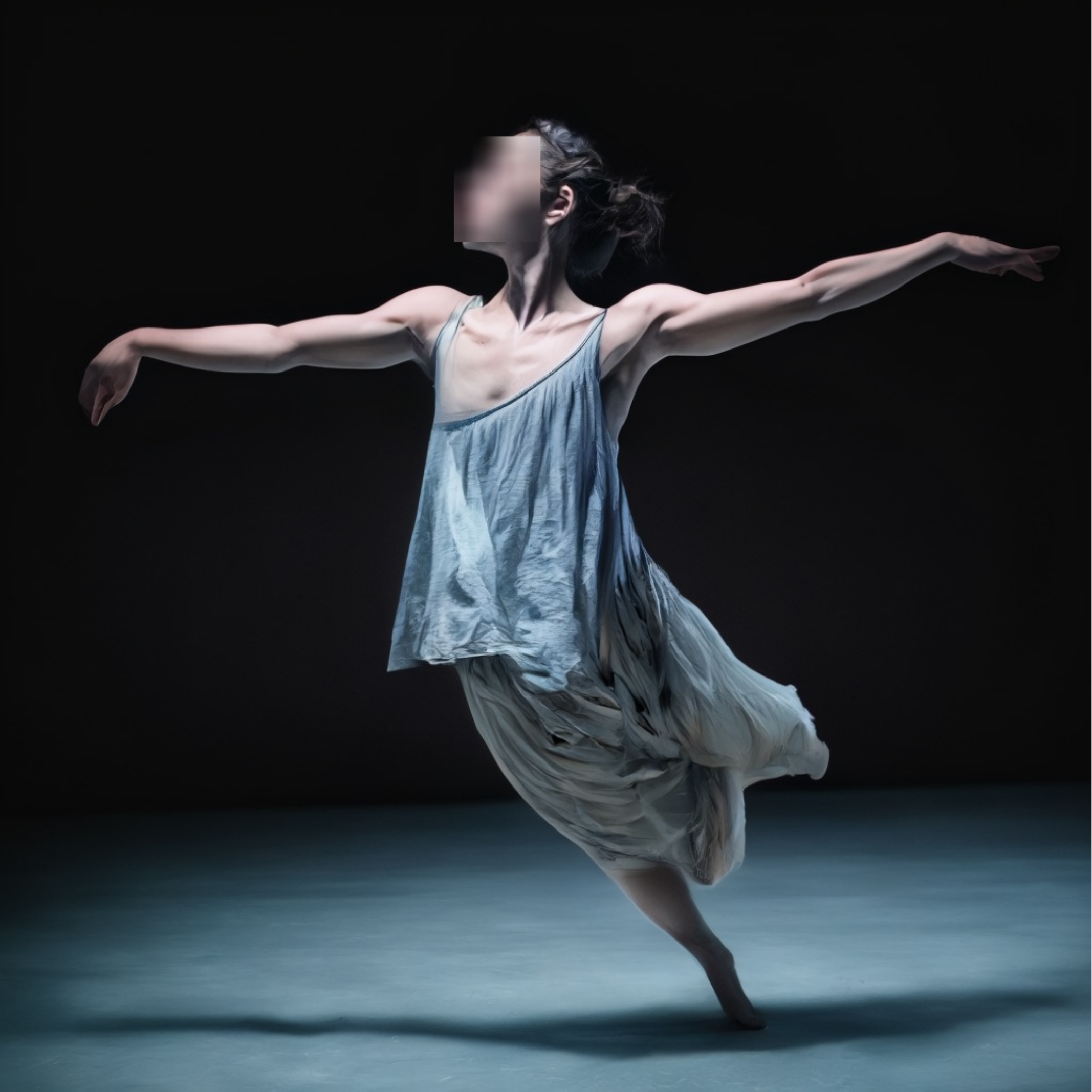} &\includegraphics[width=2.3cm]{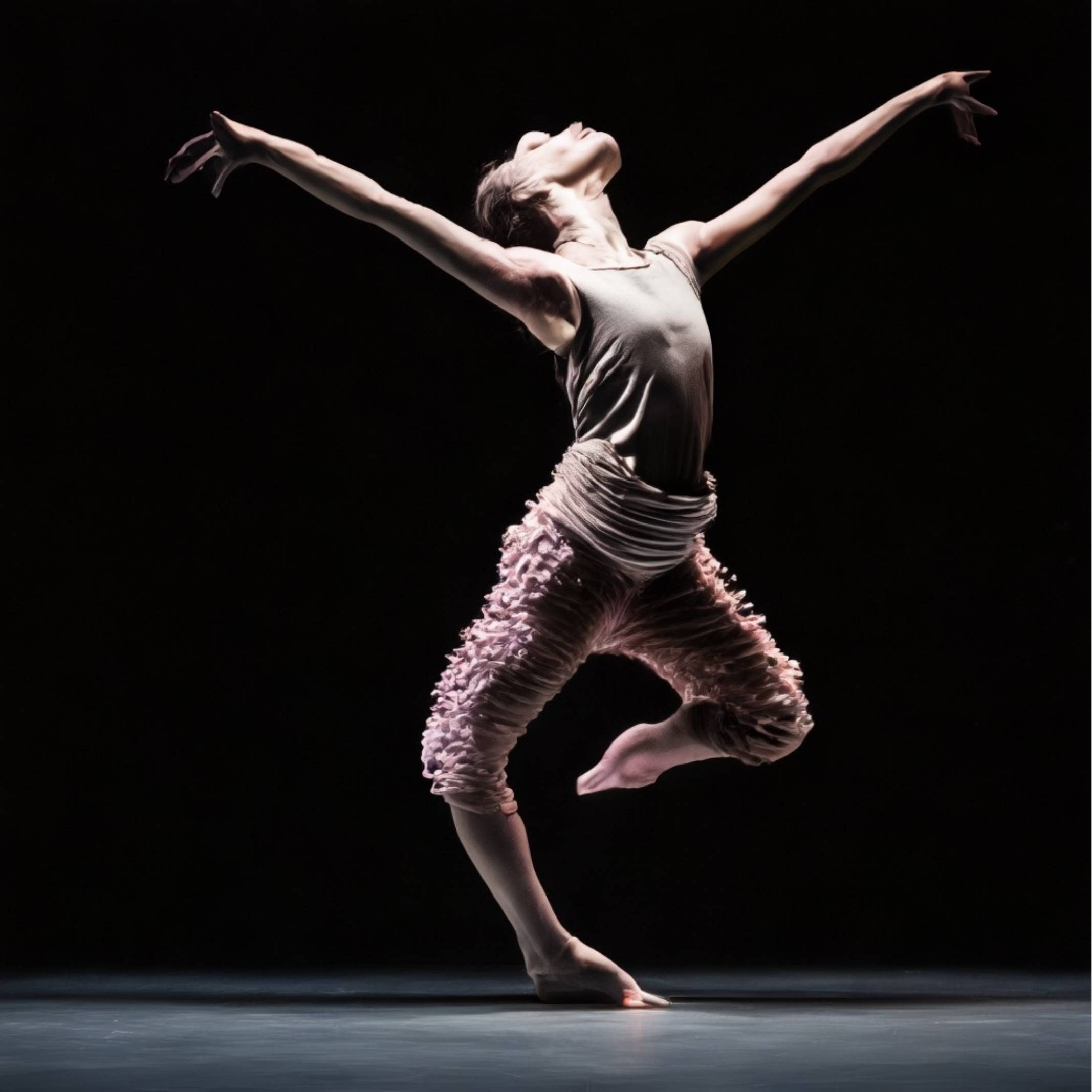} \\
        \multicolumn{5}{c}{SD-XL} \\ 
        \hline
        \hline
        \includegraphics[width=2.3cm]{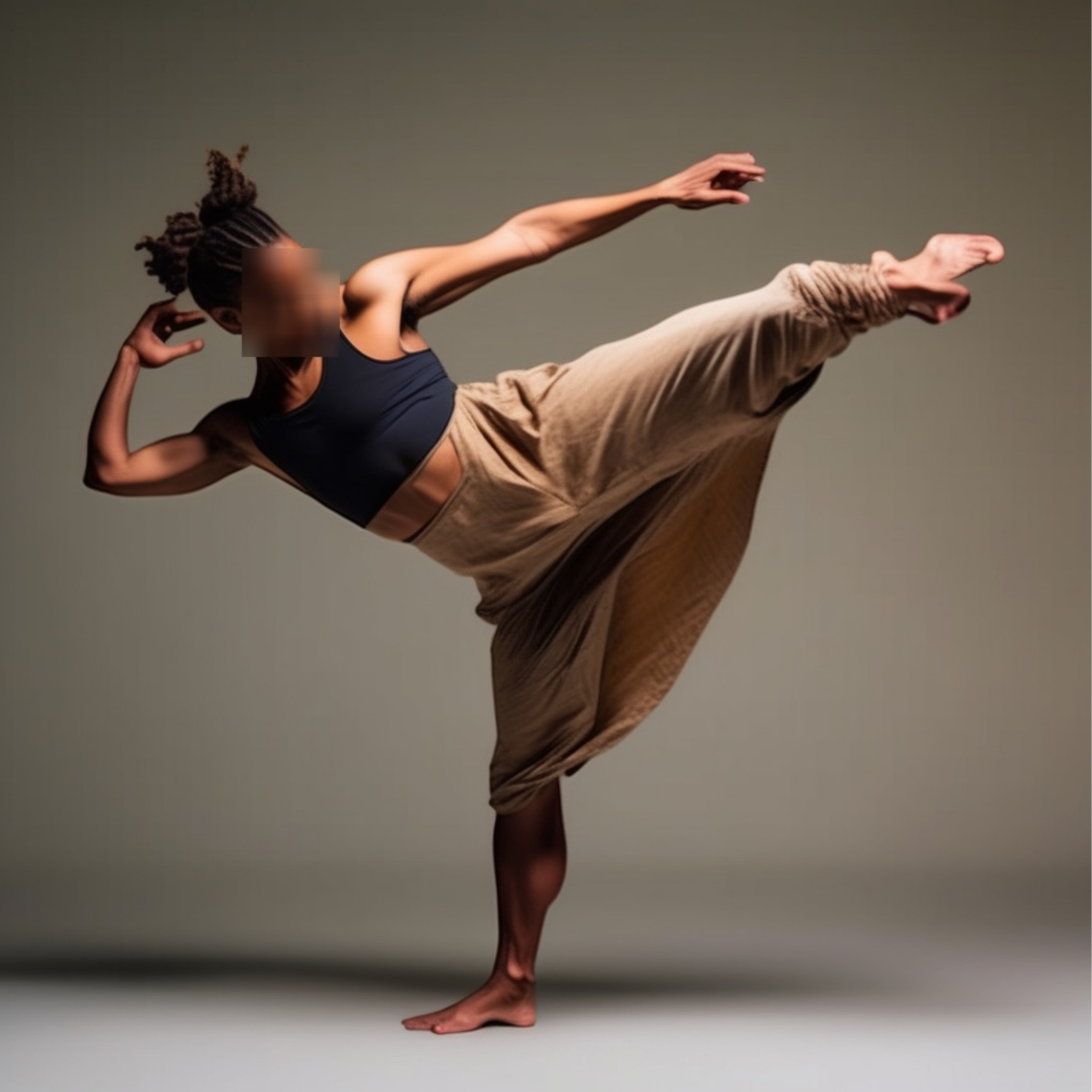} & \includegraphics[width=2.3cm]{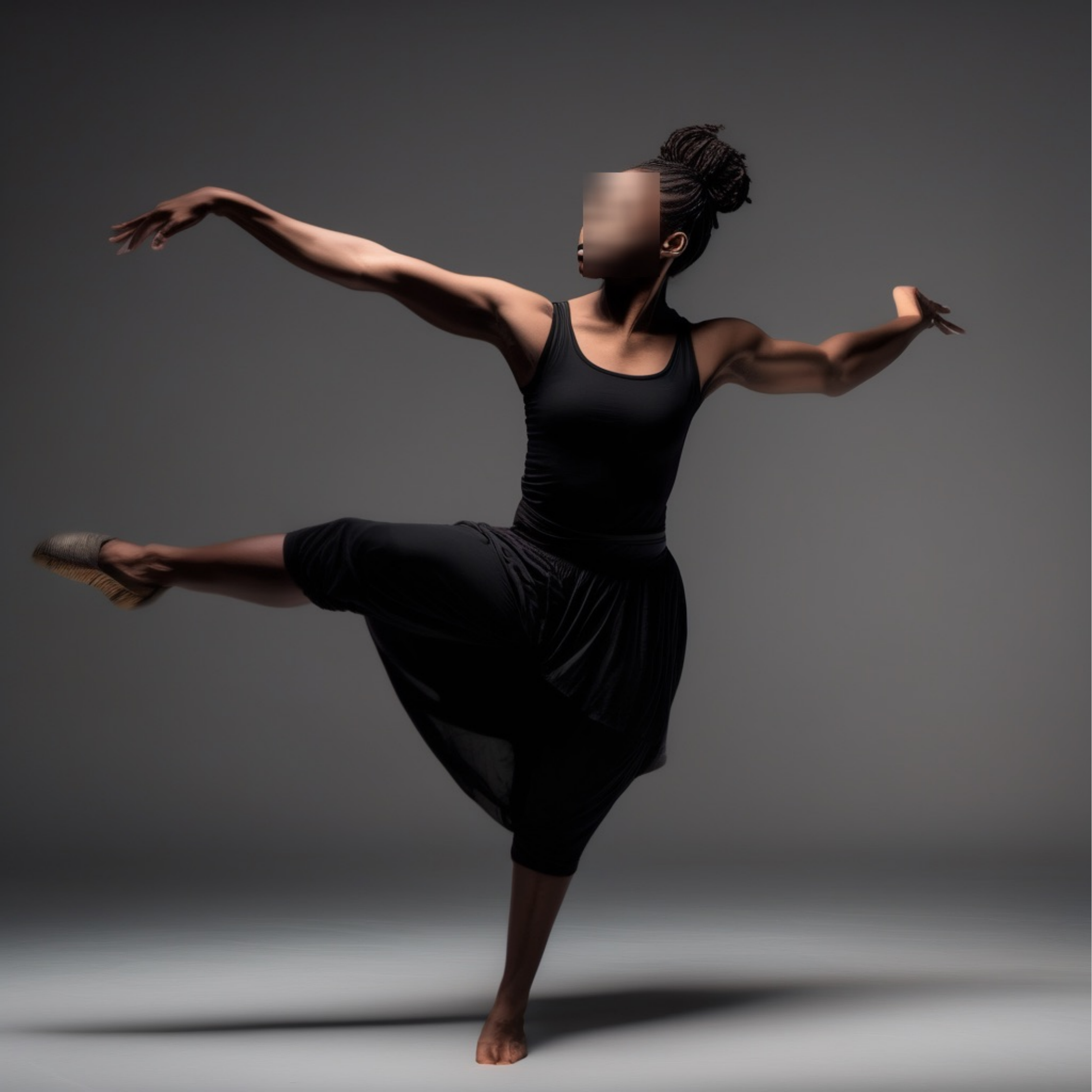} & \includegraphics[width=2.3cm]{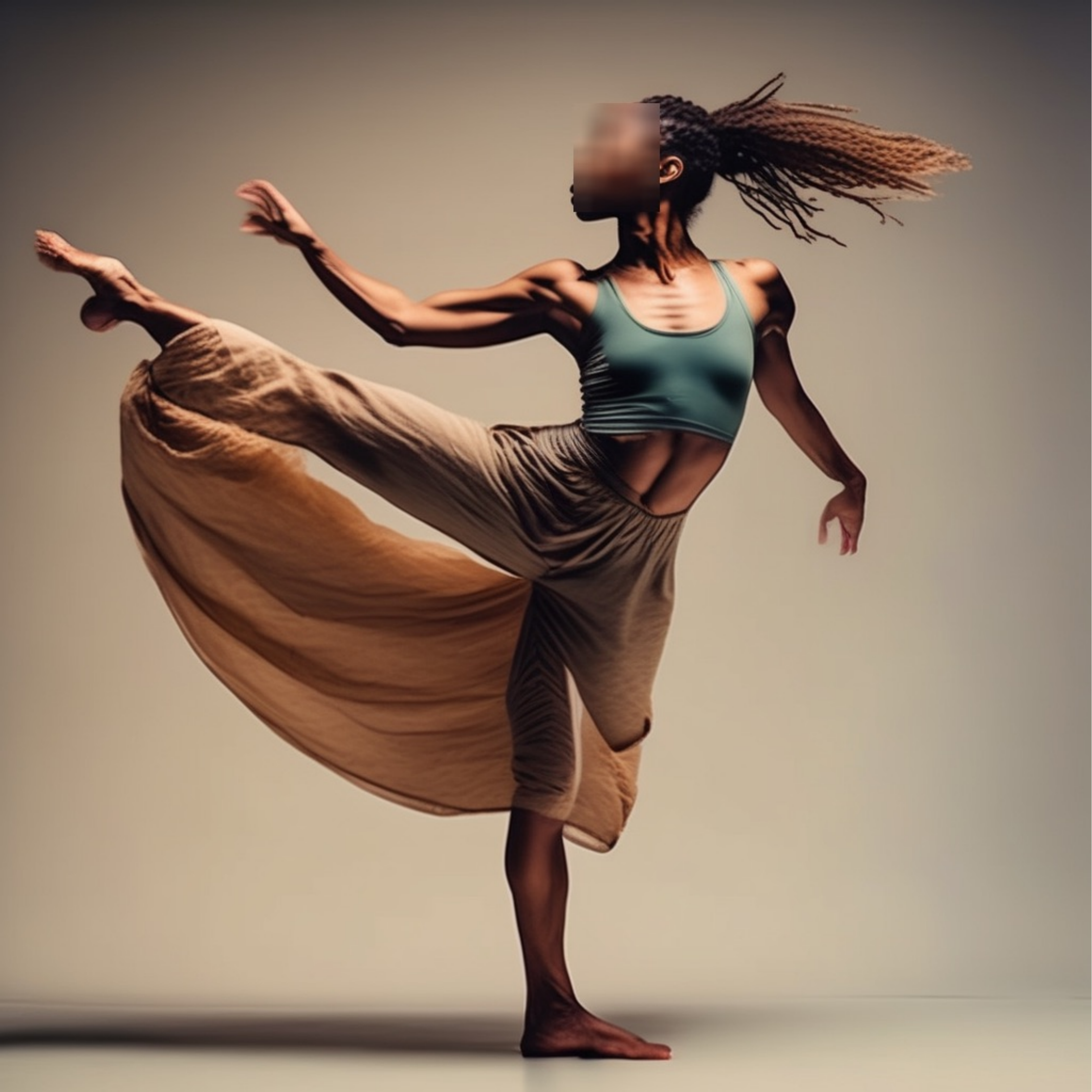} & \includegraphics[width=2.3cm]{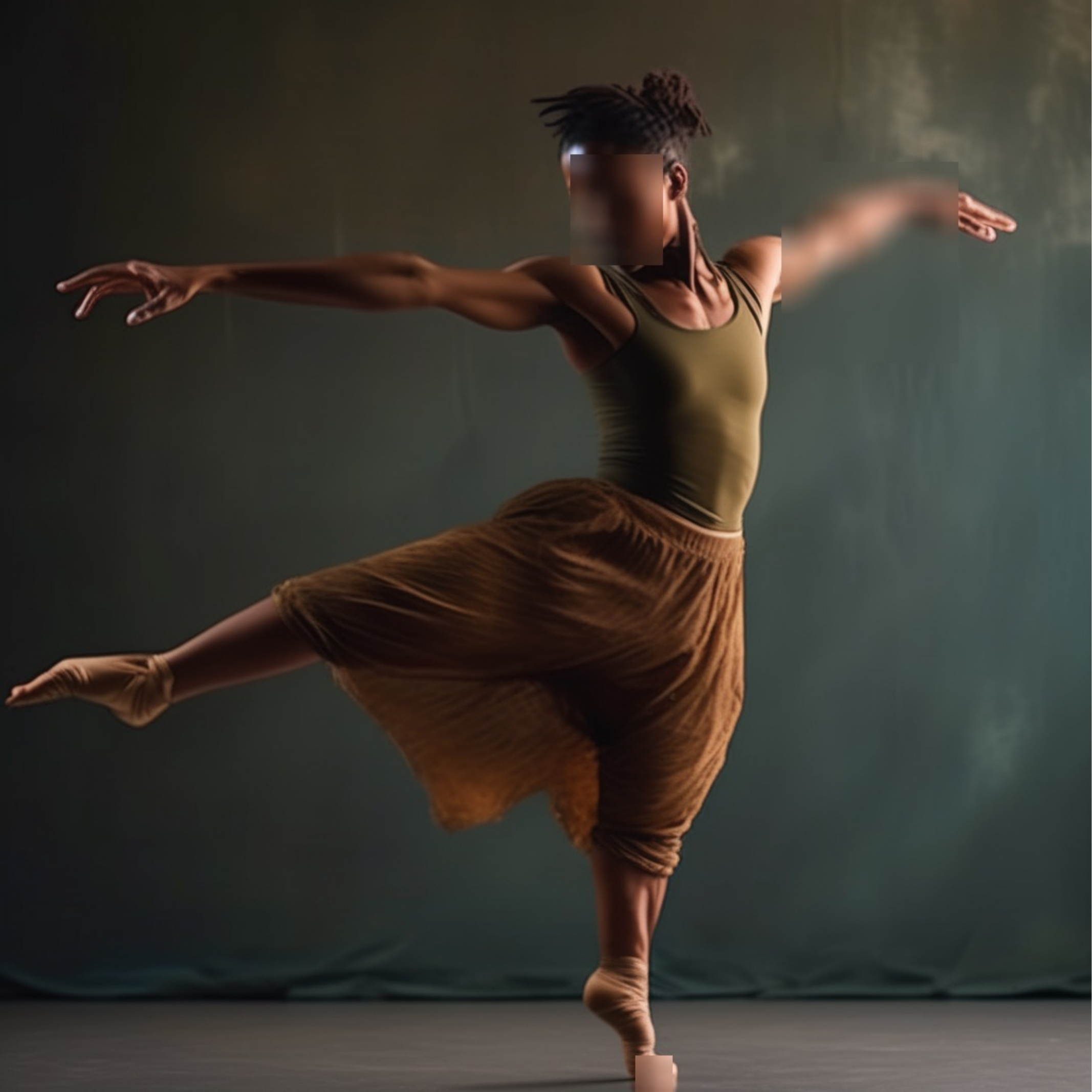} & \includegraphics[width=2.3cm]{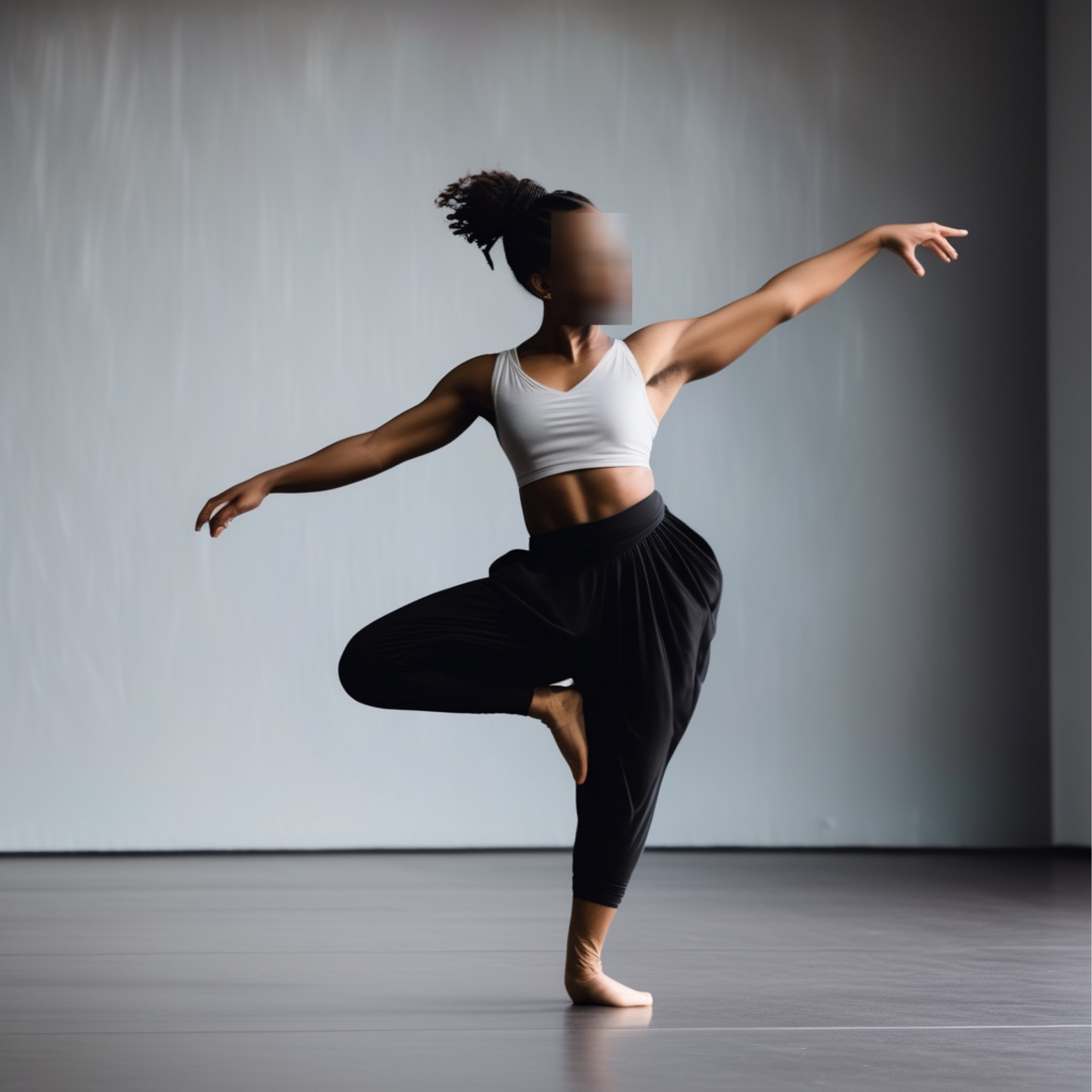} \\ 
      
    \end{tabular}
    
    \captionsetup{type=figure}
    \caption{Randomly chosen samples generated by SOTA text-to-image models.}
\label{fig:bench2}
\end{table}

\clearpage
\bibliographystyle{splncs}
\bibliography{egbib}
\end{document}